\documentclass[runningheads]{llncs}

\usepackage{eccv}

\usepackage{eccvabbrv}

\usepackage{graphicx}
\usepackage{subcaption}
\usepackage{booktabs}
\usepackage{amsmath}
\usepackage{amssymb}
\usepackage{mathtools}

\usepackage{microtype}
\usepackage{color}

\usepackage{soul}

\usepackage[dvipsnames]{xcolor}
\usepackage{colortbl}
\usepackage{makecell}
\usepackage{float}

\usepackage{adjustbox}
\usepackage{engord} 
\usepackage{caption}
\usepackage{subcaption}

\usepackage{tabularx}
\usepackage{makecell}
\usepackage{multirow}
\usepackage{wrapfig}

\usepackage[accsupp]{axessibility}  

\usepackage{hyperref}
\providecommand{\citep}[1]{\cite{#1}}
\providecommand{\citet}[1]{\textcite{#1}}

\usepackage{orcidlink}

\newcommand{\rr}[1]{{\textcolor{red}{#1}}}
\newcommand{\bb}[1]{{\textcolor{eccvblue}{#1}}}

\definecolor{sh_gray}{rgb}{0.84,0.84,0.84}
\definecolor{sh_gray2}{rgb}{1,0.89,0.75}
\definecolor{color3}{rgb}{0.95,0.95,0.95}
\definecolor{color4}{rgb}{0.96,0.96,0.86}
\definecolor{color5}{rgb}{0.90,0.90,0.90}

\begin{document}

\title{Spectral and Trajectory Regularization for Diffusion Transformer Super-Resolution} 

\titlerunning{Spectral and Trajectory Regularization for DiT SR}

\author{Jingkai Wang\inst{1}\thanks{Equal contribution.}\orcidlink{0009-0002-1278-4642} \and
Yixin Tang\inst{1}\protect\footnotemark[1]\orcidlink{0009-0005-4121-5782} \and
Jue Gong\inst{1}\orcidlink{0009-0008-4131-8951} \and
Jiatong Li\inst{1}\orcidlink{0009-0001-8822-0037} \and
Shu Li\inst{2} \and
Libo Liu\inst{2} \and
Jianliang Lan\inst{2} \and
Yutong Liu\inst{1}\thanks{Corresponding authors: Yutong Liu and Yulun Zhang.}\orcidlink{0000-0001-9386-3371} \and
Yulun Zhang\inst{1}\protect\footnotemark[2]\orcidlink{0000-0002-2288-5079}}

\authorrunning{J.~Wang et al.}

\institute{Shanghai Jiao Tong University, China \and Shenzhen Transsion Holdings Co., Ltd., China}

\maketitle

\vspace{-6mm}
\begin{abstract}
  Diffusion transformer (DiT) architectures show great potential for real-world image super-resolution (Real-ISR). However, their computationally expensive iterative sampling necessitates one-step distillation. Existing one-step distillation methods struggle with Real-ISR on DiT. They suffer from fundamental trajectory mismatch and generate severe grid-like periodic artifacts. To tackle these challenges, we propose StrSR, a novel one-step adversarial distillation framework featuring spectral and trajectory regularization. Specifically, we propose an asymmetric discriminative distillation architecture to bridge the trajectory gap. Additionally, we design a frequency distribution matching strategy to effectively suppress DiT-specific periodic artifacts caused by high-frequency spectral leakage. Extensive experiments demonstrate that StrSR achieves state-of-the-art performance in Real-ISR, across both quantitative metrics and visual perception. The code and models will be released at \url{https://github.com/jkwang28/StrSR}.
  \vspace{-4mm}
\end{abstract}

\setlength{\abovedisplayskip}{2pt}
\setlength{\belowdisplayskip}{2pt}

\vspace{-10mm}
\section{Introduction}
\vspace{-2mm}
\label{sec:intro}

Real-world image super-resolution (Real-ISR) is a typical and highly practical problem in low-level vision. It originates from the classic super-resolution (SR), \ie, the input low-resolution (LR) image is upscaled to a high-resolution (HR) image, with only Bicubic degradation~\cite{dong2014srcnn,zhang2018rcan,chen2023dat,chen2025hat}. However, in real-world scenarios, the degradation is much more complex and unknown, making Real-ISR a highly challenging and ill-posed problem~\cite{zhang2021bsrgan,wang2021realesrgan,liang2021swinir}.

Substantial efforts have been devoted to Real-ISR, with a particular focus on generative approaches like GAN-based methods~\cite{ledig2017photo,wang2018esrgan,zhang2021bsrgan,wang2021realesrgan}.
These methods continuously improve and deliver steadily better performance. 
With the emergence and rapid development of diffusion models~\cite{rombach2022ldm,podell2023sdxl,esser2024scaling}, Real-ISR reaches a new stage. Recent studies~\cite{xia2023diffir,lin2024diffbir,yu2024supir} increasingly adopt diffusion models for Real-ISR, leveraging their strong generative capability and scalability. Very recently, extensive research~\cite{henighan2020scaling} has highlighted the scaling laws of diffusion models. These studies demonstrate that diffusion transformer (DiT)~\cite{dit} architectures significantly outperform traditional UNets as model size and data volume grow. Given this superior capacity and predictable scaling behavior, adopting DiT for image SR is an essential and inevitable trend for the future of low-level vision. 

However, generating high-quality images with diffusion models usually requires dozens or hundreds of steps, which is computationally expensive. 
To tackle this issue, many works~\cite{you2025ctmsr,yue2025invsr,sun2024pisasr,dong2025tsdsr} distill pretrained text-to-image (T2I) models for image-to-image tasks. They use either 
score distillation~\cite{wu2024osediff, dong2025tsdsr}, or adversarial diffusion distillation~\cite{lin2025hypir, li2024dfosd}, and achieve significant performance.

\begin{figure}[t]
    \centering
    \begin{subfigure}{0.24\textwidth}
        \centering
        \includegraphics[width=\linewidth]{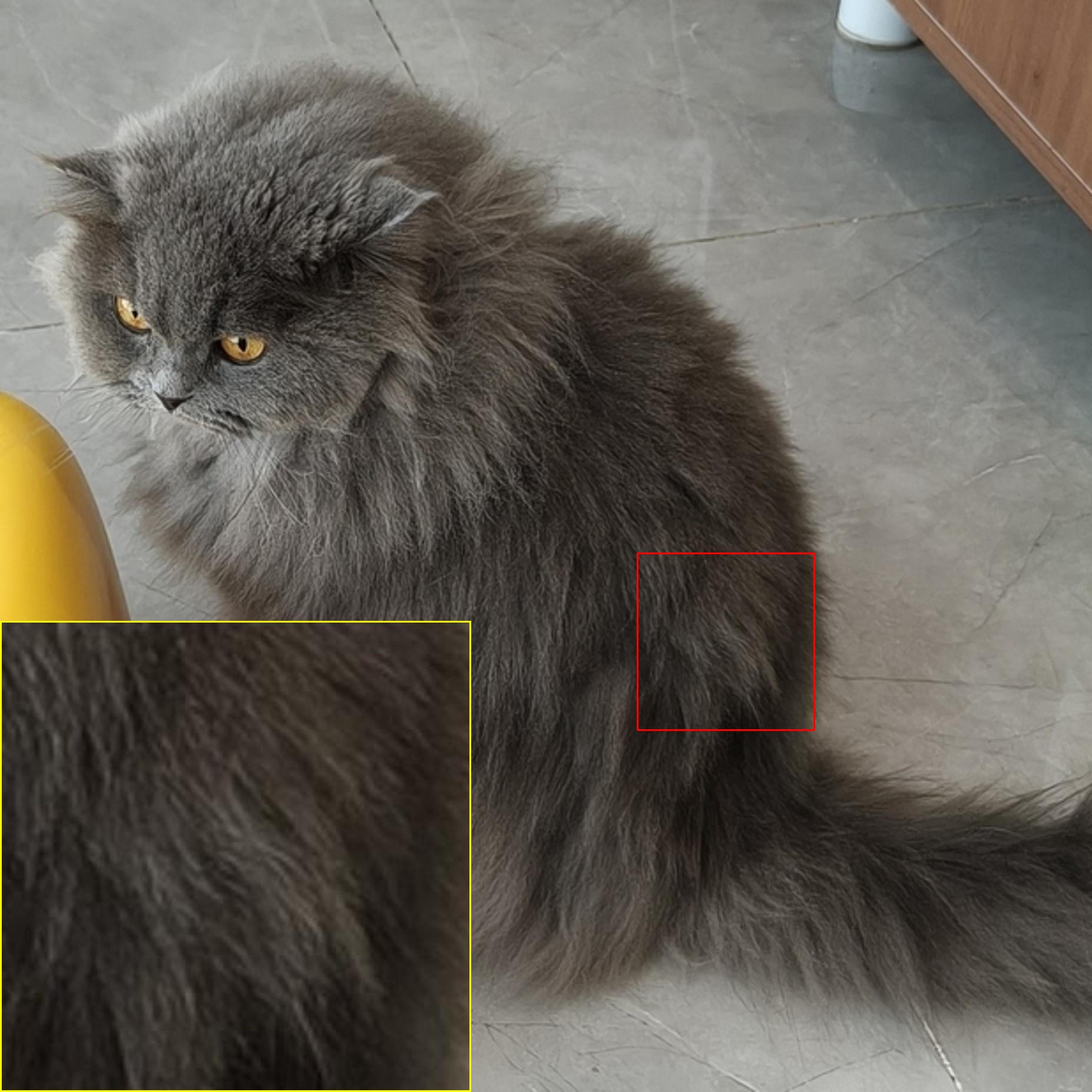}
        \caption{Input}
    \end{subfigure}
    \hfill 
    \begin{subfigure}{0.24\textwidth}
        \centering
        \includegraphics[width=\linewidth]{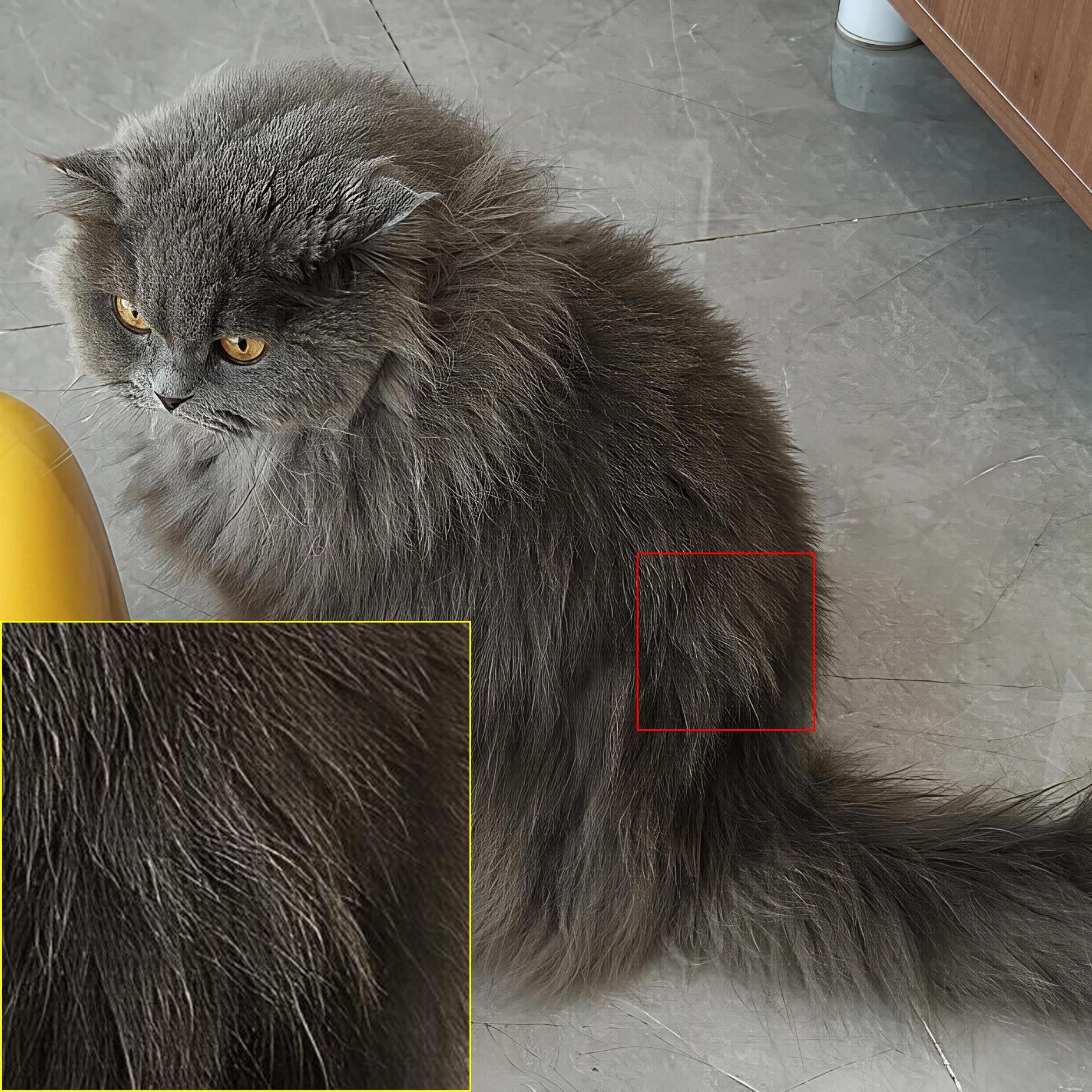}
        \caption{TSD-SR~\cite{dong2025tsdsr}}
    \end{subfigure}
    \hfill
    \begin{subfigure}{0.24\textwidth}
        \centering
        \includegraphics[width=\linewidth]{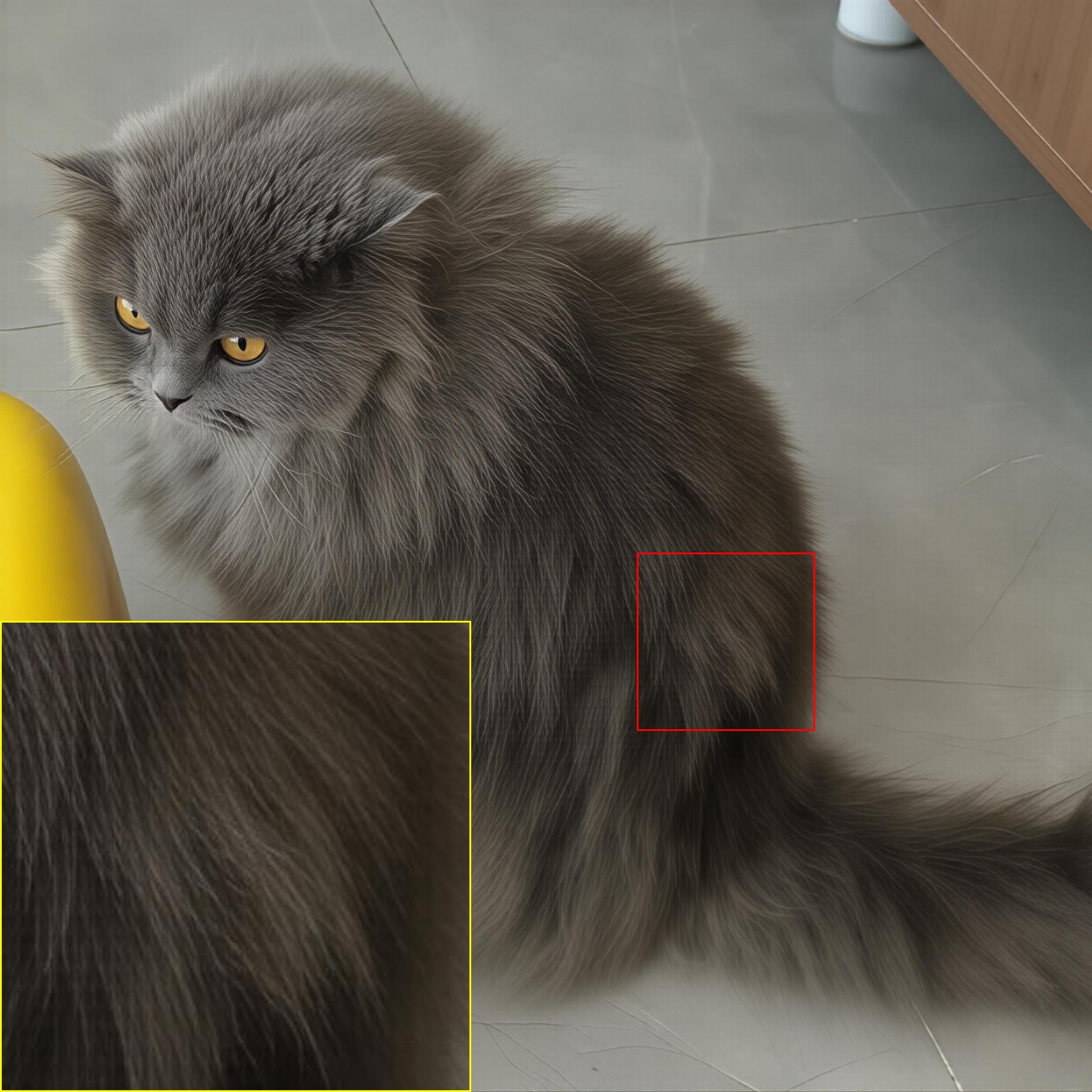}
        \caption{FluxSR~\cite{li2025fluxsr}}
    \end{subfigure}
    \hfill
    \begin{subfigure}{0.24\textwidth}
        \centering
        \includegraphics[width=\linewidth]{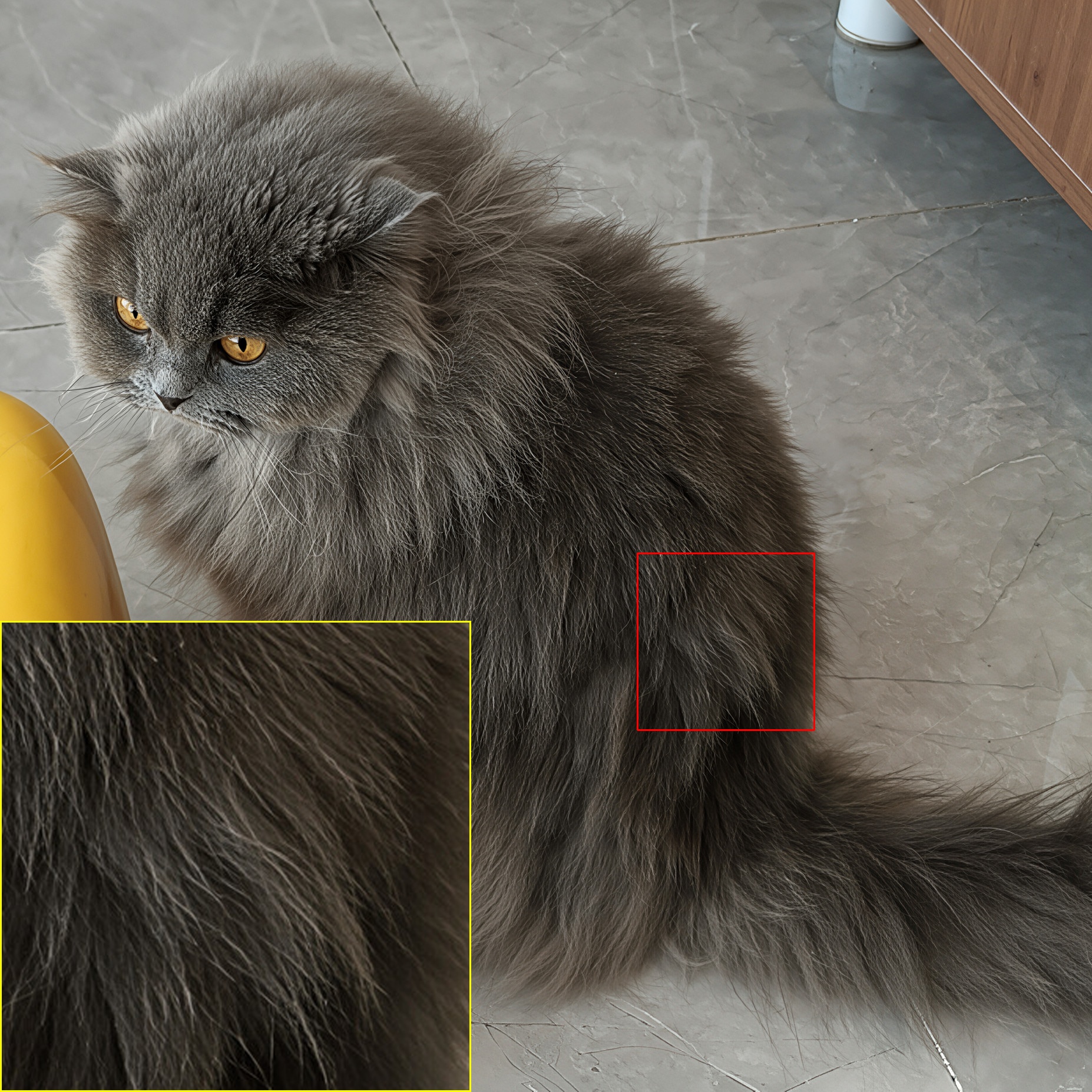}
        \caption{StrSR (ours)}
    \end{subfigure}
    \vspace{-3mm}
    \caption{For a self-captured cat image with dense fur, TSD-SR~\cite{dong2025tsdsr} and FluxSR~\cite{li2025fluxsr} suffer from heavy grid-like artifacts, whereas our StrSR restores realistic details.}
    \label{fig:artifact}
    \vspace{-6mm}
\end{figure}

Despite achieving promising results, we observe that current methods still suffer when combining DiT and one-step distillation. As shown in Fig.~\ref{fig:artifact}, those methods~\cite{dong2025tsdsr, li2025fluxsr} suffer from severe artifacts and inferior performance. 
To analyze the reason, there is a fundamental trajectory mismatch. It is difficult to quickly and stably align the new generation trajectory (from LR to HR images) with the pretrained trajectory (from pure noise to HR images). Furthermore, during this forced trajectory alignment, models struggle to maintain photo-realistic outputs. Forcing a single large evaluation step heavily exacerbates grid-like artifacts, a problem that is uniquely severe in DiT architectures.

To tackle these challenges, we propose spectral and trajectory regularization for diffusion transformer super-resolution (StrSR). This one-step adversarial distillation framework unlocks photo-realistic generation while alleviating the severe periodic artifacts inherent to DiT models. Our contributions are as follows:
\vspace{-2mm}\begin{enumerate}
    \item We propose \textbf{asymmetric discriminative distillation} to bridge the gap between multi-step and one-step generation. To bypass heavy progressive distillation and prevent model collapse, we employ a pretrained CLIP-ConvNeXt as a lightweight, texture-sensitive discriminator. Combined with a relativistic average GAN loss and approximated R1 regularization, this asymmetric design ensures stable training and superior semantic-aligned texture recovery. 
    \item We design \textbf{frequency distribution matching} for artifact suppression. To address the grid-like periodic artifacts arising from high-frequency spectral leakage in DiT patches, we introduce the frequency distribution loss (FDL). By minimizing the sliced Wasserstein distance between the predicted and target features across both amplitude and phase components, FDL provides a spectral constraint that effectively suppresses these artifacts. 
    \item We develop a \textbf{comprehensive dual-encoder architecture} that seamlessly integrates spatial adversarial distillation and spectral domain optimization. Empowered by this unified framework, StrSR achieves state-of-the-art (SOTA) performance in one-step Real-ISR.  Extensive experiments confirm its superiority across both quantitative metrics and visual quality.
\end{enumerate}

\section{Related Works}
\vspace{-2mm}
\subsection{Diffusion-based Real-ISR}
\vspace{-2mm}
Recently, the success in pretrained diffusion models~\citep{rombach2022ldm, podell2023sdxl, flux2024,wu2025qwenimage} enables researchers to leverage pretrained weights for downstream tasks like real-world image super-resolution. Built upon stable diffusion models~\citep{rombach2022ldm, podell2023sdxl} based on UNet architecture, StableSR~\cite{wang2024stablesr} and DiffBIR~\cite{lin2024diffbir} inject low-resolution information into the model through controlnet-like methods. OSEDiff~\citep{wu2024osediff} directly uses LR image as input and adopts variational score distillation (VSD)~\citep{wang2023prolificdreamer, swiftbrush} to distill a pretrained diffusion model for one-step super-resolution. InvSR~\citep{yue2025invsr} treats SR as a diffusion inversion problem to better guide image priors in large pretrained diffusion models. HYPIR~\citep{lin2025hypir} harnesses adversarial diffusion distillation (ADD) to distill SD-XL \cite{podell2023sdxl} into one-step image SR model.\par
With the prevalence of diffusion transformer (DiT)~\citep{dit} in the text-to-image (T2I) task, some works also explored the possibility to utilize DiT architecture in image SR. TSD-SR~\citep{dong2025tsdsr} proposes target score distillation (TSD) and a distribution-aware sampling module to handle the artifacts caused by VSD in training. DiT4SR~\citep{duan2025dit4sr} trains a DiT model from scratch based on SD3~\citep{esser2024scaling} MM-DiT architecture with additional LR-residual. FluxSR~\citep{li2025fluxsr} explores one-step DiT-based SR models in terms of flow trajectory~\citep{liu2022flow, lipman2023flow}.

\vspace{-2mm}
\subsection{Diffusion Model Acceleration}
\vspace{-2mm}
Diffusion models have strong generative abilities. However, they suffer from high computational costs during inference. This issue limits their practical deployment. 

Many methods have been proposed to speed up this process. 
One major category is deterministic methods using a deterministic probability flow. They aim to predict the exact output of a teacher model using fewer steps. Examples include progressive distillation~\citep{progressivedistillation}, consistency distillation~\citep{song2023consistency, song2023improved, lu2024simplifying, luo2023latent}, and rectified flow~\citep{liu2022flow}, and they are easy to train. However, they usually need more steps (\textit{e.g.}, 8 steps) to generate high-quality results. Another category is distributional methods. Instead of predicting exact outputs, they approximate the data distribution of the teacher model. This group includes adversarial training~\citep{chen2025nitrofusion, kang2024distilling, yoso, sauer2024fast, ufogen} and score distillation~\citep{luo2023diff, yin2024dmd, wang2023prolificdreamer}. Some approaches even combine both techniques~\citep{yin2024dmd2, sauer2024adversarial, chadebec2025flash}. 

Recently, these acceleration techniques have been applied to diffusion-based low-level tasks. These works~\cite{wang2025osdface,gong2025haodiff} directly fine-tune pretrained diffusion models using score or adversarial distillation. They achieve superior performance. Notable examples in image SR include OSEDiff~\citep{wu2024osediff}, TSD-SR~\citep{dong2025tsdsr}, and HYPIR~\citep{lin2025hypir}.

\begin{figure}[t]
    \centering
    \includegraphics[width=0.8\textwidth]{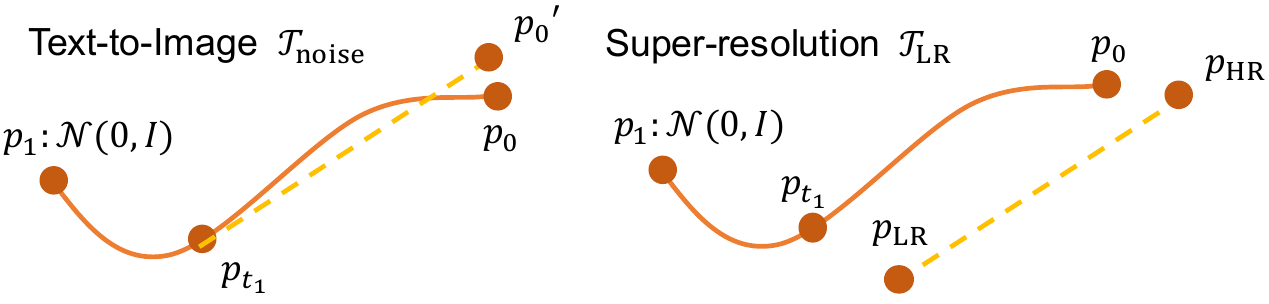}
    \caption{Two gaps exist between the multi-step T2I model and the one-step SR model. First, a direct linear trajectory from the intermediate step $t=t_1$ to $t=0$ cannot reach the real HR data distribution. Second, the LR data distribution mismatches the pretrained T2I model's distribution at $t_1$. Points in the figure represent distributions.}
    \label{fig:distribution}
\end{figure}

\vspace{-2mm}
\section{Methods}
\vspace{-2mm}
\subsection{Preliminaries and Motivation}
\vspace{-2mm}
\noindent\textit{Diffusion Models and Rectified Flow} Continuous-time generative models, such as diffusion models~\cite{ho2020denoising,song2021denoising,rombach2022ldm}, build a mapping between a target data distribution $p_0(x_0)$ and a prior noise distribution $p_1(x_1)$. This transformation is driven by an ordinary differential equation over a continuous time variable $t \in [0,1]$:
\begin{equation}
    \frac{\mathrm{d}x_t}{\mathrm{d}t} = v(x_t, t),
\end{equation}
where $v(x_t, t)$ is the time-dependent vector field. By solving this ODE from $t=1$ to $t=0$, we can transport samples from noise to the data distribution.

In practice, operating directly in the high-dimensional pixel space is computationally expensive. Therefore, modern generative models project images into a continuous latent space using a pretrained variational autoencoder (VAE)~\cite{kingma2013auto,rombach2022ldm} $\mathcal{E}_\text{VAE}, \mathcal{D}_\text{VAE}$. Let $z_0 = \mathcal{E}_\text{VAE}(x_0)$ be the latent representation of the high-quality data, and $z_1 \sim \mathcal{N}(0, \mathbf{I})$ be the standard Gaussian noise.

Rectified flow~\cite{liu2022flow} connects $z_0$ and $z_1$ using a straightforward linear trajectory. The intermediate latent state $z_t$ is defined as:
\begin{equation}
    z_t = t z_1 + (1 - t) z_0.
\end{equation}
By taking the derivative of $z_t$ with respect to time $t$, we obtain the ground-truth vector field for this linear path:
\begin{equation}
    \frac{\mathrm{d}z_t}{\mathrm{d}t} = z_1 - z_0.
\end{equation}

To enable controlled generation, we introduce a condition $c$ extracted from the input. A neural network generator $G_\theta(z_t, t, c)$ is used to predict the vector field. The network is optimized by minimizing a mean-squared error objective:
\begin{equation}
    \mathcal{L} = \mathbb{E}_{z_0, z_1, t, c} \left[ \| G_\theta(z_t, t, c) - (z_1 - z_0) \|^2 \right].
\end{equation}

By minimizing this loss, $G_\theta$ learns to simulate the linear mapping conditioned on $c$. The linear nature of rectified flow ensures a straight generation trajectory, reducing truncation errors and enabling fast inference. 

\noindent\textit{Flow Matching for Real-ISR} Building upon this theoretical foundation, we adapt rectified flow for image SR. Instead of synthesizing images from pure noise, we formulate the generative process as a translation from LR inputs $x_\text{LR}$ (assuming $t=t_1$) to HR targets $x_\text{HR}$ ($t=0$). Operating within the compressed latent space, we redefine the initial state as the degraded latent $z_{t_1} = \mathcal{E}_\text{VAE}(x_\text{LR})$ and the target state as the clean latent $z_0 = \mathcal{E}_\text{VAE}(x_\text{HR})$. To accelerate convergence and leverage rich natural image priors, it is common practice to initialize the generator $G_\theta$ with weights from pretrained text-to-image models. Several recent methods~\cite{wu2024osediff,li2024dfosd,lin2025hypir} have successfully adapted these noise-pretrained weights to this deterministic $z_{t_1} \rightarrow z_0$ setting. However, as shown in Fig.~\ref{fig:distribution}, the distribution of degraded latents differs from intermediate state along the standard noise-to-data trajectory $\mathcal{T}_\text{noise}$.

This distribution shift poses a significant challenge for efficient inference. For one-step generation, where the number of function evaluations (NFE) is exactly 1, the model assumes a perfectly linear trajectory and attempts to predict the target in a single Euler step: $\hat{z}_0 = z_{t_1} - G_\theta(z_{t_1},{t_1}, c)$. When image degradation is mild, the mapping distance between the pretrained noise trajectory $\mathcal{T}_\text{noise}$ and the new restoration trajectory $\mathcal{T}_\text{LR}$ is relatively small. In such cases, fine-tuning a multi-step model into a one-step model is straightforward. However, diffusion models are deployed in low-level vision primarily for their ability to restore severely degraded real-world images. Under severe degradation, the domain gap between $\mathcal{T}_\text{noise}$ and $\mathcal{T}_\text{LR}$ widens considerably. Consequently, forcing a single large evaluation step from a distant, out-of-distribution $z_{t_1}$ leads to substantial errors, creating a distinct performance gap between multi-step and one-step generation. 

\begin{figure}[t]
    \centering
    \begin{subfigure}{0.19\textwidth}
        \centering
        \includegraphics[width=\linewidth]{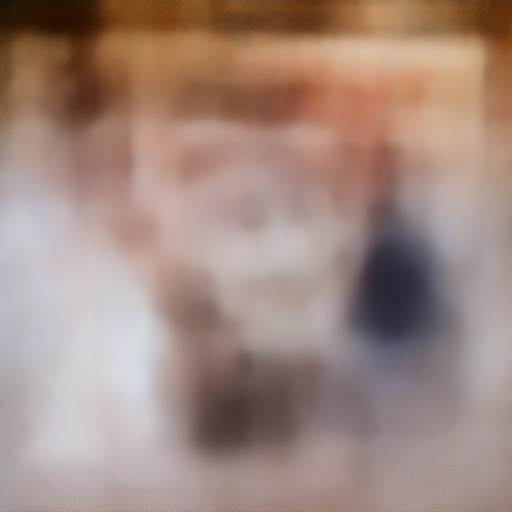}
        \caption*{SD 2.1~\cite{rombach2022ldm}}
    \end{subfigure}
    \hfill 
    \begin{subfigure}{0.19\textwidth}
        \centering
        \includegraphics[width=\linewidth]{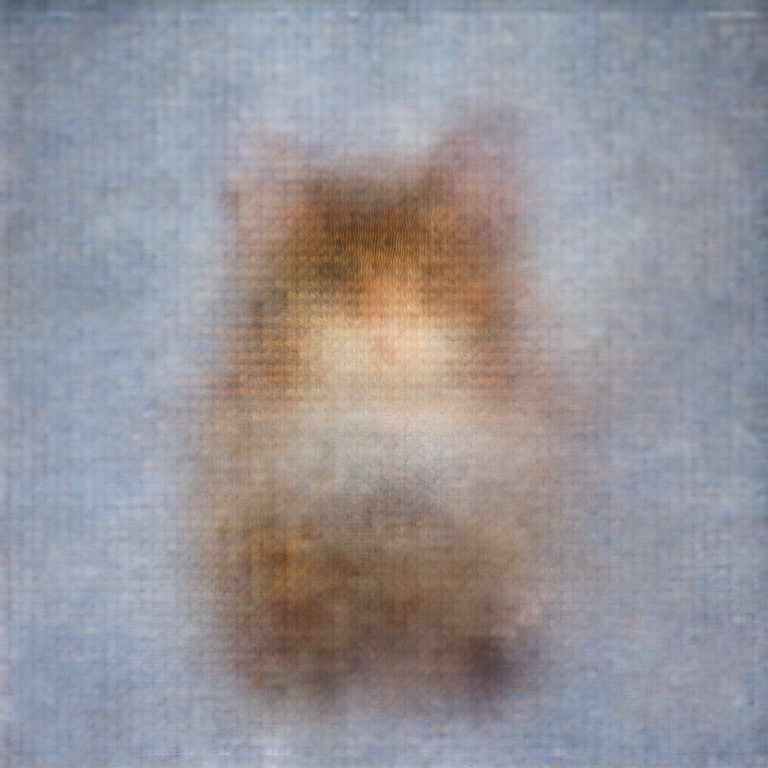}
        \caption*{SD 3~\cite{esser2024scaling}}
    \end{subfigure}
    \hfill
    \begin{subfigure}{0.19\textwidth}
        \centering
        \includegraphics[width=\linewidth]{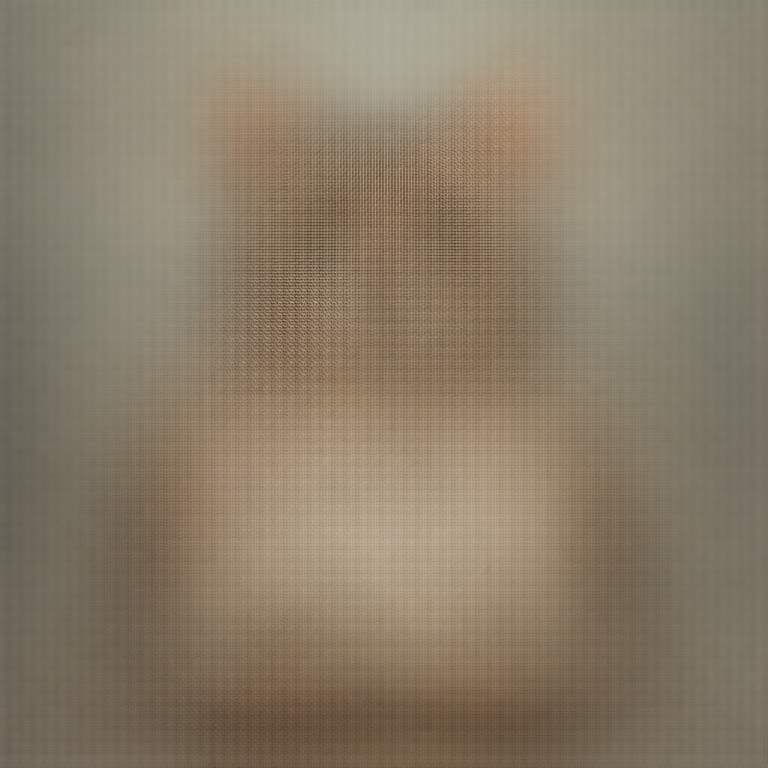}
        \caption*{FLUX.1 [dev]~\cite{flux2024}}
    \end{subfigure}
    \hfill
    \hfill 
    \begin{subfigure}{0.19\textwidth}
        \centering
        \includegraphics[width=\linewidth]{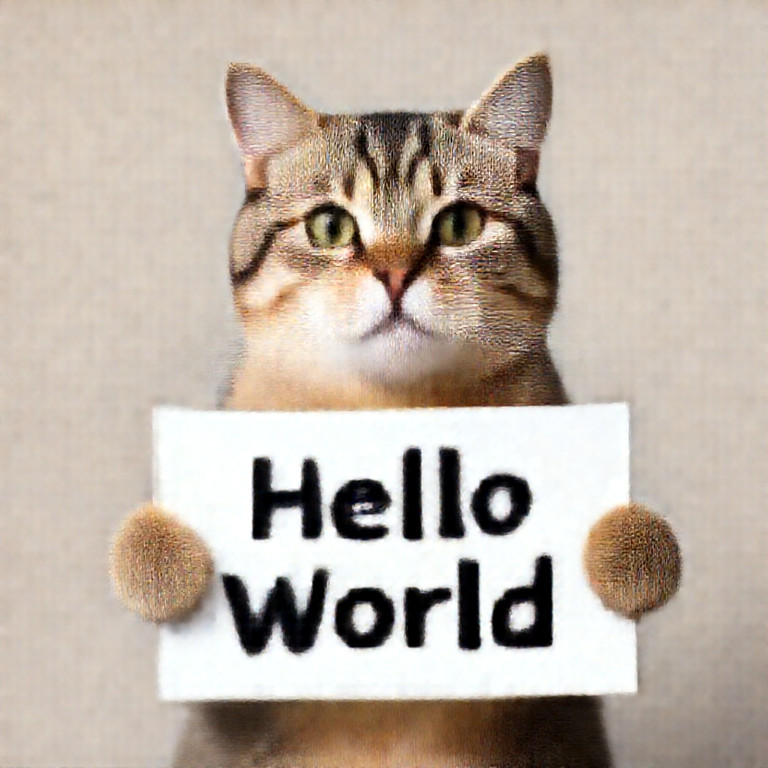}
        \caption*{Z-Image-Turbo~\cite{team2025zimage}}
    \end{subfigure}
    \hfill
    \begin{subfigure}{0.19\textwidth}
        \centering
        \includegraphics[width=\linewidth]{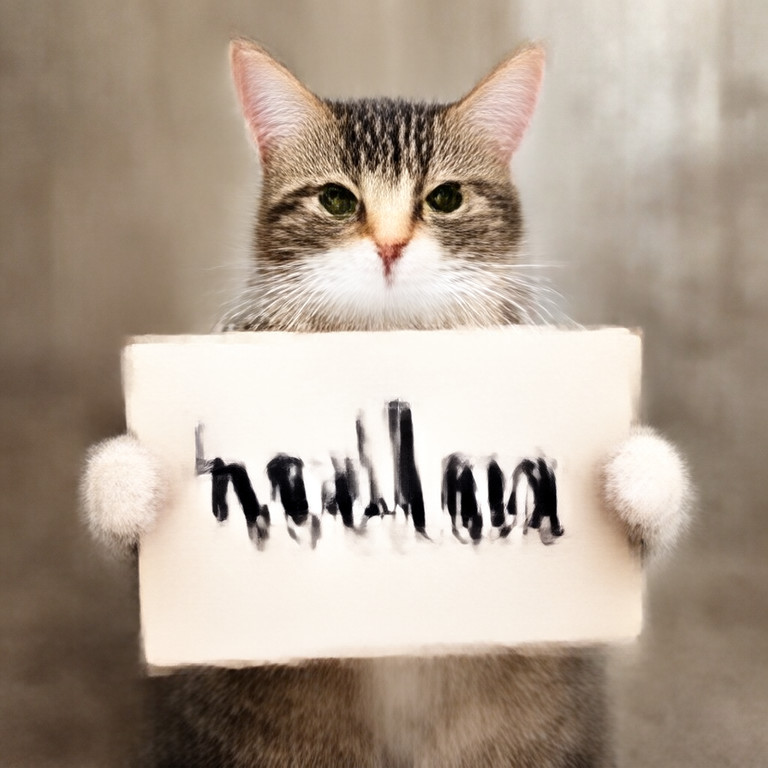}
        \caption*{FLUX.2 [klein]~\cite{FLUX.2-klein-4B}}
    \end{subfigure}
    \caption{One-step T2I inference along a linear trajectory: UNet vs. DiT. While both models produce artifacts, the DiT-based model exhibits more severe grid-like artifacts.}
    \label{fig:distribution_comparison}
\end{figure}

Furthermore, achieving one-step generation is even more challenging with diffusion transformers (DiT). Since we fine-tune from pretrained DiT weights, our method is heavily constrained by base model. The base model is trained to generate images from pure noise. At large time steps $t$ (near pure noise), the model primarily focuses on low-frequency global structures and ignores high-frequency details. In our one-step setting, we force a single large step directly to the target $z_\text{HR}$. As shown in Fig.~\ref{fig:distribution_comparison}, while taking such a large step does not cause obvious high-frequency artifacts in UNet architectures, it produces severe ones in DiT.

Overall, attempting to fine-tune a multi-step DiT model into a one-step model presents two main challenges: 1) the large mapping distance between $\mathcal{T}_\text{noise}$ and $\mathcal{T}_\text{LR}$ leads to poor results; and 2) the severe discretization error inherent in a single evaluation step worsens high-frequency artifacts in DiT architectures. To address these challenges, we need to design a novel distillation method that can effectively bridge the domain gap between multi-step and one-step trajectories while specifically mitigating DiT-related artifacts.

\begin{figure*}[t]
    \centering
    \includegraphics[width=\textwidth]{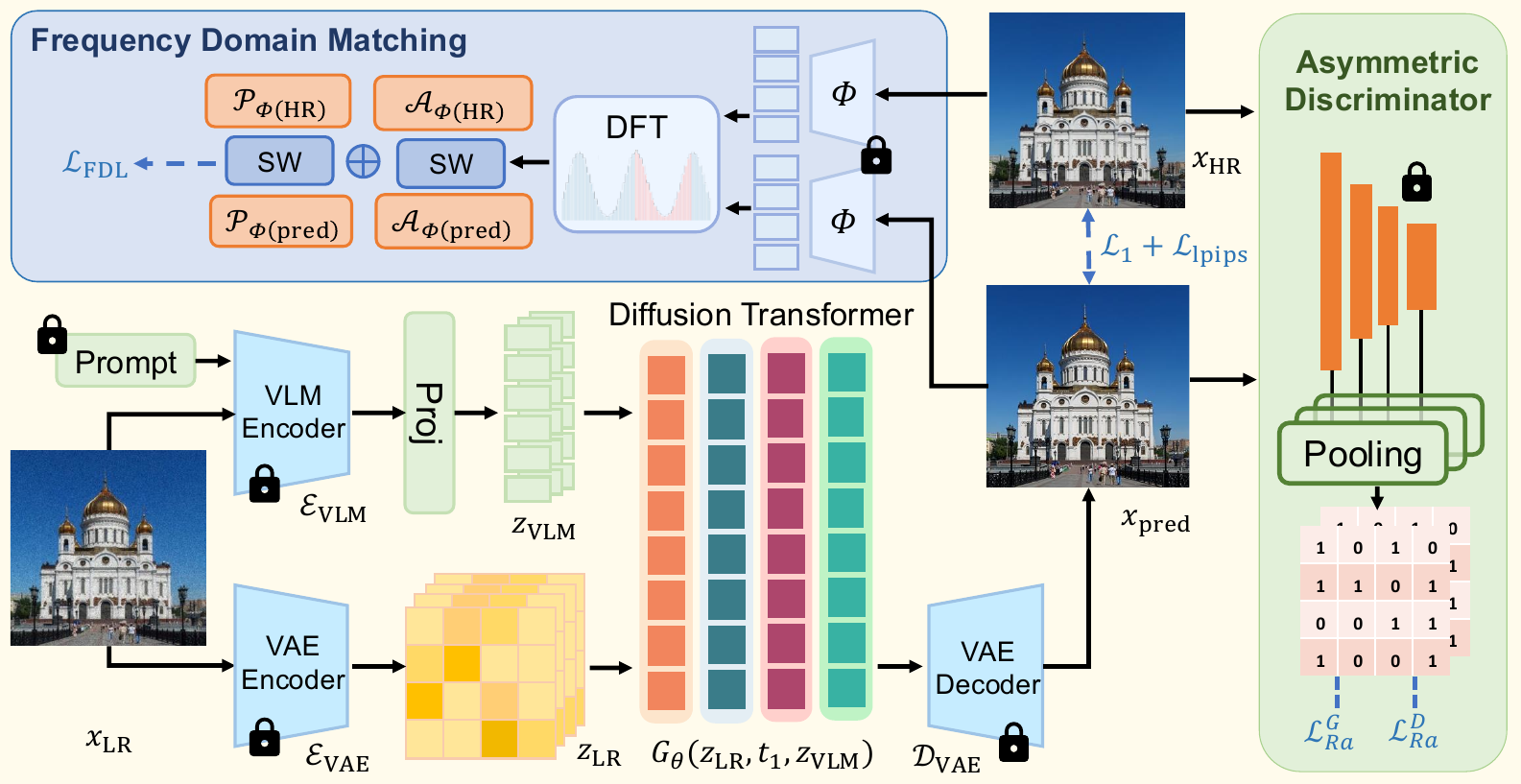}
    \vspace{-3mm}
    \caption{Training pipeline of StrSR. The LR image $x_\text{LR}$ is input into the dual-encoder $\mathcal{E}_\text{VAE}$ and $\mathcal{E}_\text{VLM}$ and following DiT $G_\theta$ to gain the predicted HR image $x_\text{pred}$. DiT is fine-tuned by LoRA. The generator and discriminator are trained alternately. }
    \label{fig:overall}
    \vspace{-6mm}
\end{figure*}

\subsection{Model Formulation}
To effectively tackle the ill-posed nature of image super-resolution, our overall architecture employs a dual-encoder design to decouple feature extraction: a vision-language model (VLM) encoder captures high-level semantic information, while a variational autoencoder extracts low-level spatial and structural representations.

As illustrated in Fig.~\ref{fig:overall}, let $x_\text{LR}$ and $x_\text{HR}$ denote the LR and corresponding HR images. Specifically, $x_\text{LR}$ is fed into a pretrained VLM encoder, $\mathcal{E}_\text{VLM}$, to extract rich semantic features. To align these features with the input space of the diffusion backbone, they are processed through an adaptive  MLP $f_\text{adapt}$:
\begin{equation}
    z_\text{VLM} = f_\text{adapt}(\mathcal{E}_\text{VLM}(x_\text{LR})).
\end{equation}
In our design, $z_\text{VLM}$ serves as a direct substitute for the conventional text condition embeddings typically used in the base foundational models.

In parallel, $x_\text{LR}$ is processed by a pretrained VAE encoder, $\mathcal{E}_\text{VAE}$, to obtain the continuous latent space representation $z_\text{VAE}$, \ie, $z_\text{LR}$:
\begin{equation}
    z_\text{LR} = \mathcal{E}_\text{VAE}(x_\text{LR}).
\end{equation}
Operating within a rectified flow framework, this deterministic latent representation $z_\text{LR}$ serves directly as the initial state of the generative process.

The core generation is driven by a DiT generator, $G_\theta$. The generator takes both the spatial latent $z_\text{VAE}$ and the semantic tokens $z_\text{VLM}$ as inputs. These representations are formulated as token sequences and processed through the transformer blocks to predict the vector field governing the generation trajectory. The final restored high-quality image $x_\text{pred}$ is obtained by decoding the integrated latent state via the VAE decoder $\mathcal{D}_\text{VAE}$.

\vspace{-3mm}
\subsection{Towards Photo-Realistic Restoration}
\vspace{-2mm}
Discriminative distillation is a widely-used, effective, and standard approach to align generation trajectories~\cite{seedvr2,li2024dfosd,lin2025hypir,APT}. Our goal is to design an efficient and stable strategy that fully exploits pretrained priors, tailored for Real-ISR.

\begin{figure*}[t]
    \centering
    \includegraphics[width=\textwidth]{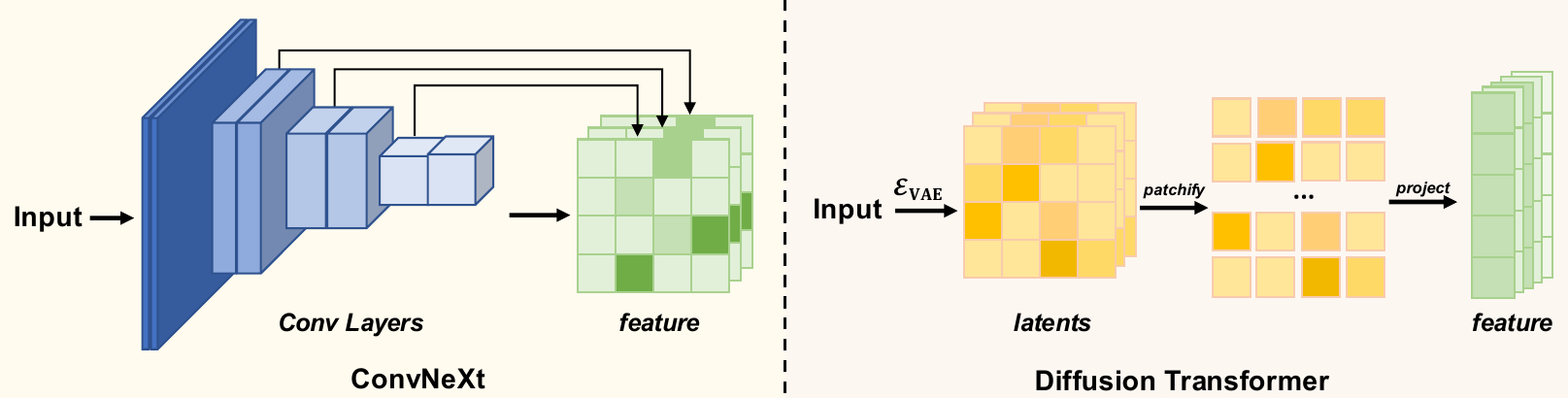}
    \caption{Discriminator comparison. ConvNeXt's large-kernel convolutions preserve multi-scale spatial information. Conversely, DiT’s patchification heavily compresses intra-patch details (edges and textures), reducing its ability to detect patch-level periodic artifacts.}
    \label{fig:discriminator}
    \vspace{-6mm}
\end{figure*}

Distribution matching-based distillation~\cite{yin2024dmd,yin2024dmd2} is proven effective in generative tasks. However, when using pretrained DiTs~\citep{dit} for both the generator and discriminator, we observe rapid model collapse during training. We attribute this collapse to the absence of a progressive distillation~\citep{progressivedistillation}, as highlighted in SeedVR2~\citep{seedvr2}. To avoid complex and lengthy training pipelines while maintaining high visual quality, we discard the DiT-based discriminator. As shown in Fig.~\ref{fig:discriminator}, we introduce a discriminator based on a pretrained CLIP-ConvNeXt~\citep{convnext, laion, lin2025hypir}, augmented with multiple trainable convolutional layers for calculating logits. As a lightweight yet powerful feature extractor, this convolutional CLIP~\citep{radford2021learning} model offers three key advantages: a smaller model size, comparable semantic understanding, and superior texture perception. Most importantly, the strong local inductive bias of ConvNeXt makes it highly sensitive to high-frequency textures and grid artifacts. This asymmetric architecture perfectly mitigates the one-step DiT generator's flaw of producing local periodic artifacts.

Specifically, our discriminator adopts a PatchGAN architecture. We extract features from different stages and the final pooling layer from the CLIP-ConvNeXt backbone. These extracted features are then processed by trainable convolutional heads to produce multi-level, patch-wise logit maps for computing the GAN loss.

We adopt the relativistic average GAN (RaGAN)~\citep{relativistic} rather than naive non-saturating GAN~\citep{goodfellow2014generative} for better visual performance and stable training convergence. The generator loss and discriminator loss are set as follows, respectively. 
\begin{equation}
\begin{aligned}
   \mathcal{L}_\mathit{Ra}^{G} &= -\mathbb{E}_{z_\text{LR}}\left[\log g\left(x_\text{pred}, x_\text{HR}\right)\right] - \mathbb{E}_{x_\text{HR}}\left[\log\left(1 - g\left(x_\text{HR}, x_\text{pred}\right)\right)\right]; \\
   \mathcal{L}_\mathit{Ra}^{D} &= -\mathbb{E}_{x_\text{HR}}\left[\log g\left(x_\text{HR}, x_\text{pred}\right)\right] - \mathbb{E}_{z_\text{LR}}\left[\log\left(1 - g\left(x_\text{pred}, x_\text{HR}\right)\right)\right].
\end{aligned}
\end{equation}
Here, $x_\text{pred}=\mathcal{D}_\text{VAE}(z_{\mathrm{LR}} - tG_\theta(z_\text{LR}, t, z_\text{VLM}))$, $g(x, y)=\varsigma(C(x) - \mathbb{E}_{y}[C(y)])$ with $\varsigma$ denoting the sigmoid function. $C(\cdot)$ is the output logits for a real or fake sample.

To better regularize the discriminator and facilitate training convergence, we also introduce approximated R1 loss~\citep{roth2017stabilizing} as:
\begin{equation}
    L_\mathit{R1}=||D(x_\text{real})-D(\mathcal{N}(x_\text{real}, \sigma \mathbf{I}))||^2.
\label{eq:r1}\end{equation}
We apply Gaussian noise with small variance $\sigma$ to real data when calculating discriminator loss, and adopt additional approximated R1 loss as Eq.~\eqref{eq:r1}. 
This regularization forces the discriminator to produce similar outputs for both the clean images and their noisy counterparts. 
Consequently, it achieves a penalizing effect similar to the standard R1 loss without the computational overhead.

In summary, our asymmetric discriminative training objectives can be: 
\begin{equation}
    \mathcal{L}_D=\mathcal{L}_\mathit{Ra}^D + \lambda_\mathit{R1}\cdot \mathcal{L}_\mathit{R1}
\end{equation}

\subsection{Frequency Distribution Matching} 
\begin{figure*}[t]
\scriptsize
\centering
\scalebox{0.99}{
    \begin{adjustbox}{valign=t}
    \begin{tabular}{cccccc}
    \includegraphics[width=0.156\textwidth]{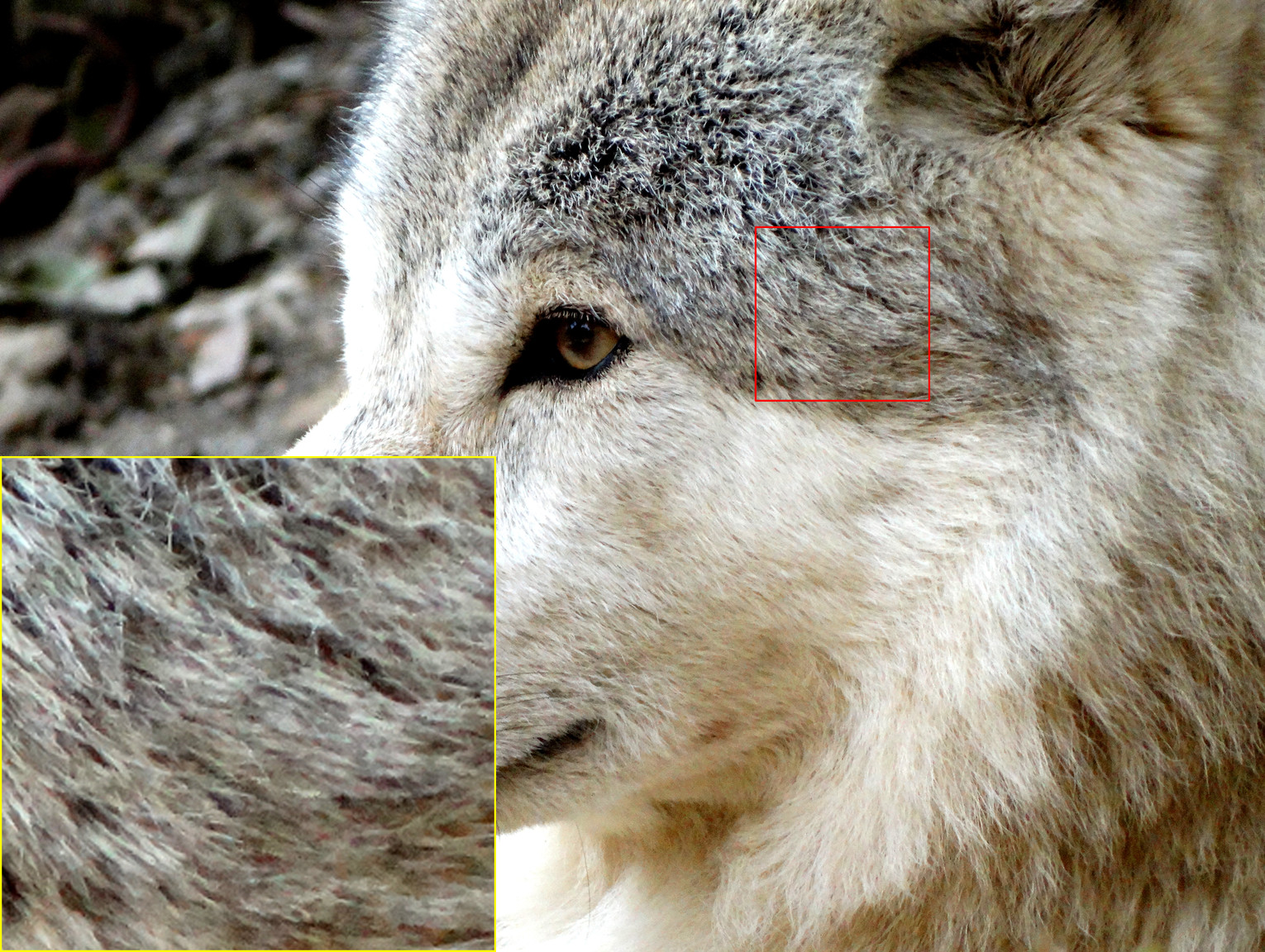} \hspace{-1.2mm} &
    \includegraphics[width=0.156\textwidth]{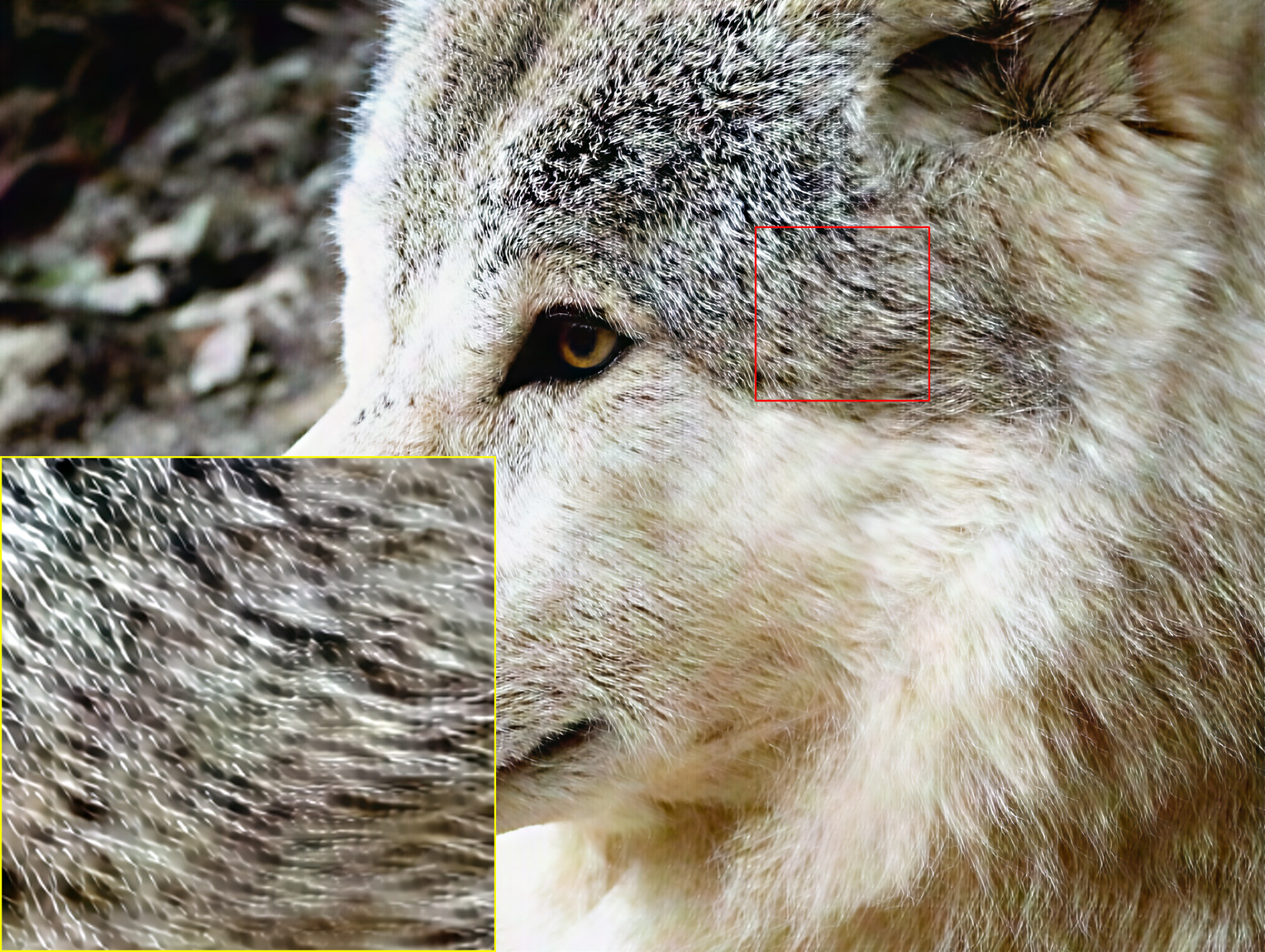} \hspace{-1.2mm} &
    \includegraphics[width=0.156\textwidth]{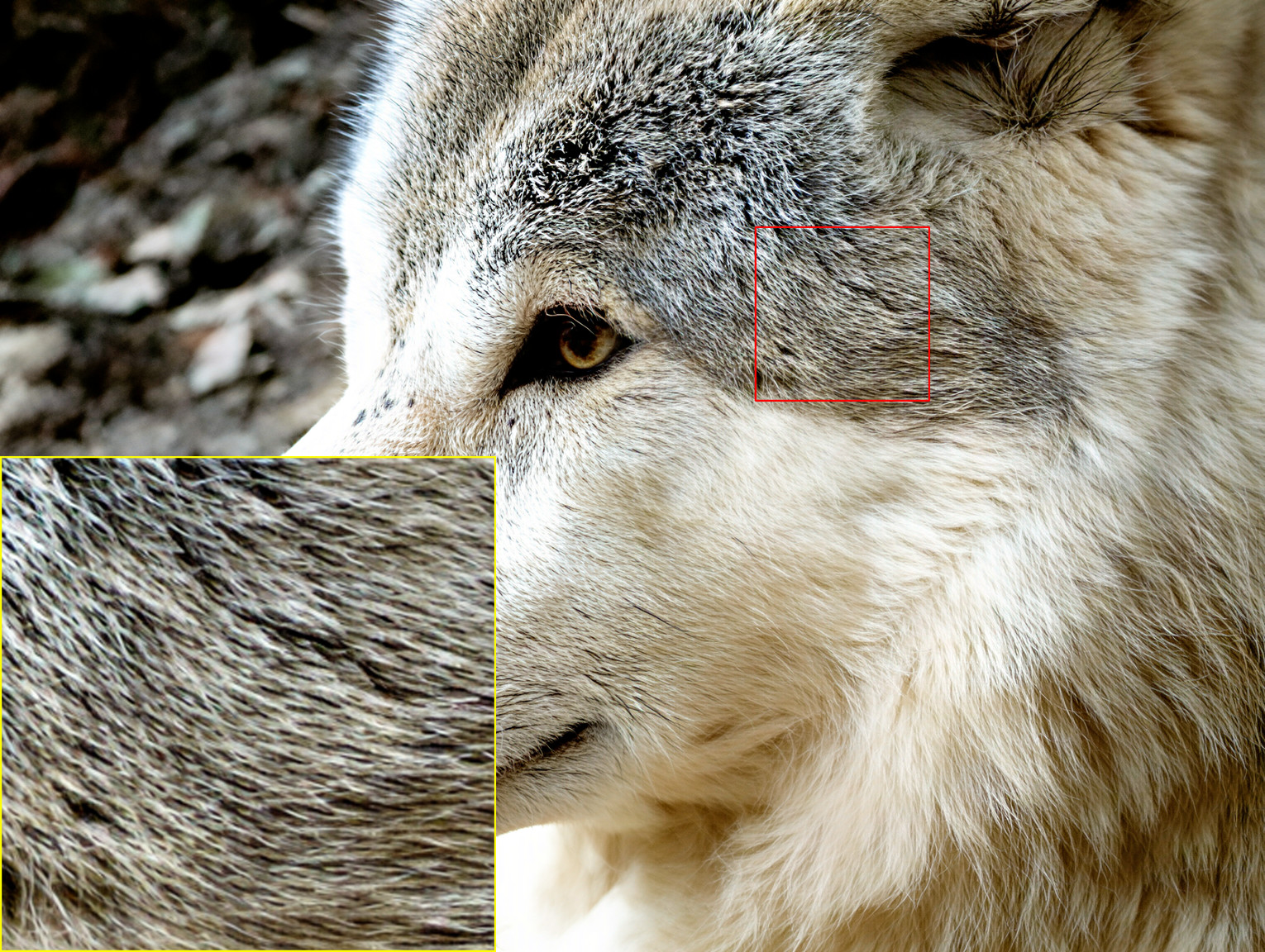} \hspace{1.0mm} &
    \includegraphics[width=0.156\textwidth]{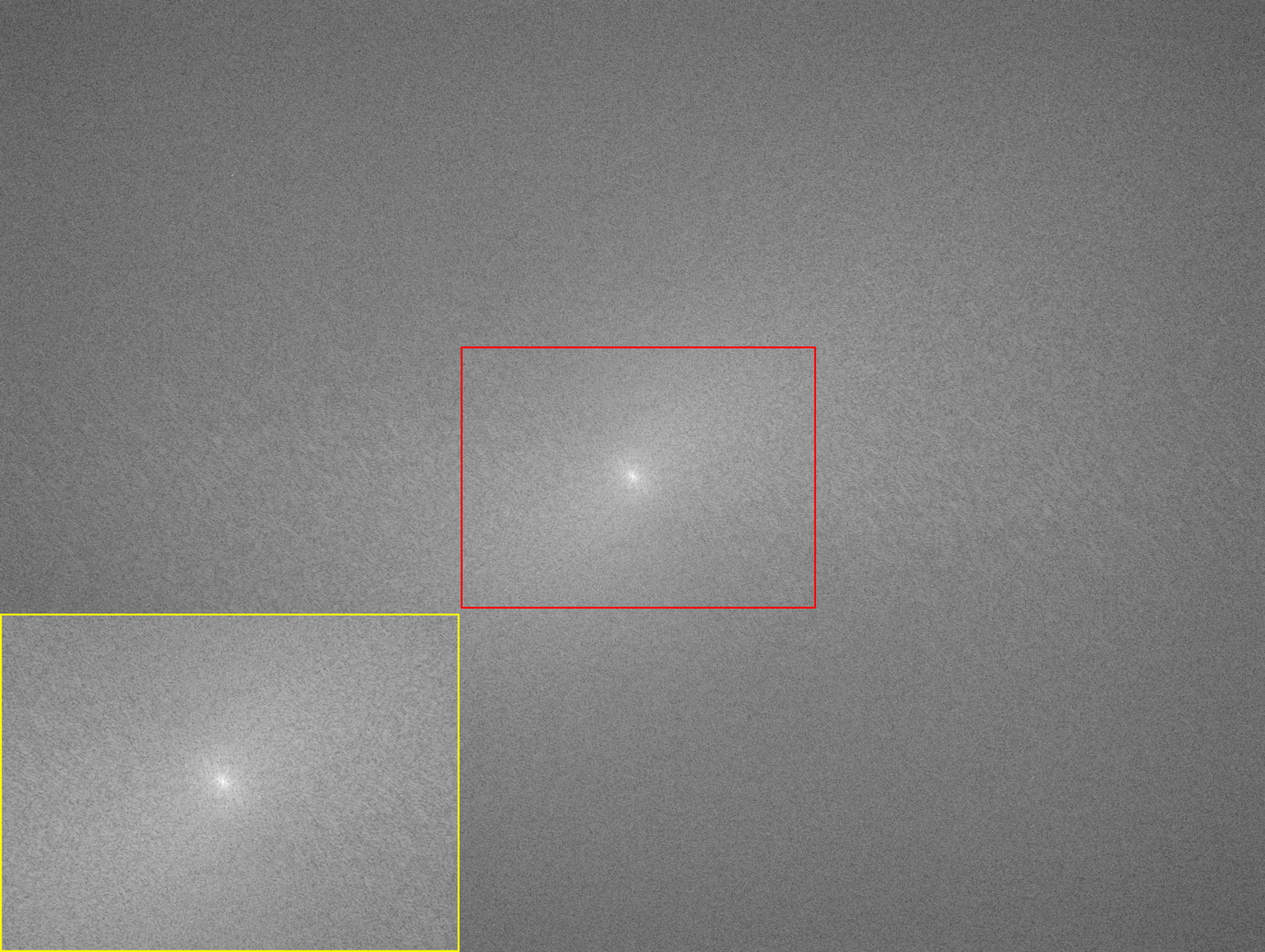}   \hspace{-1.2mm} &
    \includegraphics[width=0.156\textwidth]{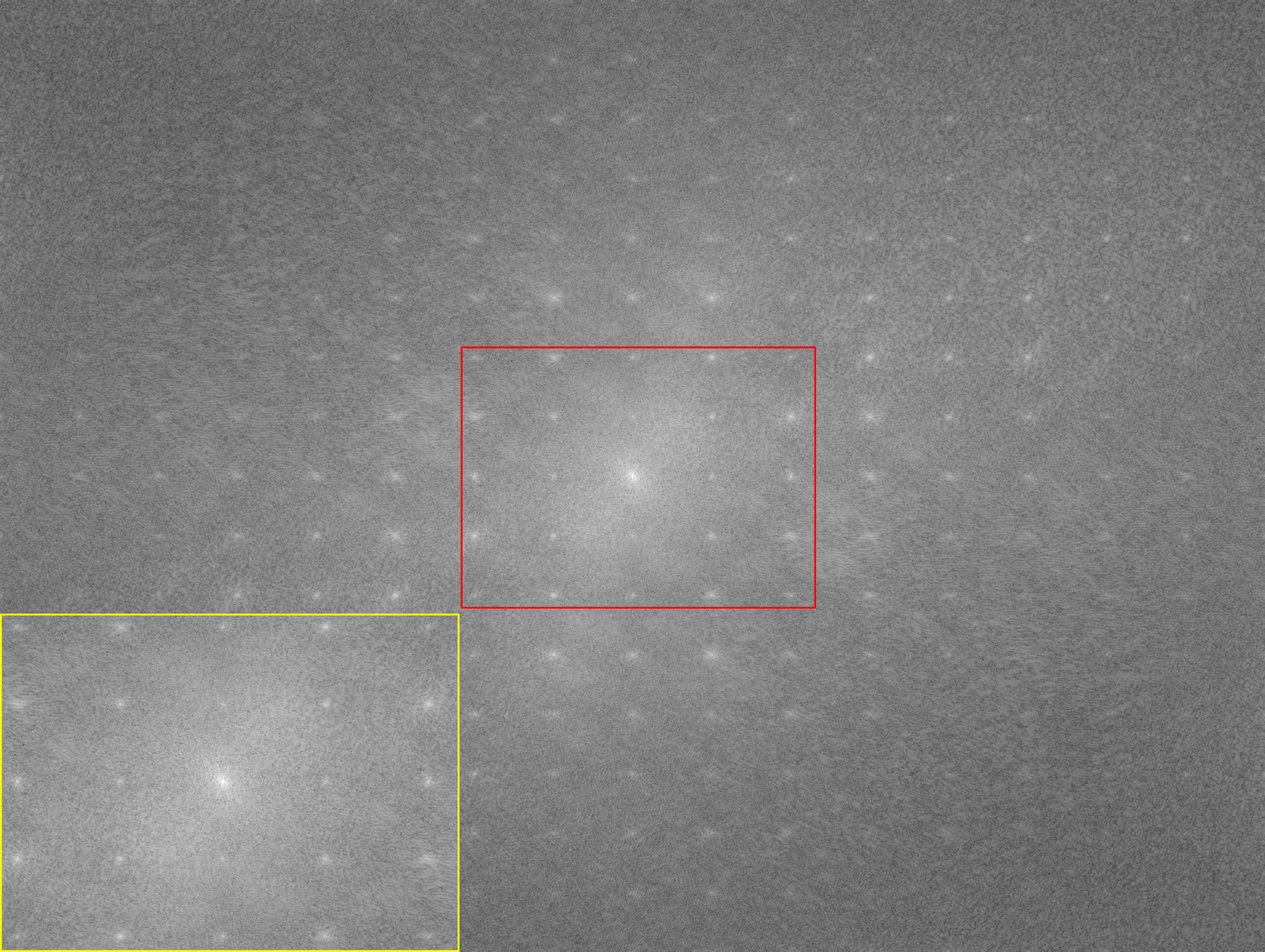}  \hspace{-1.2mm} &
    \includegraphics[width=0.156\textwidth]{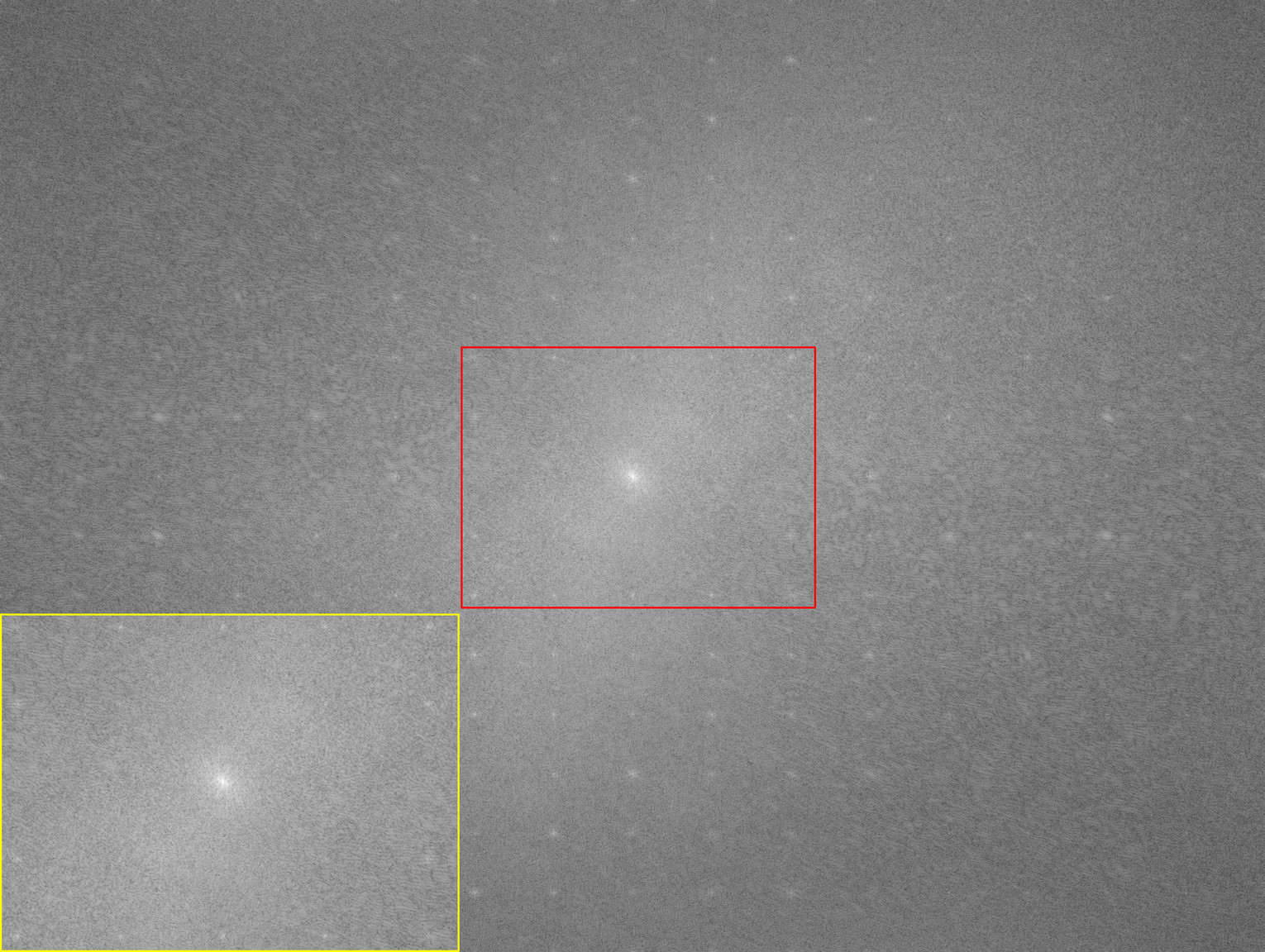}  \hspace{-1.2mm}
    \\
    HR \hspace{-1.2mm} &
    w/o FDL \hspace{-1.2mm} &
    w/ FDL \hspace{1.0mm} &
    HR \hspace{-1.2mm} &
    w/o FDL  \hspace{-1.2mm} &
    w/ FDL \hspace{-1.2mm}
    \\
    \end{tabular}
    \end{adjustbox}
}
\vspace{-1mm}
\caption{Comparison of spectral decomposition with and without FDL. Without FDL, distinct periodic artifacts are visible in the left image. After applying FDL, the spectral leakage is effectively mitigated.}
\vspace{-2mm}
\label{fig:spectrum}
\end{figure*}

Although the proposed asymmetric discriminative distillation can partially mitigate the artifacts caused by the mismatch between multi-step and one-step generation, notable periodic artifacts still remain, as shown in Fig.~\ref{fig:spectrum}. Previous studies~\cite{li2025fluxsr} have observed similar issues and analyzed them from a phenomenological perspective. They attribute these artifacts to token similarity within DiT patches, noting that the artifact period in pixel space equals the DiT patch size multiplied by the VAE scale factor. We conduct a deeper investigation into this phenomenon, showing that the fundamental cause ultimately lies in the high-frequency domain, \ie, the spectral leakage problem.

To mitigate the artifacts so as to enhance realistic quality, we decide to introduce frequency domain loss regularization, frequency distribution loss (FDL)~\citep{ni2024misalignment}, to narrow the gap between predicted and real data distribution in the frequency domain. FDL was first proposed to offset the spatially misaligned data problem in image transformation tasks. The formulation and introduction of frequency in FDL provides a natural constraint to the generated distribution, resulting in anti-periodic artifact effects and better global perception.

\begin{table*}[t]
\scriptsize 
\setlength{\tabcolsep}{0.3mm} 
\newcolumntype{C}{>{\centering\arraybackslash}X}
\begin{center}
\begin{tabularx}{\textwidth}{c|c|CC|CC|CCCCC}
\toprule[0.15em]
\rowcolor{color3} Type & Methods & PSNR$\uparrow$ & SSIM$\uparrow$ & LPIPS$\downarrow$ & DISTS$\downarrow$ & NIQE$\downarrow$ & MAN.$\uparrow$ & MUS.$\uparrow$ & QAli.$\uparrow$ & Qual.$\uparrow$ \\

\midrule[0.15em]

\multirow{6}{*}{\makecell{Multi-step\\Diffusion}} 
& SUPIR~\cite{yu2024supir} & 22.22 & 0.5748 & \bb{0.3806} & \bb{0.1660} & \bb{3.4372} & 0.5793 & \bb{63.4906} & 4.3862 & 0.6959 \\
& DiT4SR~\cite{duan2025dit4sr} & 20.76 & 0.5451 & 0.3995 & 0.1670 & \rr{3.0084} & \rr{0.6246} & \rr{65.1493} & \rr{4.4567} & \bb{0.6987} \\
& DiffBIR~\cite{lin2024diffbir} & \rr{23.49} & 0.5837 & 0.3963 & 0.2396 & 4.8019 & 0.5730 & 56.7752 & 3.9390 & 0.6607 \\
& PASD~\cite{yang2024pasd} & \bb{23.42} & \rr{0.6229} & 0.4119 & 0.1942 & 4.3308 & 0.4657 & 56.1247 & 4.1016 & 0.5530 \\
& I.R.~\cite{liu2025instructrestore} & 23.35 & 0.6001 & 0.4108 & 0.2554 & 4.3449 & 0.5221 & 52.9015 & 3.7092 & 0.5486 \\
& SeeSR~\cite{wu2024seesr} & 23.24 & \bb{0.6125} & \rr{0.3473} & \rr{0.1580} & 3.6129 & \bb{0.5945} & 61.5973 & \bb{4.4410} & \rr{0.7048} \\
\cmidrule{1-11}
\multirow{9}{*}{\makecell{One-step\\Diffusion}} 
& CTMSR~\cite{you2025ctmsr} & \rr{23.93} & 0.6200 & 0.3652 & 0.1918 & 4.4132 & 0.4721 & 60.6717 & 4.0600 & 0.6872 \\
& PiSA-SR~\cite{sun2024pisasr} & \bb{23.36} & \bb{0.6224} & 0.3442 & 0.1789 & 3.6403 & 0.5706 & 60.1926 & 4.1500 & 0.7109 \\
& TSD-SR~\cite{dong2025tsdsr} & 21.68 & 0.5549 & 0.3453 & 0.1526 & \rr{2.8397} & 0.5454 & \rr{66.5933} & 4.1558 & 0.7209 \\
& OSEDiff~\cite{wu2024osediff} & 23.27 & \rr{0.6441} & 0.3972 & 0.2179 & 4.5961 & 0.4940 & 54.0388 & 3.9813 & 0.6287 \\
& InvSR~\cite{yue2025invsr} & 21.09 & 0.5979 & 0.4255 & 0.2125 & 4.3214 & 0.5027 & 58.3533 & 3.7655 & 0.6247 \\
& HYPIR~\cite{lin2025hypir} & 21.78 & 0.5845 & 0.3551 & 0.1457 & 3.3566 & 0.5673 & 63.6403 & 4.3359 & \bb{0.7299} \\
& SinSR~\cite{wang2024sinsr} & 22.74 & 0.5356 & 0.4630 & 0.2143 & 6.1304 & 0.4830 & 60.8577 & 4.1208 & 0.5115 \\
& StrSR - Z & 21.35 & 0.5798 & \bb{0.3216} & \bb{0.1407} & 3.4738 & \bb{0.5727} & \bb{64.6547} & \bb{4.4759} & \rr{0.7316} \\
& StrSR - F & 21.70 & 0.5982 & \rr{0.2992} & \rr{0.1288} & \bb{3.0379} & \rr{0.5864} & 63.6568 & \rr{4.4763} & 0.7022 \\
\bottomrule[0.15em]
\end{tabularx}
\end{center}
\scriptsize
\vspace{-1mm}
\caption{Quantitative results ($\times$4) on DIV2K-val dataset. I.R. denotes InstructRestore; MAN., MUS., QAli., and Qual. stand for MANIQA, MUSIQ, QAlign, and QualiCLIP, respectively. StrSR - Z/F refers to the Z-Image and FLUX implementations of StrSR. } 
\label{tab:metric_DIV2K}
\vspace{-6mm}
\end{table*}

Firstly, FDL transforms predicted and target images into feature space using a pretrained feature extractor $\Phi$, then employs Sliced Wasserstein (SW) distance to measure the distribution distance between frequency components of the features after the discrete Fourier transform (DFT). The final FDL can be formulated as:
\begin{equation}
    \mathcal{L}_\text{FDL}(x_{\text{HR}},x_{\text{pred}}) = \operatorname{SW}(\mathcal{A}_{\Phi(x_{\text{HR}})}, \mathcal{A}_{\Phi(x_{\text{pred}})}) + \lambda \cdot \operatorname{SW}(\mathcal{P}_{\Phi(x_{\text{HR}})}, \mathcal{P}_{\Phi(x_{\text{pred}})}),
\end{equation}
where $\mathcal{A}$ and $\mathcal{P}$ denote the amplitude and phase components of data in the frequency domain, respectively.\par

Taking into account the reconstruction goal in both the spatial and spectral domains, we have the following training objective for the generator:
\begin{equation}
    \mathcal{L}_G=\mathcal{L}_1 + \lambda_1 \cdot \mathcal{L}_\text{lpips}+\lambda_2 \cdot \mathcal{L}^{G}_\mathit{Ra}+\lambda_3 \cdot \mathcal{L}_\text{FDL},
\end{equation}
where $\mathcal{L}_1$ and $\mathcal{L}_\text{lpips}$ represent the commonly used L1 and LPIPS losses in image restoration for perception-distortion trade-off~\citep{blau2018perception}. 

\vspace{-2mm}
\section{Experiments}
\vspace{-1mm}
\subsection{Experimental Settings}
\vspace{-1mm}
\noindent\textbf{Training Datasets} We choose widely-used LSDIR~\citep{li2023lsdir} and synthetic dataset Aesthetic-4K~\citep{zhang2025diffusion} for training. Specifically, we picked the first $60$,$000$ images from Aesthetic-4K and all the images from LSDIR, for better generative quality and faster convergence. Images are center-cropped to 1,024$\times$1,024 pixels during training, and the corresponding LR images are synthesized using the standard degradation pipeline proposed in RealESRGAN~\citep{wang2021realesrgan}.

\begin{table*}[t]
\scriptsize 
\setlength{\tabcolsep}{0.3mm} 
\newcolumntype{C}{>{\centering\arraybackslash}X}
\begin{center}
\begin{tabularx}{\textwidth}{c|c|CC|CC|CCCCC}
\toprule[0.15em]
\rowcolor{color3} Type & Methods & PSNR$\uparrow$ & SSIM$\uparrow$ & LPIPS$\downarrow$ & DISTS$\downarrow$ & NIQE$\downarrow$ & MAN.$\uparrow$ & MUS.$\uparrow$ & QAli.$\uparrow$ & Qual.$\uparrow$ \\

\midrule[0.15em]

\multirow{6}{*}{\makecell{Multi-step\\Diffusion}} 
& SUPIR~\cite{yu2024supir} & 24.44 & 0.6993 & 0.3322 & 0.1982 & 4.5734 & 0.6034 & \bb{64.9685} & \rr{4.0683} & \bb{0.6647} \\
& DiT4SR~\cite{duan2025dit4sr} & 23.24 & 0.6617 & 0.3367 & 0.1901 & \rr{4.2814} & \rr{0.6486} & \rr{67.9475} & 3.9679 & 0.6592 \\
& DiffBIR~\cite{lin2024diffbir} & \bb{26.29} & 0.7242 & 0.3219 & 0.2075 & 5.9847 & 0.5851 & 62.4423 & 3.9038 & 0.6237 \\
& PASD~\cite{yang2024pasd} & \rr{26.71} & \rr{0.7615} & \rr{0.2759} & \rr{0.1630} & 4.8252 & 0.5416 & 60.9738 & \bb{4.0344} & 0.5811 \\
& I.R.~\cite{liu2025instructrestore} & 26.10 & 0.7450 & 0.3010 & 0.2033 & 4.9780 & 0.5953 & 62.7246 & 3.9585 & 0.6122 \\
& SeeSR~\cite{wu2024seesr} & 26.21 & \bb{0.7563} & \bb{0.2786} & \bb{0.1759} & \bb{4.5428} & \bb{0.6106} & 64.3097 & 4.0009 & \rr{0.6832} \\
\cmidrule{1-11}
\multirow{9}{*}{\makecell{One-step\\Diffusion}} 
& CTMSR~\cite{you2025ctmsr} & \rr{26.18} & \rr{0.7652} & 0.2935 & 0.1867 & 4.6381 & 0.5202 & 63.5130 & 3.9233 & 0.6750 \\
& PiSA-SR~\cite{sun2024pisasr} & 25.60 & \bb{0.7509} & \bb{0.2744} & \bb{0.1745} & 4.4039 & 0.6277 & 66.4567 & 3.9946 & 0.7073 \\
& TSD-SR~\cite{dong2025tsdsr} & 23.76 & 0.6984 & 0.2874 & 0.1844 & \bb{3.8310} & 0.6139 & \rr{69.0501} & 4.0924 & \bb{0.7408} \\
& OSEDiff~\cite{wu2024osediff} & 24.29 & 0.7153 & 0.3440 & 0.2109 & 5.7609 & 0.5154 & 59.5873 & 3.8685 & 0.6694 \\
& InvSR~\cite{yue2025invsr} & 22.61 & 0.7114 & 0.3068 & 0.1862 & 4.2273 & \bb{0.6332} & 67.1131 & 4.0047 & 0.7102 \\
& HYPIR~\cite{lin2025hypir} & 22.14 & 0.6716 & 0.3313 & 0.2004 & 4.1558 & 0.6317 & 68.3347 & \bb{4.1322} & 0.7232 \\
& SinSR~\cite{wang2024sinsr} & \bb{26.02} & 0.7085 & 0.4014 & 0.2261 & 6.2482 & 0.5246 & 61.1541 & 3.8799 & 0.5252 \\
& StrSR - Z & 22.15 & 0.6606 & 0.3215 & 0.1958 & 4.4439 & 0.5941 & \bb{68.5793} & \rr{4.2371} & \rr{0.7417} \\
& StrSR - F & 23.77 & 0.7236 & \rr{0.2599} & \rr{0.1665} & \rr{3.8090} & \rr{0.6333} & 67.3318 & 4.1296 & 0.7134 \\
\bottomrule[0.15em]
\end{tabularx}
\end{center}
\scriptsize
\caption{Quantitative results ($\times$4) on RealSR. I.R. denotes InstructRestore; MAN., MUS., QAli., and Qual. stand for MANIQA, MUSIQ, QAlign, and QualiCLIP, respectively. StrSR - Z/F refers to the Z-Image and FLUX implementations of StrSR. } 
\label{tab:metric_RealSR}
\vspace{-8mm}
\end{table*}

\vspace{1mm plus 1mm minus 1mm}
\noindent\textbf{Evaluation Datasets} We test our method on synthetic dataset DIV2K-val~\citep{agustsson2017ntire} and commonly used real-world datasets: RealSR~\citep{cai2019toward} and RealLQ250~\citep{ai2024dreamclear}. We use the same degradation pipeline as RealESRGAN~\citep{wang2021realesrgan} to generate LR images of DIV2K-val~\citep{agustsson2017ntire}. Our method is tested on full-sized images to better reflect real-world performance. All evaluations are conducted on the $\times$4 SR setting.

\begin{table}[t]
\scriptsize
\setlength{\tabcolsep}{1.5mm} 
\newcolumntype{C}{>{\centering\arraybackslash}X}
\begin{center}
\begin{tabularx}{0.9\columnwidth}{c|c|CCCC}
\toprule[0.15em]
\rowcolor{color3} Type & Methods & NIQE$\downarrow$ & MANIQA$\uparrow$ & MUSIQ$\uparrow$ & QAlign$\uparrow$ \\

\midrule[0.15em]
\multirow{6}{*}{\makecell{Multi-step\\Diffusion}} 
& SUPIR~\cite{yu2024supir} & \bb{3.6651} & 0.5785 & 66.1146 & 4.1144 \\
& DiT4SR~\cite{duan2025dit4sr} & \rr{3.5479} & \rr{0.6318} & \rr{70.9467} & \rr{4.2268} \\
& DiffBIR~\cite{lin2024diffbir} & 5.0983 & 0.5877 & 66.4842 & 3.9950 \\
& PASD~\cite{yang2024pasd} & 4.5019 & 0.5152 & 61.6403 & 3.9688 \\
& InstructRestore~\cite{liu2025instructrestore} & 4.9263 & 0.5448 & 62.9681 & 3.5949 \\
& SeeSR~\cite{wu2024seesr} & 4.4334 & \bb{0.5916} & \bb{68.5174} & \bb{4.1372} \\
\cmidrule{1-6}
\multirow{9}{*}{\makecell{One-step\\Diffusion}} 
& CTMSR~\cite{you2025ctmsr} & 4.5791 & 0.5082 & 67.9592 & 3.7108 \\
& PiSA-SR~\cite{sun2024pisasr} & 3.9149 & \bb{0.6053} & 69.4146 & 4.2053 \\
& TSD-SR~\cite{dong2025tsdsr} & \bb{3.4879} & 0.5829 & \rr{71.0490} & 4.1694 \\
& OSEDiff~\cite{wu2024osediff} & 4.8754 & 0.5070 & 61.2067 & 3.7763 \\
& InvSR~\cite{yue2025invsr} & 4.4034 & 0.5719 & 64.7987 & 3.9033 \\
& HYPIR~\cite{lin2025hypir} & 3.9651 & 0.6051 & 69.7083 & 4.1993 \\
& SinSR~\cite{wang2024sinsr} & 5.7901 & 0.5156 & 65.9640 & 3.7436 \\
& StrSR - Z-Image & 4.1019 & 0.5774 & \bb{70.1176} & \bb{4.2595} \\
& StrSR - FLUX & \rr{3.4693} & \rr{0.6138} & 69.2050 & \rr{4.2851} \\

\bottomrule[0.15em]
\end{tabularx}
\end{center}
\caption{Comparison ($\times$4) of different methods on RealLQ250.} 
\label{tab:metric_RealLQ250}
\end{table}

\vspace{1mm plus 1mm minus 1mm}
\noindent\textbf{Metrics} For those full-reference metrics, we evaluate pixel-level restoration using PSNR and SSIM, and perceptual metrics include LPIPS~\cite{zhang2018lpips} and DISTS~\cite{ding2020dists}. For no-reference metrics, we choose NIQE~\cite{zhang2015niqe}, MANIQA~\cite{yang2022maniqa}, MUSIQ~\cite{ke2021musiq}, QAlign~\cite{wu2024qalign}, and QualiCLIP~\cite{agnolucci2024qualityaware}. All the metrics are evaluated by \texttt{pyiqa}~\cite{pyiqa}.

\vspace{1mm plus 1mm minus 1mm}
\noindent\textbf{Implementation Details} We adopt Z-Image-Turbo~\cite{team2025zimage} and FLUX.2 [klein] 4B~\cite{FLUX.2-klein-4B} as base generator models. The Z-Image-Turbo model has 6B parameters, and it is an 8-step model distilled from Z-Image. The FLUX.2 [klein] 4B model is distilled from FLUX.2 [klein] Base 9B, and it has 4 NFEs. 
We instantiate the reasoning vision-language module with Qwen3-VL-4B-Instruct~\citep{qwen3vltechnicalreport}. As both generator backbones are built on the Qwen3~\citep{qwen3technicalreport} text encoder, Qwen3-VL provides compatible tokenization and hidden representations, allowing us to freeze the VLM and learn only a lightweight adaptive projection layer.

The discriminator's backbone is pretrained CLIP-ConvNeXt-XXLarge~\citep{radford2021learning,convnext,laion}. We fine-tune the pretrained generator with LoRA~\citep{hu2022lora} rank$=$256. The generator and discriminator are updated at a 1:1 ratio.
The two-stage training is performed in bf16 precision for $60$,$000$ steps on 4 NVIDIA A800 GPUs with batch size$=$4. For each training stage, we use AdamW optimizer~\citep{adamw} with default betas, and set the learning rates of generator and discriminator to $5$$\times$$10^{-6}$ and $1$$\times$$10^{-6}$ respectively. The first stage takes 40$,$000 steps, and then another 20$,$000 steps are used to fine-tune with FDL loss. For the generator, we set the hyperparameters as $\lambda_1$$=$$3$, $\lambda_2$$=$$0.1$, $\lambda_3$$=$$0$ in the first stage and $\lambda_3$$=$$0.002$ in the second stage, respectively. For the discriminator, we set $\lambda_\mathit{R1}$$=$$10$.  
For approximated R1, we add Gaussian perturbation to real inputs with $\sigma$$=$$0.005$.

\begin{figure*}[t]
\scriptsize
\centering

\begin{tabular}{cccc}

\hspace{-0.4cm}
\newcommand{\imgid}{image_22}
\newcommand{\imgnote}{022}
\begin{adjustbox}{valign=t}
\begin{tabular}{c}
\includegraphics[width=0.237\textwidth,height=0.1845\textwidth]{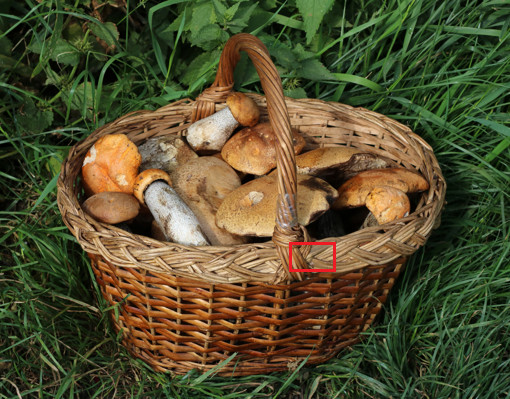} \\
DIV2K: \texttt{img\_\imgnote}
\end{tabular}
\end{adjustbox}
\hspace{-0.2cm}
\begin{adjustbox}{valign=t}
\begin{tabular}{cccccc}
\includegraphics[width=0.139\textwidth]{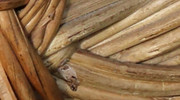} \hspace{-1mm} &
\includegraphics[width=0.139\textwidth]{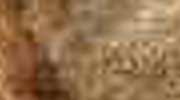} \hspace{-1mm} &
\includegraphics[width=0.139\textwidth]{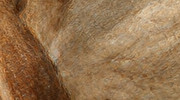} \hspace{-1mm} &
\includegraphics[width=0.139\textwidth]{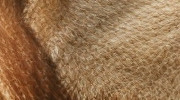} \hspace{-1mm} &
\includegraphics[width=0.139\textwidth]{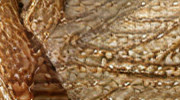} \hspace{-1mm} \\
HR \hspace{-1mm} &
Bicubic \hspace{-1mm} &
CTMSR~\cite{you2025ctmsr} \hspace{-1mm} &
PiSA-SR~\cite{sun2024pisasr} \hspace{-1mm} &
TSD-SR~\cite{dong2025tsdsr} \hspace{-1mm} \\
\includegraphics[width=0.139\textwidth]{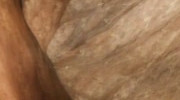} \hspace{-1mm} &
\includegraphics[width=0.139\textwidth]{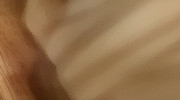} \hspace{-1mm} &
\includegraphics[width=0.139\textwidth]{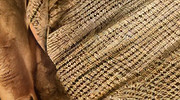} \hspace{-1mm} &
\includegraphics[width=0.139\textwidth]{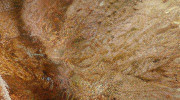} \hspace{-1mm} &
\includegraphics[width=0.139\textwidth]{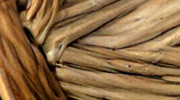} \hspace{-1mm} \\
OSEDiff~\cite{wu2024osediff} \hspace{-1mm} &
InvSR~\cite{yue2025invsr} \hspace{-1mm} &
HYPIR~\cite{lin2025hypir} \hspace{-1mm} &
SinSR~\cite{wang2024sinsr}  \hspace{-1mm} &
StrSR (ours) \hspace{-1mm} \\
\end{tabular}
\end{adjustbox}
\\

\hspace{-0.4cm}
\newcommand{\imgid}{image_34}
\newcommand{\imgnote}{034}
\begin{adjustbox}{valign=t}
\begin{tabular}{c}
\includegraphics[width=0.237\textwidth,height=0.1845\textwidth]{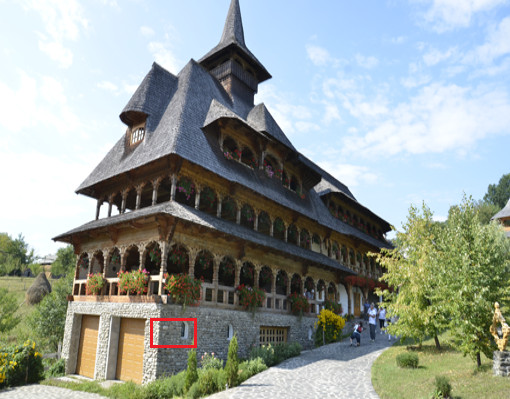} \\
DIV2K: \texttt{img\_\imgnote}
\end{tabular}
\end{adjustbox}
\hspace{-0.2cm}
\begin{adjustbox}{valign=t}
\begin{tabular}{cccccc}
\includegraphics[width=0.139\textwidth]{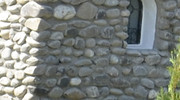} \hspace{-1mm} &
\includegraphics[width=0.139\textwidth]{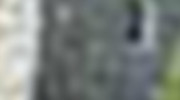} \hspace{-1mm} &
\includegraphics[width=0.139\textwidth]{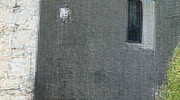} \hspace{-1mm} &
\includegraphics[width=0.139\textwidth]{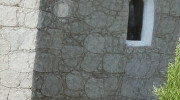} \hspace{-1mm} &
\includegraphics[width=0.139\textwidth]{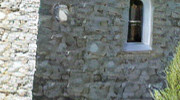} \hspace{-1mm} \\
HR \hspace{-1mm} &
Bicubic \hspace{-1mm} &
CTMSR~\cite{you2025ctmsr} \hspace{-1mm} &
PiSA-SR~\cite{sun2024pisasr} \hspace{-1mm} &
TSD-SR~\cite{dong2025tsdsr} \hspace{-1mm} \\
\includegraphics[width=0.139\textwidth]{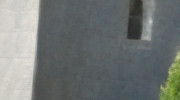} \hspace{-1mm} &
\includegraphics[width=0.139\textwidth]{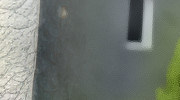} \hspace{-1mm} &
\includegraphics[width=0.139\textwidth]{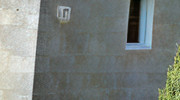} \hspace{-1mm} &
\includegraphics[width=0.139\textwidth]{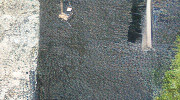} \hspace{-1mm} &
\includegraphics[width=0.139\textwidth]{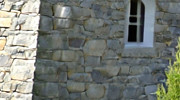} \hspace{-1mm} \\
OSEDiff~\cite{wu2024osediff} \hspace{-1mm} &
InvSR~\cite{yue2025invsr} \hspace{-1mm} &
HYPIR~\cite{lin2025hypir} \hspace{-1mm} &
SinSR~\cite{wang2024sinsr}  \hspace{-1mm} &
StrSR (ours) \hspace{-1mm} \\
\end{tabular}
\end{adjustbox}
\\

\hspace{-0.4cm}
\newcommand{\imgid}{image_70}
\newcommand{\imgnote}{070}
\begin{adjustbox}{valign=t}
\begin{tabular}{c}
\includegraphics[width=0.237\textwidth,height=0.1845\textwidth]{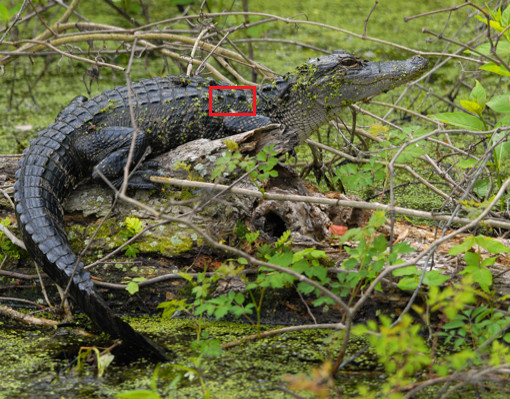} \\
DIV2K: \texttt{img\_\imgnote}
\end{tabular}
\end{adjustbox}
\hspace{-0.2cm}
\begin{adjustbox}{valign=t}
\begin{tabular}{cccccc}
\includegraphics[width=0.139\textwidth]{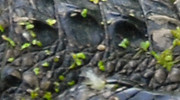} \hspace{-1mm} &
\includegraphics[width=0.139\textwidth]{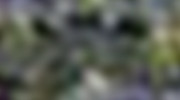} \hspace{-1mm} &
\includegraphics[width=0.139\textwidth]{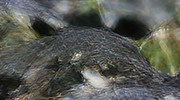} \hspace{-1mm} &
\includegraphics[width=0.139\textwidth]{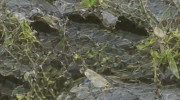} \hspace{-1mm} &
\includegraphics[width=0.139\textwidth]{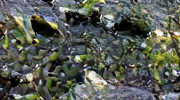} \hspace{-1mm} \\
HR \hspace{-1mm} &
Bicubic \hspace{-1mm} &
CTMSR~\cite{you2025ctmsr} \hspace{-1mm} &
PiSA-SR~\cite{sun2024pisasr} \hspace{-1mm} &
TSD-SR~\cite{dong2025tsdsr} \hspace{-1mm} \\
\includegraphics[width=0.139\textwidth]{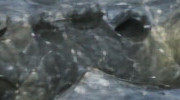} \hspace{-1mm} &
\includegraphics[width=0.139\textwidth]{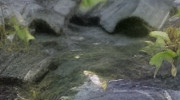} \hspace{-1mm} &
\includegraphics[width=0.139\textwidth]{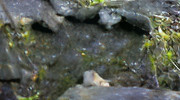} \hspace{-1mm} &
\includegraphics[width=0.139\textwidth]{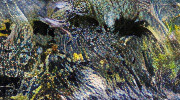} \hspace{-1mm} &
\includegraphics[width=0.139\textwidth]{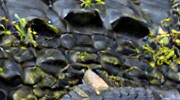} \hspace{-1mm} \\
OSEDiff~\cite{wu2024osediff} \hspace{-1mm} &
InvSR~\cite{yue2025invsr} \hspace{-1mm} &
HYPIR~\cite{lin2025hypir} \hspace{-1mm} &
SinSR~\cite{wang2024sinsr}  \hspace{-1mm} &
StrSR (ours) \hspace{-1mm} \\
\end{tabular}
\end{adjustbox}
\\

\hspace{-0.4cm}
\newcommand{\imgid}{image_86}
\newcommand{\imgnote}{086}
\begin{adjustbox}{valign=t}
\begin{tabular}{c}
\includegraphics[width=0.237\textwidth,height=0.1845\textwidth]{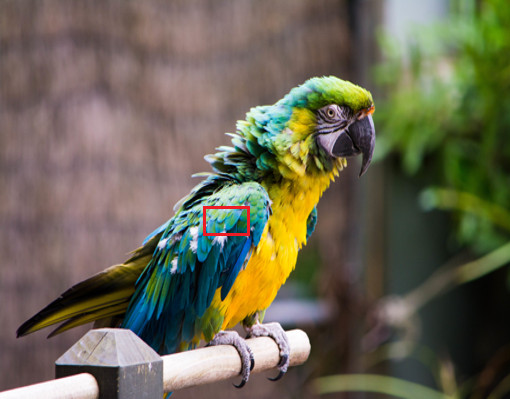} \\
DIV2K: \texttt{img\_\imgnote}
\end{tabular}
\end{adjustbox}
\hspace{-0.2cm}
\begin{adjustbox}{valign=t}
\begin{tabular}{cccccc}
\includegraphics[width=0.139\textwidth]{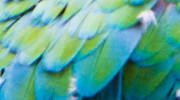} \hspace{-1mm} &
\includegraphics[width=0.139\textwidth]{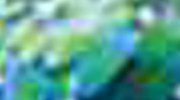} \hspace{-1mm} &
\includegraphics[width=0.139\textwidth]{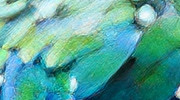} \hspace{-1mm} &
\includegraphics[width=0.139\textwidth]{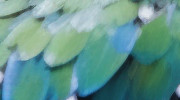} \hspace{-1mm} &
\includegraphics[width=0.139\textwidth]{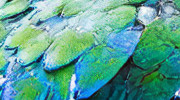} \hspace{-1mm} \\
HR \hspace{-1mm} &
Bicubic \hspace{-1mm} &
CTMSR~\cite{you2025ctmsr} \hspace{-1mm} &
PiSA-SR~\cite{sun2024pisasr} \hspace{-1mm} &
TSD-SR~\cite{dong2025tsdsr} \hspace{-1mm} \\
\includegraphics[width=0.139\textwidth]{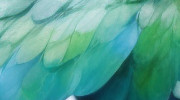} \hspace{-1mm} &
\includegraphics[width=0.139\textwidth]{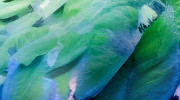} \hspace{-1mm} &
\includegraphics[width=0.139\textwidth]{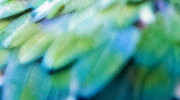} \hspace{-1mm} &
\includegraphics[width=0.139\textwidth]{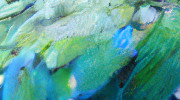} \hspace{-1mm} &
\includegraphics[width=0.139\textwidth]{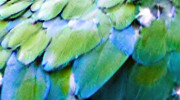} \hspace{-1mm} \\
OSEDiff~\cite{wu2024osediff} \hspace{-1mm} &
InvSR~\cite{yue2025invsr} \hspace{-1mm} &
HYPIR~\cite{lin2025hypir} \hspace{-1mm} &
SinSR~\cite{wang2024sinsr}  \hspace{-1mm} &
StrSR (ours) \hspace{-1mm} \\
\end{tabular}
\end{adjustbox}
\\

\end{tabular}

\caption{Visual comparison for image SR ($\times$4) in DIV2K-val dataset.}
\label{fig:vis-DIV2K}
\end{figure*}

\begin{figure*}[t]
\scriptsize
\centering
\scalebox{0.98}{
\begin{tabular}{cccc}

\begin{adjustbox}{valign=t}
\begin{tabular}{c}
\includegraphics[width=0.237\textwidth]{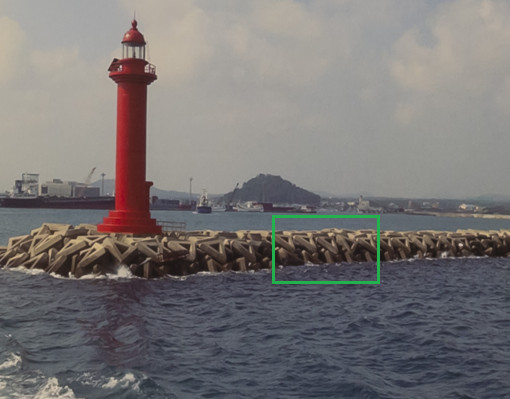}
\\
RealSR: \texttt{Canon\_030}
\end{tabular}
\end{adjustbox}
\hspace{-0.2cm}
\begin{adjustbox}{valign=t}
\begin{tabular}{cccccc}
\includegraphics[width=0.139\textwidth]{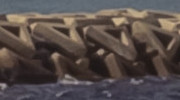} \hspace{-1mm} &
\includegraphics[width=0.139\textwidth]{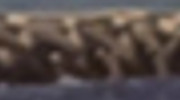} \hspace{-1mm} &
\includegraphics[width=0.139\textwidth]{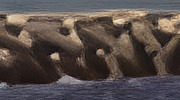} \hspace{-1mm} &
\includegraphics[width=0.139\textwidth]{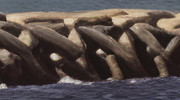} \hspace{-1mm} &
\includegraphics[width=0.139\textwidth]{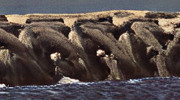} \hspace{-1mm} &
\\
HR \hspace{-1mm} &
Bicubic \hspace{-1mm} &
CTMSR~\cite{you2025ctmsr} \hspace{-1mm} &
PiSA-SR~\cite{sun2024pisasr} \hspace{-1mm} &
TSD-SR~\cite{dong2025tsdsr} \hspace{-1mm} 
\\
\includegraphics[width=0.139\textwidth]{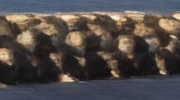} \hspace{-1mm} &
\includegraphics[width=0.139\textwidth]{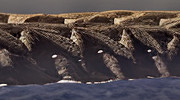} \hspace{-1mm} &
\includegraphics[width=0.139\textwidth]{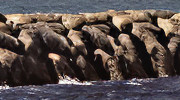} \hspace{-1mm} &
\includegraphics[width=0.139\textwidth]{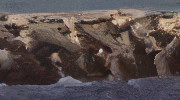} \hspace{-1mm} &
\includegraphics[width=0.139\textwidth]{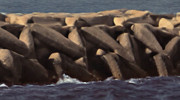} \hspace{-1mm}  
\\ 
OSEDiff~\cite{wu2024osediff} \hspace{-1mm} &
InvSR~\cite{yue2025invsr} \hspace{-1mm} &
HYPIR~\cite{lin2025hypir} \hspace{-1mm} &
SinSR~\cite{wang2024sinsr}  \hspace{-1mm} &
StrSR (ours) \hspace{-1mm}
\\
\end{tabular}
\end{adjustbox}
\\

\begin{adjustbox}{valign=t}
\begin{tabular}{c}
\includegraphics[width=0.237\textwidth]{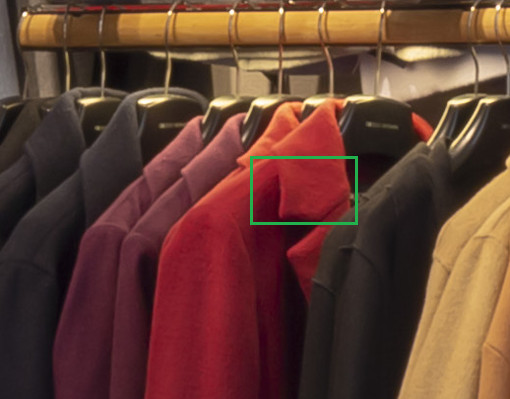}
\\
RealSR: \texttt{Canon\_039}
\end{tabular}
\end{adjustbox}
\hspace{-0.2cm}
\begin{adjustbox}{valign=t}
\begin{tabular}{cccccc}
\includegraphics[width=0.139\textwidth]{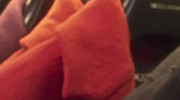} \hspace{-1mm} &
\includegraphics[width=0.139\textwidth]{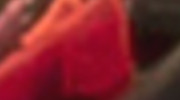} \hspace{-1mm} &
\includegraphics[width=0.139\textwidth]{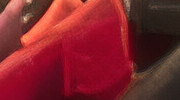} \hspace{-1mm} &
\includegraphics[width=0.139\textwidth]{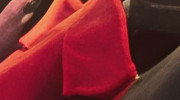} \hspace{-1mm} &
\includegraphics[width=0.139\textwidth]{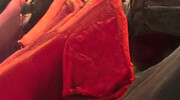} \hspace{-1mm} &
\\
HR \hspace{-1mm} &
Bicubic \hspace{-1mm} &
CTMSR~\cite{you2025ctmsr} \hspace{-1mm} &
PiSA-SR~\cite{sun2024pisasr} \hspace{-1mm} &
TSD-SR~\cite{dong2025tsdsr} \hspace{-1mm} 
\\
\includegraphics[width=0.139\textwidth]{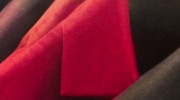} \hspace{-1mm} &
\includegraphics[width=0.139\textwidth]{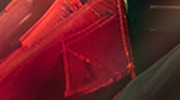} \hspace{-1mm} &
\includegraphics[width=0.139\textwidth]{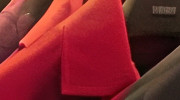} \hspace{-1mm} &
\includegraphics[width=0.139\textwidth]{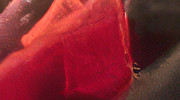} \hspace{-1mm} &
\includegraphics[width=0.139\textwidth]{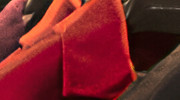} \hspace{-1mm}  
\\ 
OSEDiff~\cite{wu2024osediff} \hspace{-1mm} &
InvSR~\cite{yue2025invsr} \hspace{-1mm} &
HYPIR~\cite{lin2025hypir} \hspace{-1mm} &
SinSR~\cite{wang2024sinsr}  \hspace{-1mm} &
StrSR (ours) \hspace{-1mm}
\\
\end{tabular}
\end{adjustbox}
\\

\begin{adjustbox}{valign=t}
\begin{tabular}{c}
\includegraphics[width=0.237\textwidth]{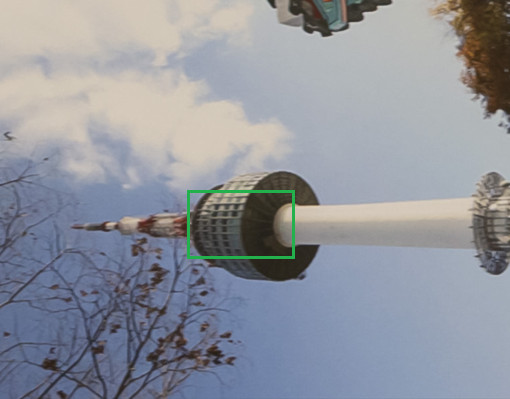}
\\
RealSR: \texttt{Canon\_018}
\end{tabular}
\end{adjustbox}
\hspace{-0.2cm}
\begin{adjustbox}{valign=t}
\begin{tabular}{cccccc}
\includegraphics[width=0.139\textwidth]{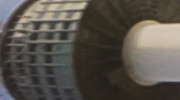} \hspace{-1mm} &
\includegraphics[width=0.139\textwidth]{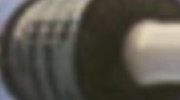} \hspace{-1mm} &
\includegraphics[width=0.139\textwidth]{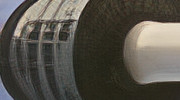} \hspace{-1mm} &
\includegraphics[width=0.139\textwidth]{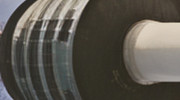} \hspace{-1mm} &
\includegraphics[width=0.139\textwidth]{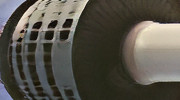} \hspace{-1mm} &
\\
HR \hspace{-1mm} &
Bicubic \hspace{-1mm} &
CTMSR~\cite{you2025ctmsr} \hspace{-1mm} &
PiSA-SR~\cite{sun2024pisasr} \hspace{-1mm} &
TSD-SR~\cite{dong2025tsdsr} \hspace{-1mm} 
\\
\includegraphics[width=0.139\textwidth]{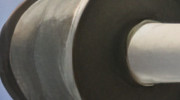} \hspace{-1mm} &
\includegraphics[width=0.139\textwidth]{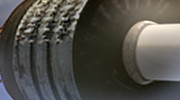} \hspace{-1mm} &
\includegraphics[width=0.139\textwidth]{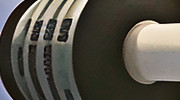} \hspace{-1mm} &
\includegraphics[width=0.139\textwidth]{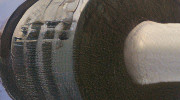} \hspace{-1mm} &
\includegraphics[width=0.139\textwidth]{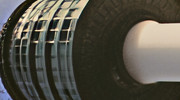} \hspace{-1mm}  
\\ 
OSEDiff~\cite{wu2024osediff} \hspace{-1mm} &
InvSR~\cite{yue2025invsr} \hspace{-1mm} &
HYPIR~\cite{lin2025hypir} \hspace{-1mm} &
SinSR~\cite{wang2024sinsr}  \hspace{-1mm} &
StrSR (ours) \hspace{-1mm}
\\
\end{tabular}
\end{adjustbox}
\\

\begin{adjustbox}{valign=t}
\begin{tabular}{c}
\includegraphics[width=0.237\textwidth]{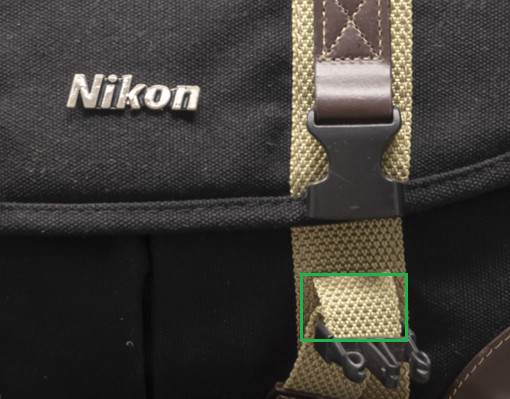}
\\
RealSR: \texttt{Nikon\_020}
\end{tabular}
\end{adjustbox}
\hspace{-0.2cm}
\begin{adjustbox}{valign=t}
\begin{tabular}{cccccc}
\includegraphics[width=0.139\textwidth]{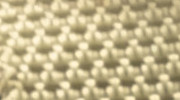} \hspace{-1mm} &
\includegraphics[width=0.139\textwidth]{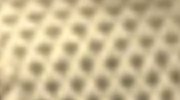} \hspace{-1mm} &
\includegraphics[width=0.139\textwidth]{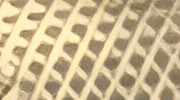} \hspace{-1mm} &
\includegraphics[width=0.139\textwidth]{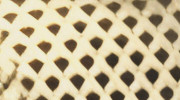} \hspace{-1mm} &
\includegraphics[width=0.139\textwidth]{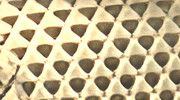} \hspace{-1mm} &
\\
HR \hspace{-1mm} &
Bicubic \hspace{-1mm} &
CTMSR~\cite{you2025ctmsr} \hspace{-1mm} &
PiSA-SR~\cite{sun2024pisasr} \hspace{-1mm} &
TSD-SR~\cite{dong2025tsdsr} \hspace{-1mm} 
\\
\includegraphics[width=0.139\textwidth]{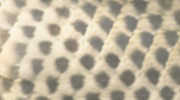} \hspace{-1mm} &
\includegraphics[width=0.139\textwidth]{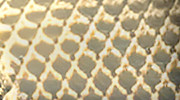} \hspace{-1mm} &
\includegraphics[width=0.139\textwidth]{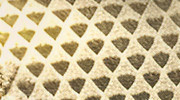} \hspace{-1mm} &
\includegraphics[width=0.139\textwidth]{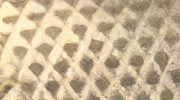} \hspace{-1mm} &
\includegraphics[width=0.139\textwidth]{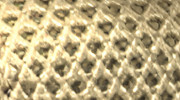} \hspace{-1mm}  
\\ 
OSEDiff~\cite{wu2024osediff} \hspace{-1mm} &
InvSR~\cite{yue2025invsr} \hspace{-1mm} &
HYPIR~\cite{lin2025hypir} \hspace{-1mm} &
SinSR~\cite{wang2024sinsr}  \hspace{-1mm} &
StrSR (ours) \hspace{-1mm}
\\
\end{tabular}
\end{adjustbox}
\\

\end{tabular}
}
\caption{Visual comparison for image SR ($\times$4) in RealSR dataset.}
\label{fig:vis-RealSR}
\vspace{-4mm}
\end{figure*}

\begin{figure*}[t] 
\scriptsize
\centering

\newcommand{\imgid}{00009}
\newcommand{\imgnote}{009}
\scalebox{1.0}{
    \begin{adjustbox}{valign=t}
    \begin{tabular}{cccccc}
    \includegraphics[width=0.158\textwidth]{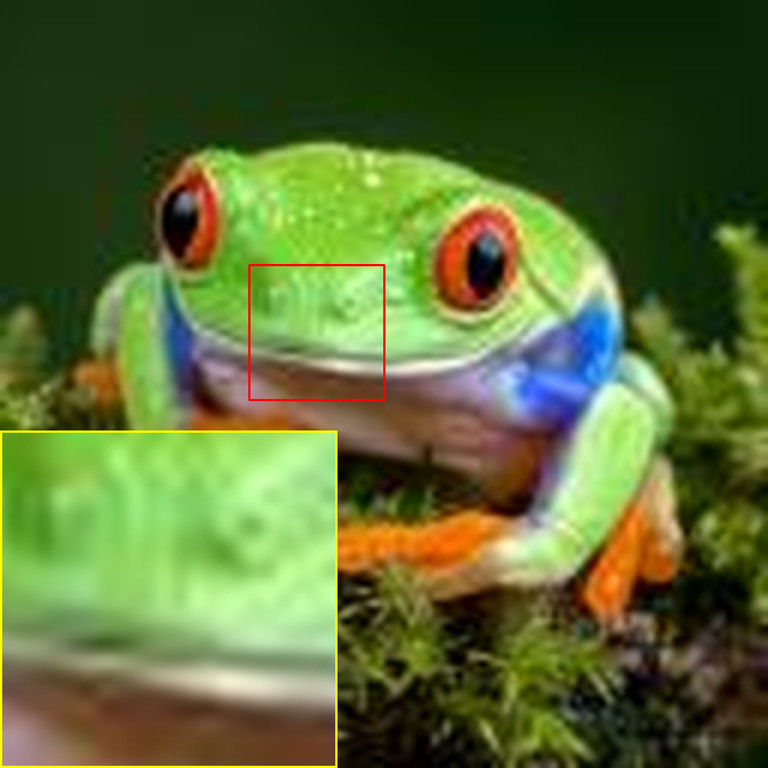} \hspace{-1.2mm} &
    \includegraphics[width=0.158\textwidth]{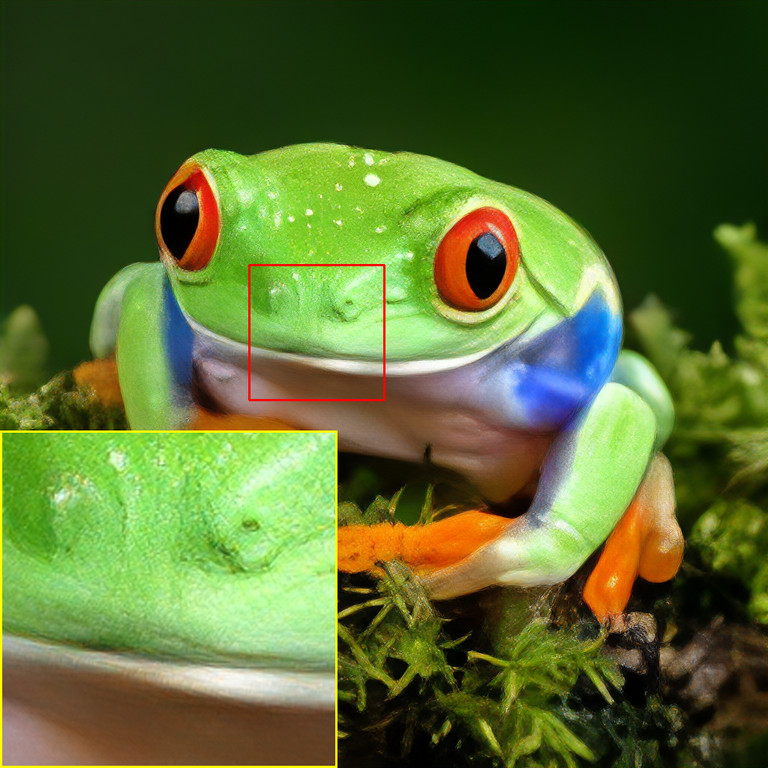} \hspace{-1.2mm} &
    \includegraphics[width=0.158\textwidth]{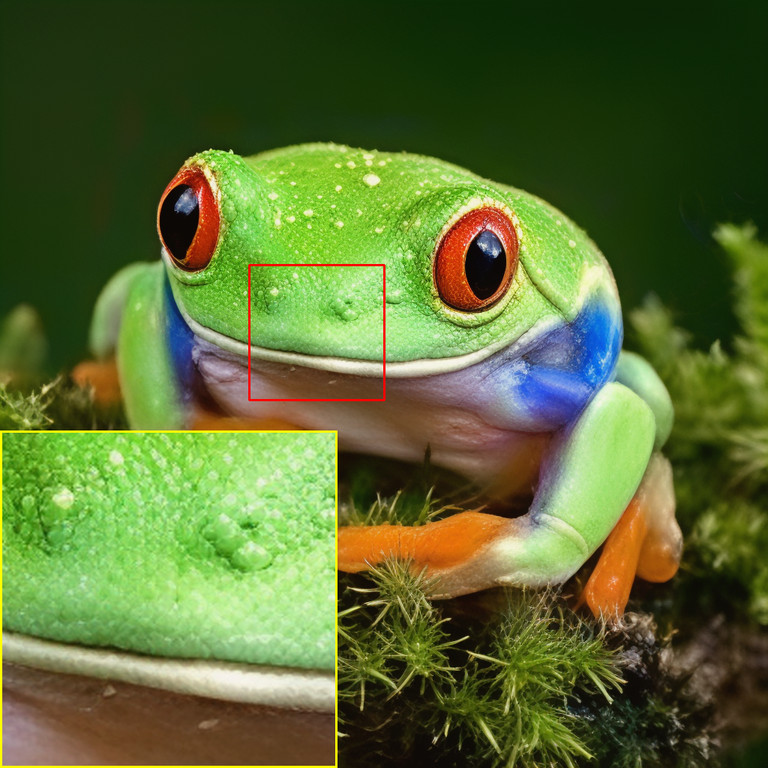} \hspace{-1.2mm} &
    \includegraphics[width=0.158\textwidth]{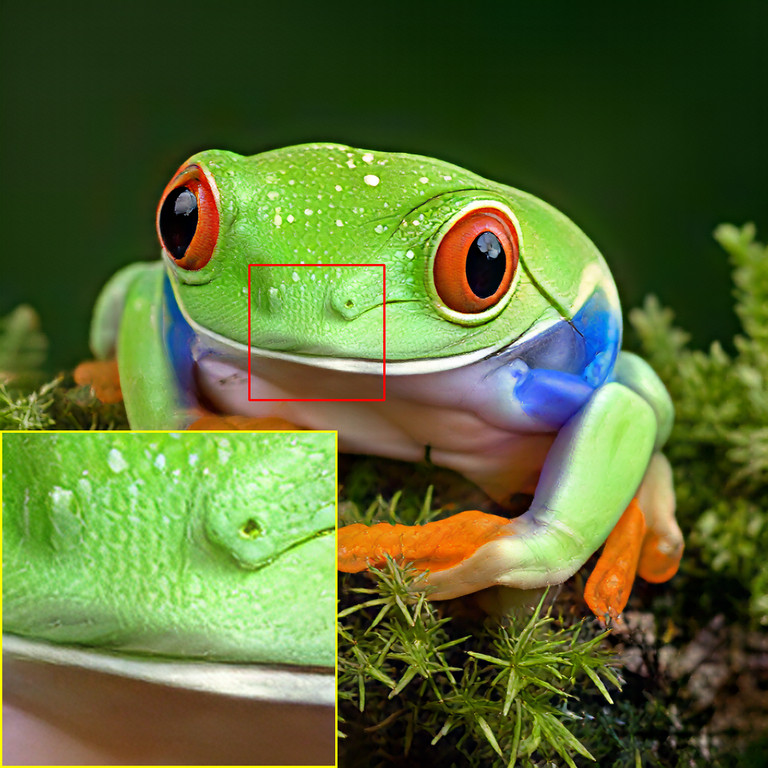}  \hspace{-1.2mm} &
    \includegraphics[width=0.158\textwidth]{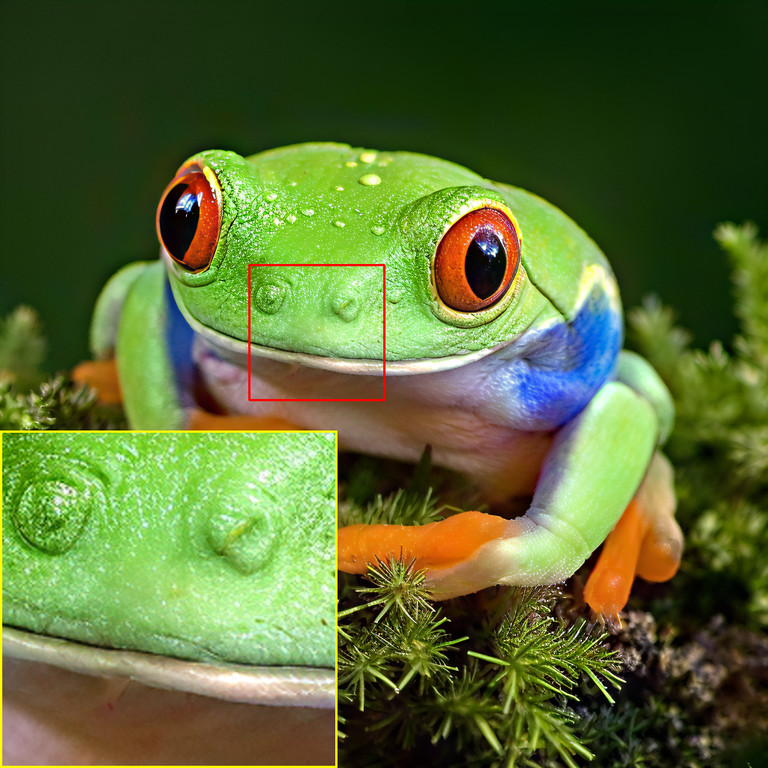} \hspace{-1.2mm} &
    \includegraphics[width=0.158\textwidth]{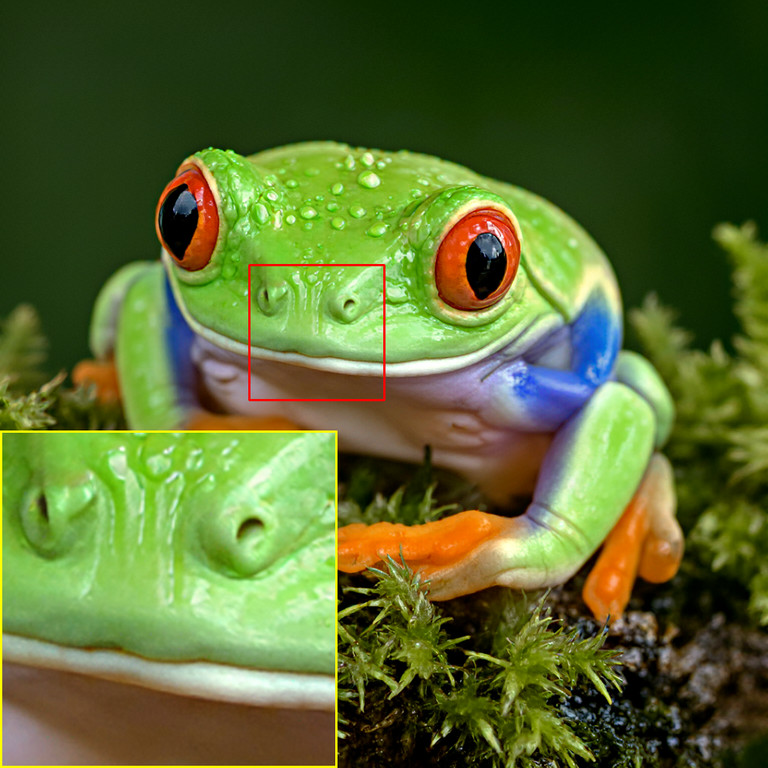} \hspace{-1.2mm}
    \\
    RealLQ250: \texttt{119} \hspace{-1.2mm} &
    CTMSR~\cite{you2025ctmsr} \hspace{-1.2mm} &
    PiSA-SR~\cite{sun2024pisasr} \hspace{-1.2mm} &
    TSD-SR~\cite{dong2025tsdsr} \hspace{-1.2mm} &
    HYPIR~\cite{lin2025hypir}  \hspace{-1.2mm} &
    StrSR (ours) \hspace{-1.2mm}
    \\
    \end{tabular}
    \end{adjustbox}
}

\renewcommand{\imgid}{00009}
\renewcommand{\imgnote}{009}
\scalebox{1.0}{
    \begin{adjustbox}{valign=t}
    \begin{tabular}{cccccc}
    \includegraphics[width=0.158\textwidth]{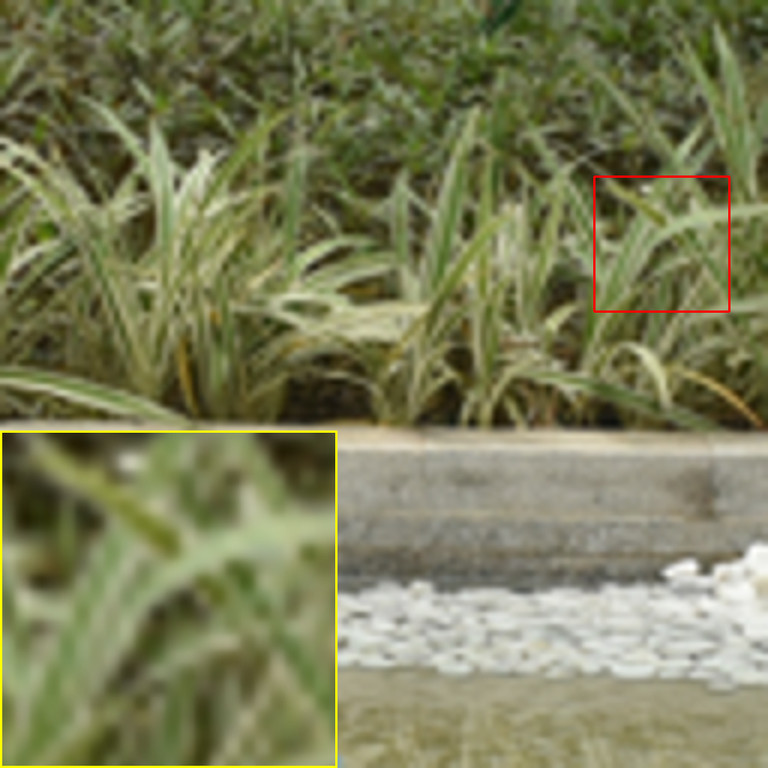} \hspace{-1.2mm} &
    \includegraphics[width=0.158\textwidth]{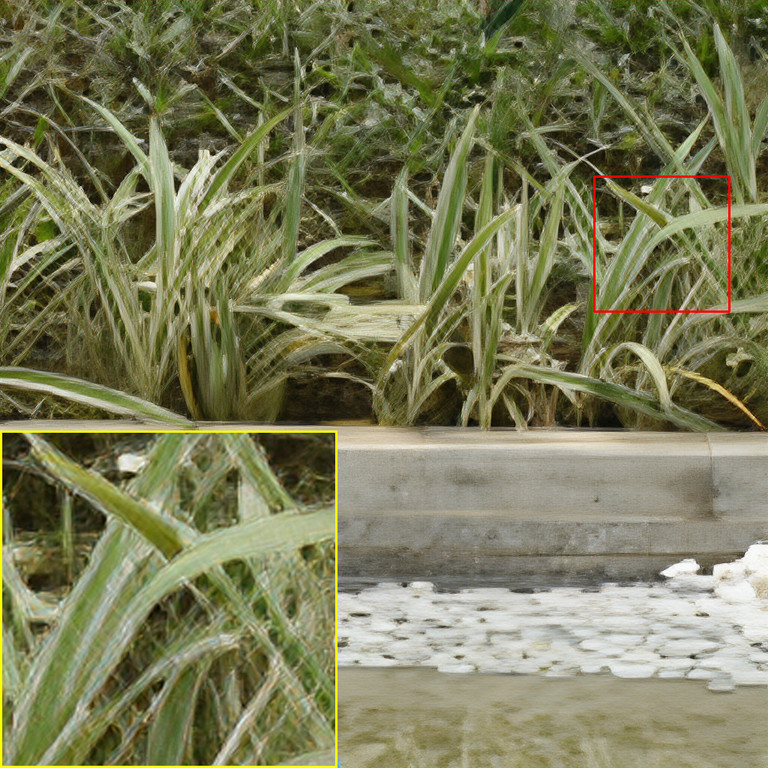} \hspace{-1.2mm} &
    \includegraphics[width=0.158\textwidth]{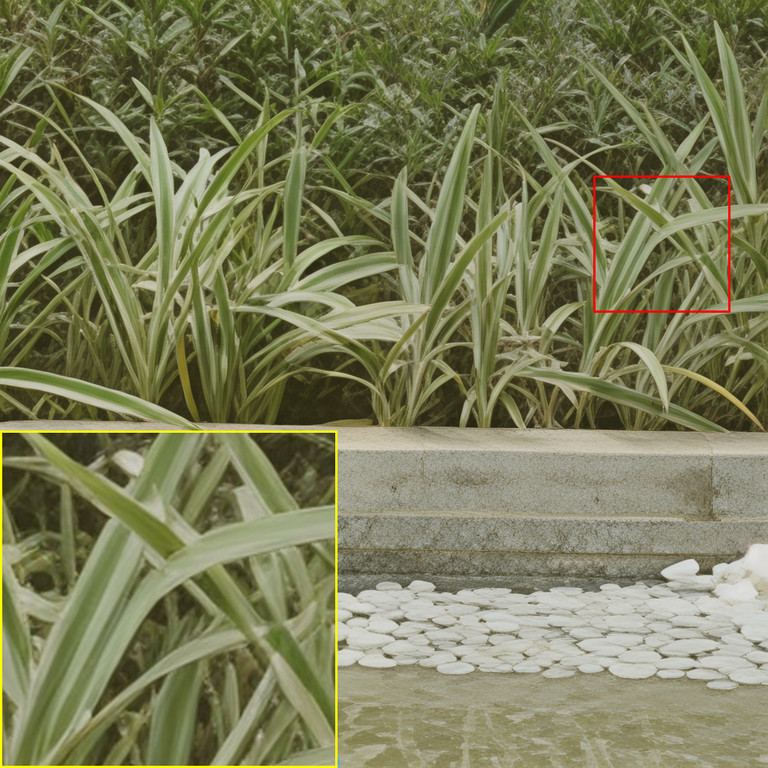} \hspace{-1.2mm} &
    \includegraphics[width=0.158\textwidth]{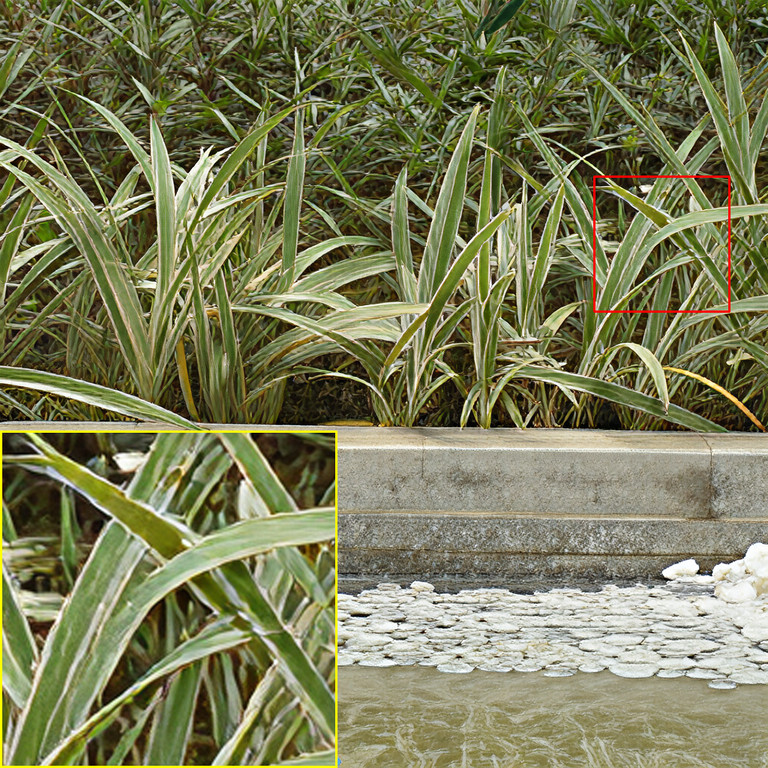} \hspace{-1.2mm} &
    \includegraphics[width=0.158\textwidth]{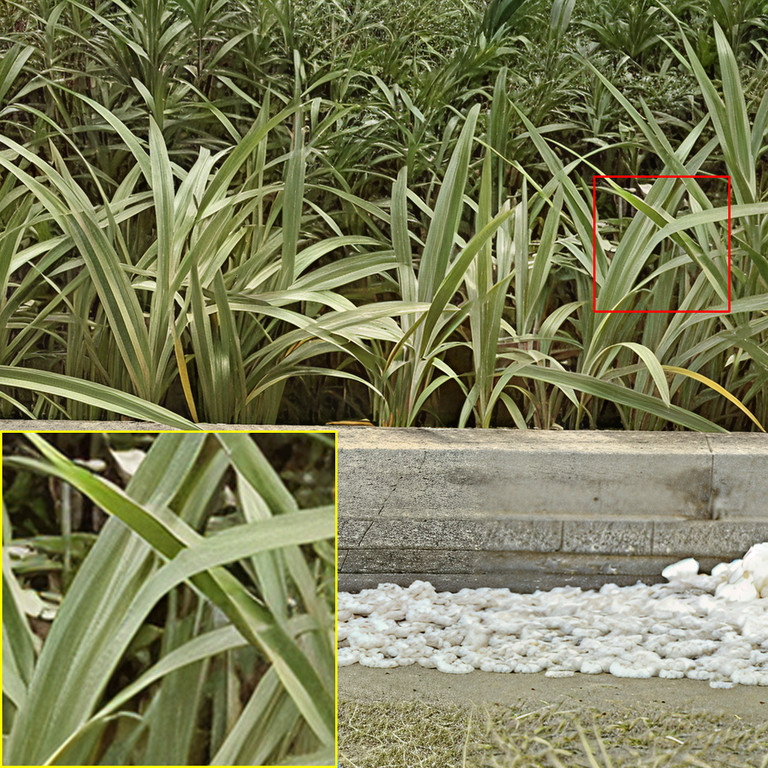} \hspace{-1.2mm} &
    \includegraphics[width=0.158\textwidth]{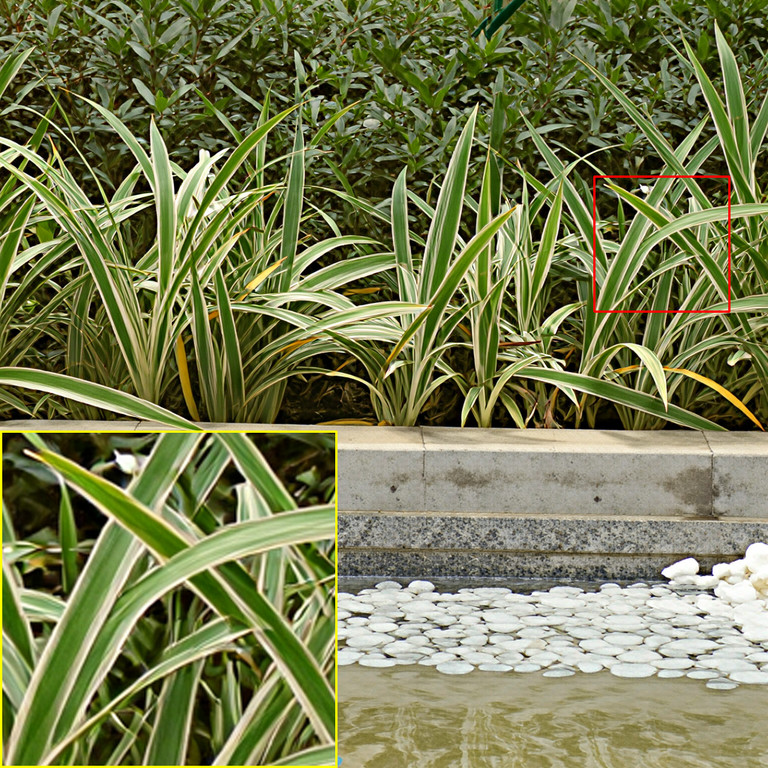} \hspace{-1.2mm}
    \\
    RealLQ250: \texttt{194} \hspace{-1.2mm} &
    CTMSR~\cite{you2025ctmsr} \hspace{-1.2mm} &
    PiSA-SR~\cite{sun2024pisasr} \hspace{-1.2mm} &
    TSD-SR~\cite{dong2025tsdsr} \hspace{-1.2mm} &
    HYPIR~\cite{lin2025hypir}  \hspace{-1.2mm} &
    StrSR (ours) \hspace{-1.2mm}
    \end{tabular}
    \end{adjustbox}
}
\renewcommand{\imgid}{00009}
\renewcommand{\imgnote}{009}
\scalebox{1.0}{
    \begin{adjustbox}{valign=t}
    \begin{tabular}{cccccc}
    \includegraphics[width=0.158\textwidth]{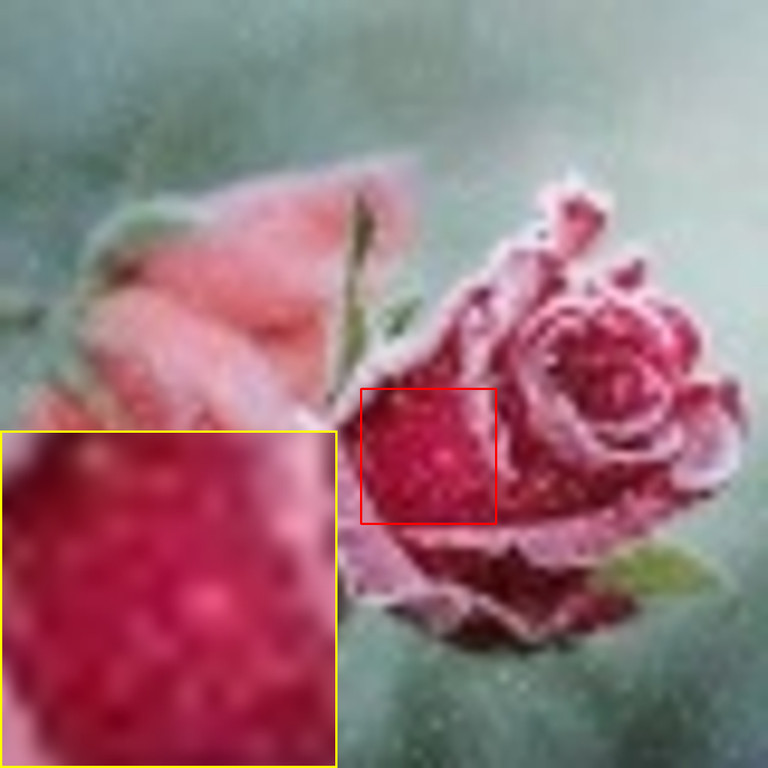} \hspace{-1.2mm} &
    \includegraphics[width=0.158\textwidth]{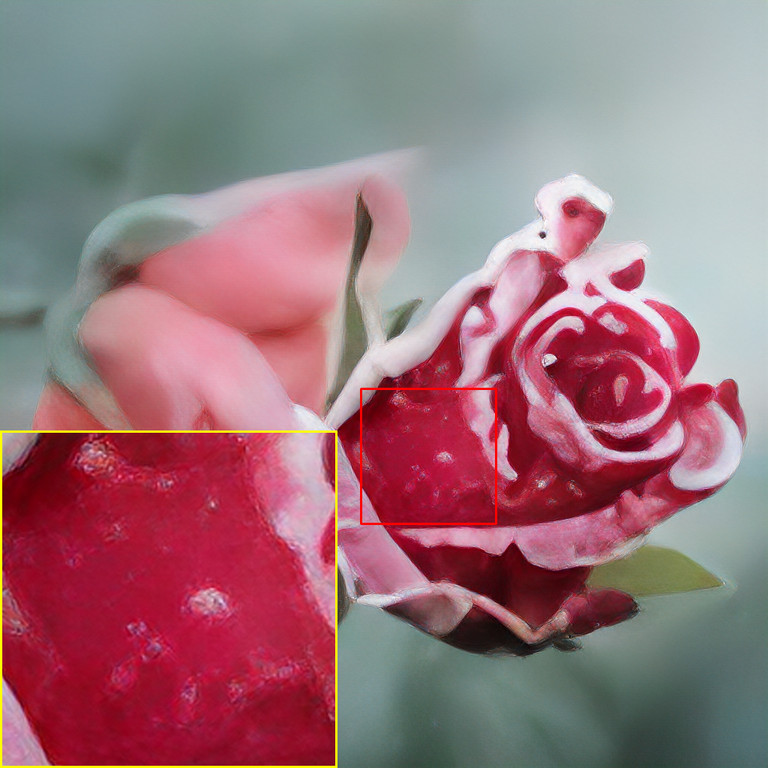} \hspace{-1.2mm} &
    \includegraphics[width=0.158\textwidth]{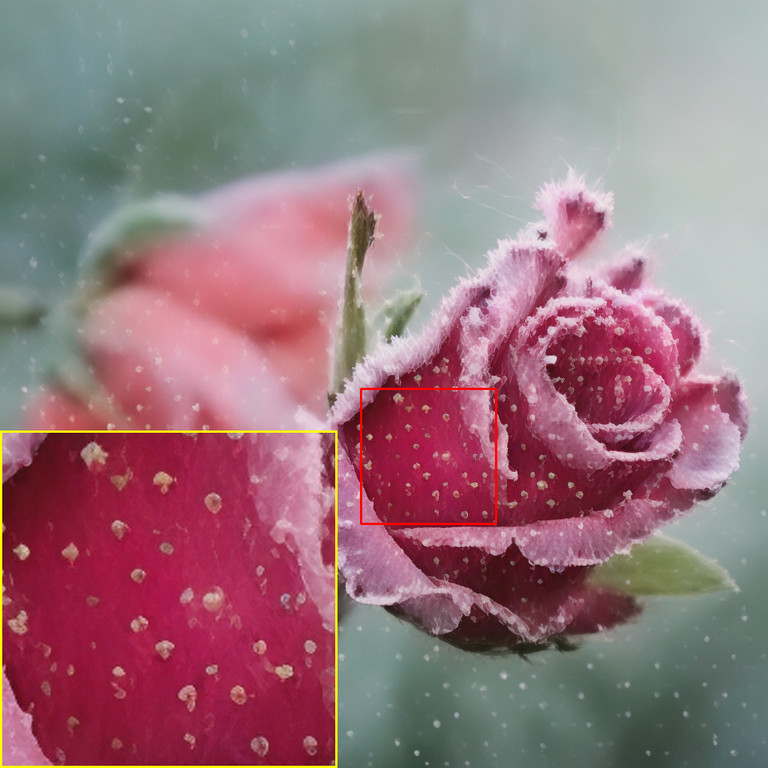} \hspace{-1.2mm} &
    \includegraphics[width=0.158\textwidth]{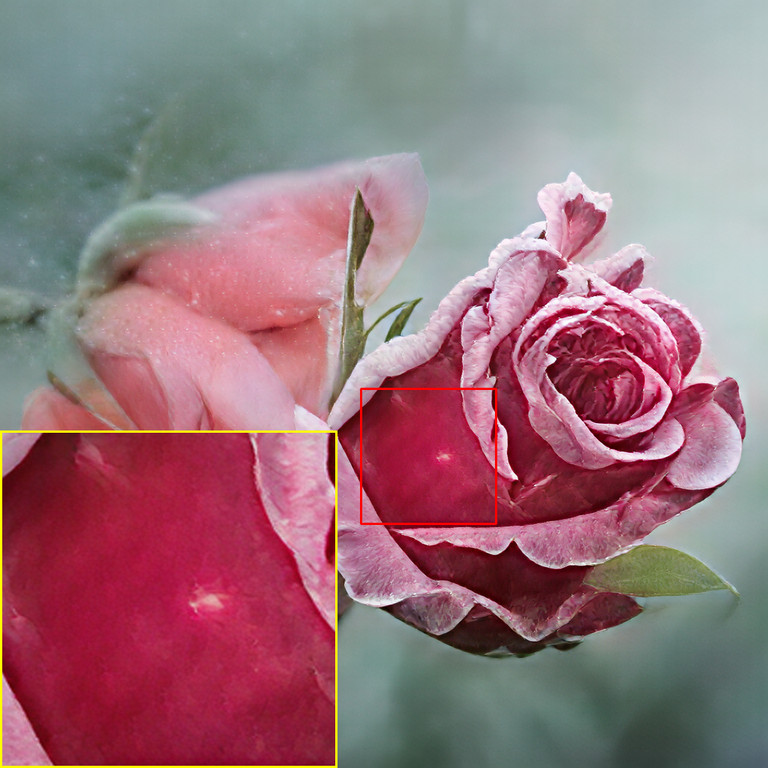} \hspace{-1.2mm} &
    \includegraphics[width=0.158\textwidth]{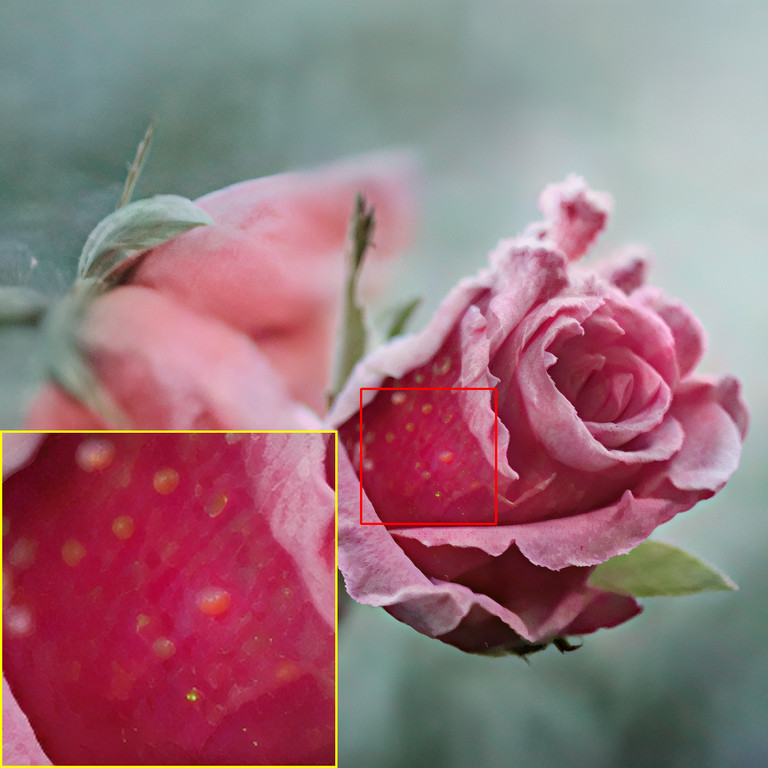} \hspace{-1.2mm} &
    \includegraphics[width=0.158\textwidth]{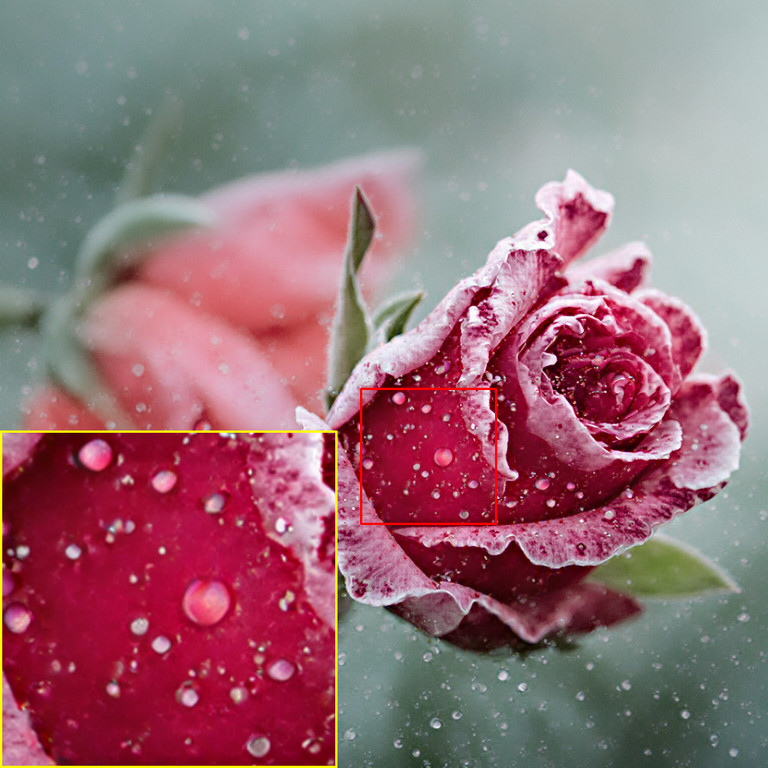} \hspace{-1.2mm}
    \\
    RealLQ250: \texttt{241} \hspace{-1.2mm} &
    CTMSR~\cite{you2025ctmsr} \hspace{-1.2mm} &
    PiSA-SR~\cite{sun2024pisasr} \hspace{-1.2mm} &
    TSD-SR~\cite{dong2025tsdsr} \hspace{-1.2mm} &
    HYPIR~\cite{lin2025hypir}  \hspace{-1.2mm} &
    StrSR (ours) \hspace{-1.2mm}
    \end{tabular}
    \end{adjustbox}
}
\caption{Visual comparison for image SR ($\times$4) in RealLQ250 dataset. The first column is the bicubic algorithm. Please zoom in for a better view. }
\label{fig:vis-reallq250}
\end{figure*}

\begin{figure*}[t]
\scriptsize
\centering

\scalebox{0.99}{
    \begin{adjustbox}{valign=t}
    \begin{tabular}{@{}c@{}} 
        \includegraphics[width=0.98\textwidth]{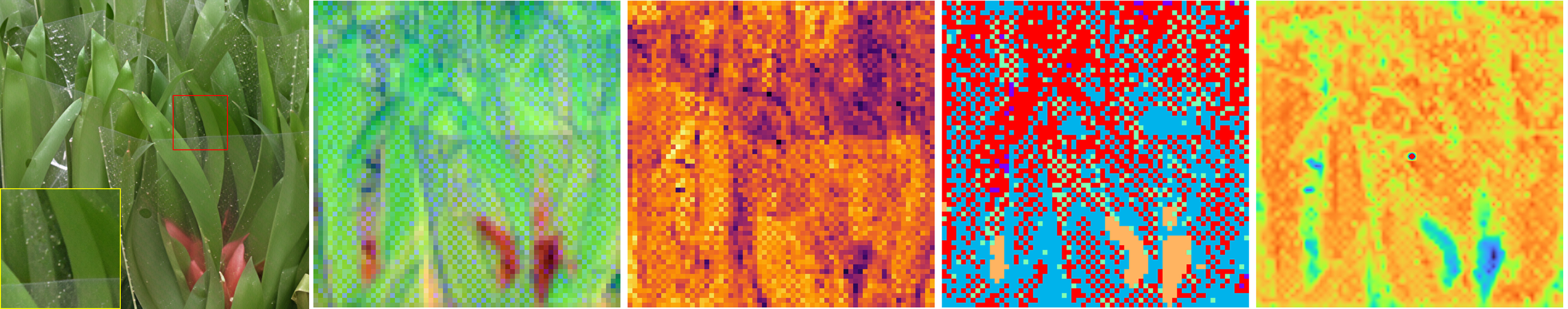} \\
        \vspace{-2.7mm} \\ 
        \makebox[0.196\textwidth]{w/o FDL}%
        \makebox[0.196\textwidth]{PCA}%
        \makebox[0.196\textwidth]{Norm}%
        \makebox[0.196\textwidth]{Cluster}%
        \makebox[0.196\textwidth]{Heatmap}
    \end{tabular}
    \end{adjustbox}
}
\scalebox{0.99}{
    \begin{adjustbox}{valign=t}
    \begin{tabular}{@{}c@{}}
        \includegraphics[width=0.98\textwidth]{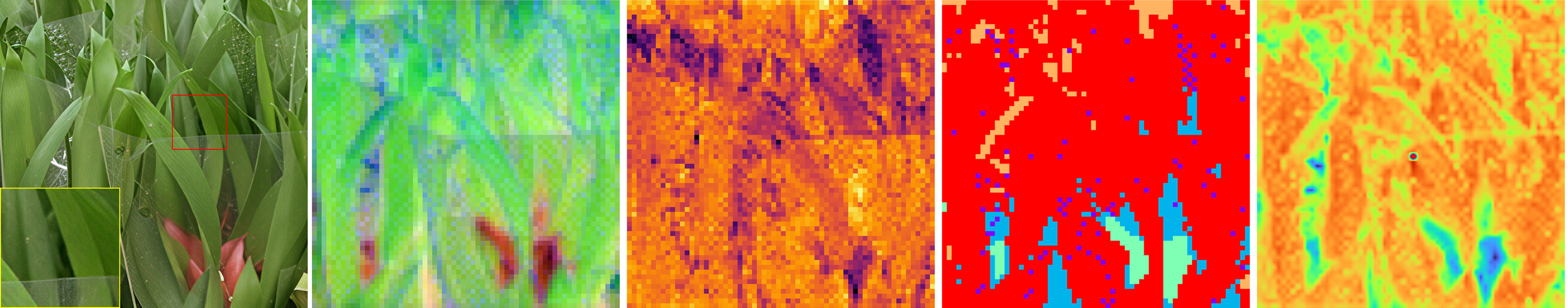} \\
        \vspace{-2.7mm} \\
        \makebox[0.196\textwidth]{w/ FDL}%
        \makebox[0.196\textwidth]{PCA}%
        \makebox[0.196\textwidth]{Norm}%
        \makebox[0.196\textwidth]{Cluster}%
        \makebox[0.196\textwidth]{Heatmap}
    \end{tabular}
    \end{adjustbox}
}

\caption{Visual comparison for extracted feature maps. We extract the features of the generator's last layer for demonstration, including four widely-used visualization methods. Our FDL loss effectively alleviates artifacts in latent level.}
\label{fig:feature}
\end{figure*}

\subsection{Comparison and Analysis}
\noindent\textbf{Compared Methods} We compare our method with several state-of-the-art diffusion-based image super-resolution methods. For those representative multi-step diffusion methods, we compare with SUPIR~\cite{yu2024supir}, DIT4SR~\cite{duan2025dit4sr}, DiffBIR~\cite{lin2024diffbir}, PASD~\cite{yang2024pasd}, InstructRestore~\cite{liu2025instructrestore} and SeeSR~\cite{wu2024seesr}. 
As for one-step diffusion methods, our comparison includes CTMSR~\cite{you2025ctmsr}, PiSA-SR~\cite{sun2024pisasr}, TSD-SR~\cite{dong2025tsdsr}, OSEDiff~\cite{wu2024osediff}, InvSR~\cite{yue2025invsr}, HYPIR~\cite{lin2025hypir}, and SinSR~\cite{wang2024sinsr}. 

\noindent\textbf{Quantitative Results} As shown in Tabs.~\ref{tab:metric_DIV2K}, \ref{tab:metric_RealSR}, and \ref{tab:metric_RealLQ250}, our methods achieve excellent overall performance. We compare StrSR on two different base models. The best and second-best results are colored in \textcolor{red}{red} and \textcolor{eccvblue}{blue}, respectively. 

Regarding the perceptual metrics LPIPS and DISTS, our two models achieve state-of-the-art (SOTA) results across all three datasets when compared with all one-step methods. Furthermore, on the DIV2K dataset, our models even outperform all multi-step diffusion methods. 
For non-reference metrics, our models also show strong competitiveness. Specifically, among all one-step methods, our FLUX model achieves the best scores in NIQE and MANIQA across the datasets. This demonstrates its ability to generate highly realistic images. Meanwhile, our Z-Image model performs exceptionally well in aesthetic and visual alignment metrics. It consistently ranks in the top two for MUSIQ and QAlign. 

These comparisons indicate that our proposed StrSR can capture more realistic local details and global structural information compared with previous methods. It significantly improves the overall perceptual and aesthetic quality while maintaining efficient one-step generation.

\noindent\textbf{Qualitative Results} Visual comparisons in Figs.~\ref{fig:vis-DIV2K}, \ref{fig:vis-RealSR}, and \ref{fig:vis-reallq250} show that StrSR yields superior visual results. We use FLUX-based StrSR for visual comparisons.

For the synthetic DIV2K-val dataset, StrSR correctly recognizes high-level semantics to generate accurate texture categories. Furthermore, it produces natural and photo-realistic fine-grained details on dense high-frequency textures, such as brick walls, scales, and feathers. Additionally, while both StrSR and TSD-SR are DiT-based, StrSR avoids generating repetitive dot-like artifacts in dense high-frequency regions (e.g., basket edges) that persist in TSD-SR.

StrSR also performs excellently on the real-world datasets RealSR and RealLQ250. It segments clothing edges accurately, as seen in \texttt{Canon\_039} from RealSR. 
Given a large gap between the LR input and HR image (\eg, \texttt{Nikon\_020} in RealSR), StrSR leverages semantic information to generate realistic belt textures. In contrast, other methods simply amplify the triangular artifacts present in the LR image. Finally, in image \texttt{241} from the RealLQ250 dataset, only our method successfully restores the semantic information of the dewdrops. Other methods fail: PiSA-SR generates meaningless white dots, while CTMSR and HYPIR treat the dewdrops as noise and completely smooth them out.

\subsection{Ablation Study}

\begin{figure*}[t]
\scriptsize
\centering
\scalebox{0.98}{
\begin{tabular}{cccc}

\begin{adjustbox}{valign=t}
\begin{tabular}{ccc}
\includegraphics[width=0.267\textwidth]{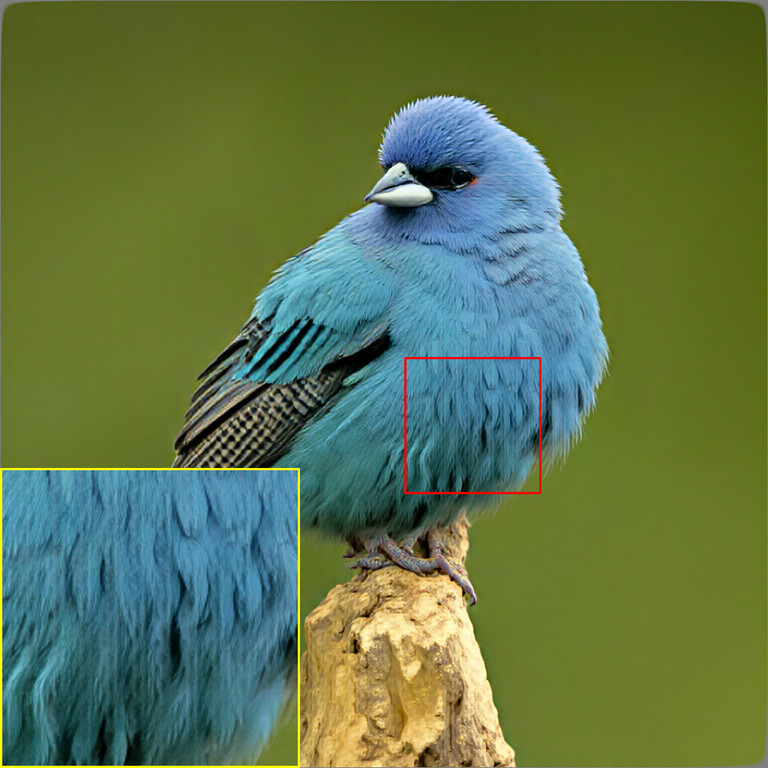} \hspace{-1mm} &
\includegraphics[width=0.267\textwidth]{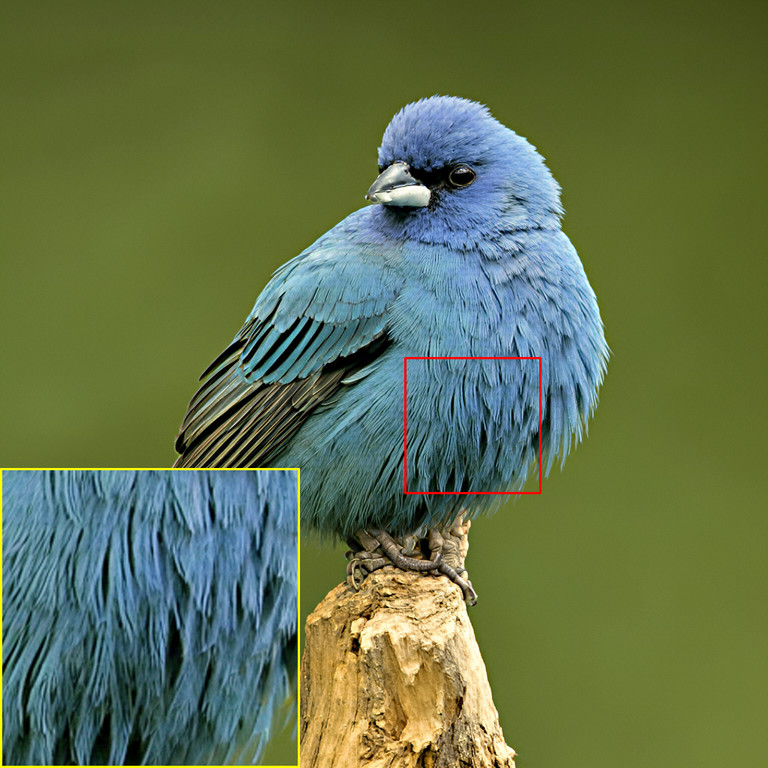} \hspace{-1mm} &
\\
w/o FDL \hspace{-1.2mm} &
w/ FDL \hspace{-1.2mm} &
\end{tabular}
\end{adjustbox}

\begin{adjustbox}{valign=t}
    \begin{tabular}{cccccc}
    \includegraphics[width=0.118\textwidth]{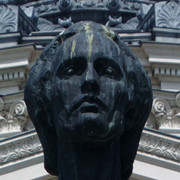} \hspace{-1.2mm} &
    \includegraphics[width=0.118\textwidth]{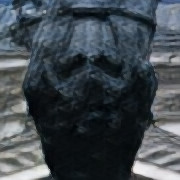}  \hspace{-1.2mm} & 
    \includegraphics[width=0.118\textwidth]{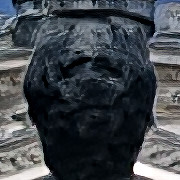} \hspace{-1.2mm}
    \\
    HR \hspace{-1.2mm} &
    w/o GAN \hspace{-1.2mm} &
    w/o Ra \hspace{-1.2mm}
    \\
    \includegraphics[width=0.118\textwidth]{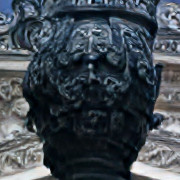} \hspace{-1.2mm} &
    \includegraphics[width=0.118\textwidth]{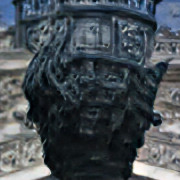} \hspace{-1.2mm} &
    \includegraphics[width=0.118\textwidth]{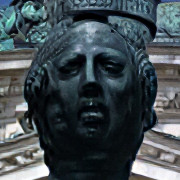} \hspace{-1.2mm}
    \\
    
    w/o VLM \hspace{-1.2mm} &
    w/o emb \hspace{-1.2mm} &
    w/ All \hspace{-1.2mm}
    \\
    \end{tabular}
    \end{adjustbox}

\end{tabular}
}
\caption{(\textit{Left})~Comparison of image textures with and without frequency distribution matching. FDL guides the model to generate more photo-realistic fine details. (\textit{Right})~Visual comparison of the ablation results for the GAN and VLM components. StrSR accurately captures both semantic information and texture details.}
\label{fig:vis-ablation}
\end{figure*}

\noindent\textbf{Frequency Distribution Matching.} We extract features from the final layer of the DiT to visualize the impact of the FDL. As shown in Fig.~\ref{fig:feature}, FDL effectively reduces the similarity between adjacent tokens, which suppresses the grid-like artifacts in the feature maps and promotes more natural details. Additionally, visual results in Fig.~\ref{fig:vis-ablation} demonstrate the effectiveness of frequency distribution matching in handling challenging high-frequency textures. These results indicate that FDL effectively mitigates the periodic artifacts common in DiT-based super-resolution models, enabling the generation of realistic fine details.

\vspace{2mm}

\begin{figure}[t]
    \centering
    \begin{minipage}[c]{0.36\textwidth}
        \centering
        \scriptsize
        \begin{tabular}{lccc}
            \toprule
            \rowcolor{color3} Methods &  MAN. & MUS. & QAli. \\
            \midrule
            w/o GAN   & 0.4720 & 54.086 & 4.1629 \\
            w/o Ra    & 0.4930 & 61.820 & 4.4570 \\
            w/o emb   & 0.5043 & 58.868 & 4.1291 \\
            w/o VLM   & 0.5300 & 62.093 & 4.3435 \\
            w/ All    & \textbf{0.5754} & \textbf{65.106} & \textbf{4.4779} \\
            \bottomrule
        \end{tabular}
        \captionof{table}{Ablation study of the GAN and VLM modules on DIV2K-val. Our full model achieves the best scores, restoring semantically faithful and photo-realistic details.}
        \label{tab:ablation}
    \end{minipage}\hfill
    \begin{minipage}[c]{0.62\textwidth}
        \centering
        \includegraphics[width=\linewidth]{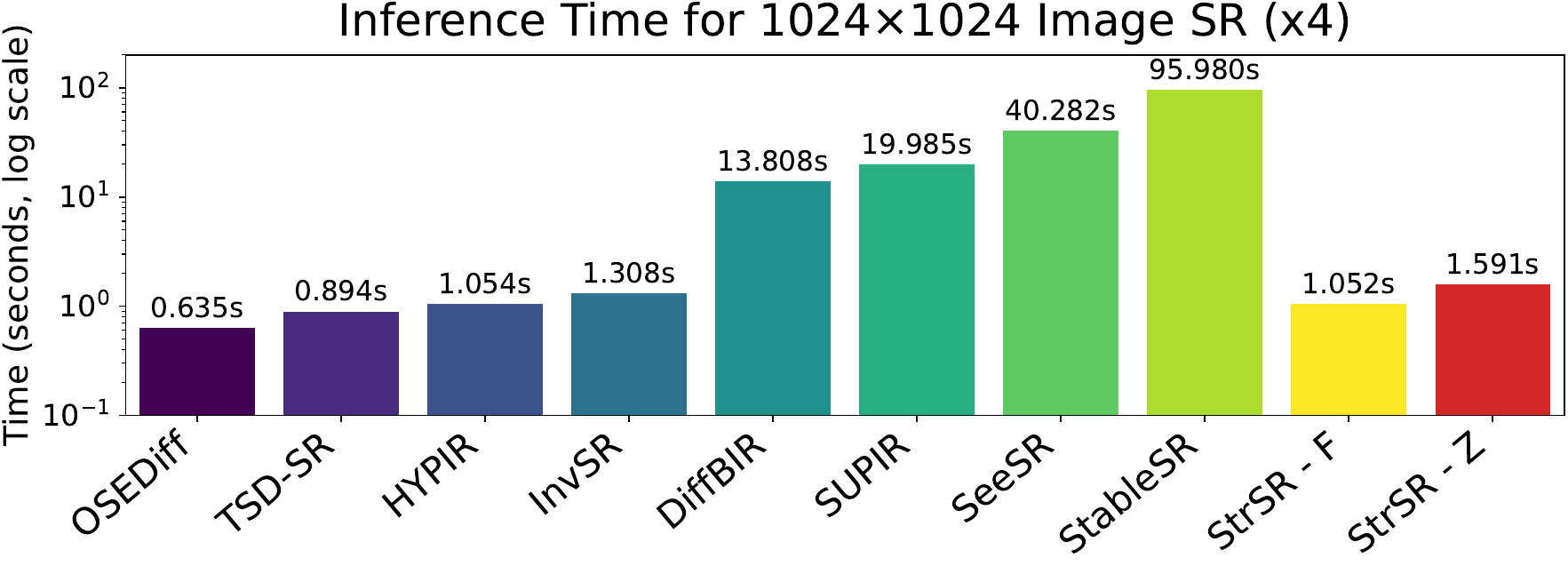}
        \vspace{-6.6mm}
        \captionof{figure}{Comparison of inference times for image super-resolution ($\times$4) across various methods. The vertical axis is displayed on a logarithmic scale. Our StrSR, although utilizing a significantly larger backbone, achieves comparable inference speeds. }
        \label{fig:speed}
    \end{minipage}
    
    \vspace{-2mm}
\end{figure}

\noindent\textbf{Asymmetric Discriminative Distillation} We evaluate the effects of various GAN settings in our model's training pipeline, including the type of GAN we used (RaGAN, non-saturated GAN, and without GAN). As shown in Tab.~\ref{tab:ablation}, our method with RaGAN outperforms other ablation settings on multiple aesthetic and semantic metrics. What's more, Fig.~\ref{fig:vis-ablation} illustrates the visual results of different training methods, demonstrating better convergence and generative quality reached with our final GAN setting. In other words, the use of RaGAN relativistic GAN loss guarantees more stable gradient flow for the generator to learn the target data distribution in the diffusion process.

\vspace{2mm}

\noindent\textbf{VLM Dual Encoder.} Specifically, we evaluate the model under three conditions: without any text embedding (w/o emb), with a naive text encoder (w/o VLM), and with our VLM-encoded embeddings (w/ VLM). As reported in Tab.~\ref{tab:ablation}, integrating the VLM yields substantial improvements across all perceptual metrics, including a notable increase in MUSIQ from 62.093 to 65.106. As supported by the visual evidence in Fig.~\ref{fig:vis-ablation}, without accurate semantic conditioning (w/o emb and w/o VLM), the model struggles to comprehend complex structures, leading to severe distortions. It is evident that the VLM effectively translates LR inputs into rich semantic priors, guiding the model to generate more semantically faithful details.

\subsection{Running Time}
We further evaluate the inference speed for 1,024$\times$1,024 input and compare it with various SR methods on an NVIDIA RTX A6000 GPU, as shown in Fig.~\ref{fig:speed}. The results show that one-step distillation significantly improves inference speed. Remarkably, compared to other one-step models, StrSR maintains a similar speed despite using a much larger model (4B for FLUX.2 or 6B for Z-Image).

\subsection{Limitations}
One-step distillation significantly reduces the sampling steps compared to multi-step diffusion methods. However, our base models, Z-Image-Turbo and FLUX.2 [klein], contain far more parameters than widely used Stable Diffusion models (<1B), which limits their practicality for on-device deployment. Although StrSR achieves strong perceptual quality, its PSNR/SSIM scores, \ie, pixel-level fidelity, still leave room for improvement. In addition, our current pipeline is less effective in text-rich regions, as it does not explicitly take text as input.

\subsection{Conclusion}
In this paper, we propose StrSR, a novel one-step adversarial distillation framework for Real-ISR. We observe that applying existing one-step distillation methods directly to DiT causes severe grid-like artifacts. This issue primarily arises from generation trajectory mismatch and high-frequency spectral leakage. To address these problems, we introduce an asymmetric discriminative distillation architecture. By employing a pretrained CLIP-ConvNeXt as the discriminator, our method ensures stable training and accurate texture recovery. Furthermore, we design frequency distribution matching to map the spectral domain, which effectively suppresses the periodic artifacts. By combining these designs in a dual-encoder architecture, StrSR achieves SOTA performance in extensive experiments. Our method successfully unlocks the generative potential of DiT architectures, enabling highly efficient and photo-realistic Real-ISR in one inference step.

\vspace{-2mm}
\section*{Acknowledgements}
\vspace{-2mm}

This work is supported by the National Natural Science Foundation of China (62501386), CCF-Tencent Rhino-Bird Open Research Fund, and CAAI-Tencent Rhino-Bird Open Research Fund. This work is also sponsored by AI Hundred Schools Program and is carried out using the Ascend AI technology stack. This work is supported in part by NSFC grant 62302299 and Huawei Explore X funds. This work is also supported by SGLabZZKT2025-09, and Oceanic Interdisciplinary Program of Shanghai Jiao Tong University, grant number SL2023ZD203.

\bibliographystyle{splncs04}
\bibliography{main}
\end{document}


\title{\textit{Supplementary Materials:}\\{Spectral and Trajectory Regularization for Diffusion Transformer Super-Resolution}}

\titlerunning{Spectral and Trajectory Regularization for DiT SR}

\author{Jingkai Wang\inst{1}\thanks{Equal contribution.}\orcidlink{0009-0002-1278-4642} \and
Yixin Tang\inst{1}\protect\footnotemark[1]\orcidlink{0009-0005-4121-5782} \and
Jue Gong\inst{1}\orcidlink{0009-0008-4131-8951} \and
Jiatong Li\inst{1}\orcidlink{0009-0001-8822-0037} \and
Shu Li\inst{2} \and
Libo Liu\inst{2} \and
Jianliang Lan\inst{2} \and
Yutong Liu\inst{1}\thanks{Corresponding authors: Yutong Liu and Yulun Zhang.}\orcidlink{0000-0001-9386-3371} \and
Yulun Zhang\inst{1}\protect\footnotemark[2]\orcidlink{0000-0002-2288-5079}}

\authorrunning{J.~Wang et al.}

\institute{Shanghai Jiao Tong University, China \and Shenzhen Transsion Holdings Co., Ltd., China}

\maketitle
\appendix
\vspace{-2mm}
\section{Overview}
\label{sec:overview}
\vspace{-1mm}
This supplementary material provides further experimental validation, more model analysis, and extensive qualitative results to support the findings in the main paper. First in Sec.~\ref{sec:flux2}, we demonstrate the superiority of our training pipeline by comparing StrSR with HYPIR~\cite{lin2025hypir} adapted to FLUX.2 base model. Next, we present a deeper analysis detailing the two-stage training strategy, additional ablations on FLUX.2 base model, and LoRA hyperparameter configurations in Sec.~\ref{sec:abl}. We further explore how NFE impact DiT artifacts in Sec.~\ref{sec:gflr}. We also extend our evaluations to include comparisons with state-of-the-art video super-resolution methods in Sec.~\ref{sec:seedvr2} and classic non-diffusion-based models in Sec.~\ref{sec:classic}. Finally, in Sec.~\ref{sec:vis}, we provide extensive visual comparisons on both synthetic and real-world datasets, highlighting StrSR's strong capabilities in semantic recovery, realistic texture generation, and color fidelity.

\begin{table*}[t]
\footnotesize
\centering
\begin{tabular}{c|cc|ccccc}
\toprule
\rowcolor{color3} Methods & LPIPS$\downarrow$ & DISTS$\downarrow$ & NIQE$\downarrow$ & MAN.$\uparrow$ & MUS.$\uparrow$ & QAli.$\uparrow$ & Qual.$\uparrow$ \\
\midrule
HYPIR~\cite{lin2025hypir} & 0.3551 & 0.1457 & 3.3566 & 0.5673 & 63.6403 & 4.3349 & 0.7299 \\
HYPIR~\cite{lin2025hypir} - FLUX & \rr{0.2966} & 0.1318 & 3.5287 & 0.5533 & 63.3139 & 4.1433 & 0.6745 \\
StrSR - FLUX (ours) & 0.2992 & \rr{0.1288} & \rr{3.0379} & \rr{0.5864} & \rr{63.6568} & \rr{4.4763} & \rr{0.7022} \\
\bottomrule
\end{tabular}
\vspace{2mm}
\scriptsize
\caption{Quantitative results of FLUX.2 based HYPIR and our method on DIV2K-val.  MAN., MUS., QAli., and Qual. stand for MANIQA, MUSIQ, QAlign, and QualiCLIP.}
\vspace{-4mm}
\label{tab:hypir-flux2}
\end{table*}

\begin{figure}[t]
    \centering
    \begin{minipage}[c]{0.4\textwidth}
        \centering
        \scriptsize
        \begin{tabular}{lccc}
            \toprule
            \rowcolor{color3} Methods &  MAN.$\uparrow$ & MUS.$\uparrow$ & QAli.$\uparrow$ \\
            \midrule
            w/o GAN   & 0.5176 & 56.7725 & 4.0288 \\
            w/o Ra    & 0.5711 & 62.0717 & 4.2555 \\
            w/o emb   & 0.5596 & 61.5075 & 4.2280 \\
            w/o VLM   & 0.5584 & 62.0344 & 4.2506 \\
            w/ All    & \rr{0.5865} & \rr{63.8108} & \rr{4.2883} \\
            \bottomrule
        \end{tabular}
        \captionof{table}{Ablation study of the GAN and VLM modules on DIV2K-val with our FLUX.2-based model, demonstrating the effectiveness of our method on different backbones.}
        \label{tab:ablation-flux}
    \end{minipage}\hfill
    \begin{minipage}[c]{0.55\textwidth}
        \centering
        \footnotesize
        \begin{tabular}{cc|cccc}
            \toprule
            \rowcolor{color3} Rank & Alpha\, &  MAN.$\uparrow$ & MUS.$\uparrow$ & Qual.$\uparrow$ & NIQE$\downarrow$ \\
            \midrule
            64 & 64   & 0.5585 & 60.7610 & 0.6654 & 3.6043 \\
            128 & 128    & \rr{0.5929} & 62.9313 & 0.7012 & 3.3484 \\
            256 & 256   & 0.5865 & \rr{63.8108} & \rr{0.7086} & \rr{3.2350} \\
            \bottomrule
        \end{tabular}
        \vspace{1.4mm}
        \captionof{table}{Ablation study of LoRA rank and alpha on DIV2K-val. Conventionally, rank and alpha values are kept equal. Higher ranks provide more trainable parameters, which generally leads to better performance across most metrics.}
        \label{tab:ablation-lora}
    \end{minipage}
    
    \vspace{-5mm}
\end{figure}

\vspace{-2mm}
\section{Performance of HYPIR using the FLUX.2 Base Model}
\label{sec:flux2}
\vspace{-1mm}
To verify the efficiency of our proposed training pipeline and isolate the impact of the base model, we retrain HYPIR~\cite{lin2025hypir}. We strictly follow its open-source training code but replace the stable diffusion 2.1~\cite{rombach2022ldm} base model with FLUX.2 [klein] 4B~\cite{FLUX.2-klein-4B}. The retraining process uses the same training data and iteration steps as StrSR. StrSR, the official HYPIR, and the retrained HYPIR models are all trained using LoRA with a rank and alpha of 256. As shown in {Tab.~\ref{tab:hypir-flux2}}, StrSR significantly outperforms the retrained HYPIR across several metrics.  This demonstrates that the training pipeline proposed in StrSR is more advantageous than HYPIR for tasks based on FLUX.2, a representative diffusion transformer base model.

\begin{figure}[t]
    \centering
    \begin{subfigure}[b]{0.49\textwidth}
        \centering
        \includegraphics[width=\textwidth]{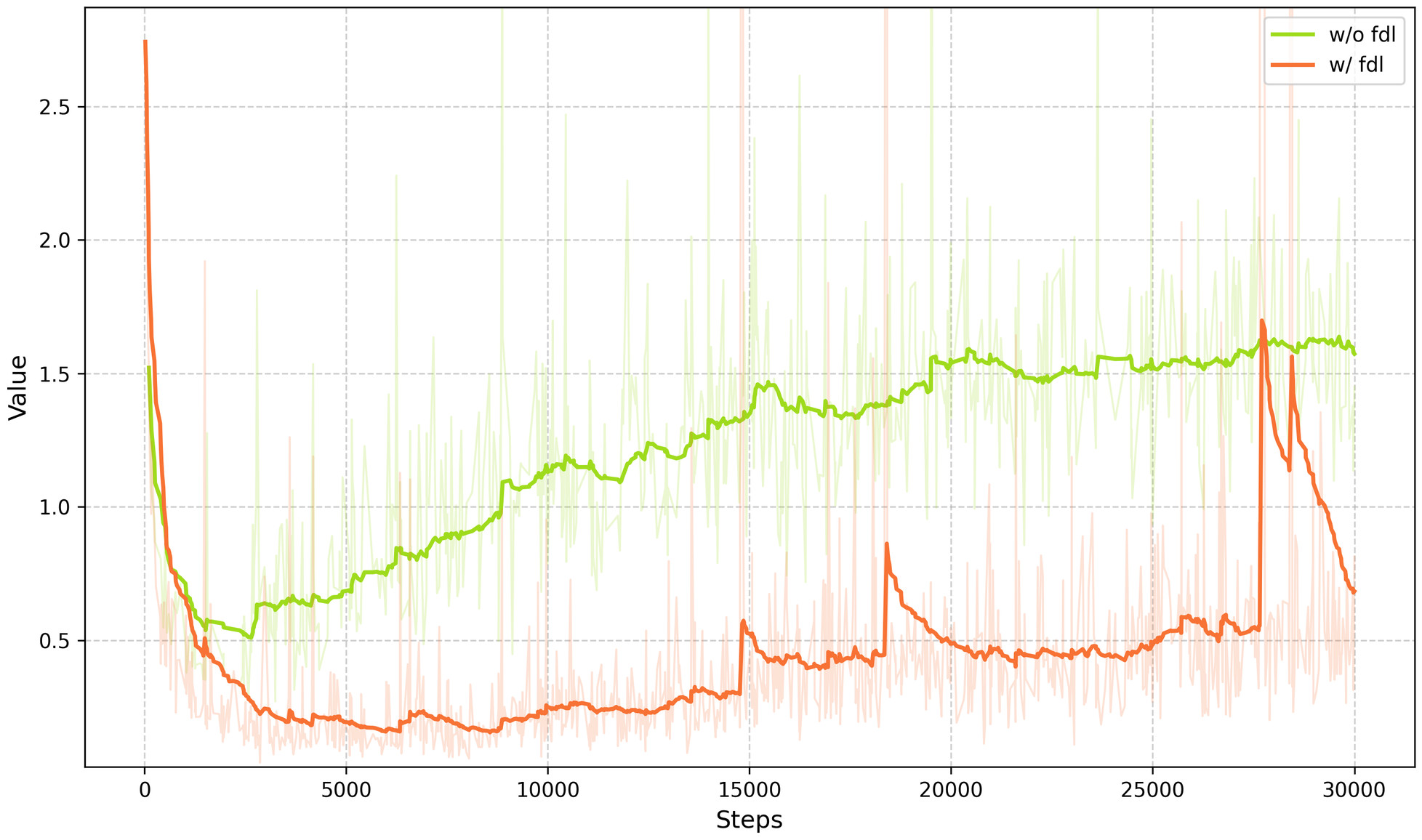}
        \caption{GAN discriminator loss}
        \label{fig:disc_loss}
    \end{subfigure}
    \hfill 
    \begin{subfigure}[b]{0.49\textwidth}
        \centering
        \includegraphics[width=\textwidth]{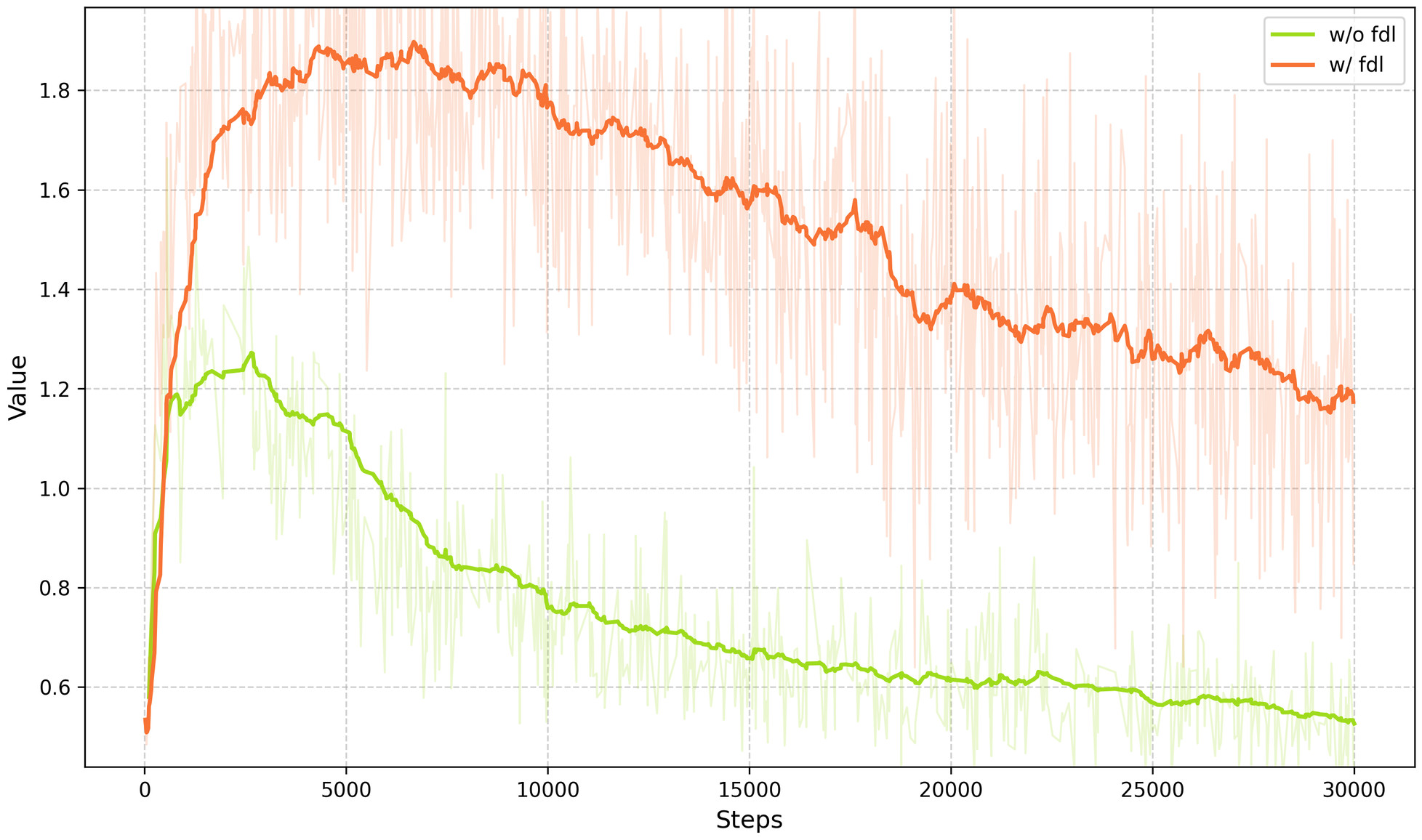} 
        \caption{GAN generator Loss}
        \label{fig:gen_loss}
    \end{subfigure}
    
    \caption{Generator and discriminator loss with and without FDL in first 30,000 iterations.}
    \label{fig:GAN_loss_curves}
\end{figure}

\begin{figure}[t]
    \centering
    \begin{subfigure}[b]{0.49\textwidth}
        \centering
        \includegraphics[width=\textwidth]{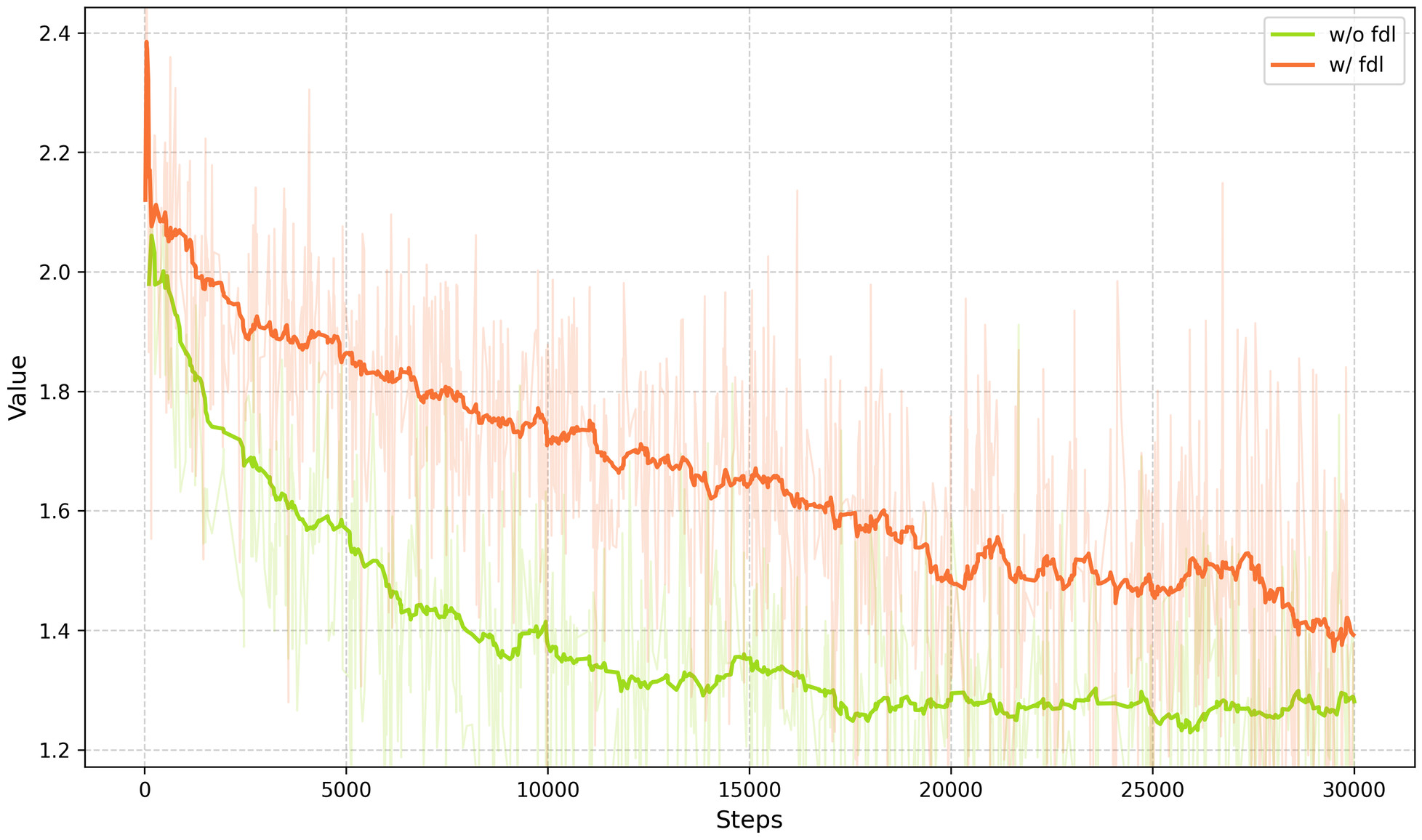}
        \caption{LPIPS loss of generator}
        \label{fig:lpips_loss}
    \end{subfigure}
    \hfill 
    \begin{subfigure}[b]{0.49\textwidth}
        \centering
        \includegraphics[width=\textwidth]{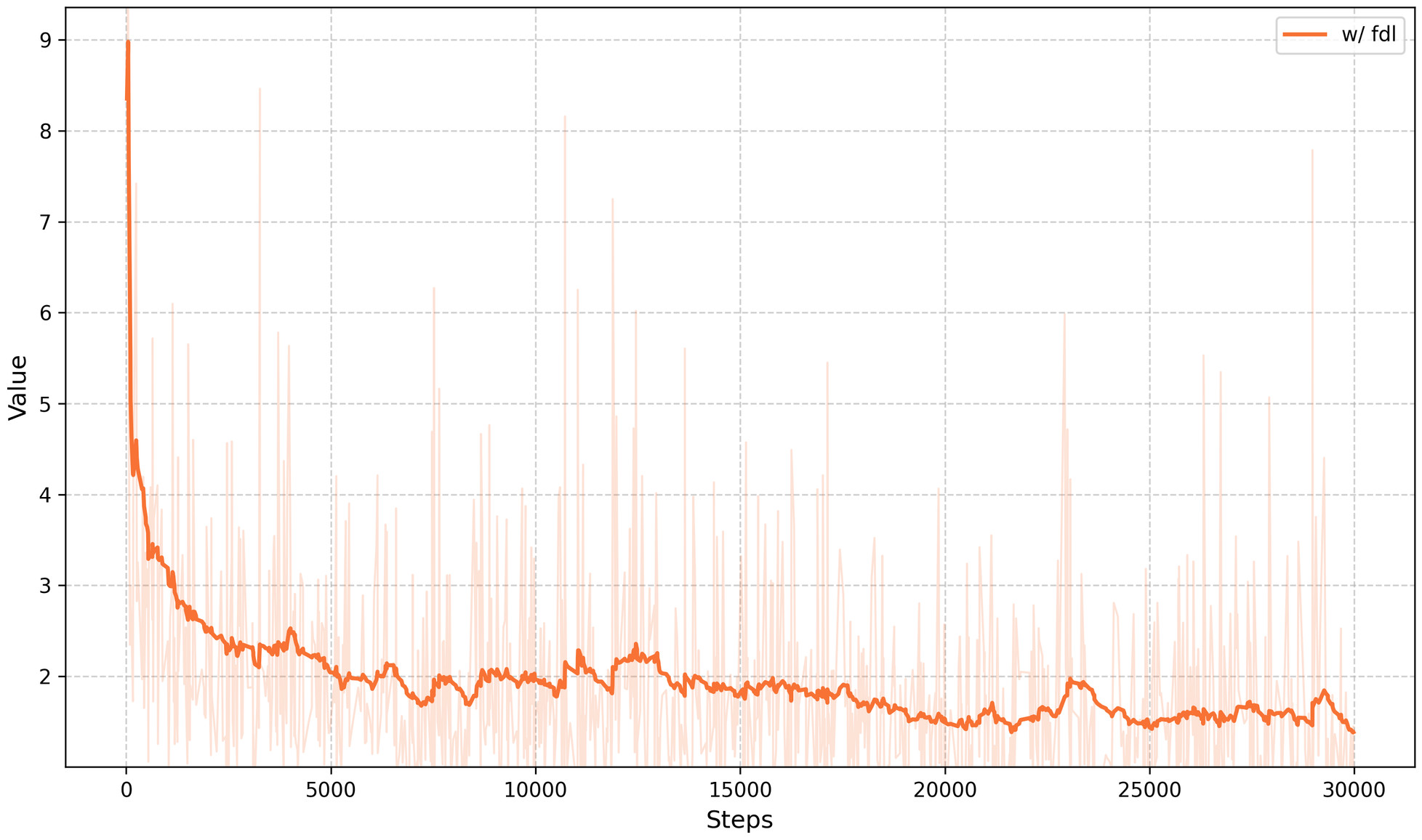} 
        \caption{FDL loss of generator}
        \label{fig:fdl_loss}
    \end{subfigure}
    \scriptsize
    \caption{Generator loss curve with and without FDL in the first 30,000 iterations.} 
    \label{fig:loss_curves}
    \vspace{-4mm}
\end{figure}

\vspace{-2mm}
\section{Further Analysis of StrSR}
\label{sec:abl}
\vspace{-1mm}

\subsection{Effect of the Two-Stage Training Strategy}
\vspace{-1mm}
In our experimental design, StrSR employs a coarse-to-fine, two-stage training strategy: 40,000 initial steps without the FDL loss, followed by 20,000 fine-tuning steps with the FDL. As illustrated in Figs.~\ref{fig:GAN_loss_curves} and~\ref{fig:loss_curves}, introducing FDL too early causes severe training instability and traps the model in local minima.

Since the training pipeline involves distilling a text-to-image model into a super-resolution model, the initial domain gap is substantial. During early training, the model must prioritize low-frequency semantics over high-frequency details. Applying the FDL too early generates excessively large and noisy loss values. This instability disrupts the generator-discriminator training dynamics and prevents perceptual loss functions, \ie, LPIPS, from taking effect quickly. Conversely, in the later stage, introducing the frequency distribution matching forces the model to align the predicted image with the ground truth in the frequency domain. This facilitates the precise correction of fine details.

\vspace{-1mm}
\subsection{Ablation on the FLUX.2 Base Model}
\vspace{-1mm}
We further conduct ablation studies on our FLUX.2 base model, following the same ablation settings as the Z-Image base model in the main text. As demonstrated in Tab.~\ref{tab:ablation-flux}, our full setting model outperforms other models in all three metrics. This further study shows the advancement of RaGAN and VLM in facilitating the model's convergence to generating photo-realistic results. 

\vspace{-1mm}
\subsection{Impact of LoRA Rank and Alpha}
\vspace{-1mm}
To evaluate the impact of trainable parameters, we analyze different LoRA rank configurations in Tab.~\ref{tab:ablation-lora}. The results demonstrate that increasing the LoRA rank from 64 to 128 significantly enhances the model's performance, while further increasing it from 128 to 256 brings somewhat limited improvement. A higher rank expands the model's trainable capacity, providing sufficient parameters for more effective fine-tuning. Therefore, selecting an appropriate rank is crucial to maximizing model capability while maintaining training efficiency. We eventually set $\text{rank}=256$ and $\text{alpha}=256$ to strike this balance.

\begin{table}[t]
\centering
\small
\setlength{\tabcolsep}{1.8pt}
\renewcommand{\arraystretch}{0.92}
\begin{tabular}{c|ccc}
\toprule
NFE & 10$\times$GFLR$_{16}$ $\downarrow$ & $\Delta_{1\to N}$ $\uparrow$ & 10$\times$GFLR$_{11}$ $\downarrow$ \\
\midrule
1 & $9.28{\pm}0.26$ & -- & $5.90{\pm}0.36$ \\
2 & $8.59{\pm}0.28$ & $0.70{\pm}0.09$ & $5.82{\pm}0.36$ \\
4 & $8.70{\pm}0.28$ & $0.59{\pm}0.09$ & $5.78{\pm}0.36$ \\
8 & $8.73{\pm}0.27$ & $0.56{\pm}0.10$ & $5.78{\pm}0.36$ \\
\bottomrule
\end{tabular}
\vspace{4mm}
\caption{Compact GFLR analysis across NFE. $T=16$ is the DiT grid period, $T=11$ is an off-grid control, and $\Delta_{1\to N}=10\times(\mathrm{GFLR}_{1}-\mathrm{GFLR}_{N})$ at $T=16$.}
\label{tab:gflr_compact}
\vspace{-4mm}
\end{table}

\vspace{-2mm}
\section{Discussion: How NFE impact DiT artifact?}\label{sec:gflr}
\vspace{-1mm}
We define the grid-frequency leakage ratio (GFLR) on DiT models to study how NFE affects artifacts.
For output $x_N$, we compute the residual $r_N$$=$$x_N$$-$$x_{\mathrm{HR}}$ and its spectrum $P_N$$=$$|\mathcal{F}(r_N)|^2$.
Let $T$ be the pixel-space grid period induced by patchification and downsampling, and $\mathcal{G}_T$ denote small windows around the grid frequency $1/T$.
We compute
\begin{equation}
\mathrm{GFLR}(N,T)=\lg
\frac{
\mathrm{Mean}_{(u,v)\in\mathcal{G}_T} P_N(u,v)
}{
\mathrm{Mean}_{(u,v)\in\mathcal{B}_T} P_N(u,v)
},
\end{equation}
where $\mathcal{B}_T$ is the remaining background excluding $\mathcal{G}_T$.
GFLR measures whether the residual error is abnormally concentrated at the DiT patch-grid frequencies. This is highly targeted since grid artifacts are structured periodic errors. 
By only varying NFE, GFLR changes reflect how sampling steps affect patch-grid spectral leakage. 
As shown in Tab.~\ref{tab:gflr_compact}, at the grid period $T{=}16$, the 1-step output has the highest GFLR, and paired comparisons show a consistent reduction after using multiple steps. But the off-grid control $T{=}11$ yields much lower GFLR. This verifies that one-step DiT artifacts arise from structured spectral leakage at patch-grid frequencies.

\begin{figure*}[t]
\scriptsize
\centering
\begin{tabular}{cccc}
\hspace{-0.4cm}
\newcommand{\imgid}{Canon\_018}
\newcommand{\imgnote}{Canon\_018}
\begin{adjustbox}{valign=t}
\begin{tabular}{c}
\includegraphics[width=0.237\textwidth,height=0.1845\textwidth]{figs/jpg/comparison/seedvr/enclosed/\imgid_HR.jpg} \\
RealSR: \texttt{\imgnote}
\end{tabular}
\end{adjustbox}
\hspace{-0.2cm}
\begin{adjustbox}{valign=t}
\begin{tabular}{cccccc}
\includegraphics[width=0.139\textwidth]{figs/jpg/comparison/seedvr/cropped/\imgid_HR.jpg} \hspace{-1mm} &
\includegraphics[width=0.139\textwidth]{figs/jpg/comparison/seedvr/cropped/\imgid_bicubic.jpg} \hspace{-1mm} &
\includegraphics[width=0.139\textwidth]{figs/jpg/comparison/seedvr/cropped/\imgid_seedvr.jpg} \hspace{-1mm} &
\includegraphics[width=0.139\textwidth]{figs/jpg/comparison/seedvr/cropped/\imgid_zimage.jpg} \hspace{-1mm} &
\includegraphics[width=0.139\textwidth]{figs/jpg/comparison/seedvr/cropped/\imgid_flux_61k.jpg} \hspace{-1mm} \\
HR \hspace{-1mm} &
Bicubic \hspace{-1mm} &
SeedVR2~\cite{seedvr2} \hspace{-1mm} &
StrSR - Z \hspace{-1mm} &
StrSR - F \hspace{-1mm} \\
\includegraphics[width=0.139\textwidth]{figs/jpg/comparison/seedvr/cropped/\imgid_HR_2.jpg} \hspace{-1mm} &
\includegraphics[width=0.139\textwidth]{figs/jpg/comparison/seedvr/cropped/\imgid_bicubic_2.jpg} \hspace{-1mm} &
\includegraphics[width=0.139\textwidth]{figs/jpg/comparison/seedvr/cropped/\imgid_seedvr_2.jpg} \hspace{-1mm} &
\includegraphics[width=0.139\textwidth]{figs/jpg/comparison/seedvr/cropped/\imgid_zimage_2.jpg} \hspace{-1mm} &
\includegraphics[width=0.139\textwidth]{figs/jpg/comparison/seedvr/cropped/\imgid_flux_61k_2.jpg} \hspace{-1mm} \\
HR \hspace{-1mm} &
Bicubic \hspace{-1mm} &
SeedVR2~\cite{seedvr2} \hspace{-1mm} &
StrSR - Z  \hspace{-1mm} &
StrSR - F \hspace{-1mm} \\
\end{tabular}
\end{adjustbox}
\\
\end{tabular}

\begin{tabular}{cccc}
\hspace{-0.4cm}
\newcommand{\imgid}{Canon\_030}
\newcommand{\imgnote}{Canon\_030}
\begin{adjustbox}{valign=t}
\begin{tabular}{c}
\includegraphics[width=0.237\textwidth,height=0.1845\textwidth]{figs/jpg/comparison/seedvr/enclosed/\imgid_HR.jpg} \\
RealSR: \texttt{\imgnote}
\end{tabular}
\end{adjustbox}
\hspace{-0.2cm}
\begin{adjustbox}{valign=t}
\begin{tabular}{cccccc}
\includegraphics[width=0.139\textwidth]{figs/jpg/comparison/seedvr/cropped/\imgid_HR.jpg} \hspace{-1mm} &
\includegraphics[width=0.139\textwidth]{figs/jpg/comparison/seedvr/cropped/\imgid_bicubic.jpg} \hspace{-1mm} &
\includegraphics[width=0.139\textwidth]{figs/jpg/comparison/seedvr/cropped/\imgid_seedvr.jpg} \hspace{-1mm} &
\includegraphics[width=0.139\textwidth]{figs/jpg/comparison/seedvr/cropped/\imgid_zimage.jpg} \hspace{-1mm} &
\includegraphics[width=0.139\textwidth]{figs/jpg/comparison/seedvr/cropped/\imgid_flux_61k.jpg} \hspace{-1mm} \\
HR \hspace{-1mm} &
Bicubic \hspace{-1mm} &
SeedVR2~\cite{seedvr2} \hspace{-1mm} &
StrSR - Z \hspace{-1mm} &
StrSR - F \hspace{-1mm} \\
\includegraphics[width=0.139\textwidth]{figs/jpg/comparison/seedvr/cropped/\imgid_HR_2.jpg} \hspace{-1mm} &
\includegraphics[width=0.139\textwidth]{figs/jpg/comparison/seedvr/cropped/\imgid_bicubic_2.jpg} \hspace{-1mm} &
\includegraphics[width=0.139\textwidth]{figs/jpg/comparison/seedvr/cropped/\imgid_seedvr_2.jpg} \hspace{-1mm} &
\includegraphics[width=0.139\textwidth]{figs/jpg/comparison/seedvr/cropped/\imgid_zimage_2.jpg} \hspace{-1mm} &
\includegraphics[width=0.139\textwidth]{figs/jpg/comparison/seedvr/cropped/\imgid_flux_61k_2.jpg} \hspace{-1mm} \\
HR \hspace{-1mm} &
Bicubic \hspace{-1mm} &
SeedVR2~\cite{seedvr2} \hspace{-1mm} &
StrSR - Z  \hspace{-1mm} &
StrSR - F \hspace{-1mm} \\
\end{tabular}
\end{adjustbox}
\\
\end{tabular}

\begin{tabular}{cccc}
\hspace{-0.4cm}
\newcommand{\imgid}{image\_22}
\newcommand{\imgnote}{image\_22}
\begin{adjustbox}{valign=t}
\begin{tabular}{c}
\includegraphics[width=0.237\textwidth,height=0.1845\textwidth]{figs/jpg/comparison/seedvr/enclosed/\imgid_HR.jpg} \\
DIV2K: \texttt{\imgnote}
\end{tabular}
\end{adjustbox}
\hspace{-0.2cm}
\begin{adjustbox}{valign=t}
\begin{tabular}{cccccc}
\includegraphics[width=0.139\textwidth]{figs/jpg/comparison/seedvr/cropped/\imgid_HR.jpg} \hspace{-1mm} &
\includegraphics[width=0.139\textwidth]{figs/jpg/comparison/seedvr/cropped/\imgid_bicubic.jpg} \hspace{-1mm} &
\includegraphics[width=0.139\textwidth]{figs/jpg/comparison/seedvr/cropped/\imgid_seedvr.jpg} \hspace{-1mm} &
\includegraphics[width=0.139\textwidth]{figs/jpg/comparison/seedvr/cropped/\imgid_zimage.jpg} \hspace{-1mm} &
\includegraphics[width=0.139\textwidth]{figs/jpg/comparison/seedvr/cropped/\imgid_flux_61k.jpg} \hspace{-1mm} \\
HR \hspace{-1mm} &
Bicubic \hspace{-1mm} &
SeedVR2~\cite{seedvr2} \hspace{-1mm} &
StrSR - Z \hspace{-1mm} &
StrSR - F \hspace{-1mm} \\
\includegraphics[width=0.139\textwidth]{figs/jpg/comparison/seedvr/cropped/\imgid_HR_2.jpg} \hspace{-1mm} &
\includegraphics[width=0.139\textwidth]{figs/jpg/comparison/seedvr/cropped/\imgid_bicubic_2.jpg} \hspace{-1mm} &
\includegraphics[width=0.139\textwidth]{figs/jpg/comparison/seedvr/cropped/\imgid_seedvr_2.jpg} \hspace{-1mm} &
\includegraphics[width=0.139\textwidth]{figs/jpg/comparison/seedvr/cropped/\imgid_zimage_2.jpg} \hspace{-1mm} &
\includegraphics[width=0.139\textwidth]{figs/jpg/comparison/seedvr/cropped/\imgid_flux_61k_2.jpg} \hspace{-1mm} \\
HR \hspace{-1mm} &
Bicubic \hspace{-1mm} &
SeedVR2~\cite{seedvr2} \hspace{-1mm} &
StrSR - Z  \hspace{-1mm} &
StrSR - F \hspace{-1mm} \\
\end{tabular}
\end{adjustbox}
\\
\end{tabular}

\begin{tabular}{cccc}
\hspace{-0.4cm}
\newcommand{\imgid}{image\_98}
\newcommand{\imgnote}{image\_98}
\begin{adjustbox}{valign=t}
\begin{tabular}{c}
\includegraphics[width=0.237\textwidth,height=0.1845\textwidth]{figs/jpg/comparison/seedvr/enclosed/\imgid_HR.jpg} \\
DIV2K: \texttt{\imgnote}
\end{tabular}
\end{adjustbox}
\hspace{-0.2cm}
\begin{adjustbox}{valign=t}
\begin{tabular}{cccccc}
\includegraphics[width=0.139\textwidth]{figs/jpg/comparison/seedvr/cropped/\imgid_HR.jpg} \hspace{-1mm} &
\includegraphics[width=0.139\textwidth]{figs/jpg/comparison/seedvr/cropped/\imgid_bicubic.jpg} \hspace{-1mm} &
\includegraphics[width=0.139\textwidth]{figs/jpg/comparison/seedvr/cropped/\imgid_seedvr.jpg} \hspace{-1mm} &
\includegraphics[width=0.139\textwidth]{figs/jpg/comparison/seedvr/cropped/\imgid_zimage.jpg} \hspace{-1mm} &
\includegraphics[width=0.139\textwidth]{figs/jpg/comparison/seedvr/cropped/\imgid_flux_61k.jpg} \hspace{-1mm} \\
HR \hspace{-1mm} &
Bicubic \hspace{-1mm} &
SeedVR2~\cite{seedvr2} \hspace{-1mm} &
StrSR - Z \hspace{-1mm} &
StrSR - F \hspace{-1mm} \\
\includegraphics[width=0.139\textwidth]{figs/jpg/comparison/seedvr/cropped/\imgid_HR_2.jpg} \hspace{-1mm} &
\includegraphics[width=0.139\textwidth]{figs/jpg/comparison/seedvr/cropped/\imgid_bicubic_2.jpg} \hspace{-1mm} &
\includegraphics[width=0.139\textwidth]{figs/jpg/comparison/seedvr/cropped/\imgid_seedvr_2.jpg} \hspace{-1mm} &
\includegraphics[width=0.139\textwidth]{figs/jpg/comparison/seedvr/cropped/\imgid_zimage_2.jpg} \hspace{-1mm} &
\includegraphics[width=0.139\textwidth]{figs/jpg/comparison/seedvr/cropped/\imgid_flux_61k_2.jpg} \hspace{-1mm} \\
HR \hspace{-1mm} &
Bicubic \hspace{-1mm} &
SeedVR2~\cite{seedvr2} \hspace{-1mm} &
StrSR - Z  \hspace{-1mm} &
StrSR - F \hspace{-1mm} \\
\end{tabular}
\end{adjustbox}
\\
\end{tabular}
\caption{Visual comparison for image SR ($\times$4) in RealSR and DIV2K-val dataset.}
\label{fig:vis-seedvr2}
\vspace{-2mm}
\end{figure*}

\vspace{-2mm}
\section{Comparison with SeedVR2}
\label{sec:seedvr2}
\vspace{-1mm}
Video super-resolution (VSR) is a rapidly developing field, and some VSR methods~\cite{seedvr2} claim to achieve good results in image SR as well. We evaluate SeedVR2~\cite{seedvr2} using their HuggingFace space demo, as shown in {Fig.~\ref{fig:vis-seedvr2}}. When processing images, SeedVR2 tends to produce hair-like artifacts, whereas StrSR performs well.

\begin{table*}[t]
\scriptsize
\setlength{\tabcolsep}{0.3mm} 
\newcolumntype{C}{>{\centering\arraybackslash}X}
\begin{center}
\begin{tabularx}{\textwidth}{c|c|CC|CC|CCCCC}
\toprule[0.15em]
\rowcolor{color3} Datasets & Methods & PSNR$\uparrow$ & SSIM$\uparrow$ & LPIPS$\downarrow$ & DISTS$\downarrow$ & NIQE$\downarrow$ & MAN.$\uparrow$ & MUS.$\uparrow$ & QAli.$\uparrow$ & Qual.$\uparrow$ \\

\midrule[0.15em]

\multirow{5}{*}{\makecell{DIV2K\\valid}} 
& BSRGAN~\cite{zhang2021bsrgan} & \rr{23.67} & 0.6240 & 0.4065 & 0.2274 & 3.8683 & 0.4662 & 57.7954 & 3.8239 & 0.5372 \\
& RealESRGAN~\cite{wang2021realesrgan} & \bb{23.47} & \rr{0.6372} & 0.3808 & 0.2126 & 3.8233 & 0.5089 & 56.7319 & 4.1489 & 0.6019 \\
& SwinIR~\cite{liang2021swinir} & 23.04 & \bb{0.6248} & 0.3851 & 0.2095 & 3.5762 & 0.4985 & 57.5313 & 4.0004 & 0.5736 \\
& StrSR-Z(ours) & 21.35 & 0.5798 & \bb{0.3216} & \bb{0.1407} & \bb{3.4738} & \bb{0.5727} & \rr{64.6547} & \bb{4.4759} & \rr{0.7316} \\
& StrSR-F(ours) & 21.70 & 0.5982 & \rr{0.2992} & \rr{0.1288} & \rr{3.0379} & \rr{0.5864} & \bb{63.6568} & \rr{4.4763} & \bb{0.7022} \\

\cmidrule{1-11}
\multirow{5}{*}{\makecell{RealSR}} 
& BSRGAN~\cite{zhang2021bsrgan} & \rr{26.51} & \bb{0.7746} & 0.2685 & 0.1761 & 4.6501 & 0.5278 & 64.0538 & 3.8304 & 0.6059 \\
& RealESRGAN~\cite{wang2021realesrgan} & 25.85 & 0.7734 & 0.2729 & 0.1685 & 4.6766 & 0.5385 & 62.8190 & 3.9198 & 0.6152 \\
& SwinIR~\cite{liang2021swinir} & \bb{26.03} & \rr{0.7801} & \rr{0.2593} & \rr{0.1609} & 4.6370 & 0.5122 & 62.2812 & 3.9205 & 0.5791 \\
& StrSR - Z-Image (ours) & 22.15 & 0.6606 & 0.3215 & 0.1958 & \bb{4.4439} & \bb{0.5941} & \rr{68.5793} & \rr{4.2371} & \rr{0.7417} \\
& StrSR - FLUX (ours) & 23.77 & 0.7236 & \bb{0.2599} & \bb{0.1665} & \rr{3.8090} & \rr{0.6333} & \bb{67.3318} & \bb{4.1296} & \bb{0.7134} \\
\bottomrule[0.15em]
\end{tabularx}
\end{center}
\scriptsize
\vspace{-1mm}
\caption{Quantitative results ($\times$4) on RealSR of non-diffusion based methods. MAN., MUS., QAli., and Qual. stand for MANIQA, MUSIQ, QAlign, and QualiCLIP,.  } 
\label{tab:tradition_RealSR}
\vspace{-6mm}
\end{table*}

\begin{table}[t]
\scriptsize
\setlength{\tabcolsep}{1.5mm} 
\newcolumntype{C}{>{\centering\arraybackslash}X}
\begin{center}
\begin{tabularx}{0.9\columnwidth}{c|c|CCCC}
\toprule[0.15em]
\rowcolor{color3} Dataset & Methods & NIQE$\downarrow$ & MANIQA$\uparrow$ & MUSIQ$\uparrow$ & QAlign$\uparrow$ \\

\midrule[0.15em]
\multirow{5}{*}{\makecell{RealLQ250}} 
& BSRGAN~\cite{zhang2021bsrgan} & 4.5434 & 0.5006 & 63.6450 & 3.5707 \\
& RealESRGAN~\cite{wang2021realesrgan} & 4.1294 & 0.5239 & 63.9639 & 3.7326 \\
& SwinIR~\cite{liang2021swinir} & 4.1624 & 0.5104 & 62.9853 & 3.6342 \\
& StrSR - Z-Image (ours) & \bb{4.1019} & \bb{0.5774} & \rr{70.1176} & \bb{4.2595} \\
& StrSR - FLUX (ours) & \rr{3.4693} & \rr{0.6138} & \bb{69.2050} & \rr{4.2851} \\
\bottomrule[0.15em]
\end{tabularx}
\end{center}
\vspace{-1mm}
\caption{Comparison ($\times$4) of non-diffusion based methods with ours on RealLQ250.} 
\label{tab:tradition_RealLQ250}
\vspace{-6mm}
\end{table}

\vspace{-2mm}
\section{Comparison with Non-Diffusion-Based Methods}
\label{sec:classic}
\vspace{-1mm}
We further evaluate the performance of StrSR against three classic non-diffusion-based models: BSRGAN~\cite{zhang2021bsrgan}, RealESRGAN~\cite{wang2021realesrgan}, and SwinIR~\cite{liang2021swinir}. As shown in Tabs.~\ref{tab:tradition_RealSR} and \ref{tab:tradition_RealLQ250}, although non-diffusion-based methods tend to perform better in pixel-level metrics like PSNR and SSIM, StrSR exhibits significantly superior performance across all semantic and aesthetic metrics.

\vspace{-2mm}
\section{More Visual Comparisons}
\label{sec:vis}
\begin{figure*}[t]
\scriptsize
\centering

\begin{tabular}{cccc}

\hspace{-0.4cm}
\newcommand{\imgid}{image_5}
\newcommand{\imgnote}{005}
\begin{adjustbox}{valign=t}
\begin{tabular}{c}
\includegraphics[width=0.237\textwidth,height=0.1845\textwidth]{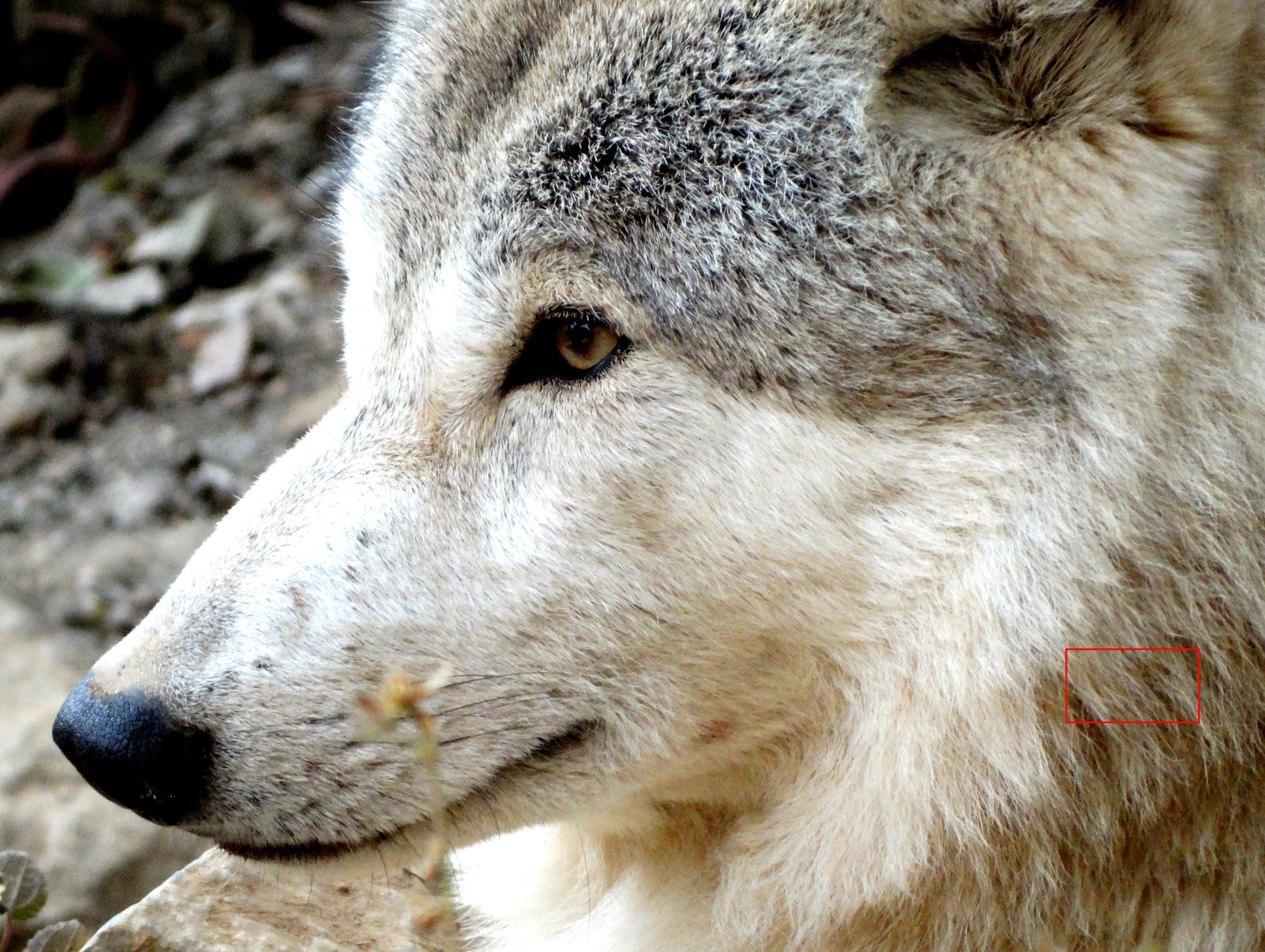} \\
DIV2K: \texttt{img\_\imgnote}
\end{tabular}
\end{adjustbox}
\hspace{-0.2cm}
\begin{adjustbox}{valign=t}
\begin{tabular}{cccccc}
\includegraphics[width=0.139\textwidth]{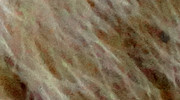} \hspace{-1mm} &
\includegraphics[width=0.139\textwidth]{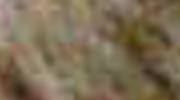} \hspace{-1mm} &
\includegraphics[width=0.139\textwidth]{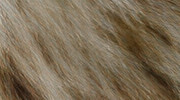} \hspace{-1mm} &
\includegraphics[width=0.139\textwidth]{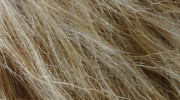} \hspace{-1mm} &
\includegraphics[width=0.139\textwidth]{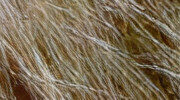} \hspace{-1mm} \\
HR \hspace{-1mm} &
Bicubic \hspace{-1mm} &
CTMSR~\cite{you2025ctmsr} \hspace{-1mm} &
PiSA-SR~\cite{sun2024pisasr} \hspace{-1mm} &
TSD-SR~\cite{dong2025tsdsr} \hspace{-1mm} \\
\includegraphics[width=0.139\textwidth]{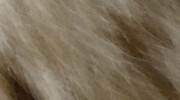} \hspace{-1mm} &
\includegraphics[width=0.139\textwidth]{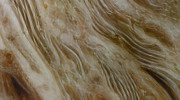} \hspace{-1mm} &
\includegraphics[width=0.139\textwidth]{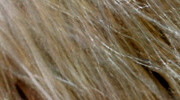} \hspace{-1mm} &
\includegraphics[width=0.139\textwidth]{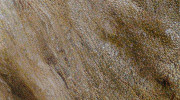} \hspace{-1mm} &
\includegraphics[width=0.139\textwidth]{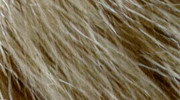} \hspace{-1mm} \\
OSEDiff~\cite{wu2024osediff} \hspace{-1mm} &
InvSR~\cite{yue2025invsr} \hspace{-1mm} &
HYPIR~\cite{lin2025hypir} \hspace{-1mm} &
SinSR~\cite{wang2024sinsr}  \hspace{-1mm} &
StrSR (ours) \hspace{-1mm} \\
\end{tabular}
\end{adjustbox}
\\

\hspace{-0.4cm}
\newcommand{\imgid}{image_36}
\newcommand{\imgnote}{036}
\begin{adjustbox}{valign=t}
\begin{tabular}{c}
\includegraphics[width=0.237\textwidth,height=0.1845\textwidth]{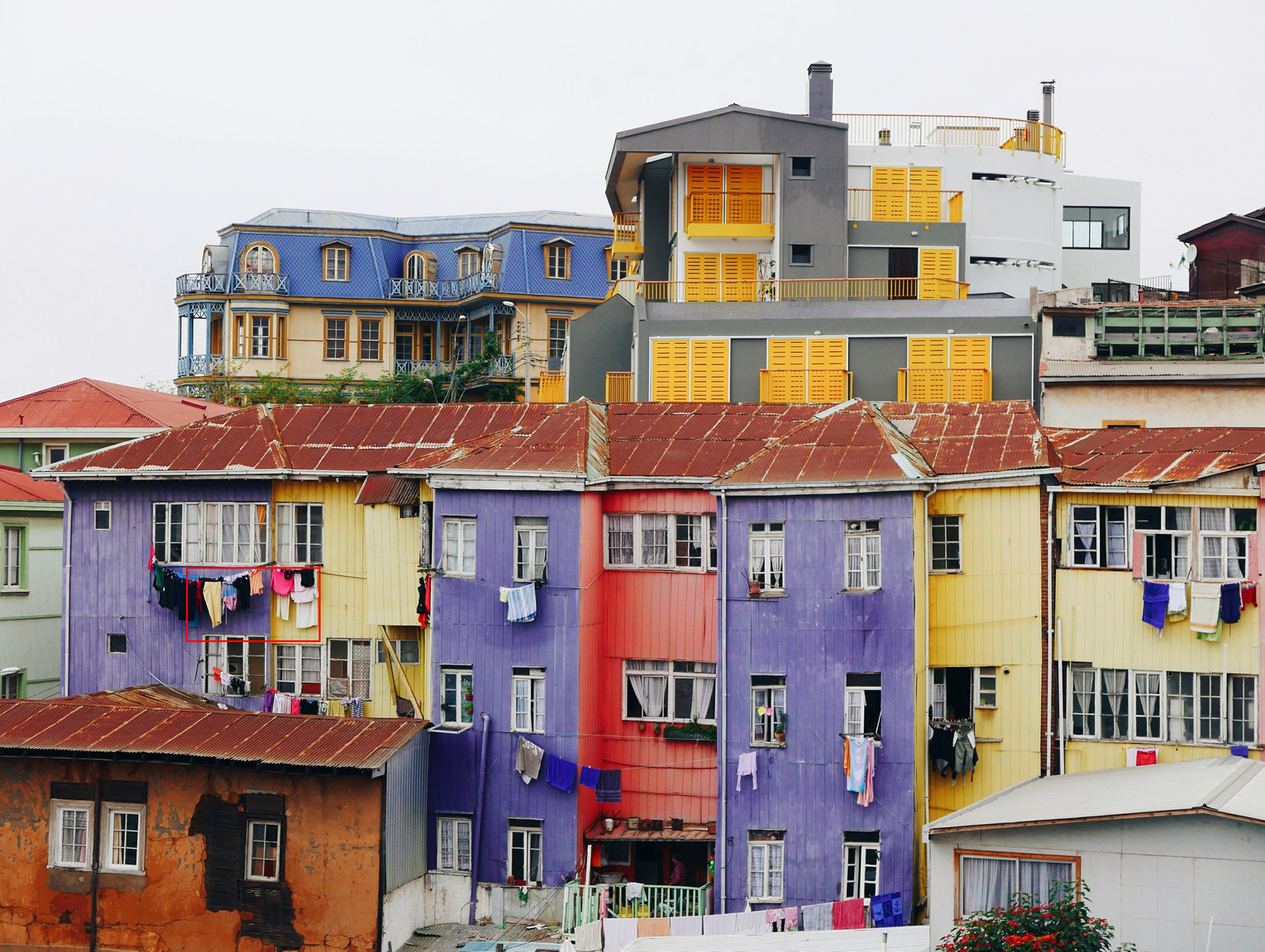} \\
DIV2K: \texttt{img\_\imgnote}
\end{tabular}
\end{adjustbox}
\hspace{-0.2cm}
\begin{adjustbox}{valign=t}
\begin{tabular}{cccccc}
\includegraphics[width=0.139\textwidth]{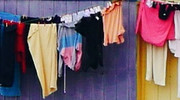} \hspace{-1mm} &
\includegraphics[width=0.139\textwidth]{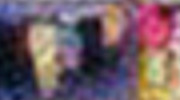} \hspace{-1mm} &
\includegraphics[width=0.139\textwidth]{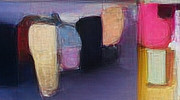} \hspace{-1mm} &
\includegraphics[width=0.139\textwidth]{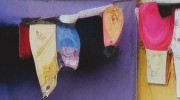} \hspace{-1mm} &
\includegraphics[width=0.139\textwidth]{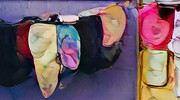} \hspace{-1mm} \\
HR \hspace{-1mm} &
Bicubic \hspace{-1mm} &
CTMSR~\cite{you2025ctmsr} \hspace{-1mm} &
PiSA-SR~\cite{sun2024pisasr} \hspace{-1mm} &
TSD-SR~\cite{dong2025tsdsr} \hspace{-1mm} \\
\includegraphics[width=0.139\textwidth]{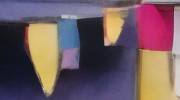} \hspace{-1mm} &
\includegraphics[width=0.139\textwidth]{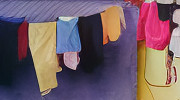} \hspace{-1mm} &
\includegraphics[width=0.139\textwidth]{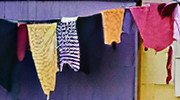} \hspace{-1mm} &
\includegraphics[width=0.139\textwidth]{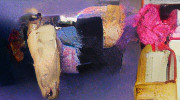} \hspace{-1mm} &
\includegraphics[width=0.139\textwidth]{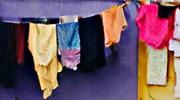} \hspace{-1mm} \\
OSEDiff~\cite{wu2024osediff} \hspace{-1mm} &
InvSR~\cite{yue2025invsr} \hspace{-1mm} &
HYPIR~\cite{lin2025hypir} \hspace{-1mm} &
SinSR~\cite{wang2024sinsr}  \hspace{-1mm} &
{StrSR} (ours) \hspace{-1mm} \\
\end{tabular}
\end{adjustbox}
\\

\hspace{-0.4cm}
\newcommand{\imgid}{image_39}
\newcommand{\imgnote}{039}
\begin{adjustbox}{valign=t}
\begin{tabular}{c}
\includegraphics[width=0.237\textwidth,height=0.1845\textwidth]{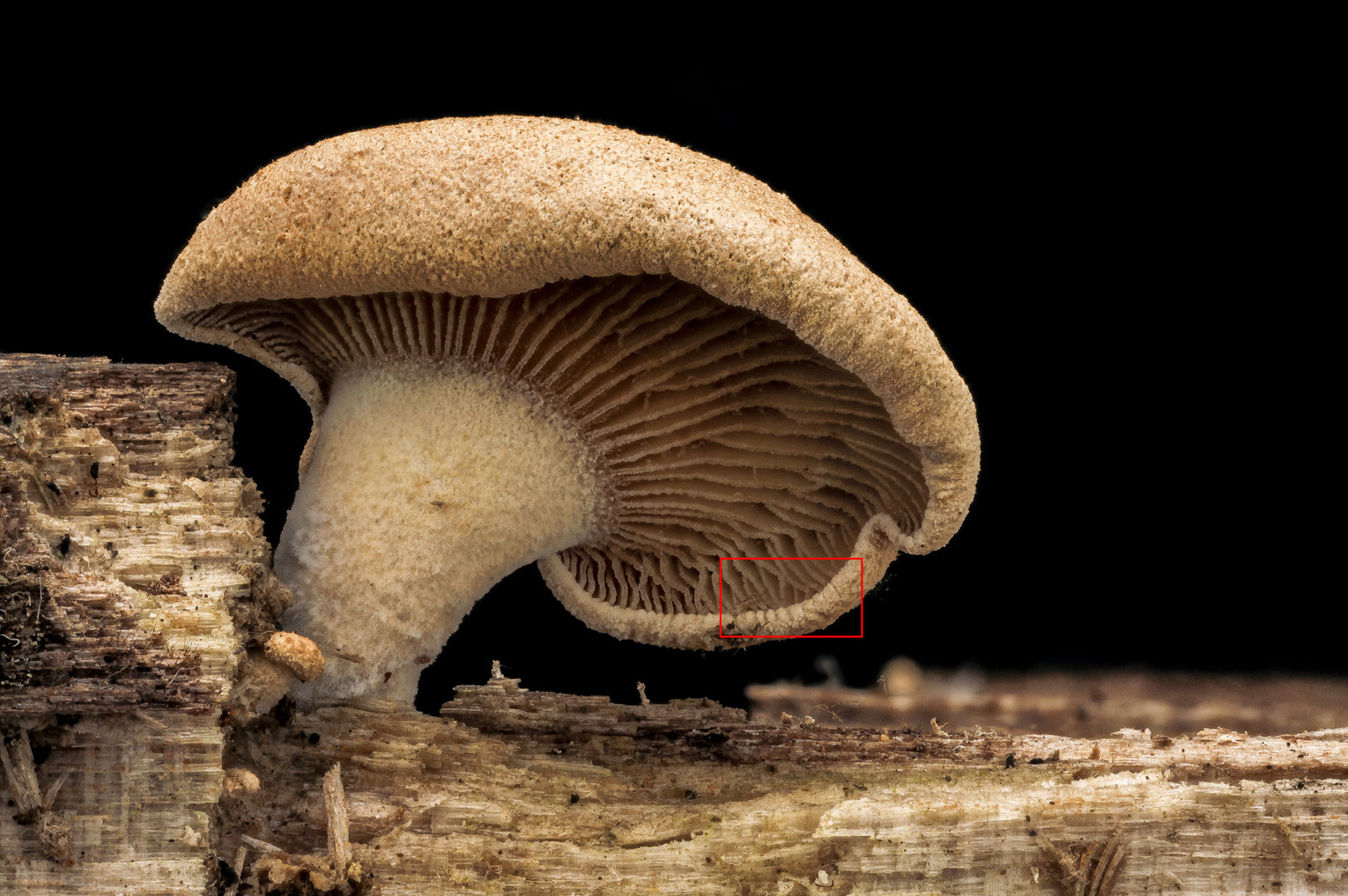} \\
DIV2K: \texttt{img\_\imgnote}
\end{tabular}
\end{adjustbox}
\hspace{-0.2cm}
\begin{adjustbox}{valign=t}
\begin{tabular}{cccccc}
\includegraphics[width=0.139\textwidth]{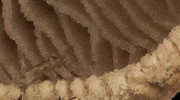} \hspace{-1mm} &
\includegraphics[width=0.139\textwidth]{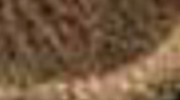} \hspace{-1mm} &
\includegraphics[width=0.139\textwidth]{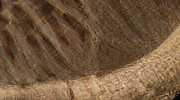} \hspace{-1mm} &
\includegraphics[width=0.139\textwidth]{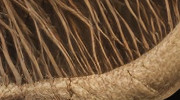} \hspace{-1mm} &
\includegraphics[width=0.139\textwidth]{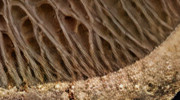} \hspace{-1mm} \\
HR \hspace{-1mm} &
Bicubic \hspace{-1mm} &
CTMSR~\cite{you2025ctmsr} \hspace{-1mm} &
PiSA-SR~\cite{sun2024pisasr} \hspace{-1mm} &
TSD-SR~\cite{dong2025tsdsr} \hspace{-1mm} \\
\includegraphics[width=0.139\textwidth]{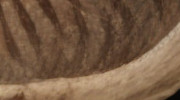} \hspace{-1mm} &
\includegraphics[width=0.139\textwidth]{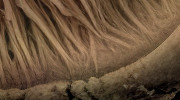} \hspace{-1mm} &
\includegraphics[width=0.139\textwidth]{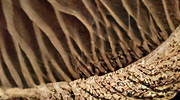} \hspace{-1mm} &
\includegraphics[width=0.139\textwidth]{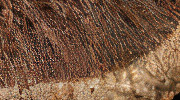} \hspace{-1mm} &
\includegraphics[width=0.139\textwidth]{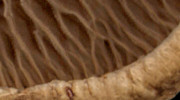} \hspace{-1mm} \\
OSEDiff~\cite{wu2024osediff} \hspace{-1mm} &
InvSR~\cite{yue2025invsr} \hspace{-1mm} &
HYPIR~\cite{lin2025hypir} \hspace{-1mm} &
SinSR~\cite{wang2024sinsr}  \hspace{-1mm} &
{StrSR} (ours) \hspace{-1mm} \\
\end{tabular}
\end{adjustbox}
\\

\hspace{-0.4cm}
\newcommand{\imgid}{image_40}
\newcommand{\imgnote}{040}
\begin{adjustbox}{valign=t}
\begin{tabular}{c}
\includegraphics[width=0.237\textwidth,height=0.1845\textwidth]{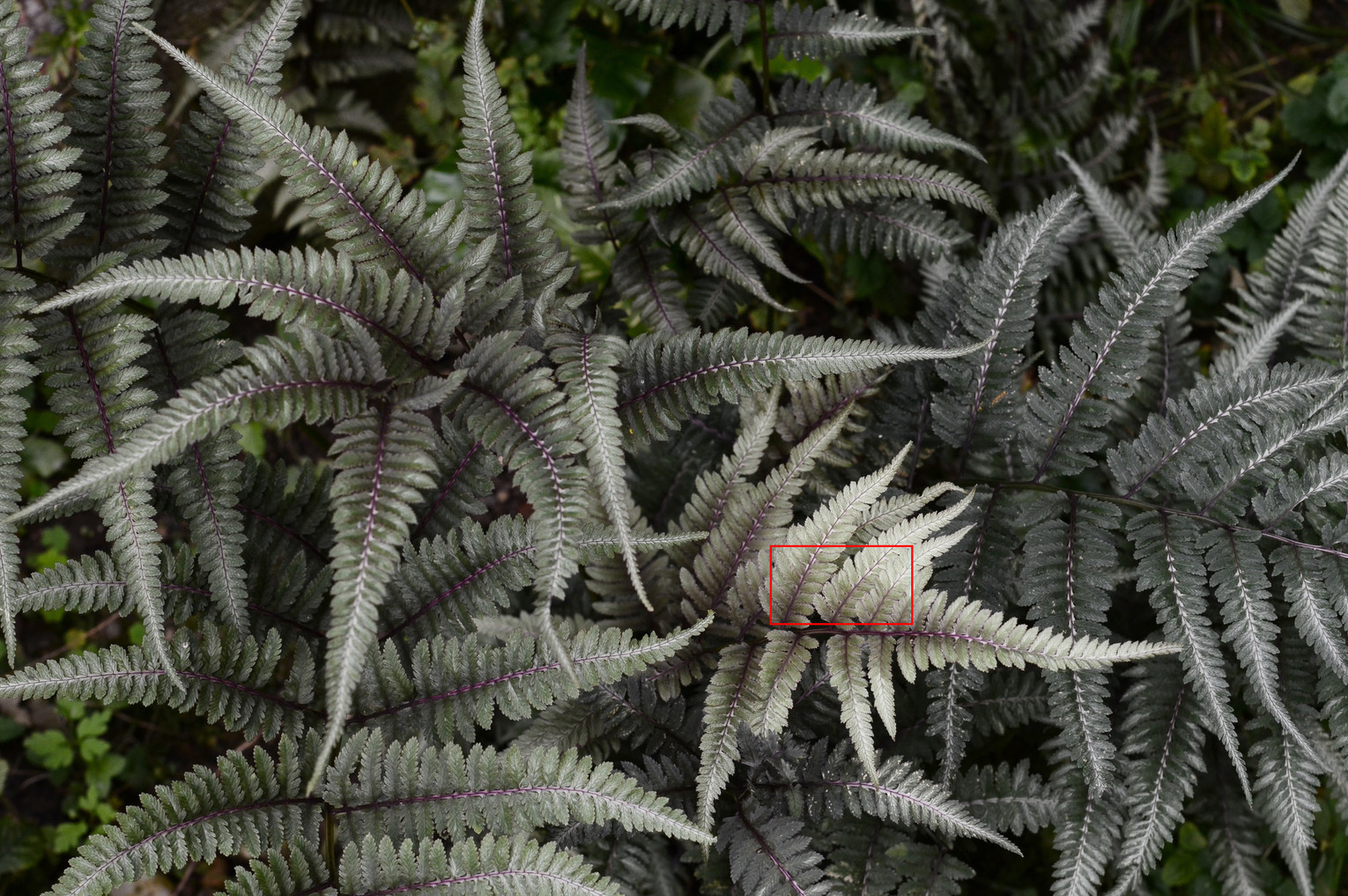} \\
DIV2K: \texttt{img\_\imgnote}
\end{tabular}
\end{adjustbox}
\hspace{-0.2cm}
\begin{adjustbox}{valign=t}
\begin{tabular}{cccccc}
\includegraphics[width=0.139\textwidth]{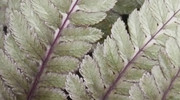} \hspace{-1mm} &
\includegraphics[width=0.139\textwidth]{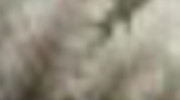} \hspace{-1mm} &
\includegraphics[width=0.139\textwidth]{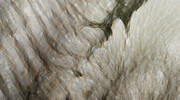} \hspace{-1mm} &
\includegraphics[width=0.139\textwidth]{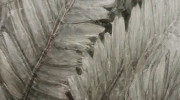} \hspace{-1mm} &
\includegraphics[width=0.139\textwidth]{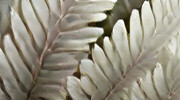} \hspace{-1mm} \\
HR \hspace{-1mm} &
Bicubic \hspace{-1mm} &
CTMSR~\cite{you2025ctmsr} \hspace{-1mm} &
PiSA-SR~\cite{sun2024pisasr} \hspace{-1mm} &
TSD-SR~\cite{dong2025tsdsr} \hspace{-1mm} \\
\includegraphics[width=0.139\textwidth]{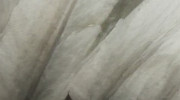} \hspace{-1mm} &
\includegraphics[width=0.139\textwidth]{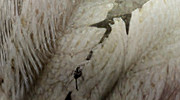} \hspace{-1mm} &
\includegraphics[width=0.139\textwidth]{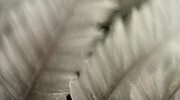} \hspace{-1mm} &
\includegraphics[width=0.139\textwidth]{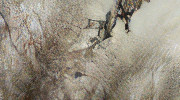} \hspace{-1mm} &
\includegraphics[width=0.139\textwidth]{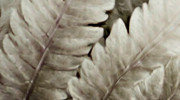} \hspace{-1mm} \\
OSEDiff~\cite{wu2024osediff} \hspace{-1mm} &
InvSR~\cite{yue2025invsr} \hspace{-1mm} &
HYPIR~\cite{lin2025hypir} \hspace{-1mm} &
SinSR~\cite{wang2024sinsr}  \hspace{-1mm} &
StrSR (ours) \hspace{-1mm} \\
\end{tabular}
\end{adjustbox}
\\

\hspace{-0.4cm}
\newcommand{\imgid}{image_42}
\newcommand{\imgnote}{042}
\begin{adjustbox}{valign=t}
\begin{tabular}{c}
\includegraphics[width=0.237\textwidth,height=0.1845\textwidth]{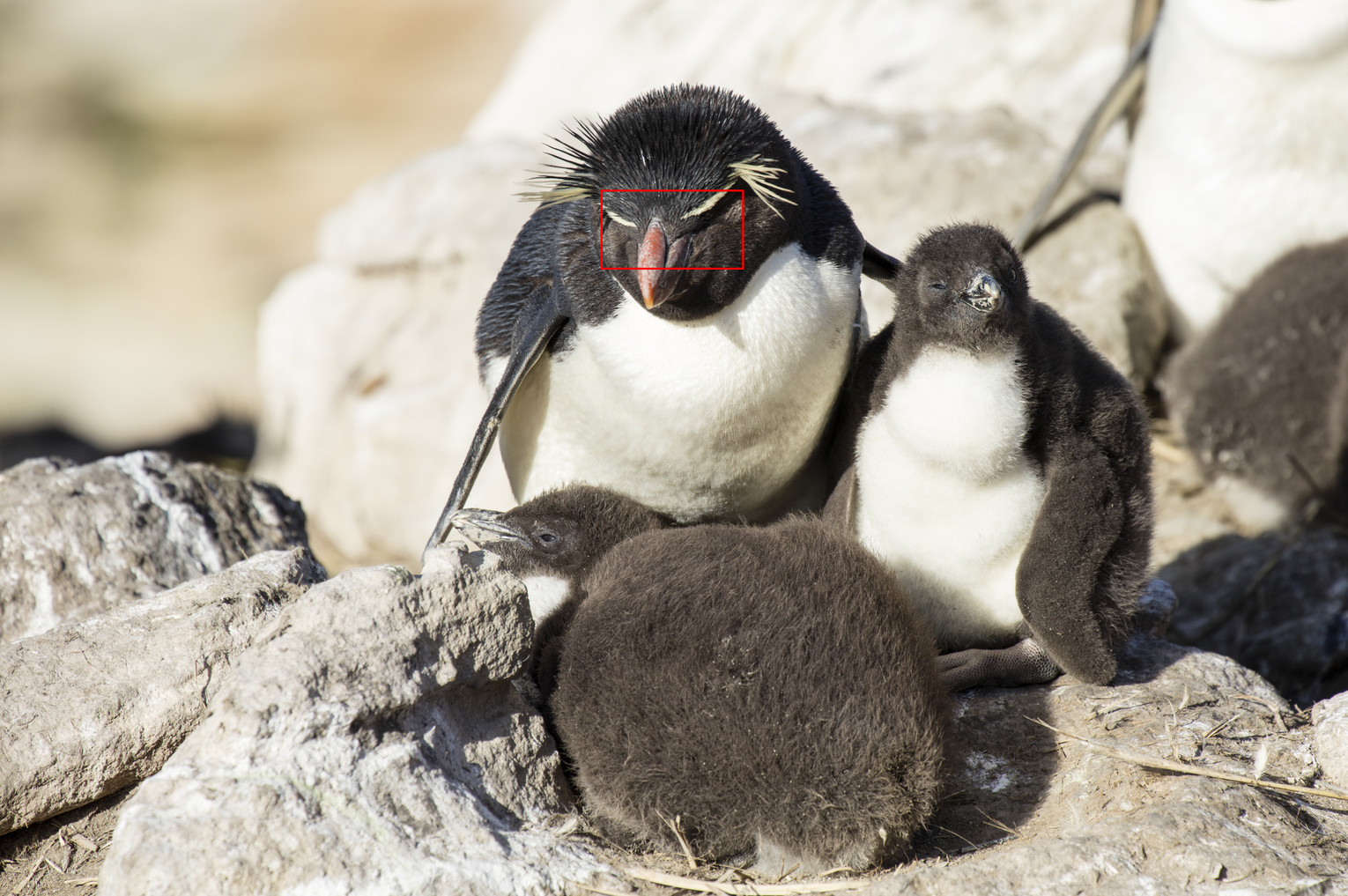} \\
DIV2K: \texttt{img\_\imgnote}
\end{tabular}
\end{adjustbox}
\hspace{-0.2cm}
\begin{adjustbox}{valign=t}
\begin{tabular}{cccccc}
\includegraphics[width=0.139\textwidth]{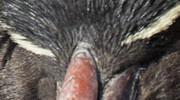} \hspace{-1mm} &
\includegraphics[width=0.139\textwidth]{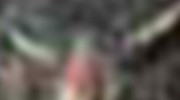} \hspace{-1mm} &
\includegraphics[width=0.139\textwidth]{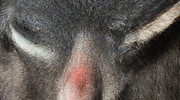} \hspace{-1mm} &
\includegraphics[width=0.139\textwidth]{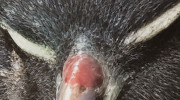} \hspace{-1mm} &
\includegraphics[width=0.139\textwidth]{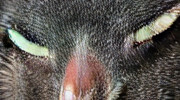} \hspace{-1mm} \\
HR \hspace{-1mm} &
Bicubic \hspace{-1mm} &
CTMSR~\cite{you2025ctmsr} \hspace{-1mm} &
PiSA-SR~\cite{sun2024pisasr} \hspace{-1mm} &
TSD-SR~\cite{dong2025tsdsr} \hspace{-1mm} \\
\includegraphics[width=0.139\textwidth]{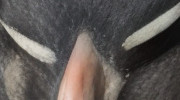} \hspace{-1mm} &
\includegraphics[width=0.139\textwidth]{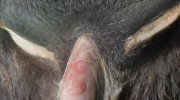} \hspace{-1mm} &
\includegraphics[width=0.139\textwidth]{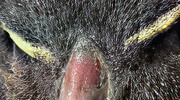} \hspace{-1mm} &
\includegraphics[width=0.139\textwidth]{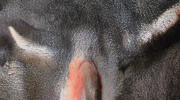} \hspace{-1mm} &
\includegraphics[width=0.139\textwidth]{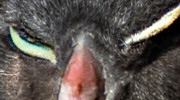} \hspace{-1mm} \\
OSEDiff~\cite{wu2024osediff} \hspace{-1mm} &
InvSR~\cite{yue2025invsr} \hspace{-1mm} &
HYPIR~\cite{lin2025hypir} \hspace{-1mm} &
SinSR~\cite{wang2024sinsr}  \hspace{-1mm} &
StrSR (ours) \hspace{-1mm} \\
\end{tabular}
\end{adjustbox}
\\

\hspace{-0.4cm}
\newcommand{\imgid}{image_47}
\newcommand{\imgnote}{047}
\begin{adjustbox}{valign=t}
\begin{tabular}{c}
\includegraphics[width=0.237\textwidth,height=0.1845\textwidth]{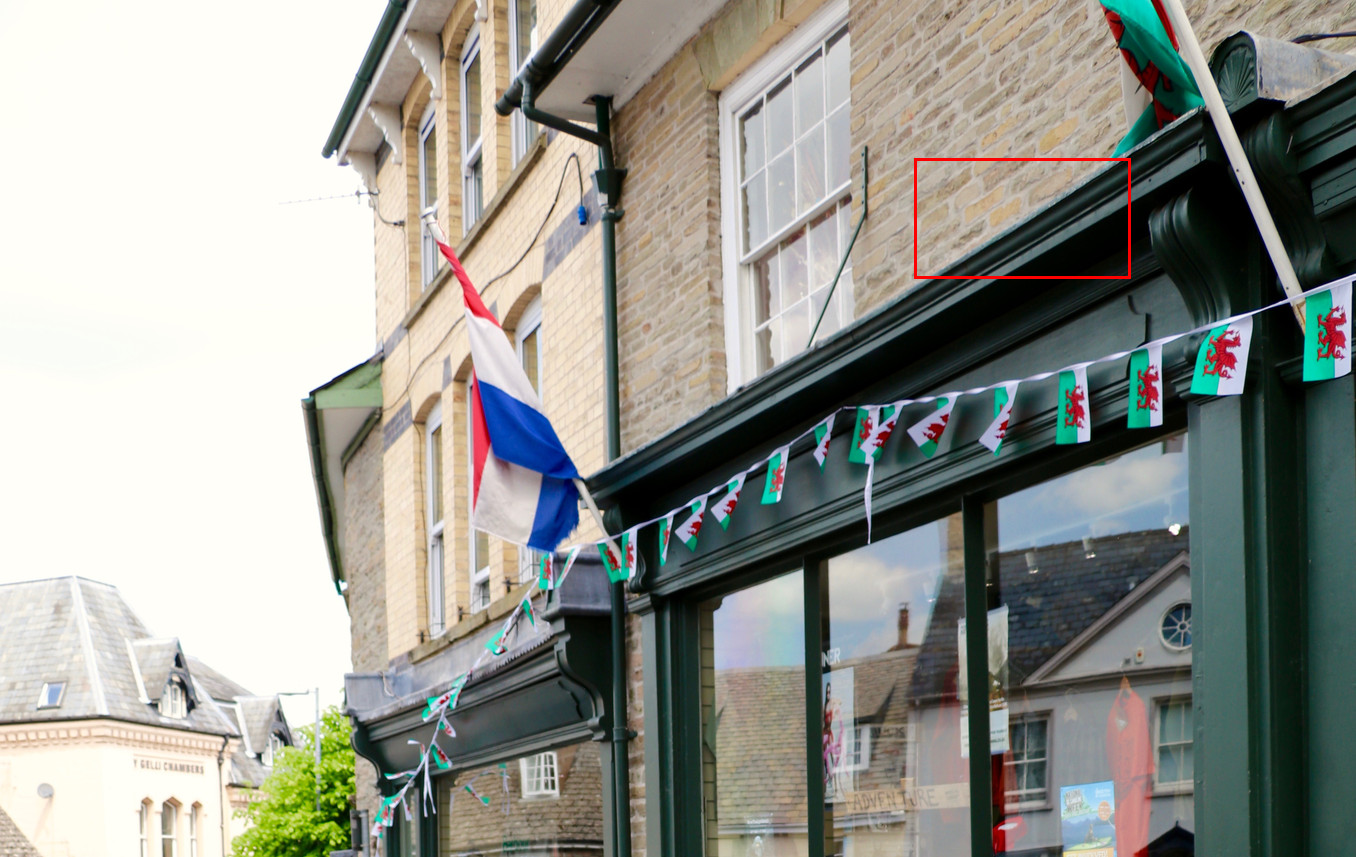} \\
DIV2K: \texttt{img\_\imgnote}
\end{tabular}
\end{adjustbox}
\hspace{-0.2cm}
\begin{adjustbox}{valign=t}
\begin{tabular}{cccccc}
\includegraphics[width=0.139\textwidth]{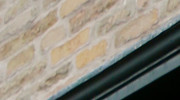} \hspace{-1mm} &
\includegraphics[width=0.139\textwidth]{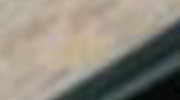} \hspace{-1mm} &
\includegraphics[width=0.139\textwidth]{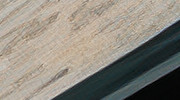} \hspace{-1mm} &
\includegraphics[width=0.139\textwidth]{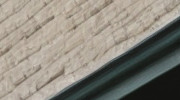} \hspace{-1mm} &
\includegraphics[width=0.139\textwidth]{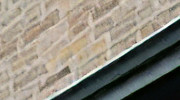} \hspace{-1mm} \\
HR \hspace{-1mm} &
Bicubic \hspace{-1mm} &
CTMSR~\cite{you2025ctmsr} \hspace{-1mm} &
PiSA-SR~\cite{sun2024pisasr} \hspace{-1mm} &
TSD-SR~\cite{dong2025tsdsr} \hspace{-1mm} \\
\includegraphics[width=0.139\textwidth]{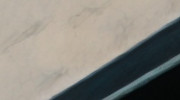} \hspace{-1mm} &
\includegraphics[width=0.139\textwidth]{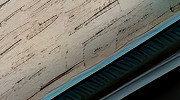} \hspace{-1mm} &
\includegraphics[width=0.139\textwidth]{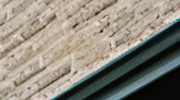} \hspace{-1mm} &
\includegraphics[width=0.139\textwidth]{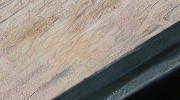} \hspace{-1mm} &
\includegraphics[width=0.139\textwidth]{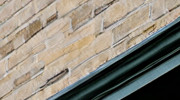} \hspace{-1mm} \\
OSEDiff~\cite{wu2024osediff} \hspace{-1mm} &
InvSR~\cite{yue2025invsr} \hspace{-1mm} &
HYPIR~\cite{lin2025hypir} \hspace{-1mm} &
SinSR~\cite{wang2024sinsr}  \hspace{-1mm} &
StrSR (ours) \hspace{-1mm} \\
\end{tabular}
\end{adjustbox}
\\

\hspace{-0.4cm}
\newcommand{\imgid}{image_55}
\newcommand{\imgnote}{055}
\begin{adjustbox}{valign=t}
\begin{tabular}{c}
\includegraphics[width=0.237\textwidth,height=0.1845\textwidth]{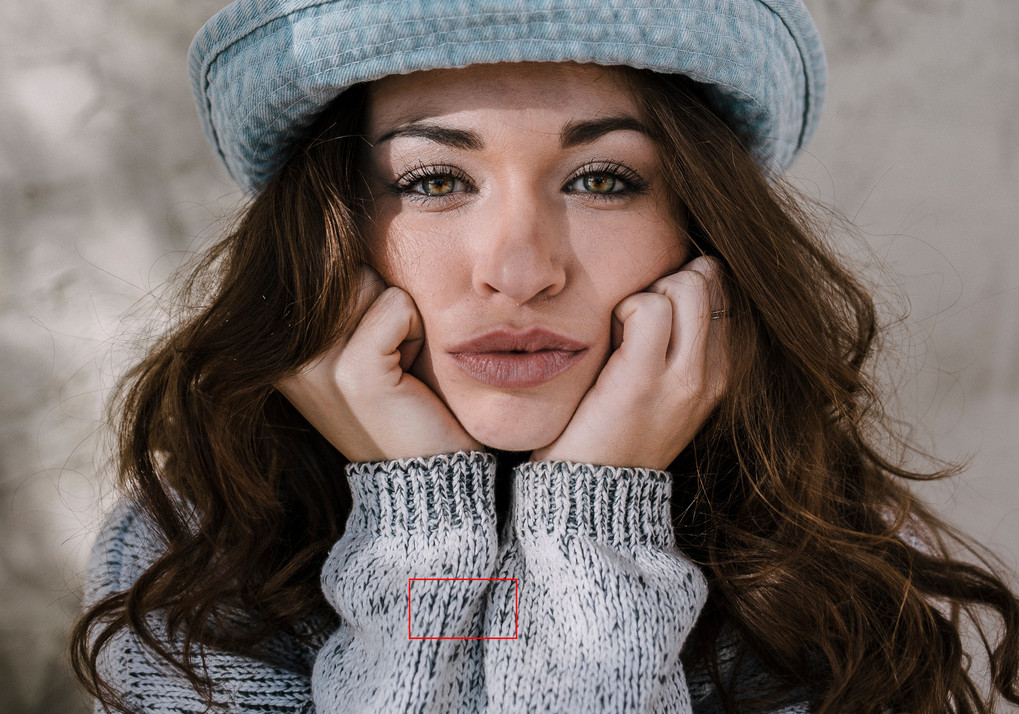} \\
DIV2K: \texttt{img\_\imgnote}
\end{tabular}
\end{adjustbox}
\hspace{-0.2cm}
\begin{adjustbox}{valign=t}
\begin{tabular}{cccccc}
\includegraphics[width=0.139\textwidth]{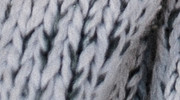} \hspace{-1mm} &
\includegraphics[width=0.139\textwidth]{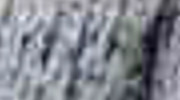} \hspace{-1mm} &
\includegraphics[width=0.139\textwidth]{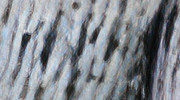} \hspace{-1mm} &
\includegraphics[width=0.139\textwidth]{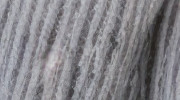} \hspace{-1mm} &
\includegraphics[width=0.139\textwidth]{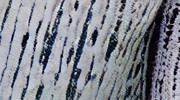} \hspace{-1mm} \\
HR \hspace{-1mm} &
Bicubic \hspace{-1mm} &
CTMSR~\cite{you2025ctmsr} \hspace{-1mm} &
PiSA-SR~\cite{sun2024pisasr} \hspace{-1mm} &
TSD-SR~\cite{dong2025tsdsr} \hspace{-1mm} \\
\includegraphics[width=0.139\textwidth]{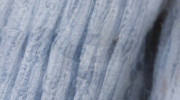} \hspace{-1mm} &
\includegraphics[width=0.139\textwidth]{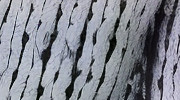} \hspace{-1mm} &
\includegraphics[width=0.139\textwidth]{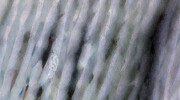} \hspace{-1mm} &
\includegraphics[width=0.139\textwidth]{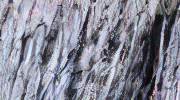} \hspace{-1mm} &
\includegraphics[width=0.139\textwidth]{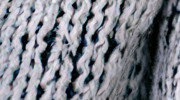} \hspace{-1mm} \\
OSEDiff~\cite{wu2024osediff} \hspace{-1mm} &
InvSR~\cite{yue2025invsr} \hspace{-1mm} &
HYPIR~\cite{lin2025hypir} \hspace{-1mm} &
SinSR~\cite{wang2024sinsr}  \hspace{-1mm} &
StrSR (ours) \hspace{-1mm} \\
\end{tabular}
\end{adjustbox}
\\

\end{tabular}

\caption{More visual comparison for image SR ($\times$4) in DIV2K-val dataset.}
\label{fig:vis-DIV2K-1}
\vspace{-4mm}
\end{figure*}

\begin{figure*}[t]
\scriptsize
\centering

\begin{tabular}{cccc}

\hspace{-0.4cm}
\newcommand{\imgid}{image_57}
\newcommand{\imgnote}{057}
\begin{adjustbox}{valign=t}
\begin{tabular}{c}
\includegraphics[width=0.237\textwidth,height=0.1845\textwidth]{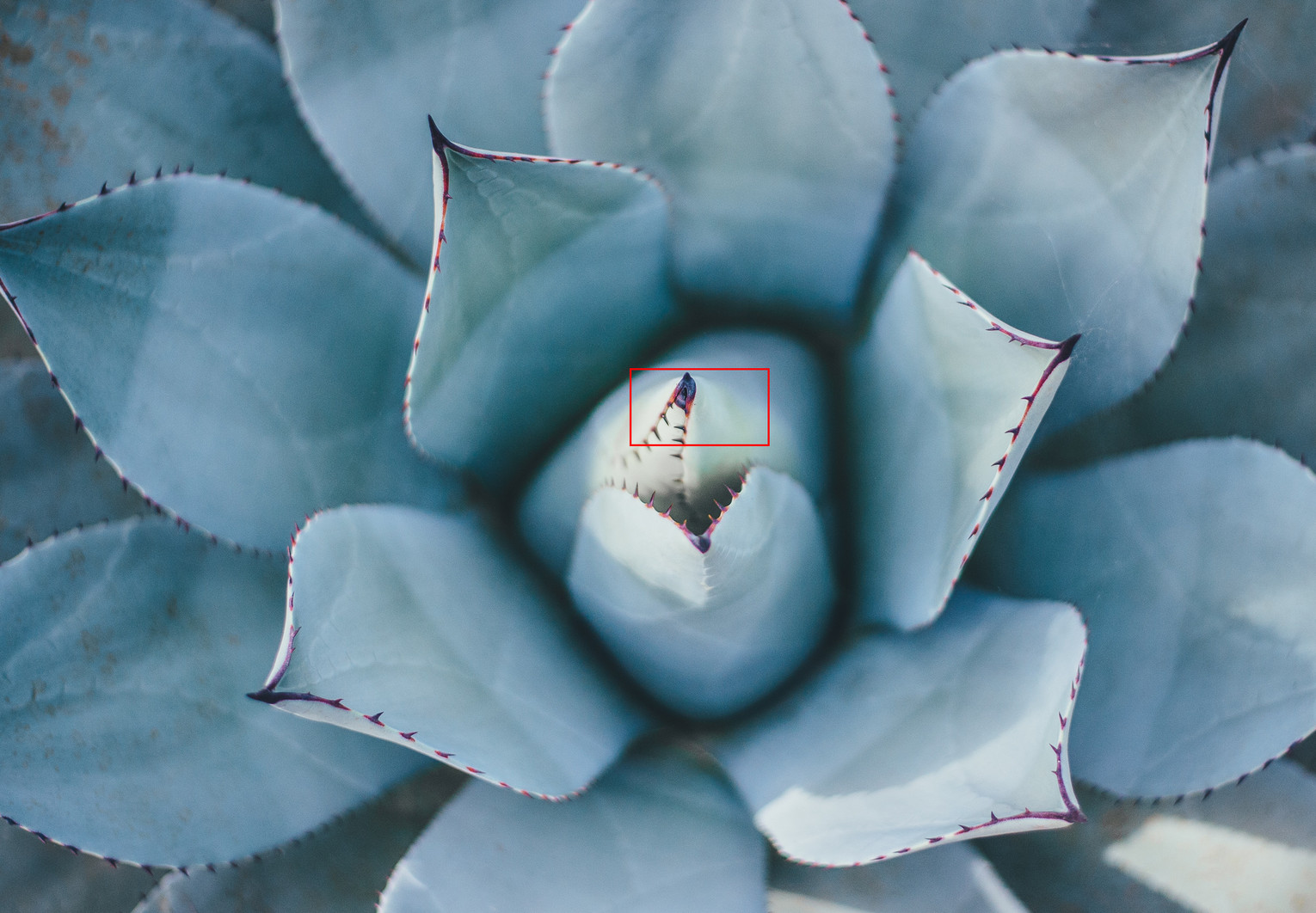} \\
DIV2K: \texttt{img\_\imgnote}
\end{tabular}
\end{adjustbox}
\hspace{-0.2cm}
\begin{adjustbox}{valign=t}
\begin{tabular}{cccccc}
\includegraphics[width=0.139\textwidth]{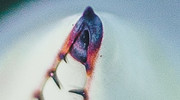} \hspace{-1mm} &
\includegraphics[width=0.139\textwidth]{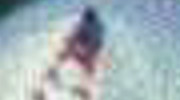} \hspace{-1mm} &
\includegraphics[width=0.139\textwidth]{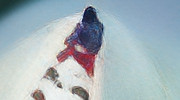} \hspace{-1mm} &
\includegraphics[width=0.139\textwidth]{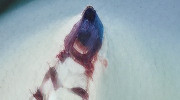} \hspace{-1mm} &
\includegraphics[width=0.139\textwidth]{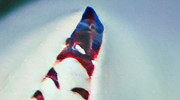} \hspace{-1mm} \\
HR \hspace{-1mm} &
Bicubic \hspace{-1mm} &
CTMSR~\cite{you2025ctmsr} \hspace{-1mm} &
PiSA-SR~\cite{sun2024pisasr} \hspace{-1mm} &
TSD-SR~\cite{dong2025tsdsr} \hspace{-1mm} \\
\includegraphics[width=0.139\textwidth]{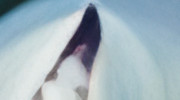} \hspace{-1mm} &
\includegraphics[width=0.139\textwidth]{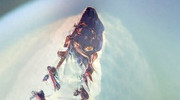} \hspace{-1mm} &
\includegraphics[width=0.139\textwidth]{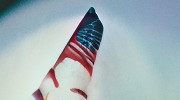} \hspace{-1mm} &
\includegraphics[width=0.139\textwidth]{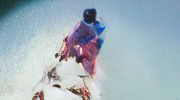} \hspace{-1mm} &
\includegraphics[width=0.139\textwidth]{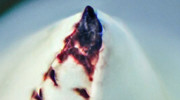} \hspace{-1mm} \\
OSEDiff~\cite{wu2024osediff} \hspace{-1mm} &
InvSR~\cite{yue2025invsr} \hspace{-1mm} &
HYPIR~\cite{lin2025hypir} \hspace{-1mm} &
SinSR~\cite{wang2024sinsr}  \hspace{-1mm} &
StrSR (ours) \hspace{-1mm} \\
\end{tabular}
\end{adjustbox}
\\

\hspace{-0.4cm}
\newcommand{\imgid}{image_59}
\newcommand{\imgnote}{059}
\begin{adjustbox}{valign=t}
\begin{tabular}{c}
\includegraphics[width=0.237\textwidth,height=0.1845\textwidth]{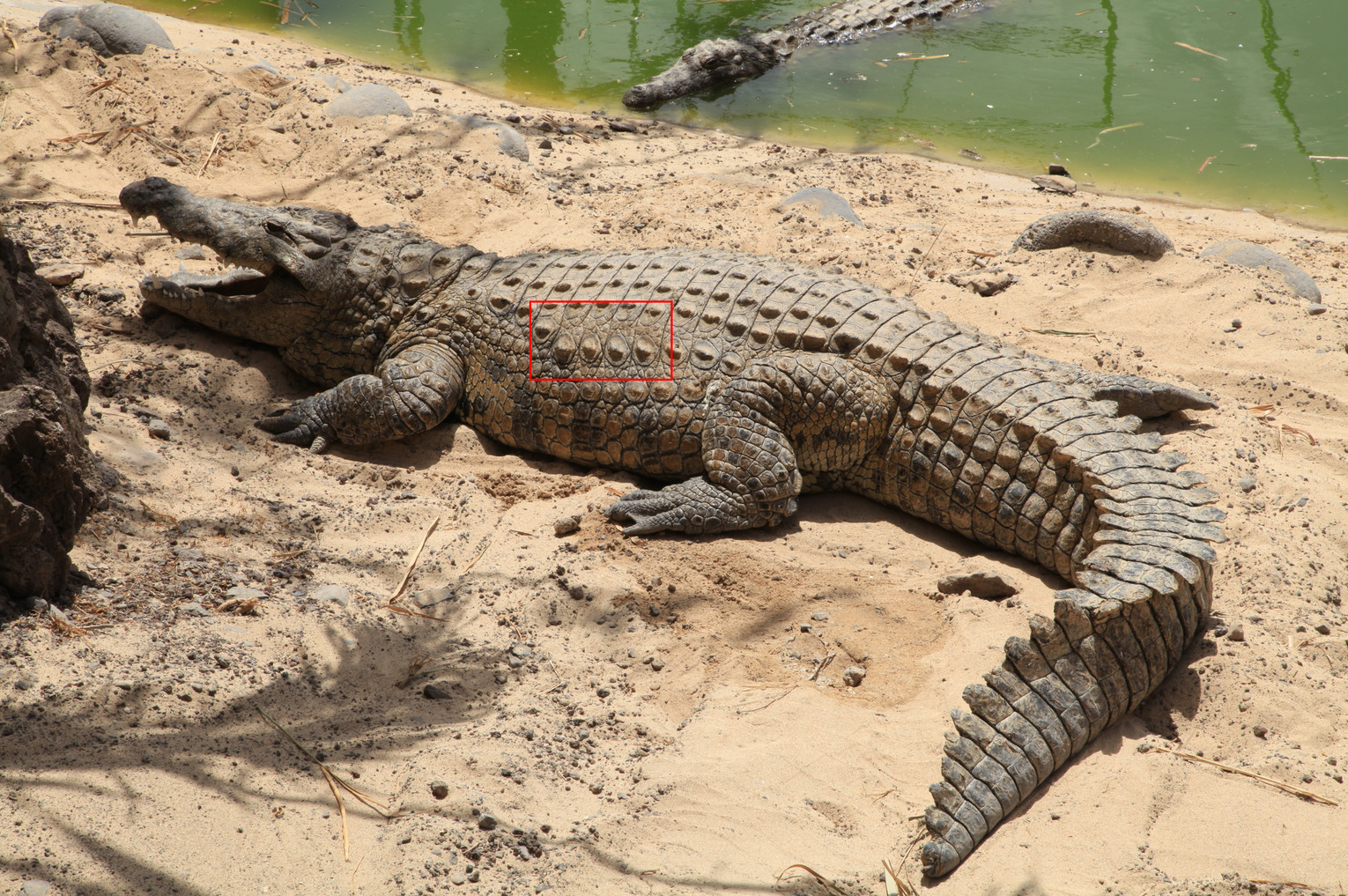} \\
DIV2K: \texttt{img\_\imgnote}
\end{tabular}
\end{adjustbox}
\hspace{-0.2cm}
\begin{adjustbox}{valign=t}
\begin{tabular}{cccccc}
\includegraphics[width=0.139\textwidth]{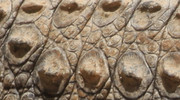} \hspace{-1mm} &
\includegraphics[width=0.139\textwidth]{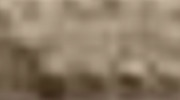} \hspace{-1mm} &
\includegraphics[width=0.139\textwidth]{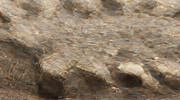} \hspace{-1mm} &
\includegraphics[width=0.139\textwidth]{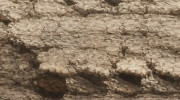} \hspace{-1mm} &
\includegraphics[width=0.139\textwidth]{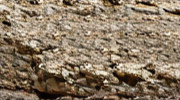} \hspace{-1mm} \\
HR \hspace{-1mm} &
Bicubic \hspace{-1mm} &
CTMSR~\cite{you2025ctmsr} \hspace{-1mm} &
PiSA-SR~\cite{sun2024pisasr} \hspace{-1mm} &
TSD-SR~\cite{dong2025tsdsr} \hspace{-1mm} \\
\includegraphics[width=0.139\textwidth]{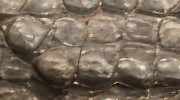} \hspace{-1mm} &
\includegraphics[width=0.139\textwidth]{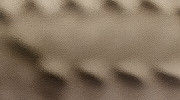} \hspace{-1mm} &
\includegraphics[width=0.139\textwidth]{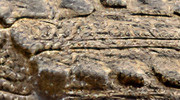} \hspace{-1mm} &
\includegraphics[width=0.139\textwidth]{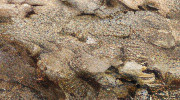} \hspace{-1mm} &
\includegraphics[width=0.139\textwidth]{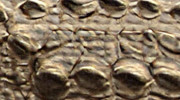} \hspace{-1mm} \\
OSEDiff~\cite{wu2024osediff} \hspace{-1mm} &
InvSR~\cite{yue2025invsr} \hspace{-1mm} &
HYPIR~\cite{lin2025hypir} \hspace{-1mm} &
SinSR~\cite{wang2024sinsr}  \hspace{-1mm} &
StrSR (ours) \hspace{-1mm} \\
\end{tabular}
\end{adjustbox}
\\

\hspace{-0.4cm}
\newcommand{\imgid}{image_73}
\newcommand{\imgnote}{073}
\begin{adjustbox}{valign=t}
\begin{tabular}{c}
\includegraphics[width=0.237\textwidth,height=0.1845\textwidth]{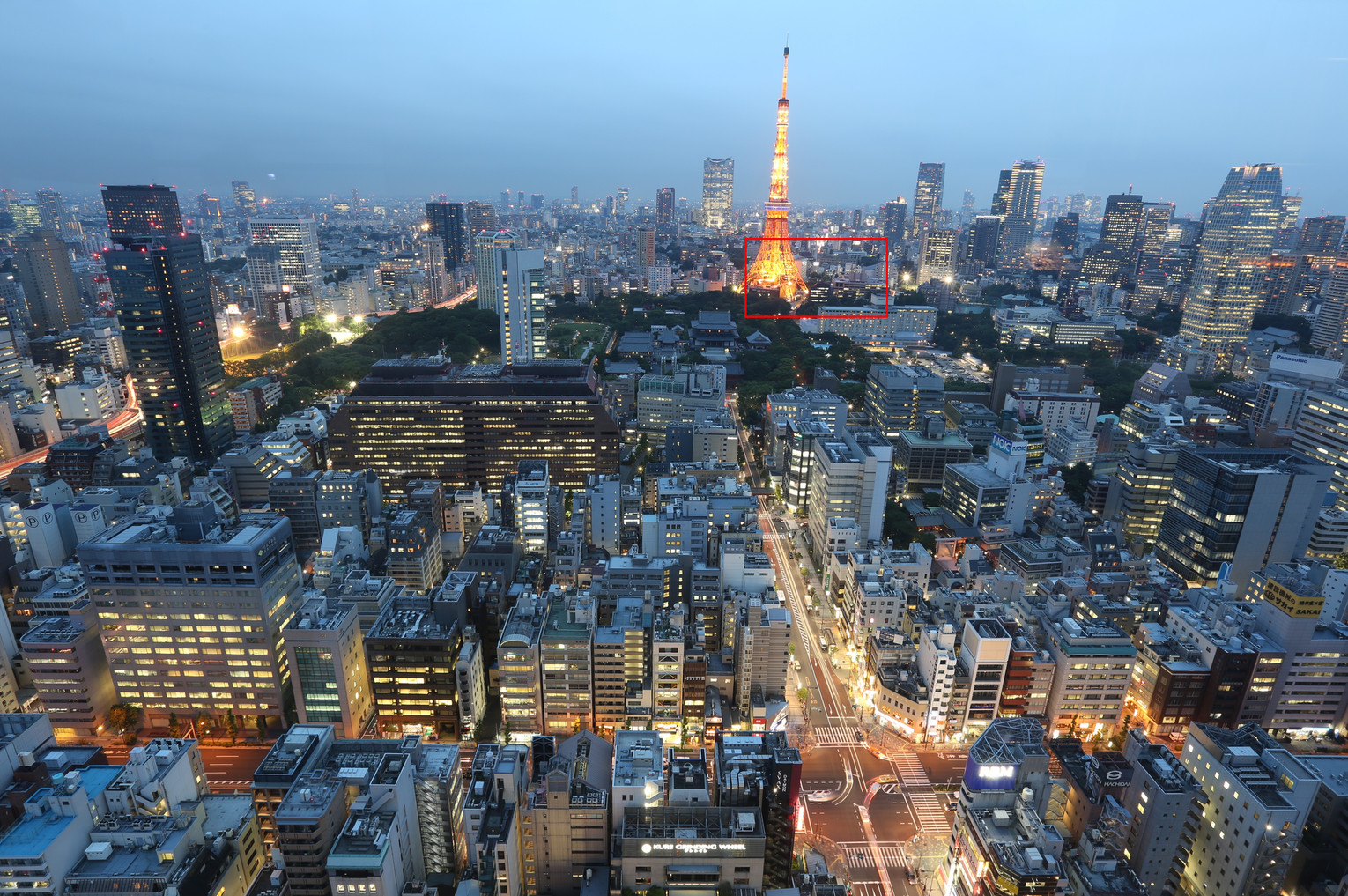} \\
DIV2K: \texttt{img\_\imgnote}
\end{tabular}
\end{adjustbox}
\hspace{-0.2cm}
\begin{adjustbox}{valign=t}
\begin{tabular}{cccccc}
\includegraphics[width=0.139\textwidth]{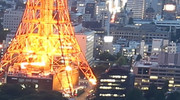} \hspace{-1mm} &
\includegraphics[width=0.139\textwidth]{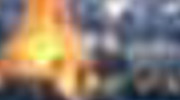} \hspace{-1mm} &
\includegraphics[width=0.139\textwidth]{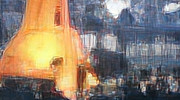} \hspace{-1mm} &
\includegraphics[width=0.139\textwidth]{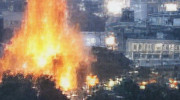} \hspace{-1mm} &
\includegraphics[width=0.139\textwidth]{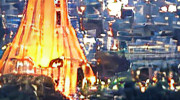} \hspace{-1mm} \\
HR \hspace{-1mm} &
Bicubic \hspace{-1mm} &
CTMSR~\cite{you2025ctmsr} \hspace{-1mm} &
PiSA-SR~\cite{sun2024pisasr} \hspace{-1mm} &
TSD-SR~\cite{dong2025tsdsr} \hspace{-1mm} \\
\includegraphics[width=0.139\textwidth]{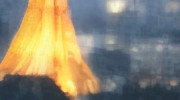} \hspace{-1mm} &
\includegraphics[width=0.139\textwidth]{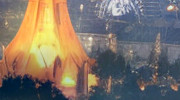} \hspace{-1mm} &
\includegraphics[width=0.139\textwidth]{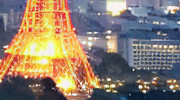} \hspace{-1mm} &
\includegraphics[width=0.139\textwidth]{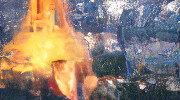} \hspace{-1mm} &
\includegraphics[width=0.139\textwidth]{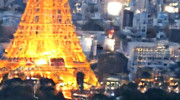} \hspace{-1mm} \\
OSEDiff~\cite{wu2024osediff} \hspace{-1mm} &
InvSR~\cite{yue2025invsr} \hspace{-1mm} &
HYPIR~\cite{lin2025hypir} \hspace{-1mm} &
SinSR~\cite{wang2024sinsr}  \hspace{-1mm} &
{StrSR} (ours) \hspace{-1mm} \\
\end{tabular}
\end{adjustbox}
\\

\hspace{-0.4cm}
\newcommand{\imgid}{image_82}
\newcommand{\imgnote}{082}
\begin{adjustbox}{valign=t}
\begin{tabular}{c}
\includegraphics[width=0.237\textwidth,height=0.1845\textwidth]{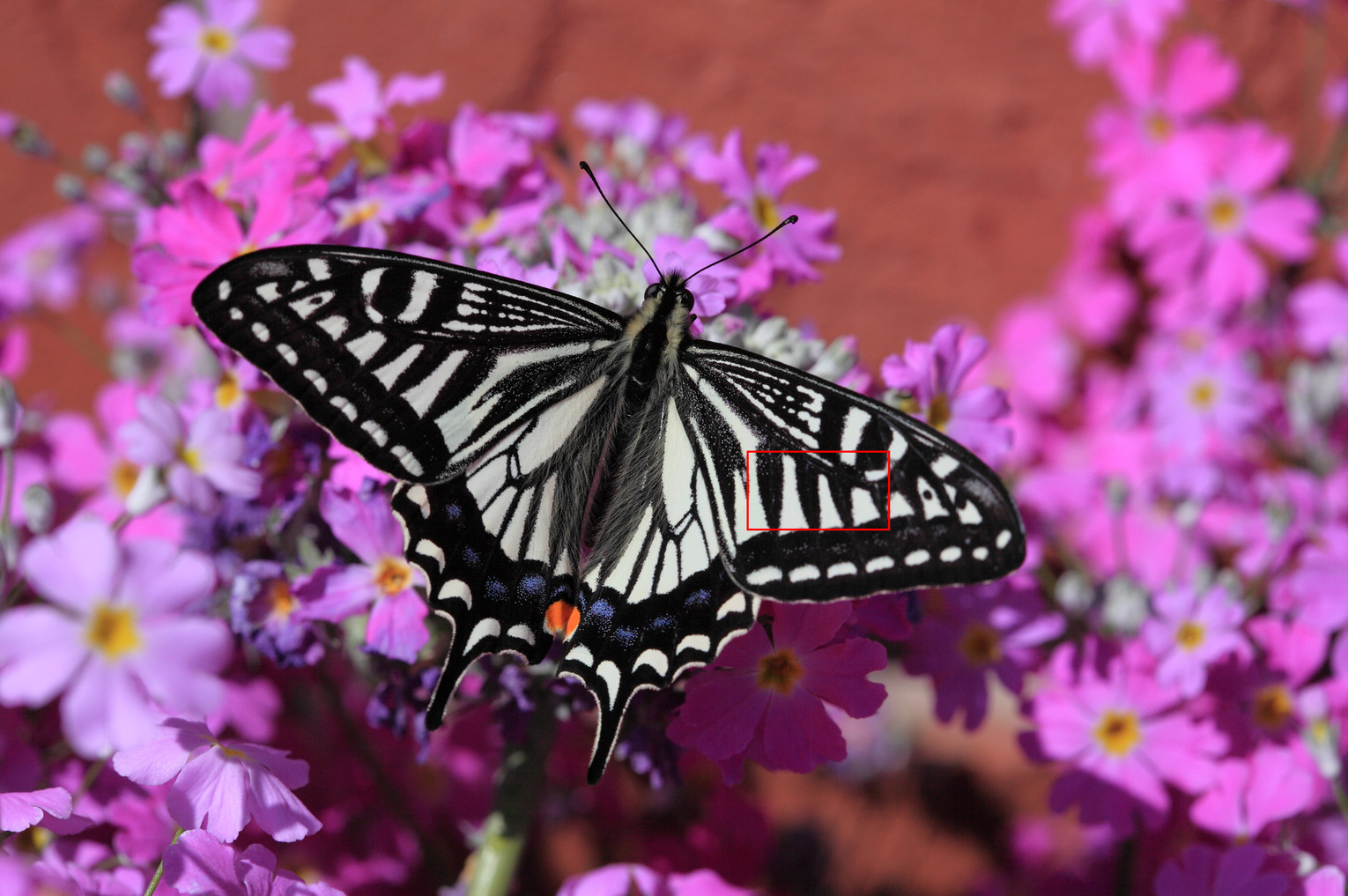} \\
DIV2K: \texttt{img\_\imgnote}
\end{tabular}
\end{adjustbox}
\hspace{-0.2cm}
\begin{adjustbox}{valign=t}
\begin{tabular}{cccccc}
\includegraphics[width=0.139\textwidth]{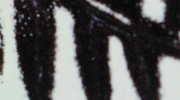} \hspace{-1mm} &
\includegraphics[width=0.139\textwidth]{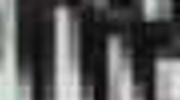} \hspace{-1mm} &
\includegraphics[width=0.139\textwidth]{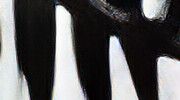} \hspace{-1mm} &
\includegraphics[width=0.139\textwidth]{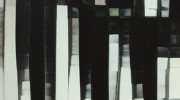} \hspace{-1mm} &
\includegraphics[width=0.139\textwidth]{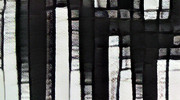} \hspace{-1mm} \\
HR \hspace{-1mm} &
Bicubic \hspace{-1mm} &
CTMSR~\cite{you2025ctmsr} \hspace{-1mm} &
PiSA-SR~\cite{sun2024pisasr} \hspace{-1mm} &
TSD-SR~\cite{dong2025tsdsr} \hspace{-1mm} \\
\includegraphics[width=0.139\textwidth]{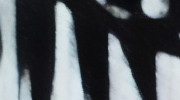} \hspace{-1mm} &
\includegraphics[width=0.139\textwidth]{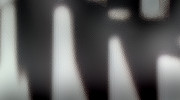} \hspace{-1mm} &
\includegraphics[width=0.139\textwidth]{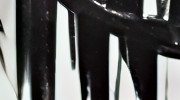} \hspace{-1mm} &
\includegraphics[width=0.139\textwidth]{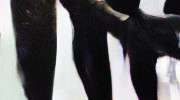} \hspace{-1mm} &
\includegraphics[width=0.139\textwidth]{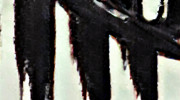} \hspace{-1mm} \\
OSEDiff~\cite{wu2024osediff} \hspace{-1mm} &
InvSR~\cite{yue2025invsr} \hspace{-1mm} &
HYPIR~\cite{lin2025hypir} \hspace{-1mm} &
SinSR~\cite{wang2024sinsr}  \hspace{-1mm} &
StrSR (ours) \hspace{-1mm} \\
\end{tabular}
\end{adjustbox}
\\

\hspace{-0.4cm}
\newcommand{\imgid}{image_94}
\newcommand{\imgnote}{094}
\begin{adjustbox}{valign=t}
\begin{tabular}{c}
\includegraphics[width=0.237\textwidth,height=0.1845\textwidth]{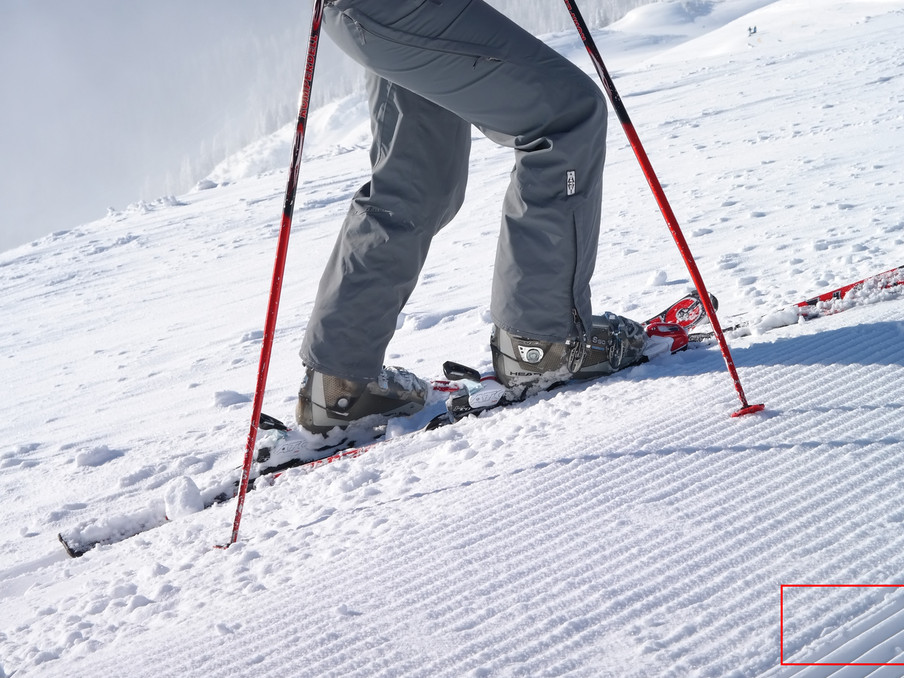} \\
DIV2K: \texttt{img\_\imgnote}
\end{tabular}
\end{adjustbox}
\hspace{-0.2cm}
\begin{adjustbox}{valign=t}
\begin{tabular}{cccccc}
\includegraphics[width=0.139\textwidth]{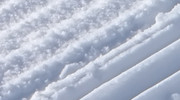} \hspace{-1mm} &
\includegraphics[width=0.139\textwidth]{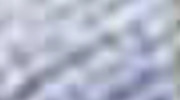} \hspace{-1mm} &
\includegraphics[width=0.139\textwidth]{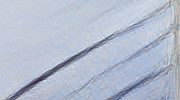} \hspace{-1mm} &
\includegraphics[width=0.139\textwidth]{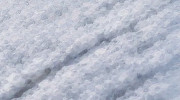} \hspace{-1mm} &
\includegraphics[width=0.139\textwidth]{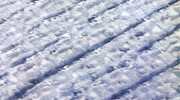} \hspace{-1mm} \\
HR \hspace{-1mm} &
Bicubic \hspace{-1mm} &
CTMSR~\cite{you2025ctmsr} \hspace{-1mm} &
PiSA-SR~\cite{sun2024pisasr} \hspace{-1mm} &
TSD-SR~\cite{dong2025tsdsr} \hspace{-1mm} \\
\includegraphics[width=0.139\textwidth]{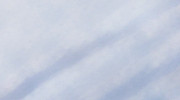} \hspace{-1mm} &
\includegraphics[width=0.139\textwidth]{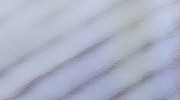} \hspace{-1mm} &
\includegraphics[width=0.139\textwidth]{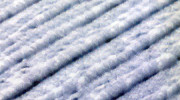} \hspace{-1mm} &
\includegraphics[width=0.139\textwidth]{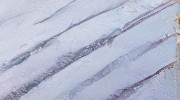} \hspace{-1mm} &
\includegraphics[width=0.139\textwidth]{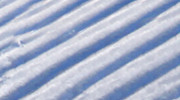} \hspace{-1mm} \\
OSEDiff~\cite{wu2024osediff} \hspace{-1mm} &
InvSR~\cite{yue2025invsr} \hspace{-1mm} &
HYPIR~\cite{lin2025hypir} \hspace{-1mm} &
SinSR~\cite{wang2024sinsr}  \hspace{-1mm} &
StrSR (ours) \hspace{-1mm} \\
\end{tabular}
\end{adjustbox}
\\

\hspace{-0.4cm}
\newcommand{\imgid}{image_96}
\newcommand{\imgnote}{096}
\begin{adjustbox}{valign=t}
\begin{tabular}{c}
\includegraphics[width=0.237\textwidth,height=0.1845\textwidth]{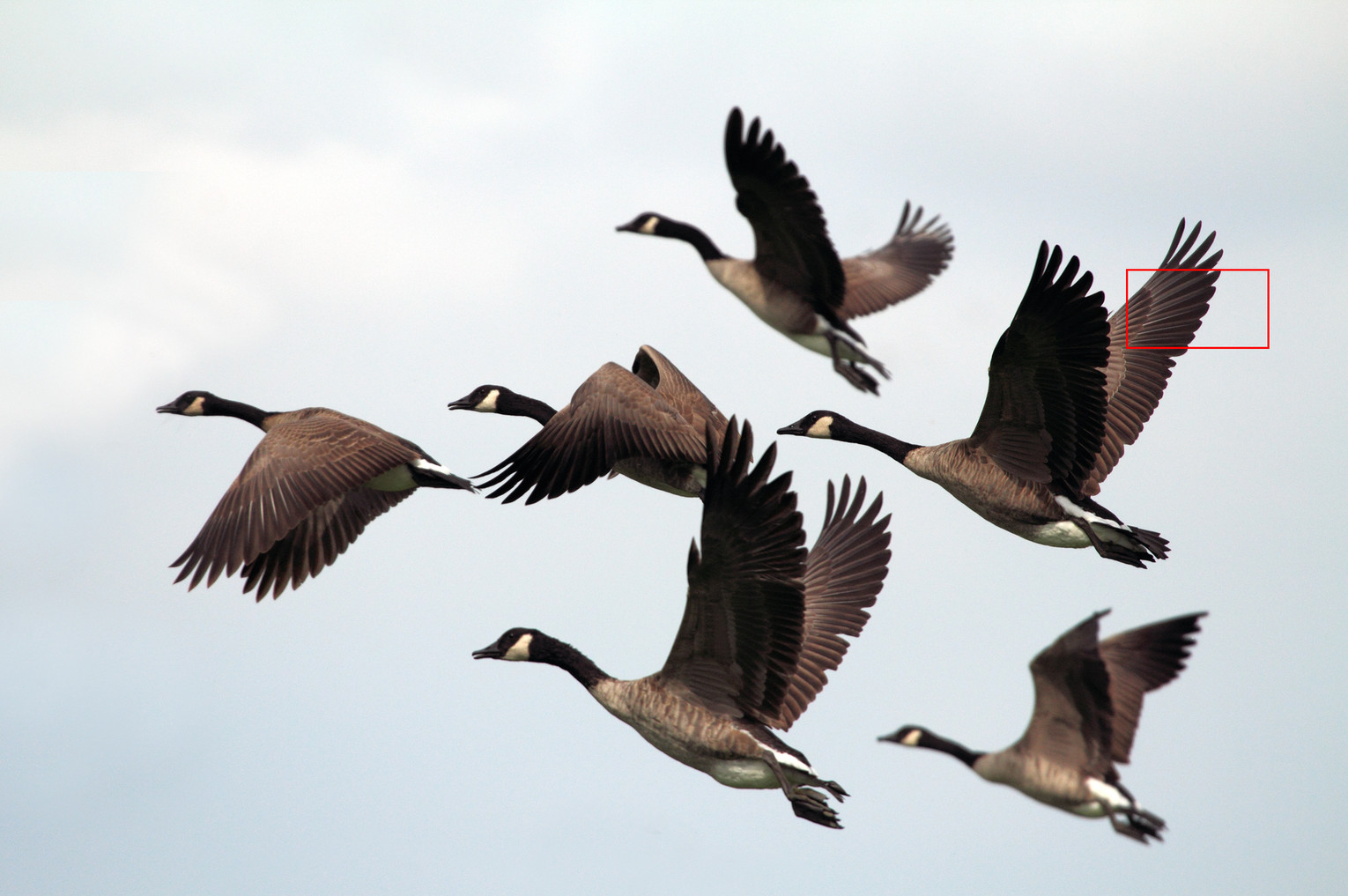} \\
DIV2K: \texttt{img\_\imgnote}
\end{tabular}
\end{adjustbox}
\hspace{-0.2cm}
\begin{adjustbox}{valign=t}
\begin{tabular}{cccccc}
\includegraphics[width=0.139\textwidth]{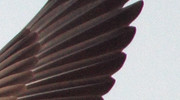} \hspace{-1mm} &
\includegraphics[width=0.139\textwidth]{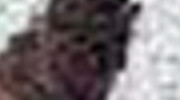} \hspace{-1mm} &
\includegraphics[width=0.139\textwidth]{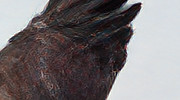} \hspace{-1mm} &
\includegraphics[width=0.139\textwidth]{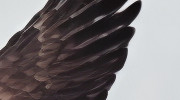} \hspace{-1mm} &
\includegraphics[width=0.139\textwidth]{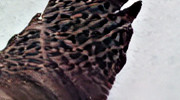} \hspace{-1mm} \\
HR \hspace{-1mm} &
Bicubic \hspace{-1mm} &
CTMSR~\cite{you2025ctmsr} \hspace{-1mm} &
PiSA-SR~\cite{sun2024pisasr} \hspace{-1mm} &
TSD-SR~\cite{dong2025tsdsr} \hspace{-1mm} \\
\includegraphics[width=0.139\textwidth]{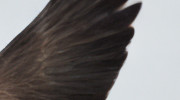} \hspace{-1mm} &
\includegraphics[width=0.139\textwidth]{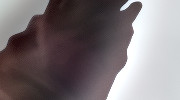} \hspace{-1mm} &
\includegraphics[width=0.139\textwidth]{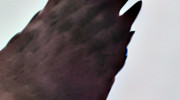} \hspace{-1mm} &
\includegraphics[width=0.139\textwidth]{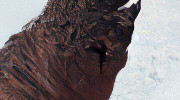} \hspace{-1mm} &
\includegraphics[width=0.139\textwidth]{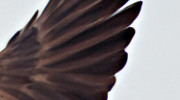} \hspace{-1mm} \\
OSEDiff~\cite{wu2024osediff} \hspace{-1mm} &
InvSR~\cite{yue2025invsr} \hspace{-1mm} &
HYPIR~\cite{lin2025hypir} \hspace{-1mm} &
SinSR~\cite{wang2024sinsr}  \hspace{-1mm} &
StrSR (ours) \hspace{-1mm} \\
\end{tabular}
\end{adjustbox}
\\

\hspace{-0.4cm}
\newcommand{\imgid}{image_98}
\newcommand{\imgnote}{098}
\begin{adjustbox}{valign=t}
\begin{tabular}{c}
\includegraphics[width=0.237\textwidth,height=0.1845\textwidth]{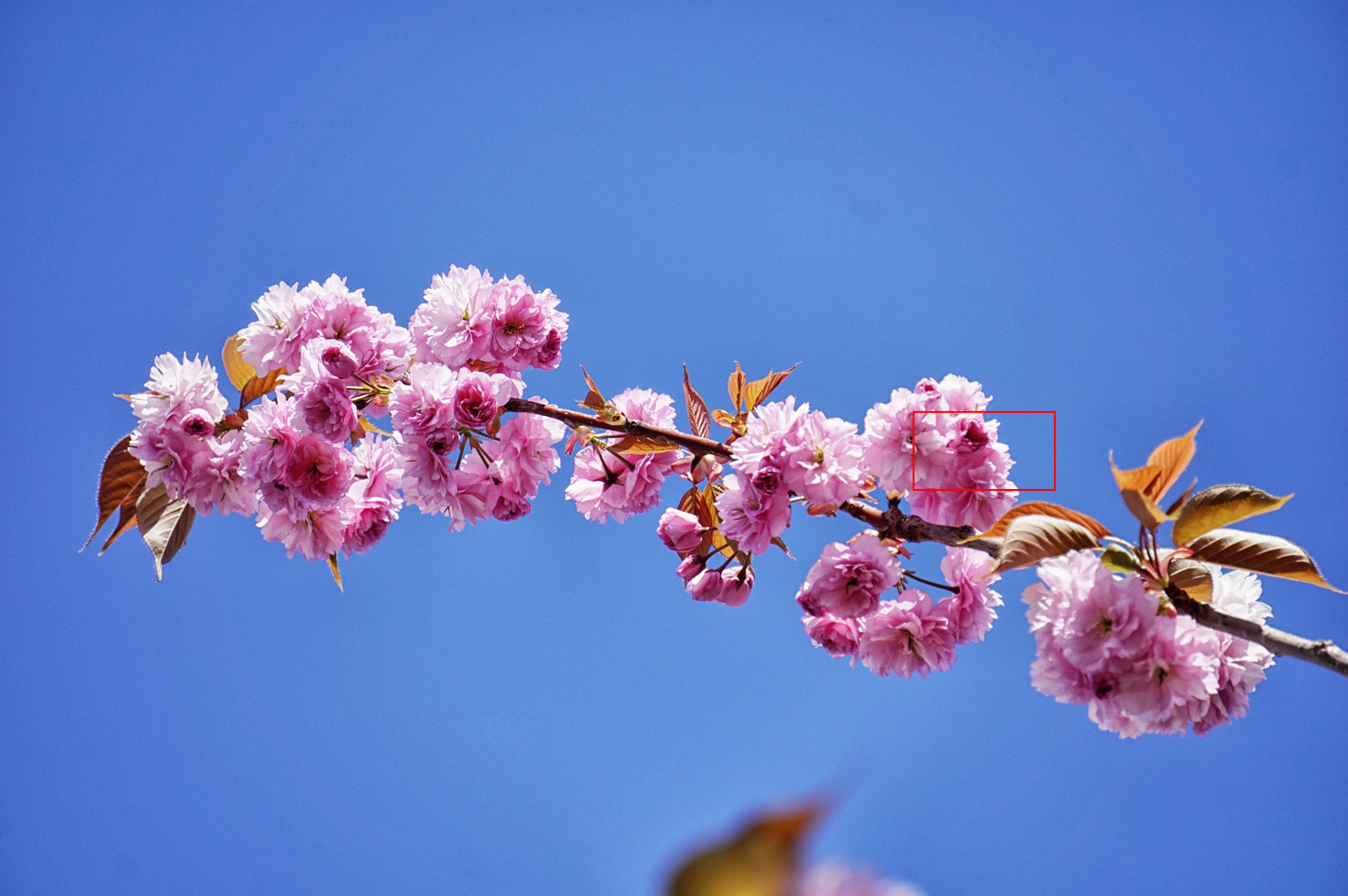} \\
DIV2K: \texttt{img\_\imgnote}
\end{tabular}
\end{adjustbox}
\hspace{-0.2cm}
\begin{adjustbox}{valign=t}
\begin{tabular}{cccccc}
\includegraphics[width=0.139\textwidth]{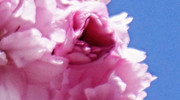} \hspace{-1mm} &
\includegraphics[width=0.139\textwidth]{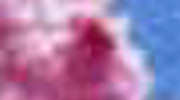} \hspace{-1mm} &
\includegraphics[width=0.139\textwidth]{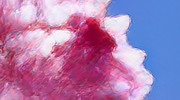} \hspace{-1mm} &
\includegraphics[width=0.139\textwidth]{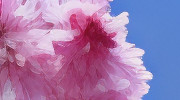} \hspace{-1mm} &
\includegraphics[width=0.139\textwidth]{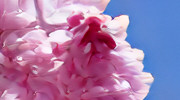} \hspace{-1mm} \\
HR \hspace{-1mm} &
Bicubic \hspace{-1mm} &
CTMSR~\cite{you2025ctmsr} \hspace{-1mm} &
PiSA-SR~\cite{sun2024pisasr} \hspace{-1mm} &
TSD-SR~\cite{dong2025tsdsr} \hspace{-1mm} \\
\includegraphics[width=0.139\textwidth]{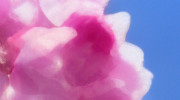} \hspace{-1mm} &
\includegraphics[width=0.139\textwidth]{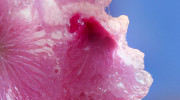} \hspace{-1mm} &
\includegraphics[width=0.139\textwidth]{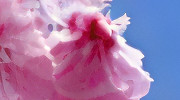} \hspace{-1mm} &
\includegraphics[width=0.139\textwidth]{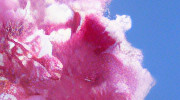} \hspace{-1mm} &
\includegraphics[width=0.139\textwidth]{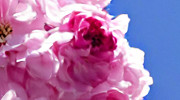} \hspace{-1mm} \\
OSEDiff~\cite{wu2024osediff} \hspace{-1mm} &
InvSR~\cite{yue2025invsr} \hspace{-1mm} &
HYPIR~\cite{lin2025hypir} \hspace{-1mm} &
SinSR~\cite{wang2024sinsr}  \hspace{-1mm} &
StrSR (ours) \hspace{-1mm} \\
\end{tabular}
\end{adjustbox}
\\

\end{tabular}

\caption{More visual comparison for image SR ($\times$4) in DIV2K-val dataset.}
\label{fig:vis-DIV2K-2}
\vspace{-4mm}
\end{figure*}

\vspace{-1mm}
For the synthetic dataset DIV2K-val in Figs.~\ref{fig:vis-DIV2K-1} and~\ref{fig:vis-DIV2K-2}, StrSR shows clear advantages in both semantics and details. Semantically, StrSR correctly restores objects such as clothing (\texttt{img\_036}), animals (\texttt{img\_042}), wall tiles (\texttt{img\_047}), butterfly wings (\texttt{img\_082}), and snow (\texttt{img\_094}). The restored images do not contain meaningless color patches. Instead, the model uses actual semantic information to recover detailed textures. Regarding fine details, StrSR fully utilizes the semantic information from the pretrained model to generate realistic textures for objects like the sweater (\texttt{img\_055}), buildings and towers (\texttt{img\_073}), bird wings (\texttt{img\_096}), and petals (\texttt{img\_098}). Notably, some HR ground-truth images lack sharp textures in certain areas due to focus limits. Our StrSR generates textures that are even clearer than the original HR images in areas like the wall in \texttt{img\_047} and the sweater in \texttt{img\_055}.

\begin{figure*}[t]
\scriptsize
\centering

\begin{tabular}{cccc}

\hspace{-0.4cm}
\newcommand{\imgid}{Canon\_006}
\newcommand{\imgnote}{Canon\_006}
\begin{adjustbox}{valign=t}
\begin{tabular}{c}
\includegraphics[width=0.237\textwidth,height=0.1845\textwidth]{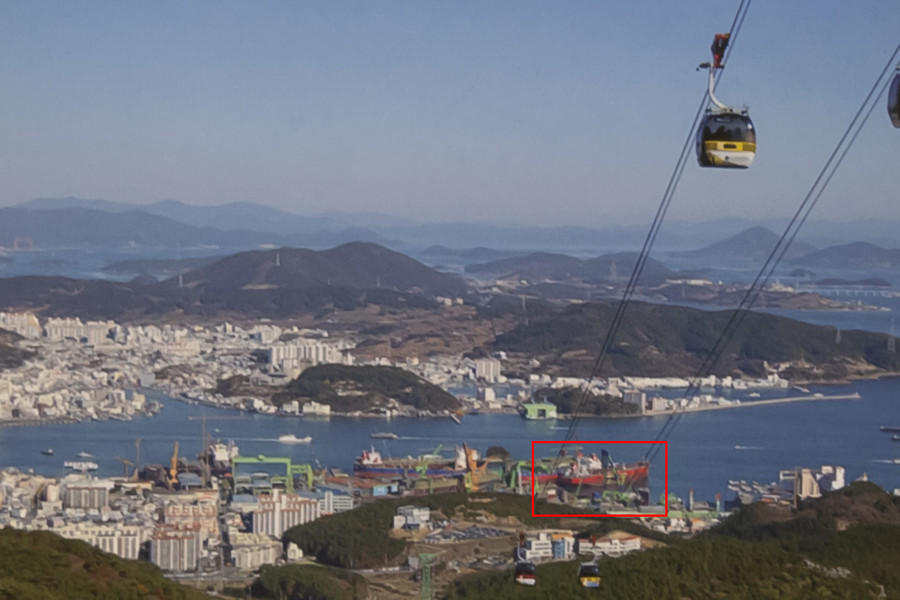} \\
RealSR: \texttt{\imgnote}
\end{tabular}
\end{adjustbox}
\hspace{-0.2cm}
\begin{adjustbox}{valign=t}
\begin{tabular}{cccccc}
\includegraphics[width=0.139\textwidth]{figs/jpg/comparison/RealSR_cropped/\imgid_HR.jpg} \hspace{-1mm} &
\includegraphics[width=0.139\textwidth]{figs/jpg/comparison/RealSR_cropped/\imgid.jpg} \hspace{-1mm} &
\includegraphics[width=0.139\textwidth]{figs/jpg/comparison/RealSR_cropped/\imgid_CTMSR.jpg} \hspace{-1mm} &
\includegraphics[width=0.139\textwidth]{figs/jpg/comparison/RealSR_cropped/\imgid_PiSASR.jpg} \hspace{-1mm} &
\includegraphics[width=0.139\textwidth]{figs/jpg/comparison/RealSR_cropped/\imgid_tsdsr.jpg} \hspace{-1mm} \\
HR \hspace{-1mm} &
Bicubic \hspace{-1mm} &
CTMSR~\cite{you2025ctmsr} \hspace{-1mm} &
PiSA-SR~\cite{sun2024pisasr} \hspace{-1mm} &
TSD-SR~\cite{dong2025tsdsr} \hspace{-1mm} \\
\includegraphics[width=0.139\textwidth]{figs/jpg/comparison/RealSR_cropped/\imgid_OSEDiff.jpg} \hspace{-1mm} &
\includegraphics[width=0.139\textwidth]{figs/jpg/comparison/RealSR_cropped/\imgid_InvSR.jpg} \hspace{-1mm} &
\includegraphics[width=0.139\textwidth]{figs/jpg/comparison/RealSR_cropped/\imgid_HYPIR.jpg} \hspace{-1mm} &
\includegraphics[width=0.139\textwidth]{figs/jpg/comparison/RealSR_cropped/\imgid_SinSR.jpg} \hspace{-1mm} &
\includegraphics[width=0.139\textwidth]{figs/jpg/comparison/RealSR_cropped/\imgid_flux_61k.jpg} \hspace{-1mm} \\
OSEDiff~\cite{wu2024osediff} \hspace{-1mm} &
InvSR~\cite{yue2025invsr} \hspace{-1mm} &
HYPIR~\cite{lin2025hypir} \hspace{-1mm} &
SinSR~\cite{wang2024sinsr}  \hspace{-1mm} &
StrSR (ours) \hspace{-1mm} \\
\end{tabular}
\end{adjustbox}
\\

\hspace{-0.4cm}
\newcommand{\imgid}{Canon\_007}
\newcommand{\imgnote}{Canon\_007}
\begin{adjustbox}{valign=t}
\begin{tabular}{c}
\includegraphics[width=0.237\textwidth,height=0.1845\textwidth]{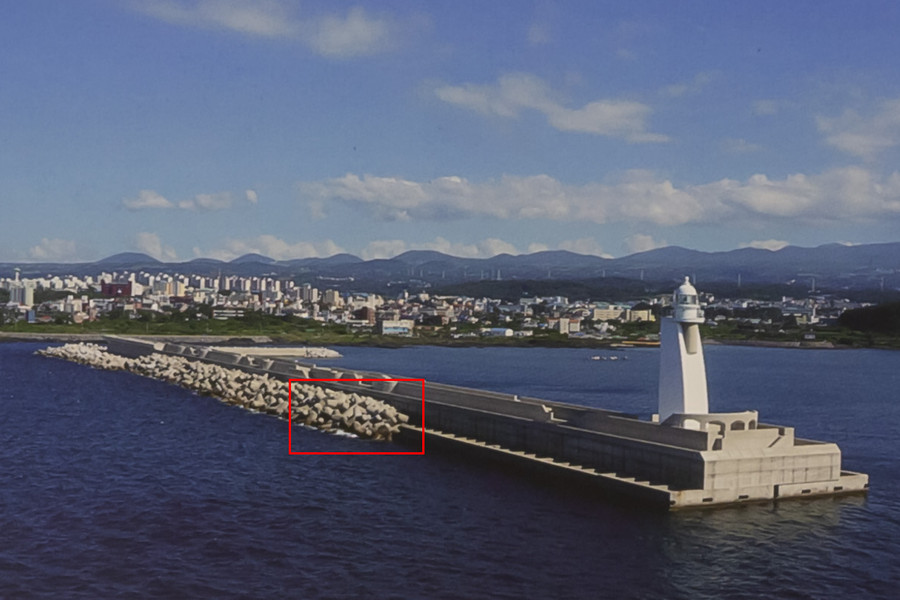} \\
RealSR: \texttt{\imgnote}
\end{tabular}
\end{adjustbox}
\hspace{-0.2cm}
\begin{adjustbox}{valign=t}
\begin{tabular}{cccccc}
\includegraphics[width=0.139\textwidth]{figs/jpg/comparison/RealSR_cropped/\imgid_HR.jpg} \hspace{-1mm} &
\includegraphics[width=0.139\textwidth]{figs/jpg/comparison/RealSR_cropped/\imgid.jpg} \hspace{-1mm} &
\includegraphics[width=0.139\textwidth]{figs/jpg/comparison/RealSR_cropped/\imgid_CTMSR.jpg} \hspace{-1mm} &
\includegraphics[width=0.139\textwidth]{figs/jpg/comparison/RealSR_cropped/\imgid_PiSASR.jpg} \hspace{-1mm} &
\includegraphics[width=0.139\textwidth]{figs/jpg/comparison/RealSR_cropped/\imgid_tsdsr.jpg} \hspace{-1mm} \\
HR \hspace{-1mm} &
Bicubic \hspace{-1mm} &
CTMSR~\cite{you2025ctmsr} \hspace{-1mm} &
PiSA-SR~\cite{sun2024pisasr} \hspace{-1mm} &
TSD-SR~\cite{dong2025tsdsr} \hspace{-1mm} \\
\includegraphics[width=0.139\textwidth]{figs/jpg/comparison/RealSR_cropped/\imgid_OSEDiff.jpg} \hspace{-1mm} &
\includegraphics[width=0.139\textwidth]{figs/jpg/comparison/RealSR_cropped/\imgid_InvSR.jpg} \hspace{-1mm} &
\includegraphics[width=0.139\textwidth]{figs/jpg/comparison/RealSR_cropped/\imgid_HYPIR.jpg} \hspace{-1mm} &
\includegraphics[width=0.139\textwidth]{figs/jpg/comparison/RealSR_cropped/\imgid_SinSR.jpg} \hspace{-1mm} &
\includegraphics[width=0.139\textwidth]{figs/jpg/comparison/RealSR_cropped/\imgid_flux_61k.jpg} \hspace{-1mm} \\
OSEDiff~\cite{wu2024osediff} \hspace{-1mm} &
InvSR~\cite{yue2025invsr} \hspace{-1mm} &
HYPIR~\cite{lin2025hypir} \hspace{-1mm} &
SinSR~\cite{wang2024sinsr}  \hspace{-1mm} &
StrSR (ours) \hspace{-1mm} \\
\end{tabular}
\end{adjustbox}
\\

\hspace{-0.4cm}
\newcommand{\imgid}{Canon\_046}
\newcommand{\imgnote}{Canon\_046}
\begin{adjustbox}{valign=t}
\begin{tabular}{c}
\includegraphics[width=0.237\textwidth,height=0.1845\textwidth]{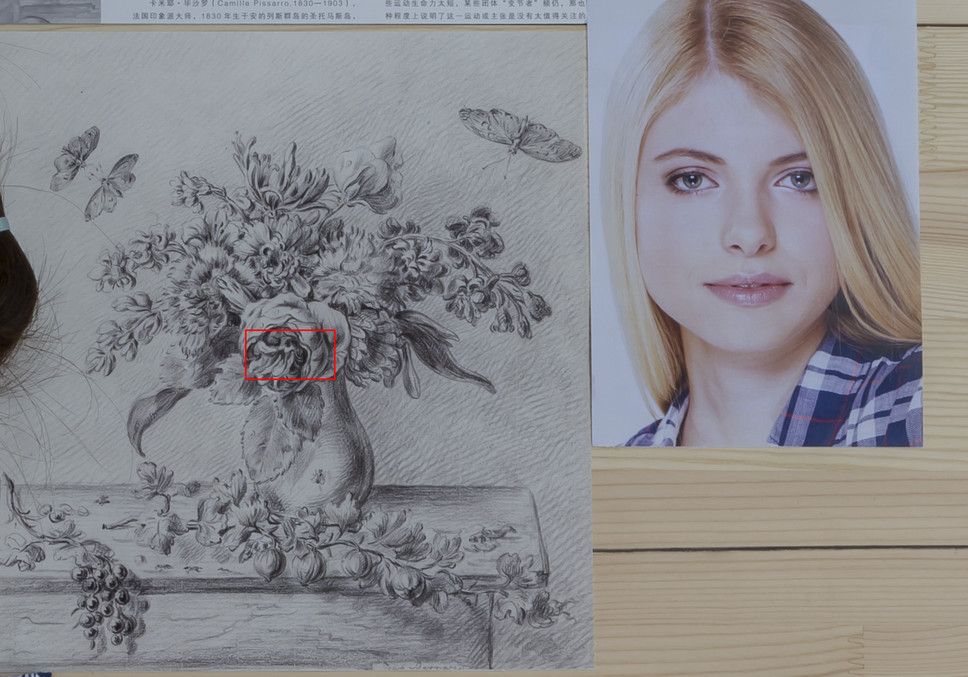} \\
RealSR: \texttt{\imgnote}
\end{tabular}
\end{adjustbox}
\hspace{-0.2cm}
\begin{adjustbox}{valign=t}
\begin{tabular}{cccccc}
\includegraphics[width=0.139\textwidth]{figs/jpg/comparison/RealSR_cropped/\imgid_HR.jpg} \hspace{-1mm} &
\includegraphics[width=0.139\textwidth]{figs/jpg/comparison/RealSR_cropped/\imgid.jpg} \hspace{-1mm} &
\includegraphics[width=0.139\textwidth]{figs/jpg/comparison/RealSR_cropped/\imgid_CTMSR.jpg} \hspace{-1mm} &
\includegraphics[width=0.139\textwidth]{figs/jpg/comparison/RealSR_cropped/\imgid_PiSASR.jpg} \hspace{-1mm} &
\includegraphics[width=0.139\textwidth]{figs/jpg/comparison/RealSR_cropped/\imgid_tsdsr.jpg} \hspace{-1mm} \\
HR \hspace{-1mm} &
Bicubic \hspace{-1mm} &
CTMSR~\cite{you2025ctmsr} \hspace{-1mm} &
PiSA-SR~\cite{sun2024pisasr} \hspace{-1mm} &
TSD-SR~\cite{dong2025tsdsr} \hspace{-1mm} \\
\includegraphics[width=0.139\textwidth]{figs/jpg/comparison/RealSR_cropped/\imgid_OSEDiff.jpg} \hspace{-1mm} &
\includegraphics[width=0.139\textwidth]{figs/jpg/comparison/RealSR_cropped/\imgid_InvSR.jpg} \hspace{-1mm} &
\includegraphics[width=0.139\textwidth]{figs/jpg/comparison/RealSR_cropped/\imgid_HYPIR.jpg} \hspace{-1mm} &
\includegraphics[width=0.139\textwidth]{figs/jpg/comparison/RealSR_cropped/\imgid_SinSR.jpg} \hspace{-1mm} &
\includegraphics[width=0.139\textwidth]{figs/jpg/comparison/RealSR_cropped/\imgid_flux_61k.jpg} \hspace{-1mm} \\
OSEDiff~\cite{wu2024osediff} \hspace{-1mm} &
InvSR~\cite{yue2025invsr} \hspace{-1mm} &
HYPIR~\cite{lin2025hypir} \hspace{-1mm} &
SinSR~\cite{wang2024sinsr}  \hspace{-1mm} &
StrSR (ours) \hspace{-1mm} \\
\end{tabular}
\end{adjustbox}
\\

\hspace{-0.4cm}
\newcommand{\imgid}{Nikon\_048}
\newcommand{\imgnote}{Nikon\_048}
\begin{adjustbox}{valign=t}
\begin{tabular}{c}
\includegraphics[width=0.237\textwidth,height=0.1845\textwidth]{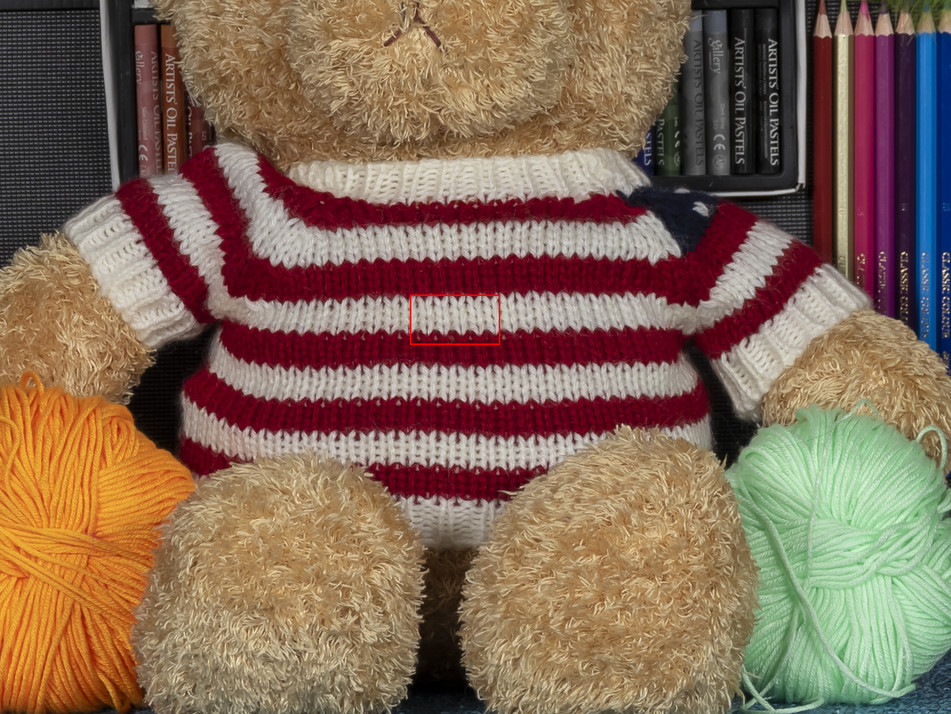} \\
RealSR: \texttt{\imgnote}
\end{tabular}
\end{adjustbox}
\hspace{-0.2cm}
\begin{adjustbox}{valign=t}
\begin{tabular}{cccccc}
\includegraphics[width=0.139\textwidth]{figs/jpg/comparison/RealSR_cropped/\imgid_HR.jpg} \hspace{-1mm} &
\includegraphics[width=0.139\textwidth]{figs/jpg/comparison/RealSR_cropped/\imgid.jpg} \hspace{-1mm} &
\includegraphics[width=0.139\textwidth]{figs/jpg/comparison/RealSR_cropped/\imgid_CTMSR.jpg} \hspace{-1mm} &
\includegraphics[width=0.139\textwidth]{figs/jpg/comparison/RealSR_cropped/\imgid_PiSASR.jpg} \hspace{-1mm} &
\includegraphics[width=0.139\textwidth]{figs/jpg/comparison/RealSR_cropped/\imgid_tsdsr.jpg} \hspace{-1mm} \\
HR \hspace{-1mm} &
Bicubic \hspace{-1mm} &
CTMSR~\cite{you2025ctmsr} \hspace{-1mm} &
PiSA-SR~\cite{sun2024pisasr} \hspace{-1mm} &
TSD-SR~\cite{dong2025tsdsr} \hspace{-1mm} \\
\includegraphics[width=0.139\textwidth]{figs/jpg/comparison/RealSR_cropped/\imgid_OSEDiff.jpg} \hspace{-1mm} &
\includegraphics[width=0.139\textwidth]{figs/jpg/comparison/RealSR_cropped/\imgid_InvSR.jpg} \hspace{-1mm} &
\includegraphics[width=0.139\textwidth]{figs/jpg/comparison/RealSR_cropped/\imgid_HYPIR.jpg} \hspace{-1mm} &
\includegraphics[width=0.139\textwidth]{figs/jpg/comparison/RealSR_cropped/\imgid_SinSR.jpg} \hspace{-1mm} &
\includegraphics[width=0.139\textwidth]{figs/jpg/comparison/RealSR_cropped/\imgid_flux_61k.jpg} \hspace{-1mm} \\
OSEDiff~\cite{wu2024osediff} \hspace{-1mm} &
InvSR~\cite{yue2025invsr} \hspace{-1mm} &
HYPIR~\cite{lin2025hypir} \hspace{-1mm} &
SinSR~\cite{wang2024sinsr}  \hspace{-1mm} &
StrSR (ours) \hspace{-1mm} \\
\end{tabular}
\end{adjustbox}
\\

\end{tabular}
\vspace{-3.2mm}
\caption{More visual comparison for image SR ($\times$4) in RealSR dataset.}
\label{fig:vis-RealSR-1}
\vspace{-3mm}
\end{figure*}

As shown in Fig.~\ref{fig:vis-RealSR-1}, our method also has strong advantages on real-world data. For the container ship in \texttt{Canon\_006}, most compared methods fail to restore the ship. They either produce over-smoothed results like InvSR, create painterly effects like TSD-SR, or introduce heavy noise like CTMSR, PiSA-SR, HYPIR, and SinSR. Meanwhile, StrSR handles semantic details well, as seen in the stones of \texttt{Canon\_007}. StrSR clearly maintains the individual shapes of the stones and renders the shadows effectively. For texture preservation, StrSR perfectly maintains the sweater's texture in \texttt{Nikon\_048}. Each yarn of the sweater is twisted from smaller threads. Only StrSR perfectly presents this material texture without over-smoothing or introducing semantic errors.

\begin{figure}[t]
\scriptsize
\centering

\newcommand{\imgid}{201}
\scalebox{1.0}{
    \hspace{-0.4cm}
    \begin{adjustbox}{valign=t}
    \begin{tabular}{cccccc}
    \includegraphics[width=0.188\textwidth]{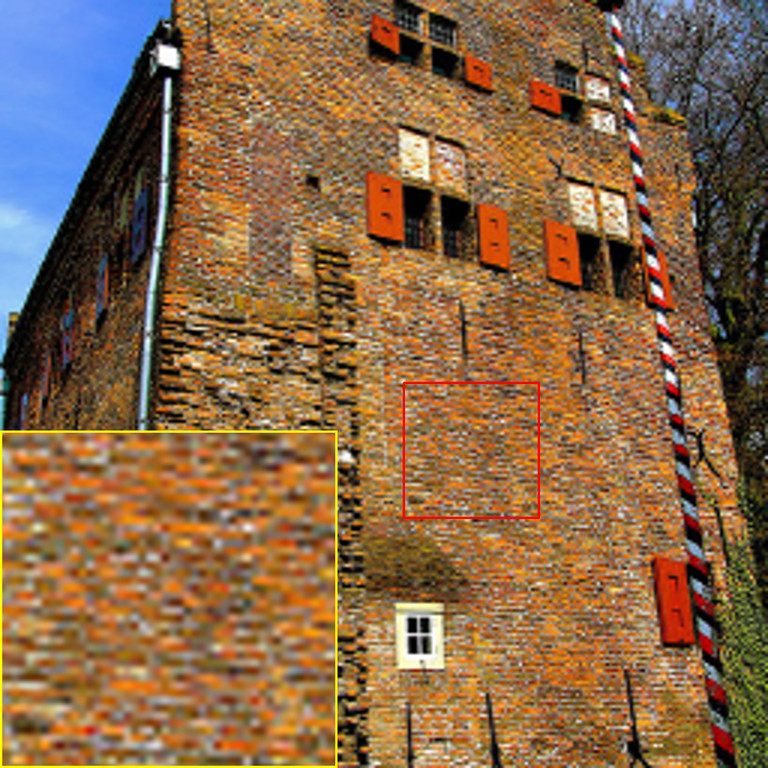} \hspace{-1.2mm} &
    \includegraphics[width=0.188\textwidth]{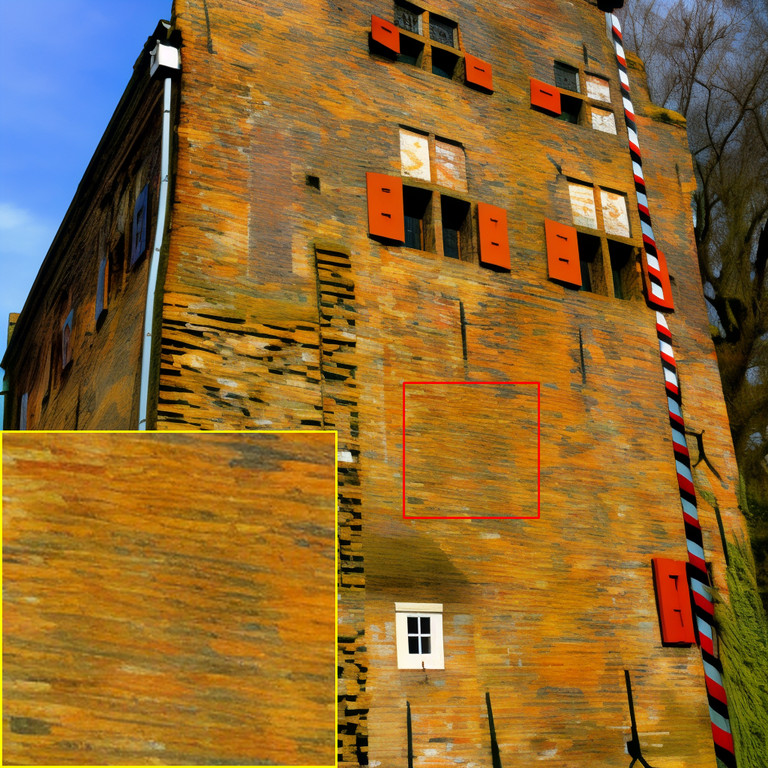} \hspace{-1.2mm} &
    \includegraphics[width=0.188\textwidth]{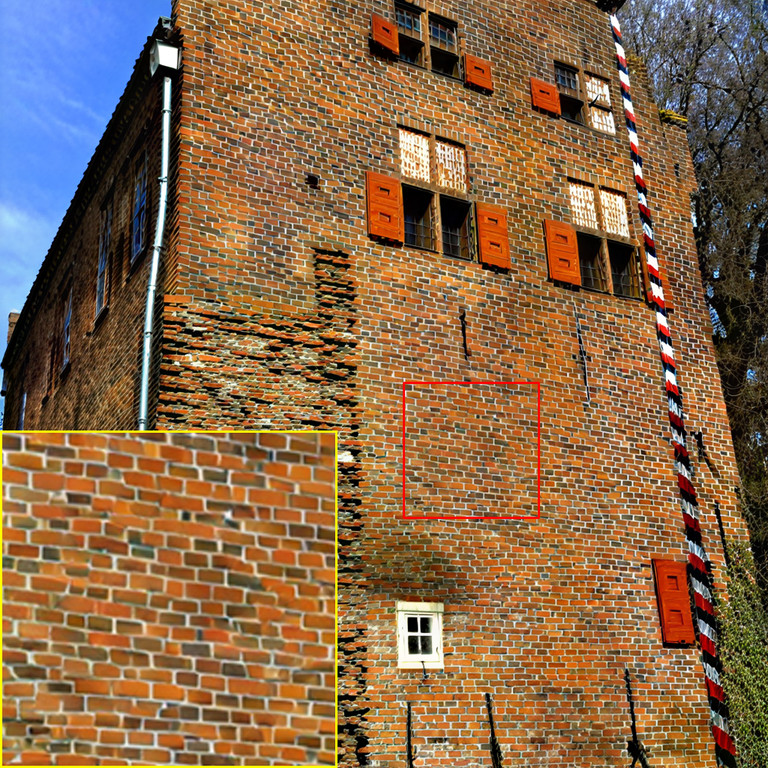} \hspace{-1.2mm} &
    \includegraphics[width=0.188\textwidth]{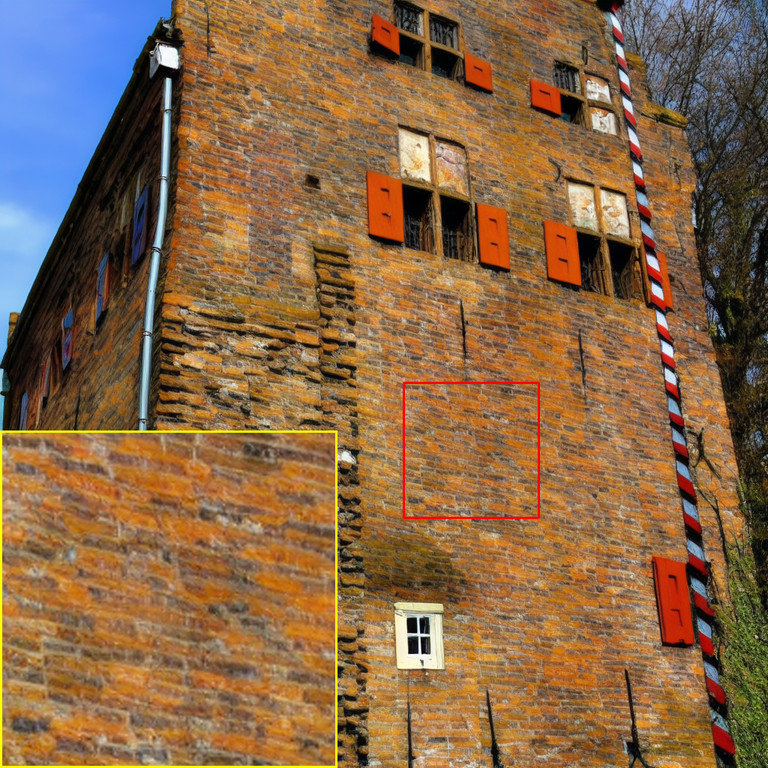} \hspace{-1.2mm} &
    \includegraphics[width=0.188\textwidth]{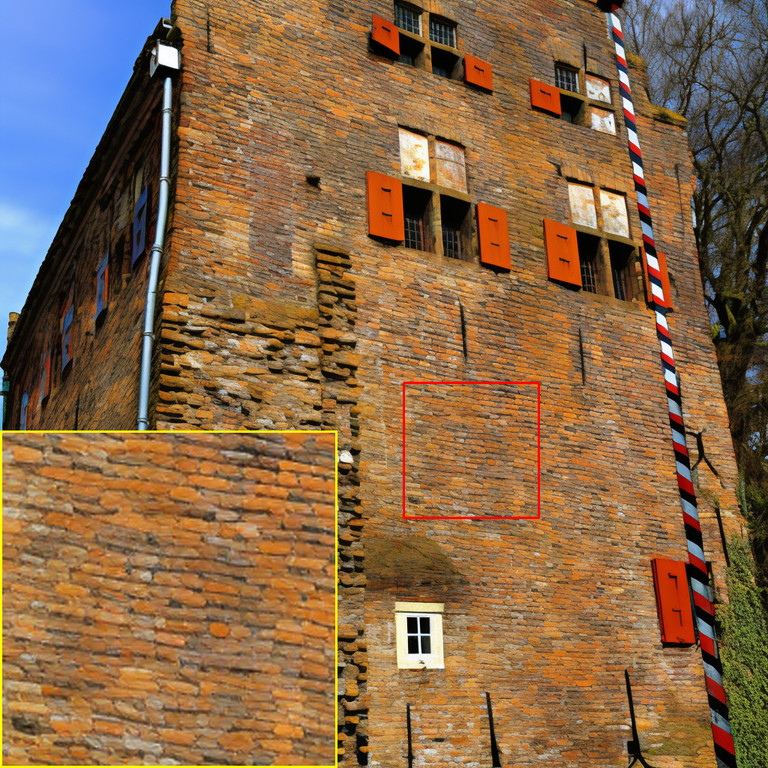} \hspace{-1.2mm} \\
    RealLQ250: \texttt{\imgid} \hspace{-1.2mm} &
    DiffBIR~\cite{lin2024diffbir} \hspace{-1.2mm} &
    SUPIR~\cite{yu2024supir} \hspace{-1.2mm} &
    PASD~\cite{yang2024pasd} \hspace{-1.2mm} &
    SeeSR~\cite{wu2024seesr}  \hspace{-1.2mm} \\
    
    \includegraphics[width=0.188\textwidth]{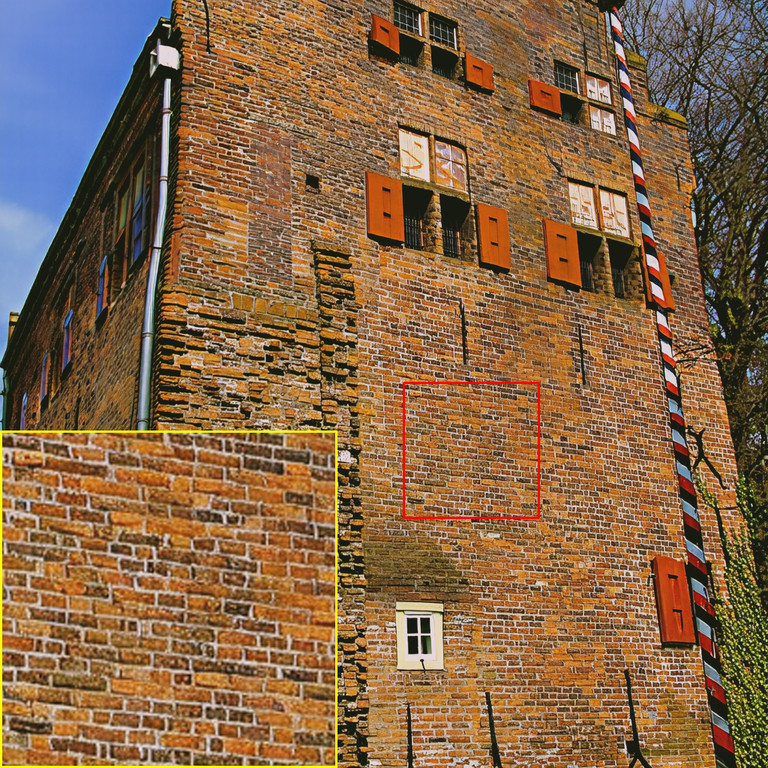} \hspace{-1.2mm} &
    \includegraphics[width=0.188\textwidth]{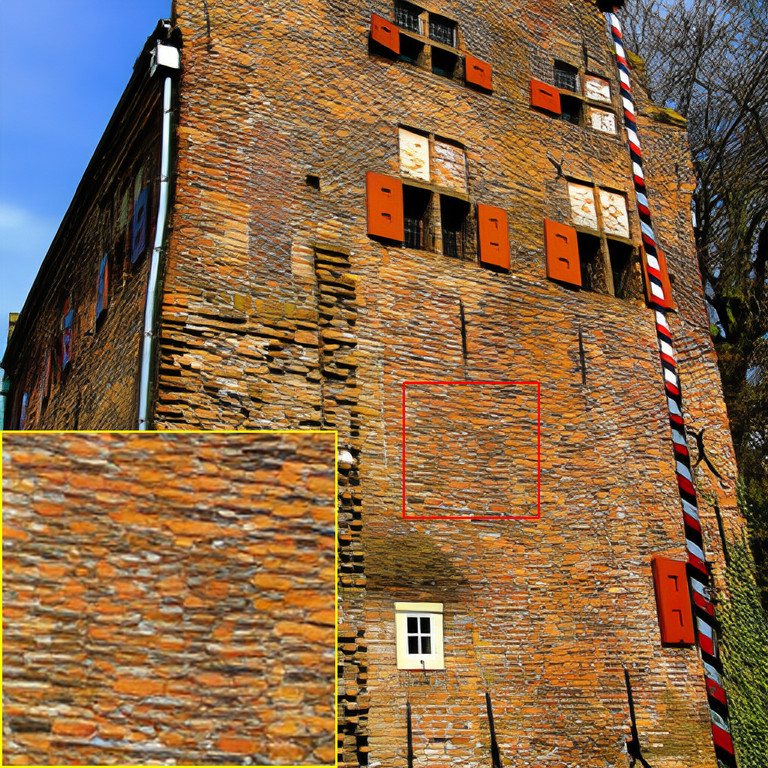} \hspace{-1.2mm} &
    \includegraphics[width=0.188\textwidth]{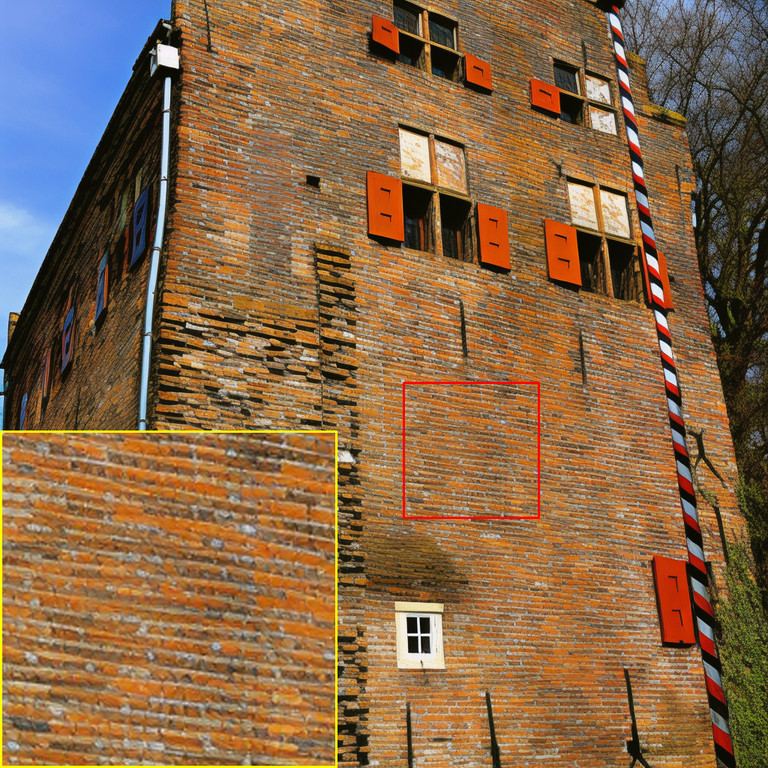} \hspace{-1.2mm} &
    \includegraphics[width=0.188\textwidth]{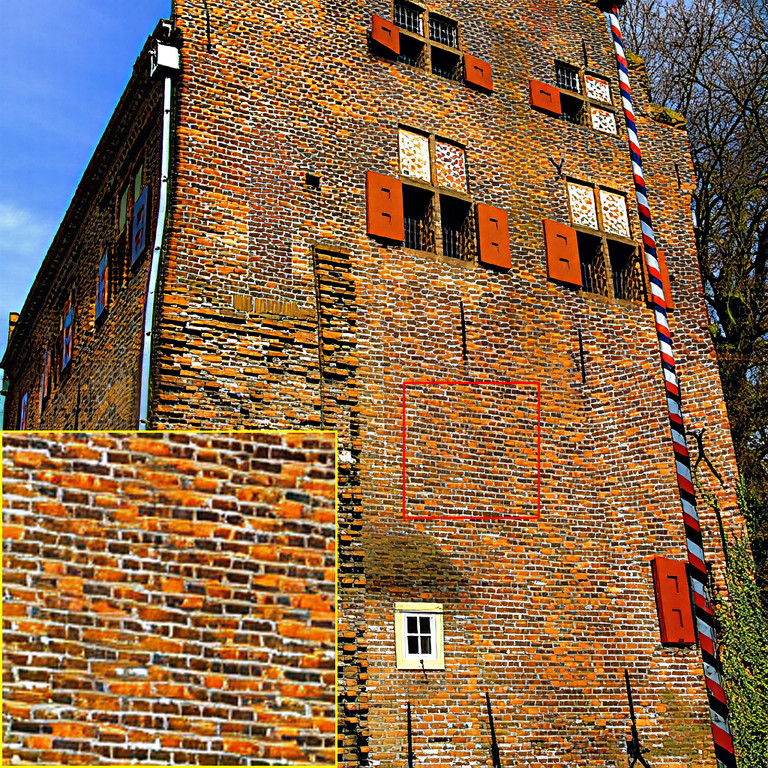} \hspace{-1.2mm} &
    \includegraphics[width=0.188\textwidth]{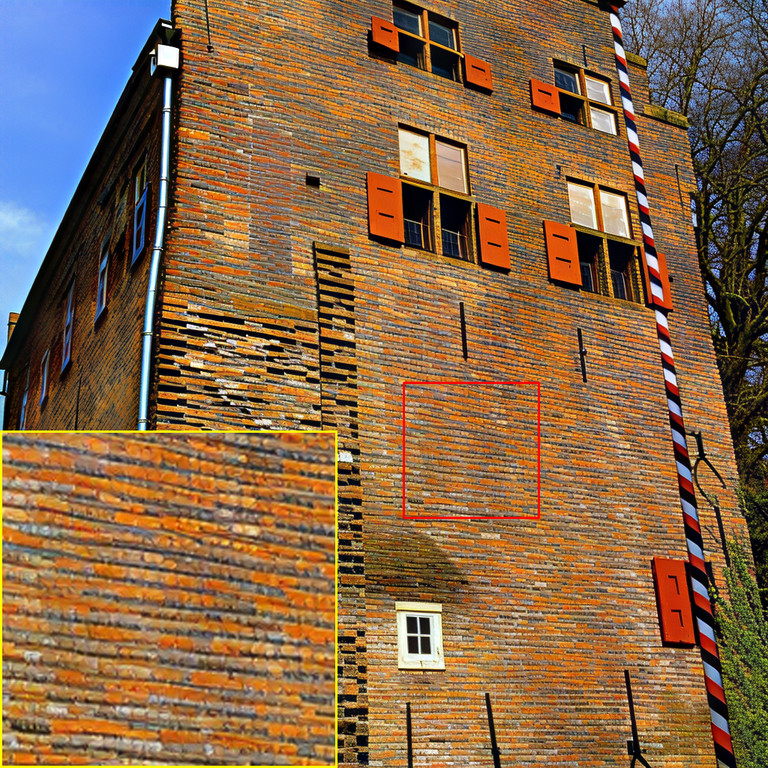} \hspace{-1.2mm} \\
    DiT4SR\cite{duan2025dit4sr} \hspace{-1.2mm} &
    CTMSR~\cite{you2025ctmsr} \hspace{-1.2mm} &
    PiSA-SR~\cite{sun2024pisasr} \hspace{-1.2mm} &
    TSD-SR~\cite{dong2025tsdsr} \hspace{-1.2mm} &
    HYPIR~\cite{lin2025hypir}  \hspace{-1.2mm} \\

    \includegraphics[width=0.188\textwidth]{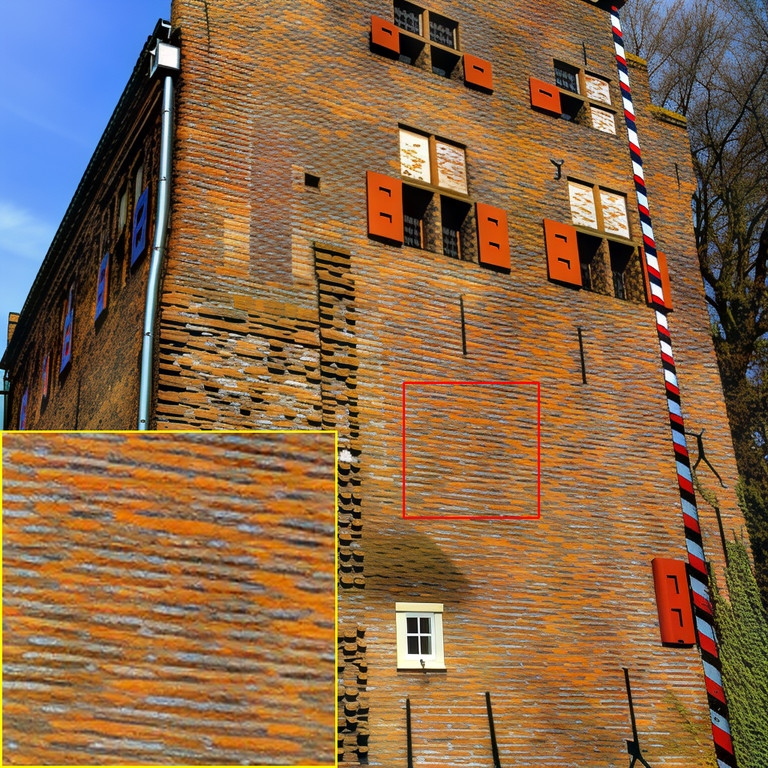} \hspace{-1.2mm} &
    \includegraphics[width=0.188\textwidth]{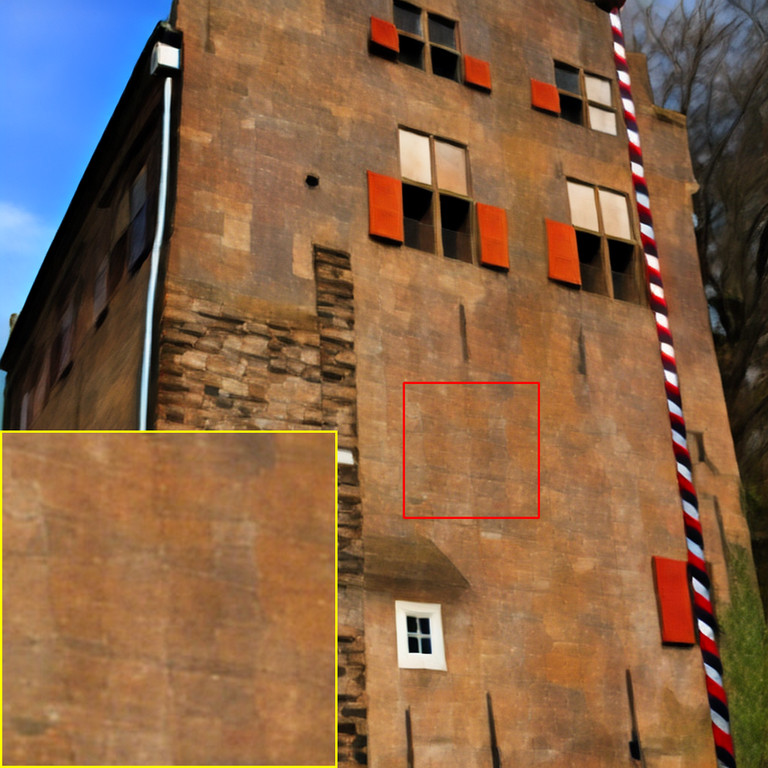} \hspace{-1.2mm} &
    \includegraphics[width=0.188\textwidth]{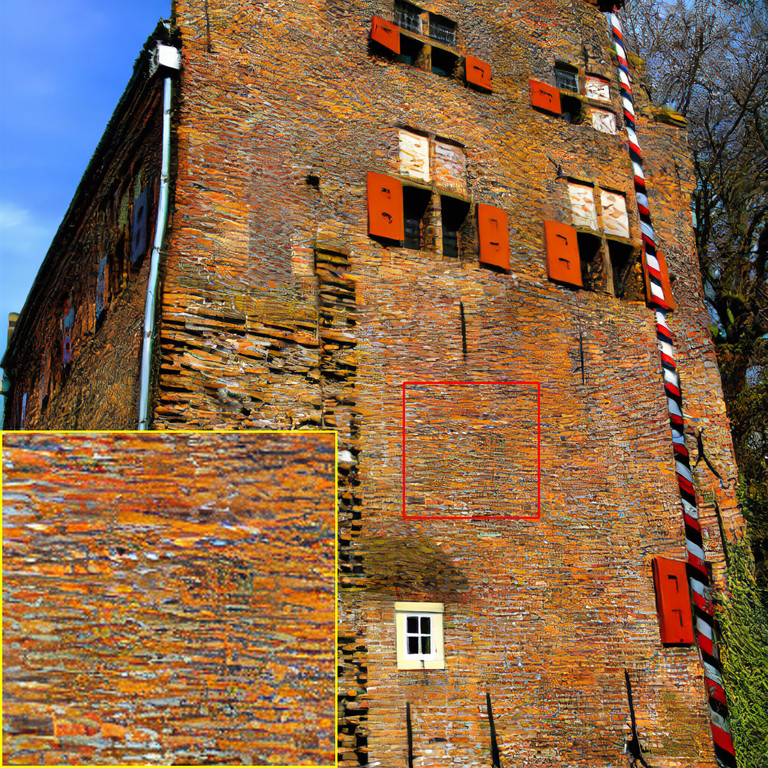} \hspace{-1.2mm} &
    \includegraphics[width=0.188\textwidth]{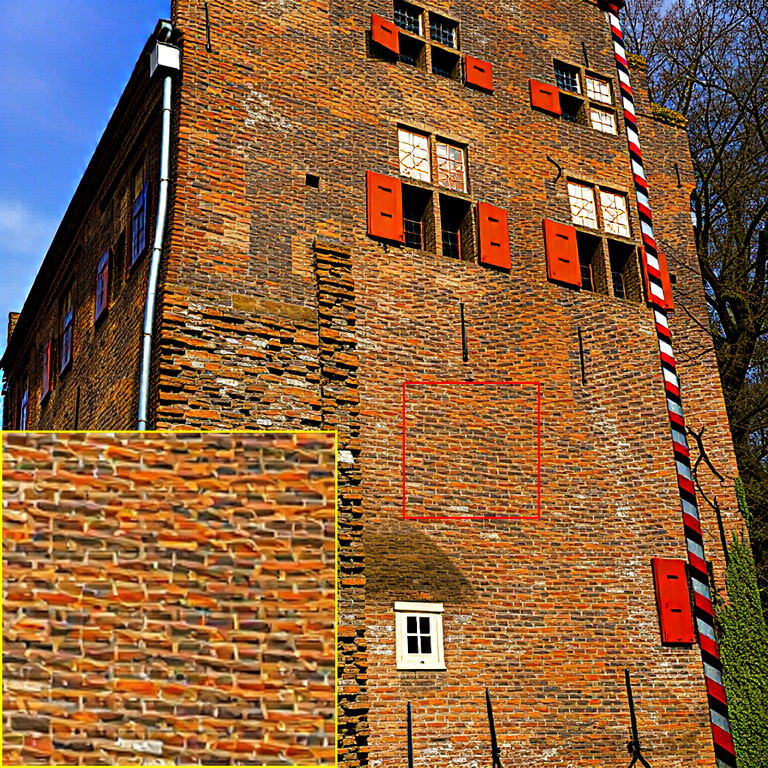} \hspace{-1.2mm} &
    \includegraphics[width=0.188\textwidth]{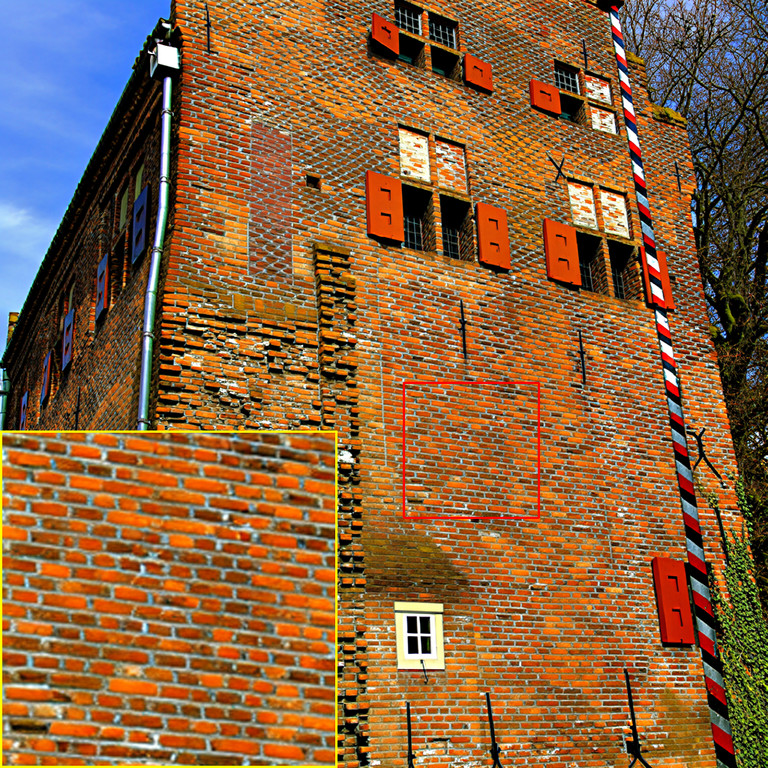} \hspace{-1.2mm} \\
    InvSR~\cite{yue2025invsr} \hspace{-1.2mm} &
    OSEDiff~\cite{wu2024osediff} \hspace{-1.2mm} &
    SinSR~\cite{wang2024sinsr} \hspace{-1.2mm} &
    StrSR - Z-Image  \hspace{-1.2mm} &
    StrSR - FLUX \hspace{-1.2mm}  \\

    \end{tabular}
    \end{adjustbox}
}

\vspace{-3.2mm}
\caption{More visual comparison for image SR ($\times$4) in RealLQ250 dataset. }
\label{fig:vis-reallq250-1}
\end{figure}

\begin{figure*}[t]
\scriptsize
\centering

\newcommand{\imgid}{041}
\scalebox{1.0}{
    \hspace{-0.4cm}
    \begin{adjustbox}{valign=t}
    \begin{tabular}{cccccc}
    \includegraphics[width=0.188\textwidth]{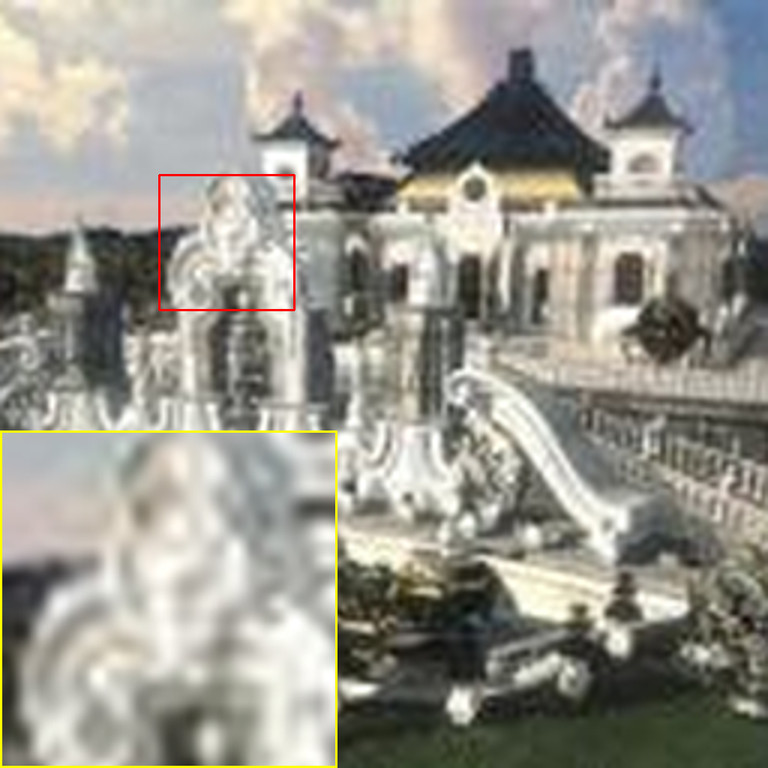} \hspace{-1.2mm} &
    \includegraphics[width=0.188\textwidth]{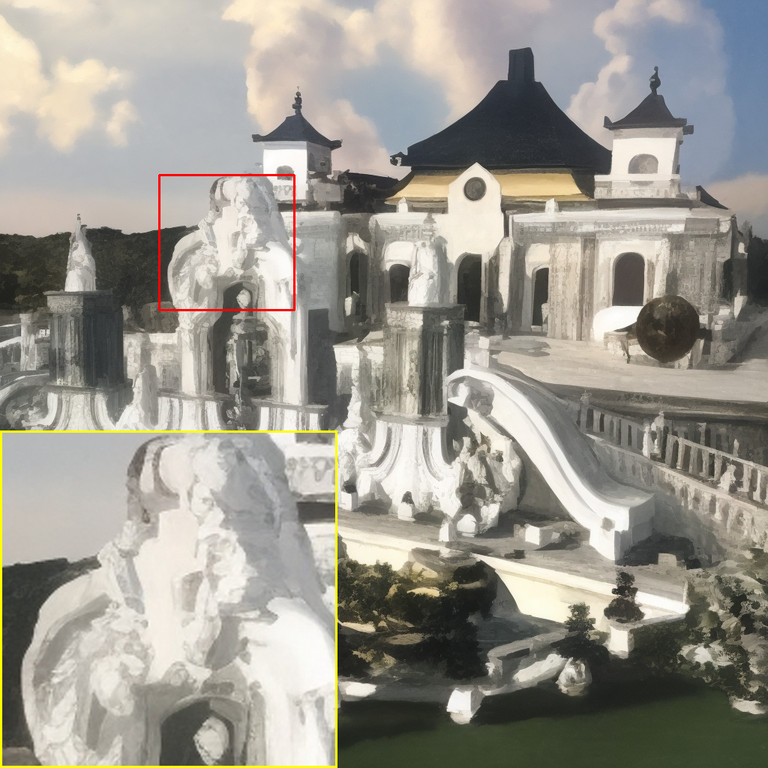} \hspace{-1.2mm} &
    \includegraphics[width=0.188\textwidth]{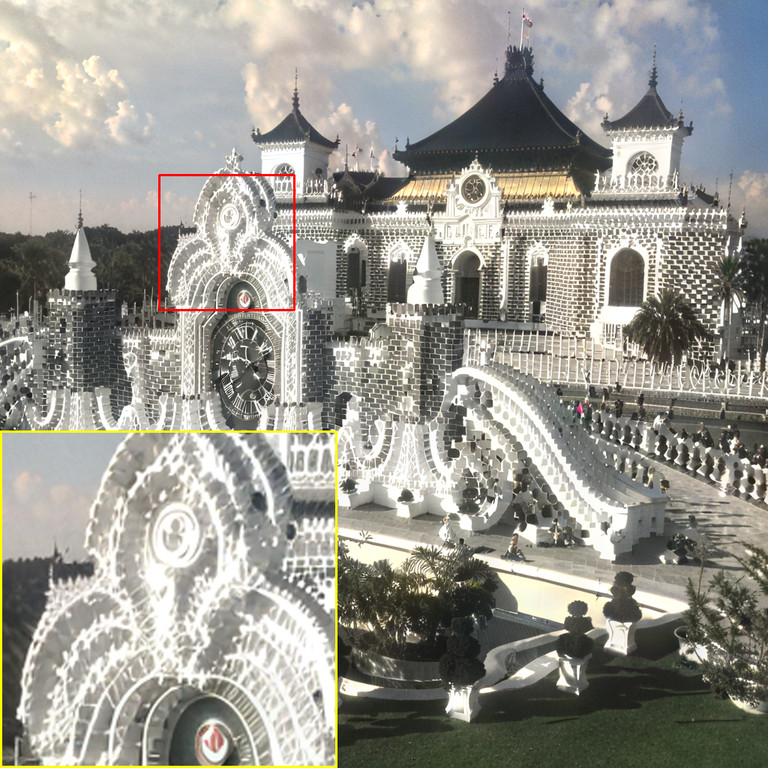} \hspace{-1.2mm} &
    \includegraphics[width=0.188\textwidth]{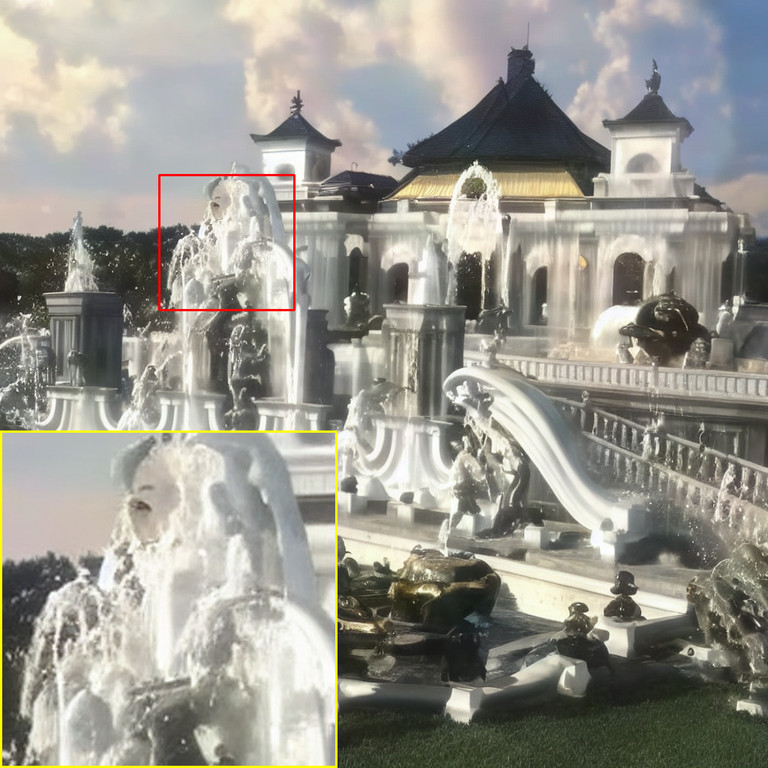} \hspace{-1.2mm} &
    \includegraphics[width=0.188\textwidth]{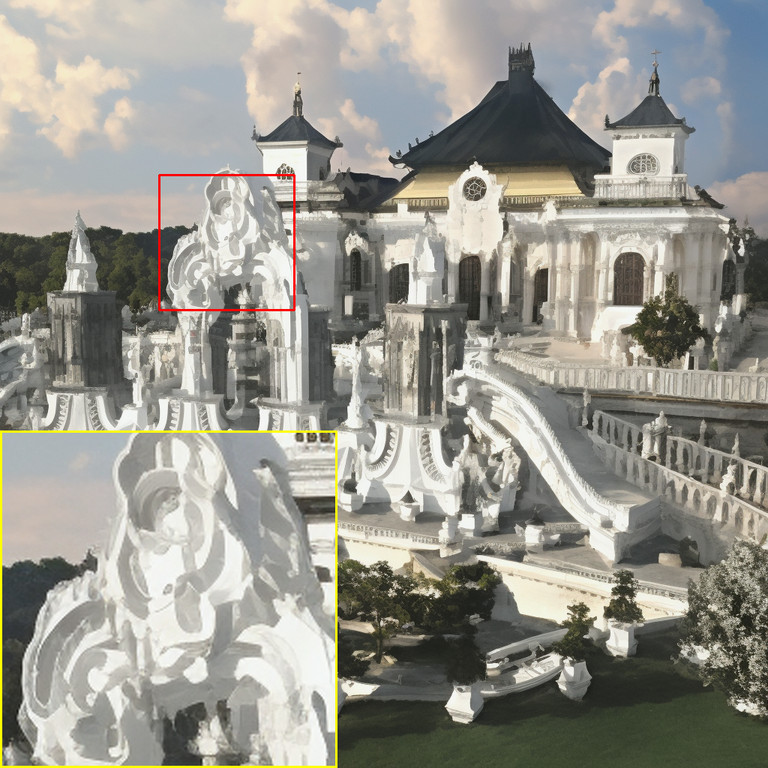} \hspace{-1.2mm} \\
    RealLQ250: \texttt{\imgid} \hspace{-1.2mm} &
    DiffBIR~\cite{lin2024diffbir} \hspace{-1.2mm} &
    SUPIR~\cite{yu2024supir} \hspace{-1.2mm} &
    PASD~\cite{yang2024pasd} \hspace{-1.2mm} &
    SeeSR~\cite{wu2024seesr}  \hspace{-1.2mm} \\
    
    \includegraphics[width=0.188\textwidth]{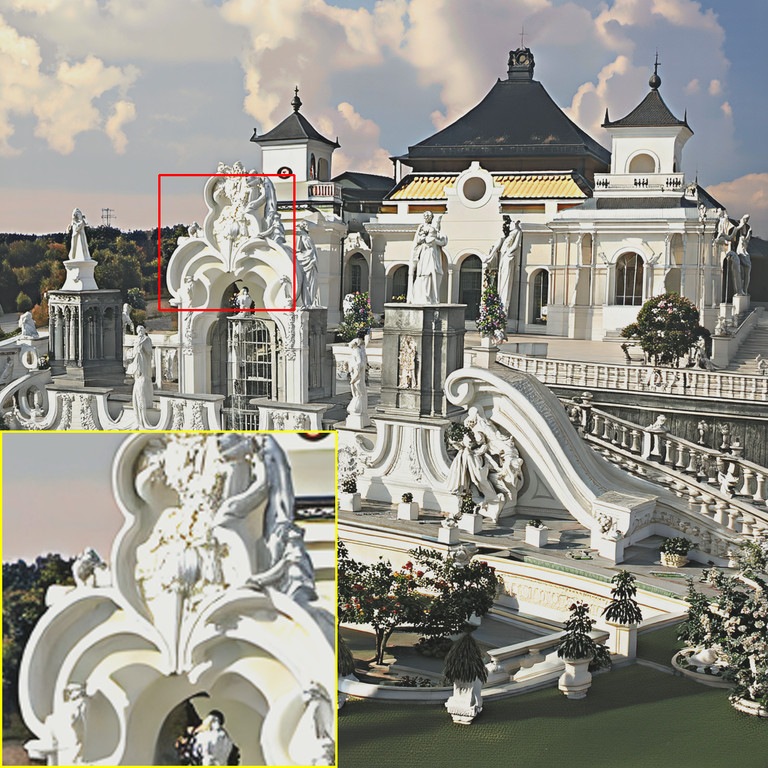} \hspace{-1.2mm} &
    \includegraphics[width=0.188\textwidth]{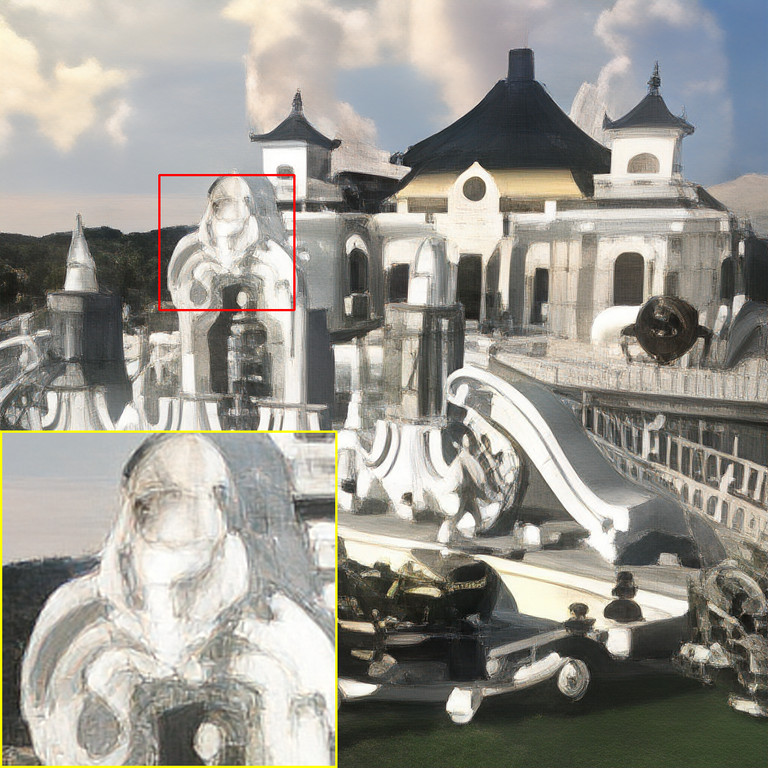} \hspace{-1.2mm} &
    \includegraphics[width=0.188\textwidth]{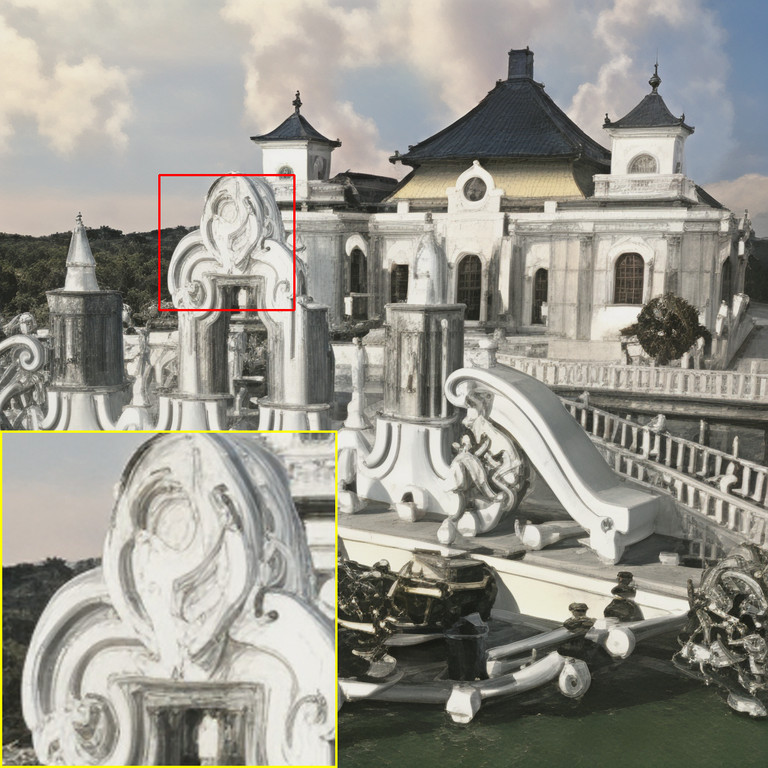} \hspace{-1.2mm} &
    \includegraphics[width=0.188\textwidth]{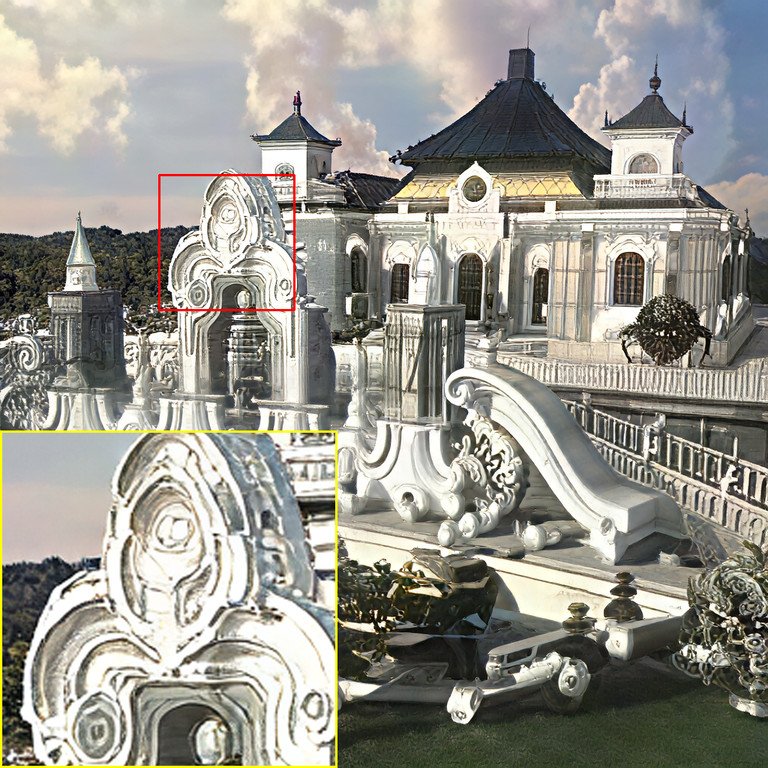} \hspace{-1.2mm} &
    \includegraphics[width=0.188\textwidth]{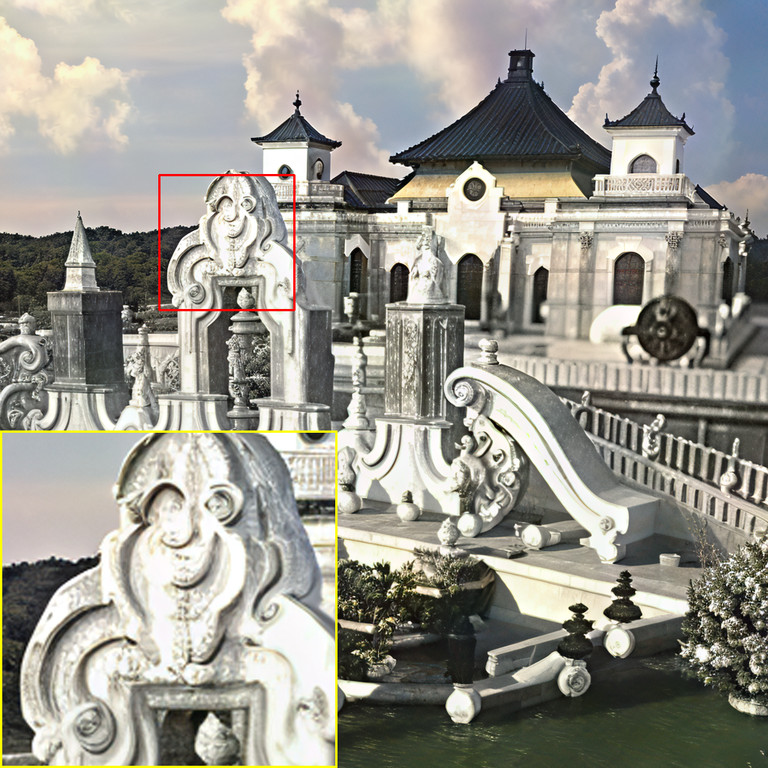} \hspace{-1.2mm} \\
    DiT4SR\cite{duan2025dit4sr} \hspace{-1.2mm} &
    CTMSR~\cite{you2025ctmsr} \hspace{-1.2mm} &
    PiSA-SR~\cite{sun2024pisasr} \hspace{-1.2mm} &
    TSD-SR~\cite{dong2025tsdsr} \hspace{-1.2mm} &
    HYPIR~\cite{lin2025hypir}  \hspace{-1.2mm} \\

    \includegraphics[width=0.188\textwidth]{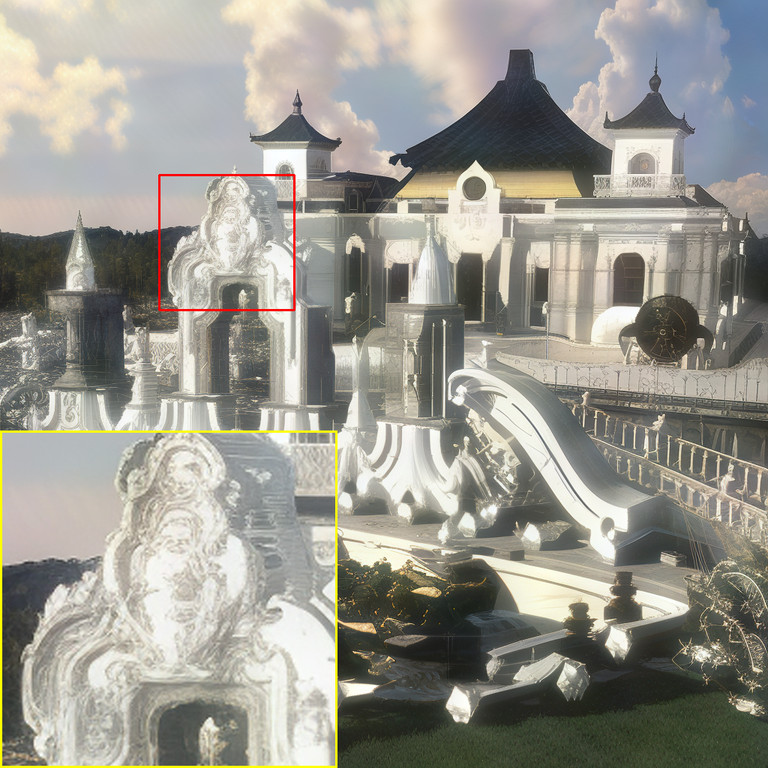} \hspace{-1.2mm} &
    \includegraphics[width=0.188\textwidth]{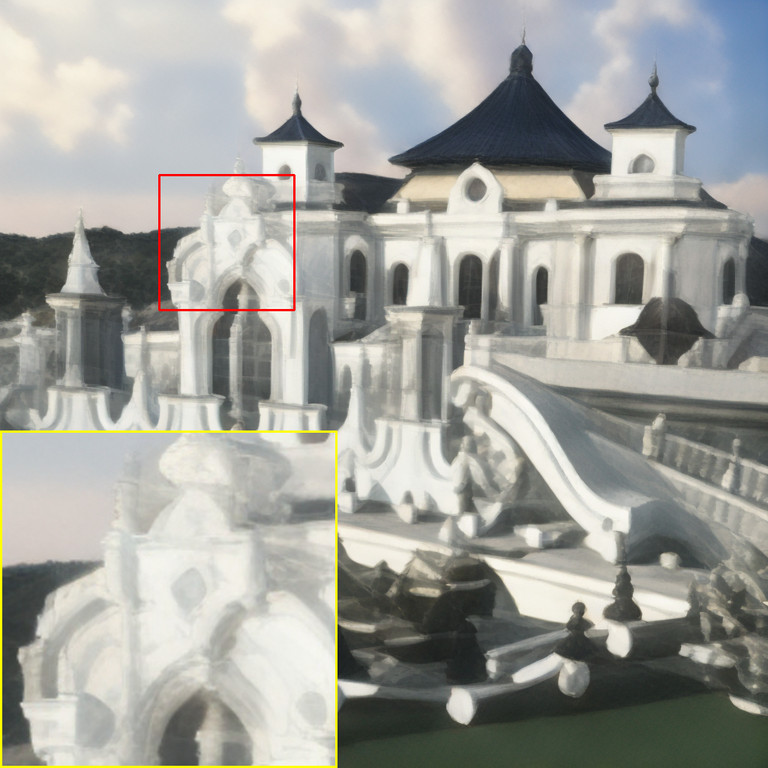} \hspace{-1.2mm} &
    \includegraphics[width=0.188\textwidth]{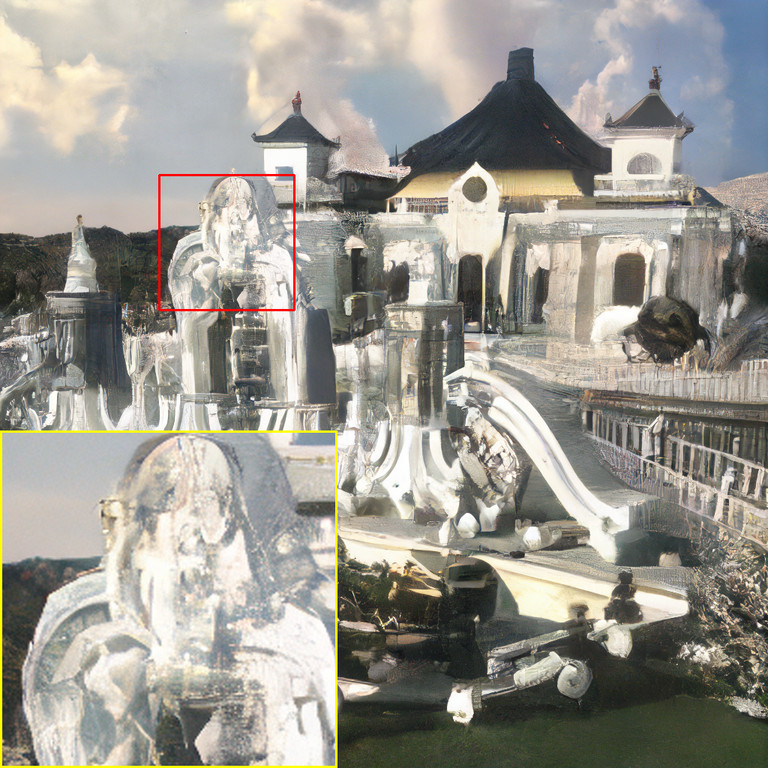} \hspace{-1.2mm} &
    \includegraphics[width=0.188\textwidth]{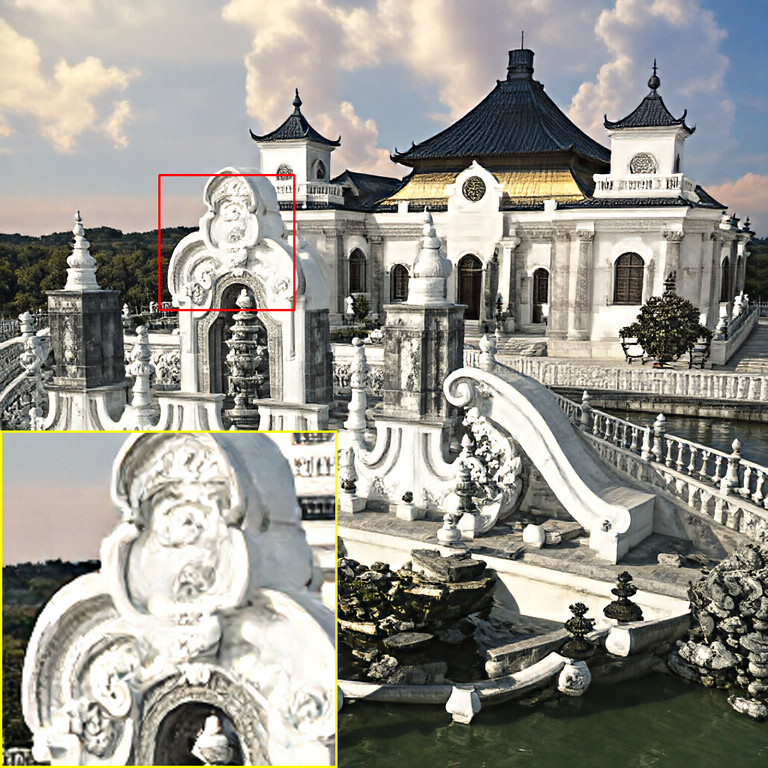} \hspace{-1.2mm} &
    \includegraphics[width=0.188\textwidth]{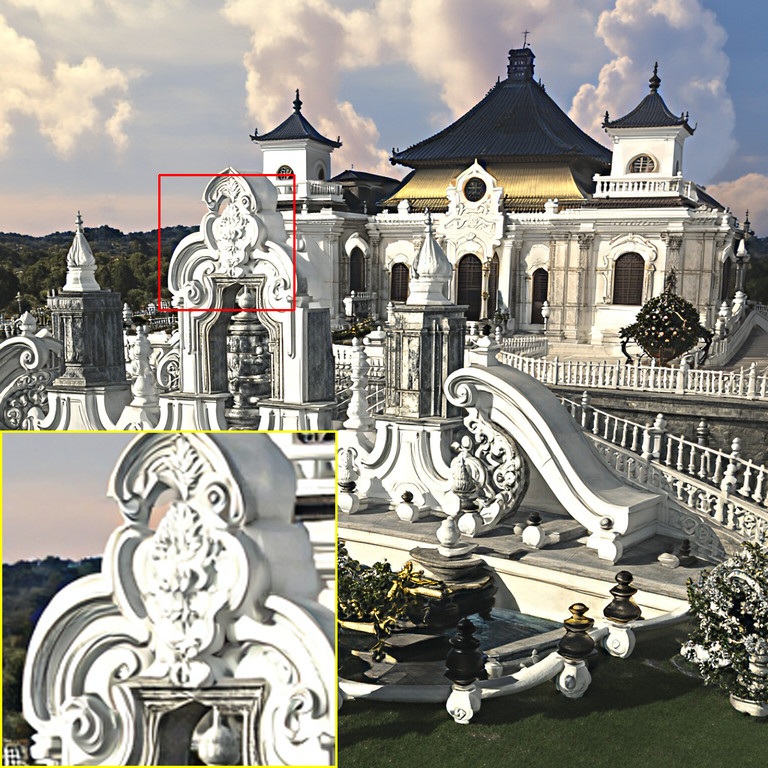} \hspace{-1.2mm} \\
    InvSR~\cite{yue2025invsr} \hspace{-1.2mm} &
    OSEDiff~\cite{wu2024osediff} \hspace{-1.2mm} &
    SinSR~\cite{wang2024sinsr} \hspace{-1.2mm} &
    StrSR - Z-Image  \hspace{-1.2mm} &
    StrSR - FLUX \hspace{-1.2mm}  \\

    \end{tabular}
    \end{adjustbox}
}

\vspace{6mm}

\renewcommand{\imgid}{184}
\scalebox{1.0}{
    \hspace{-0.4cm}
    \begin{adjustbox}{valign=t}
    \begin{tabular}{cccccc}
    \includegraphics[width=0.188\textwidth]{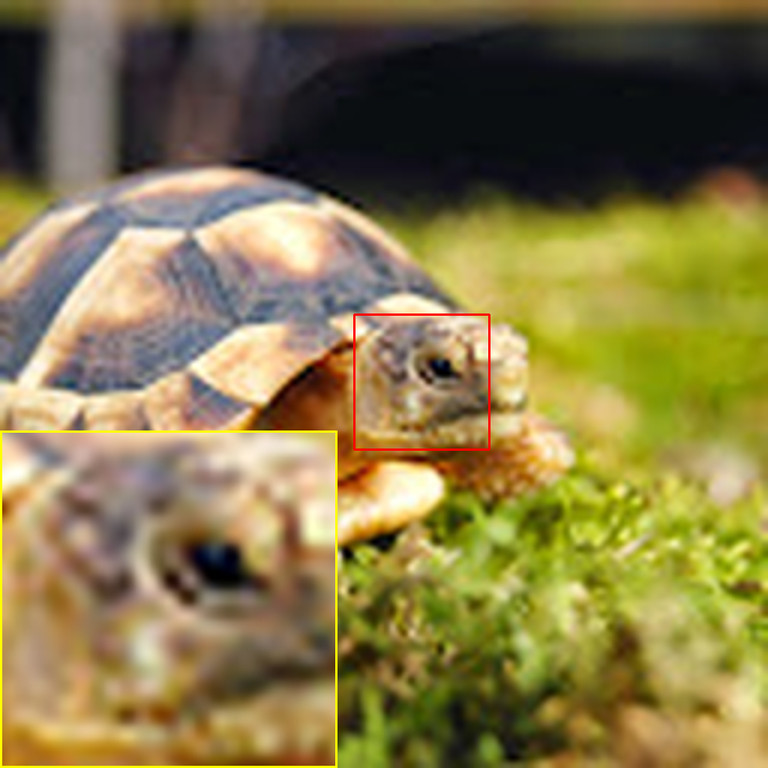} \hspace{-1.2mm} &
    \includegraphics[width=0.188\textwidth]{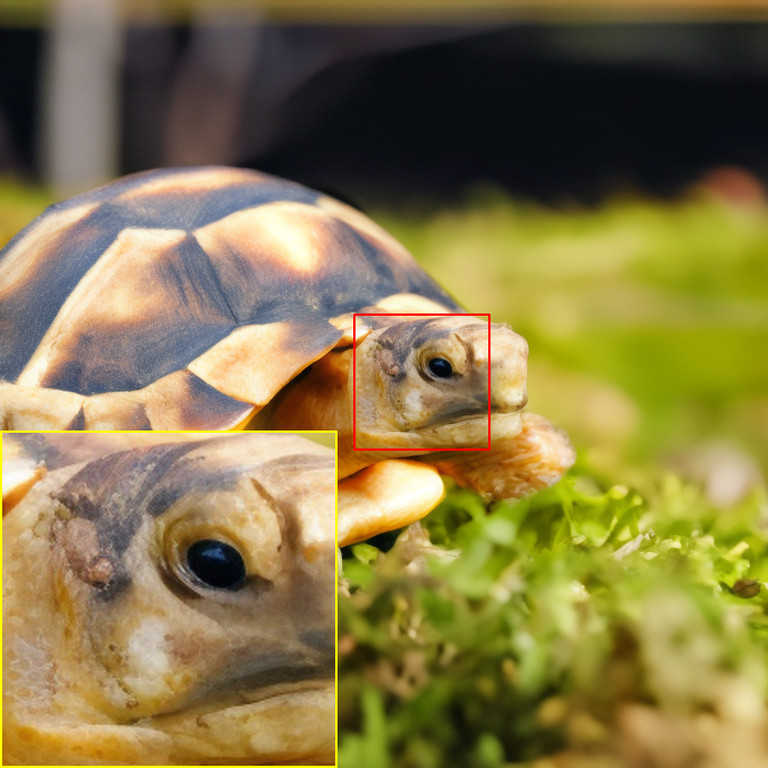} \hspace{-1.2mm} &
    \includegraphics[width=0.188\textwidth]{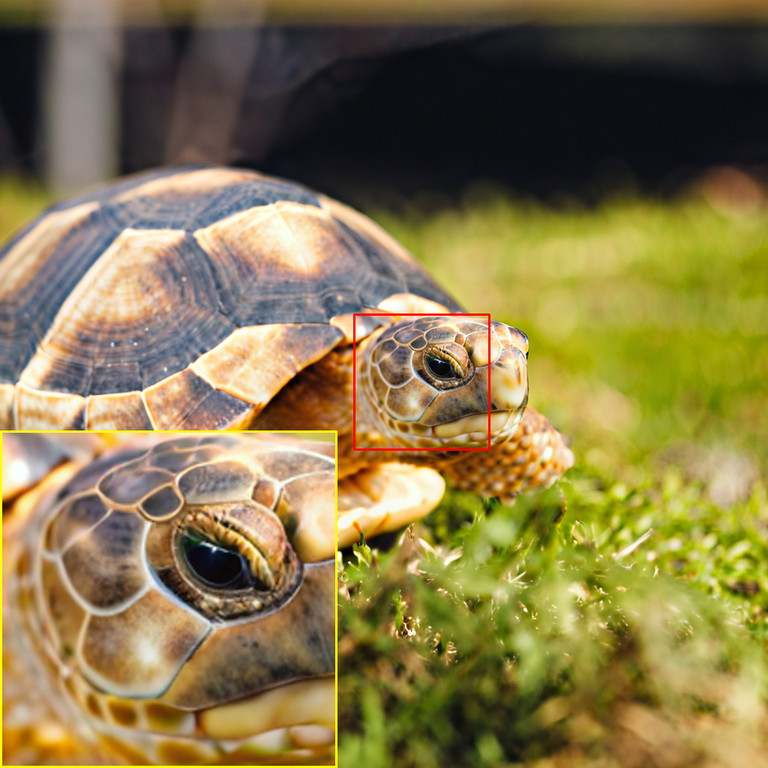} \hspace{-1.2mm} &
    \includegraphics[width=0.188\textwidth]{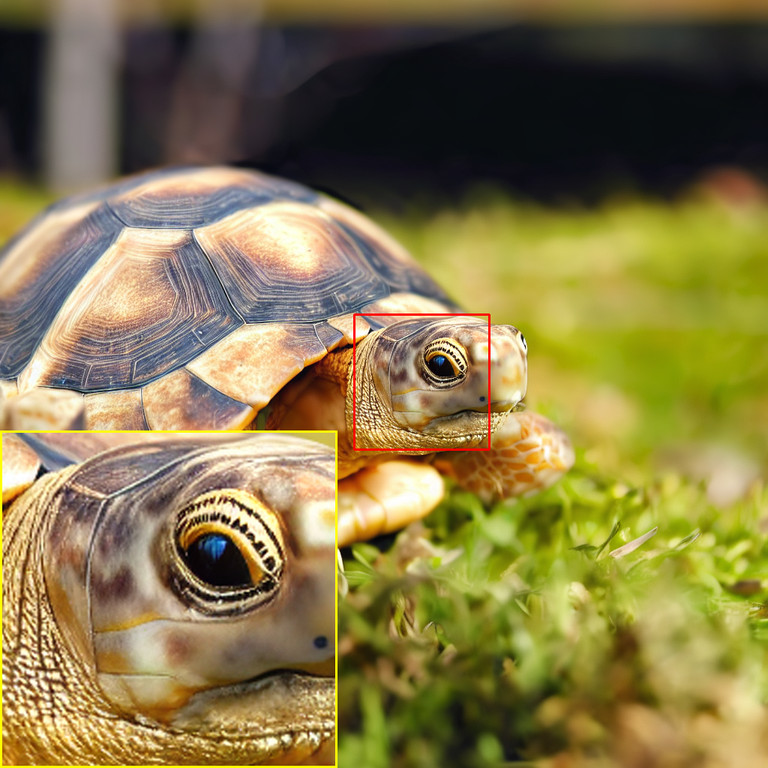} \hspace{-1.2mm} &
    \includegraphics[width=0.188\textwidth]{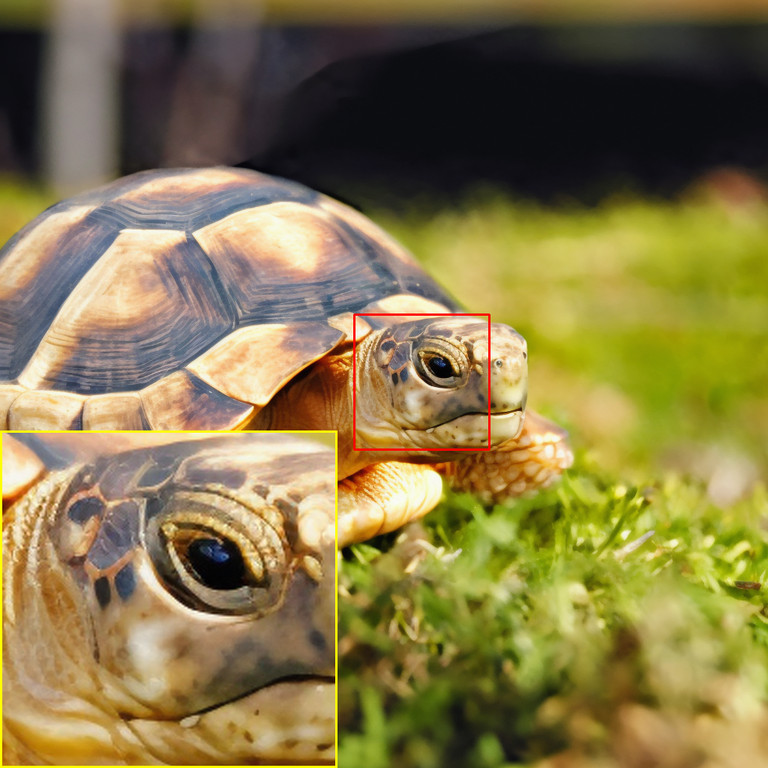} \hspace{-1.2mm} \\
    RealLQ250: \texttt{\imgid} \hspace{-1.2mm} &
    DiffBIR~\cite{lin2024diffbir} \hspace{-1.2mm} &
    SUPIR~\cite{yu2024supir} \hspace{-1.2mm} &
    PASD~\cite{yang2024pasd} \hspace{-1.2mm} &
    SeeSR~\cite{wu2024seesr}  \hspace{-1.2mm} \\
    
    \includegraphics[width=0.188\textwidth]{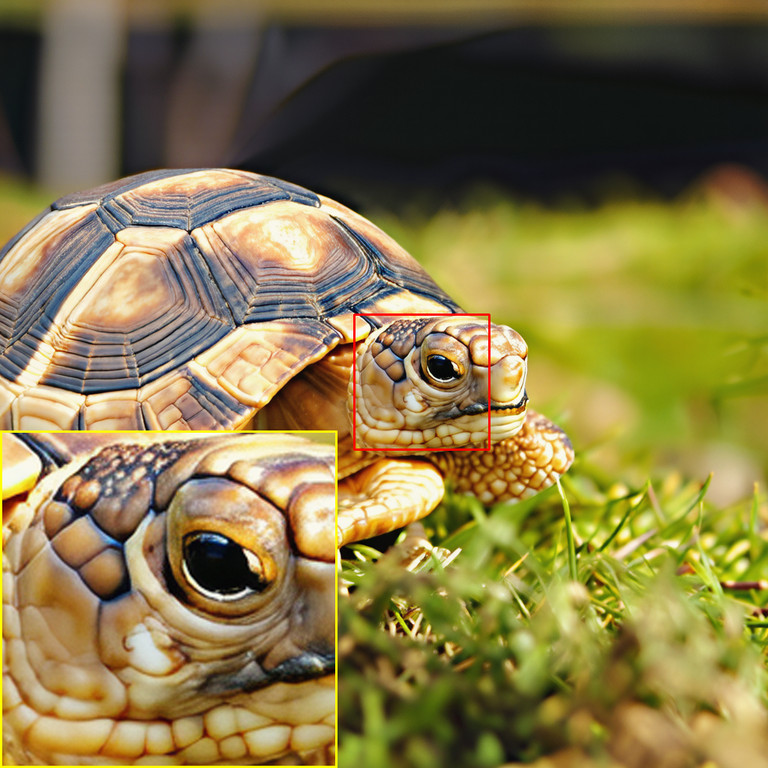} \hspace{-1.2mm} &
    \includegraphics[width=0.188\textwidth]{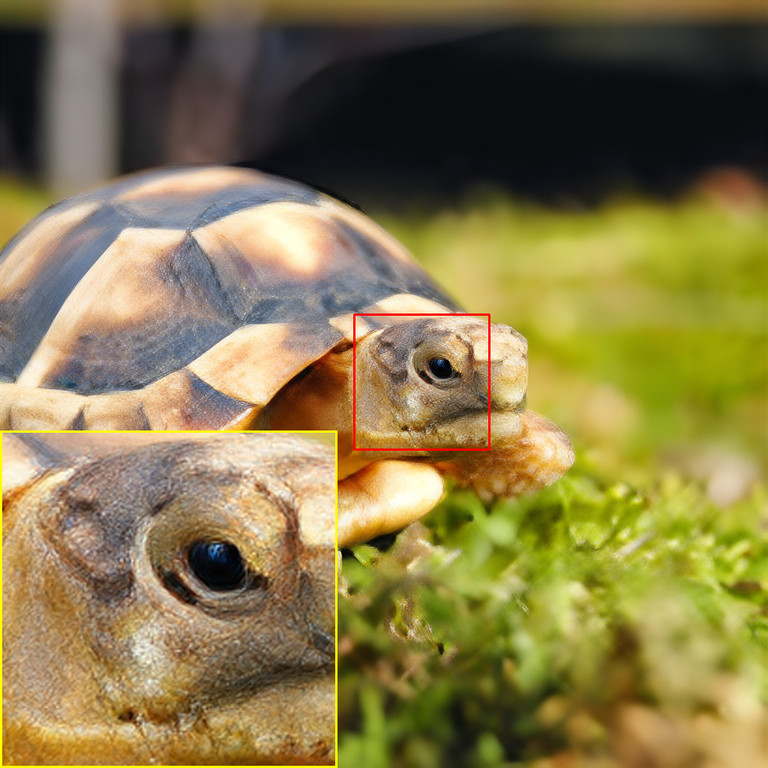} \hspace{-1.2mm} &
    \includegraphics[width=0.188\textwidth]{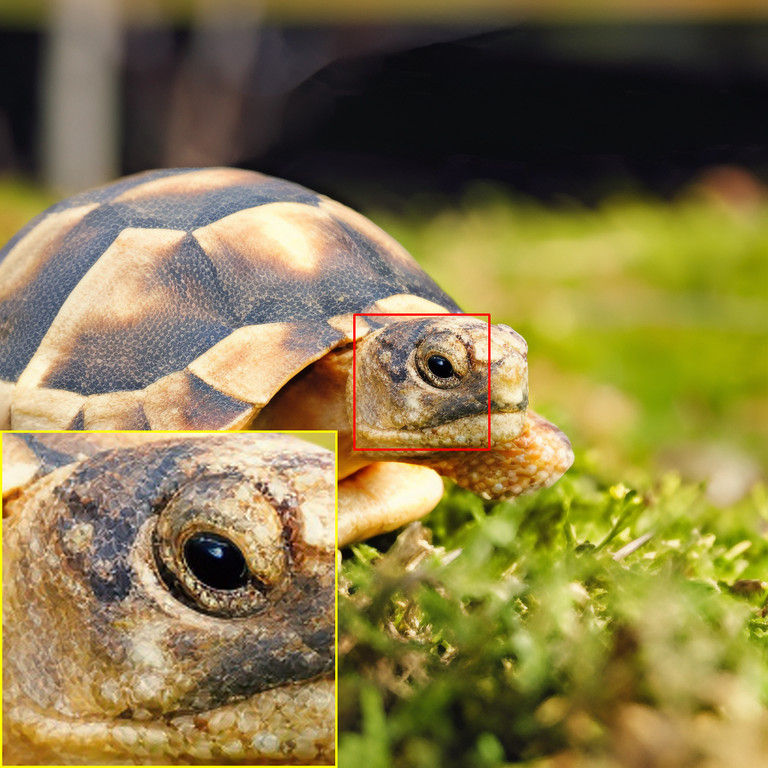} \hspace{-1.2mm} &
    \includegraphics[width=0.188\textwidth]{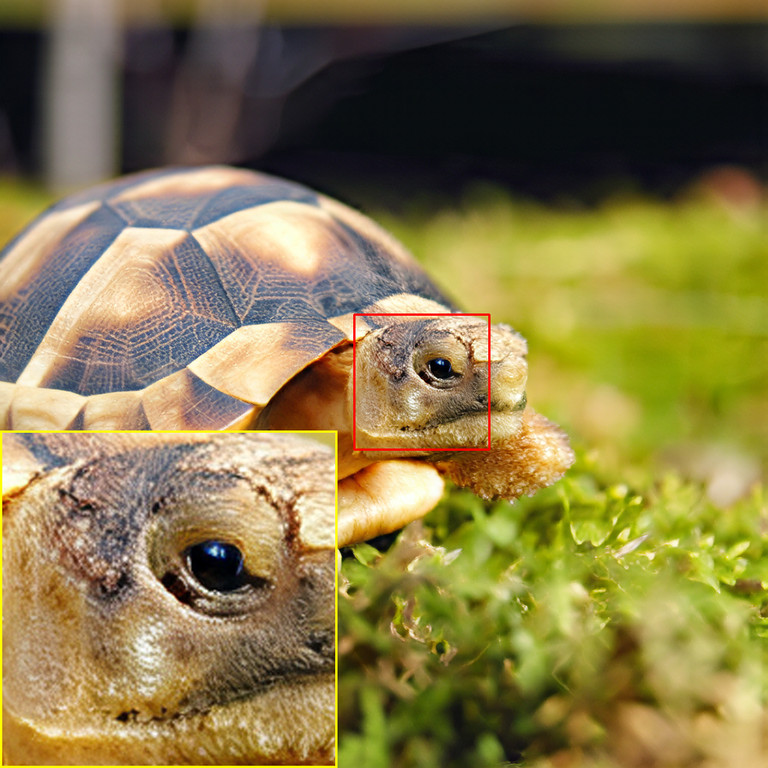} \hspace{-1.2mm} &
    \includegraphics[width=0.188\textwidth]{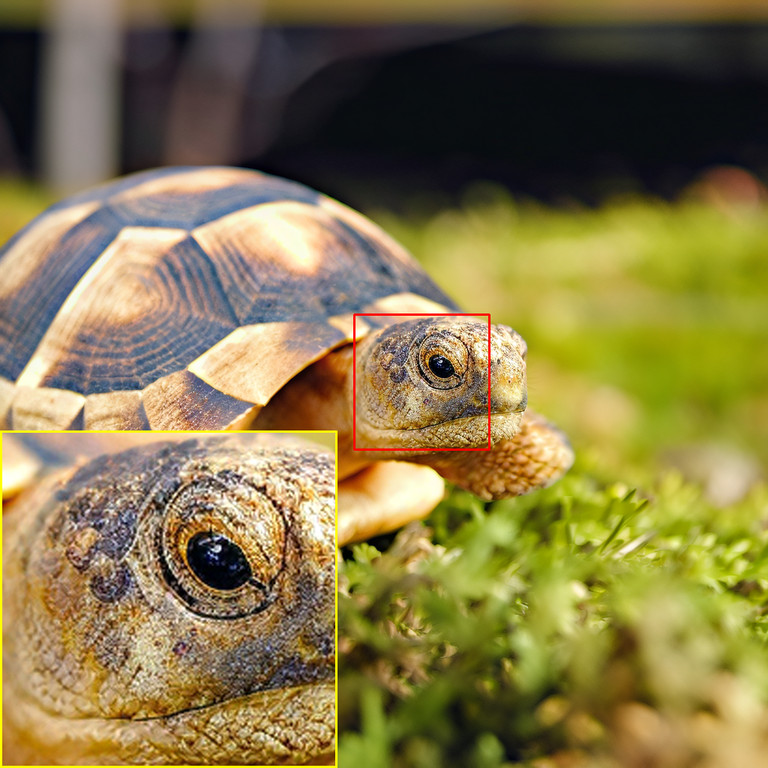} \hspace{-1.2mm} \\
    DiT4SR\cite{duan2025dit4sr} \hspace{-1.2mm} &
    CTMSR~\cite{you2025ctmsr} \hspace{-1.2mm} &
    PiSA-SR~\cite{sun2024pisasr} \hspace{-1.2mm} &
    TSD-SR~\cite{dong2025tsdsr} \hspace{-1.2mm} &
    HYPIR~\cite{lin2025hypir}  \hspace{-1.2mm} \\

    \includegraphics[width=0.188\textwidth]{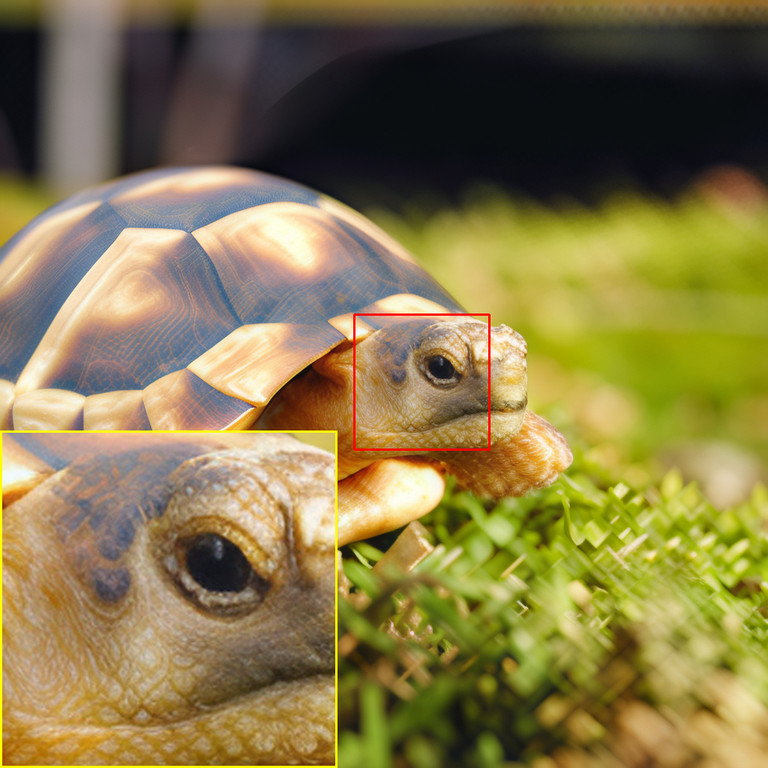} \hspace{-1.2mm} &
    \includegraphics[width=0.188\textwidth]{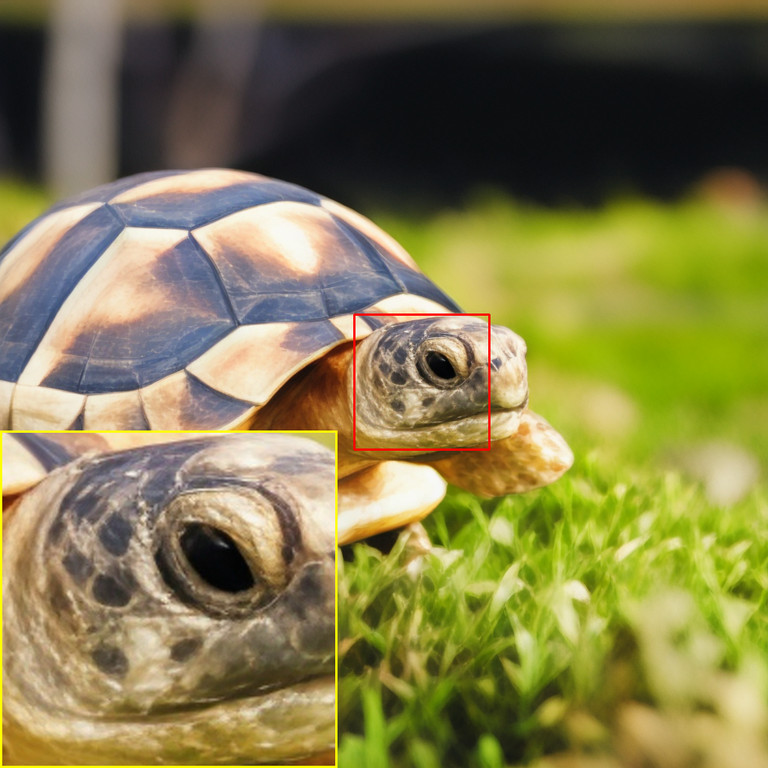} \hspace{-1.2mm} &
    \includegraphics[width=0.188\textwidth]{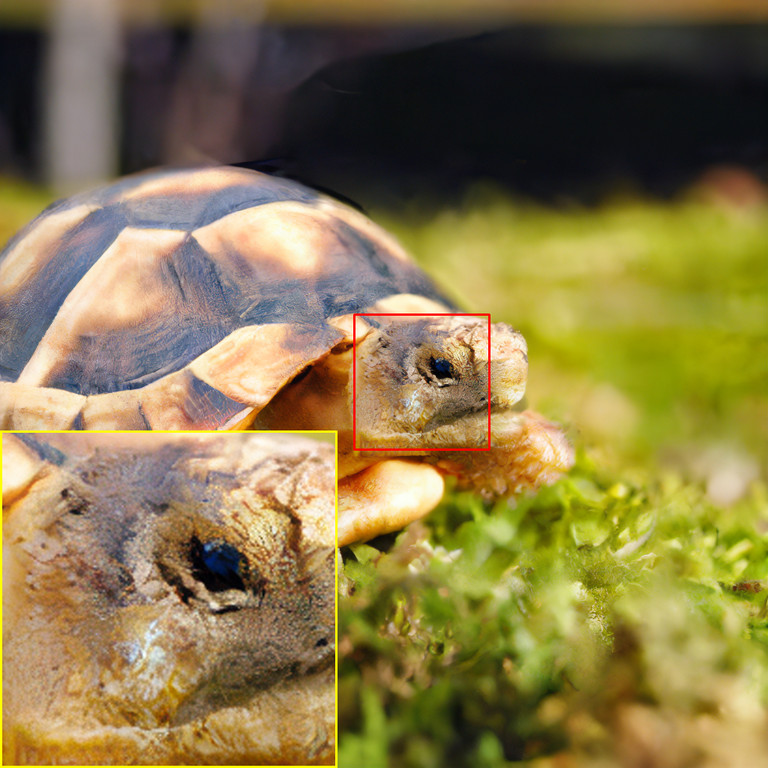} \hspace{-1.2mm} &
    \includegraphics[width=0.188\textwidth]{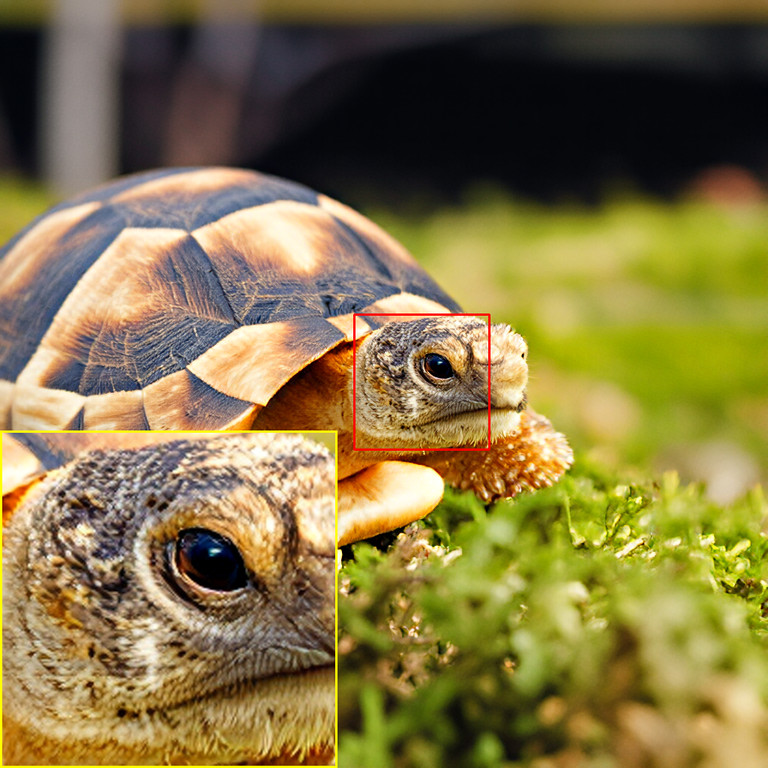} \hspace{-1.2mm} &
    \includegraphics[width=0.188\textwidth]{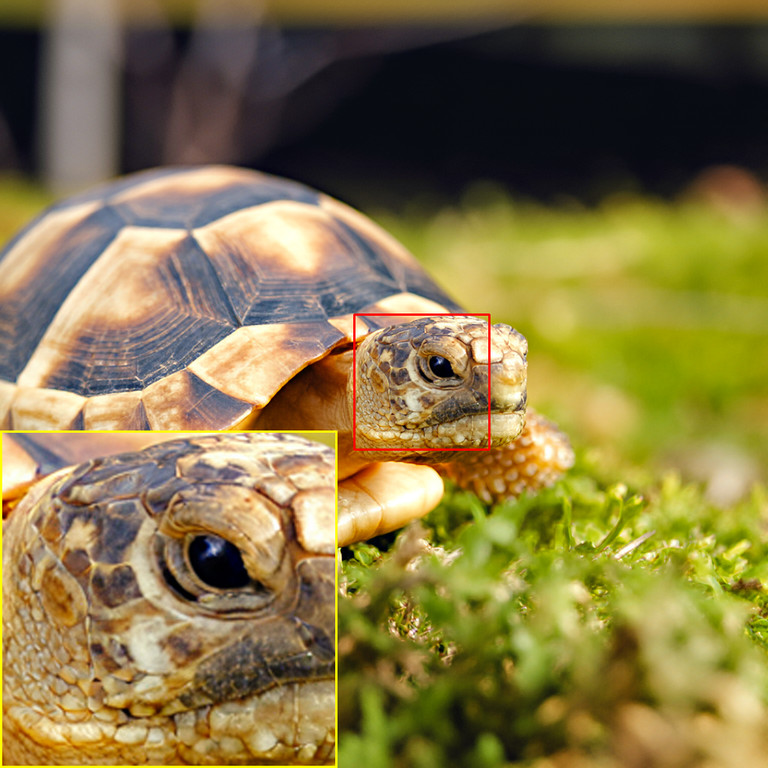} \hspace{-1.2mm} \\
    InvSR~\cite{yue2025invsr} \hspace{-1.2mm} &
    OSEDiff~\cite{wu2024osediff} \hspace{-1.2mm} &
    SinSR~\cite{wang2024sinsr} \hspace{-1.2mm} &
    StrSR - Z-Image  \hspace{-1.2mm} &
    StrSR - FLUX \hspace{-1.2mm}  \\

    \end{tabular}
    \end{adjustbox}
}

\vspace{6mm}

\caption{More visual comparison for image SR ($\times$4) in RealLQ250 dataset. }
\label{fig:vis-reallq250-2}
\end{figure*}

\begin{figure*}[t]
\scriptsize
\centering

\newcommand{\imgid}{190}
\scalebox{1.0}{
    \hspace{-0.4cm}
    \begin{adjustbox}{valign=t}
    \begin{tabular}{cccccc}
    \includegraphics[width=0.188\textwidth]{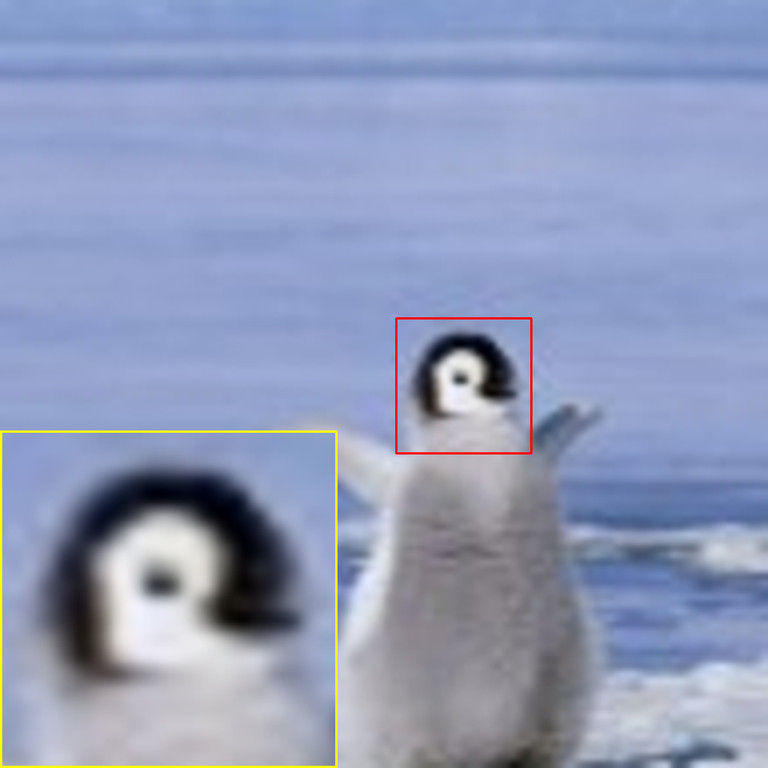} \hspace{-1.2mm} &
    \includegraphics[width=0.188\textwidth]{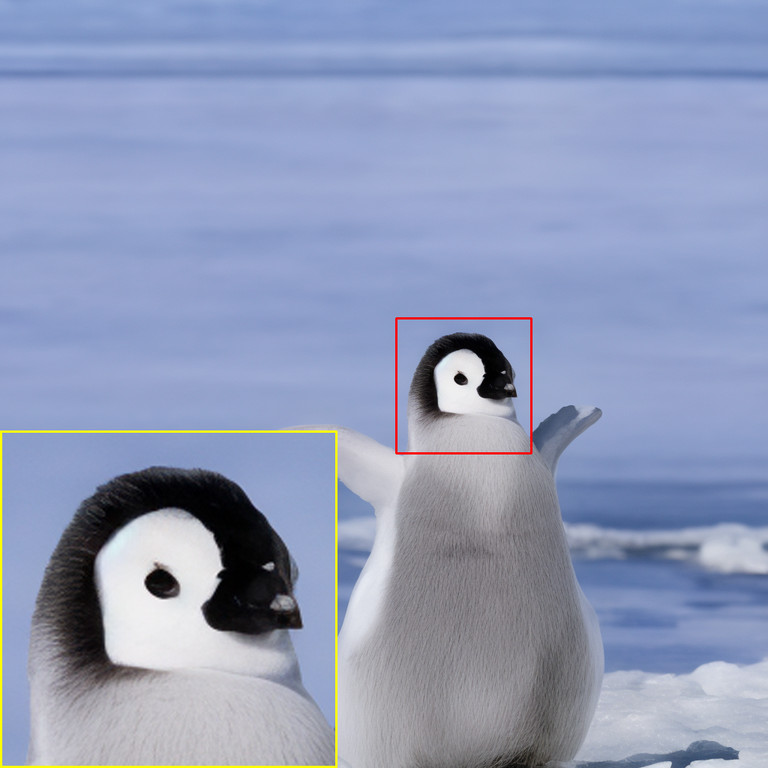} \hspace{-1.2mm} &
    \includegraphics[width=0.188\textwidth]{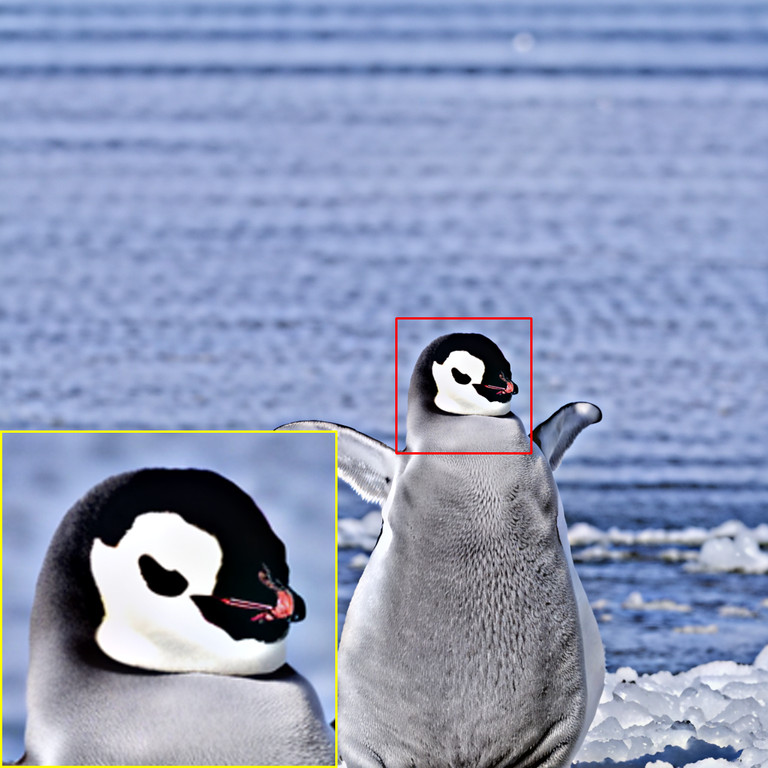} \hspace{-1.2mm} &
    \includegraphics[width=0.188\textwidth]{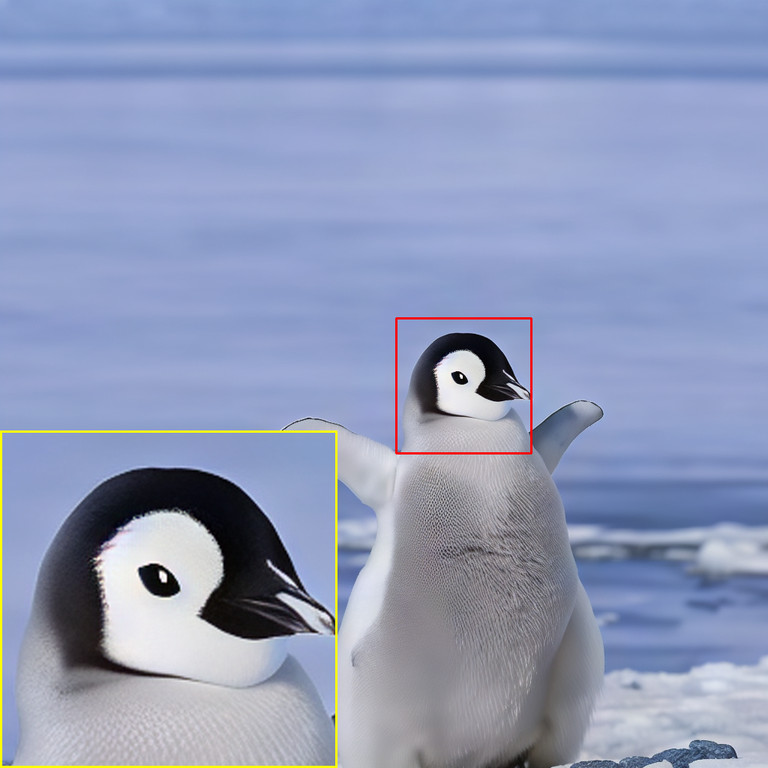} \hspace{-1.2mm} &
    \includegraphics[width=0.188\textwidth]{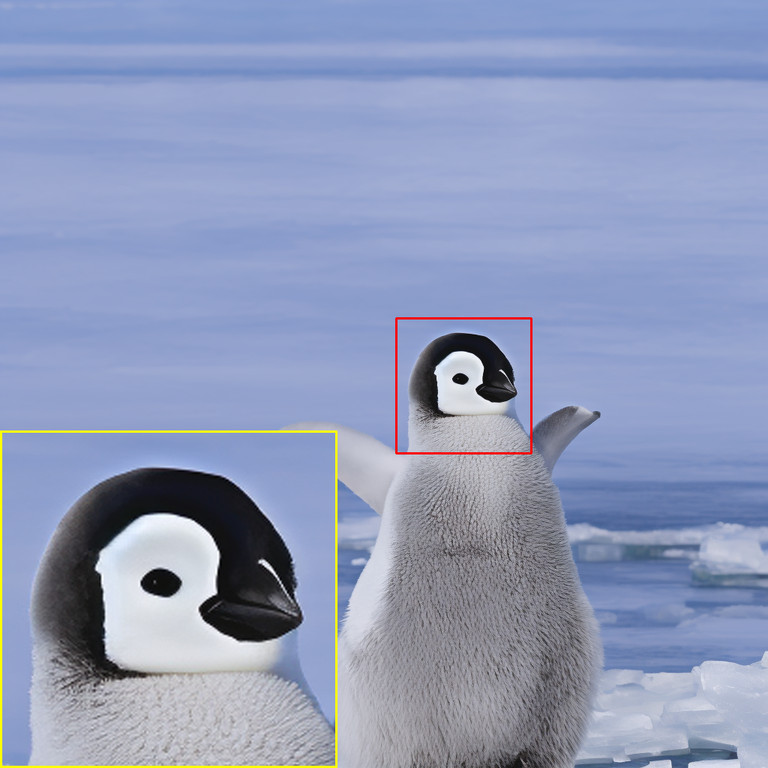} \hspace{-1.2mm} \\
    RealLQ250: \texttt{\imgid} \hspace{-1.2mm} &
    DiffBIR~\cite{lin2024diffbir} \hspace{-1.2mm} &
    SUPIR~\cite{yu2024supir} \hspace{-1.2mm} &
    PASD~\cite{yang2024pasd} \hspace{-1.2mm} &
    SeeSR~\cite{wu2024seesr}  \hspace{-1.2mm} \\
    
    \includegraphics[width=0.188\textwidth]{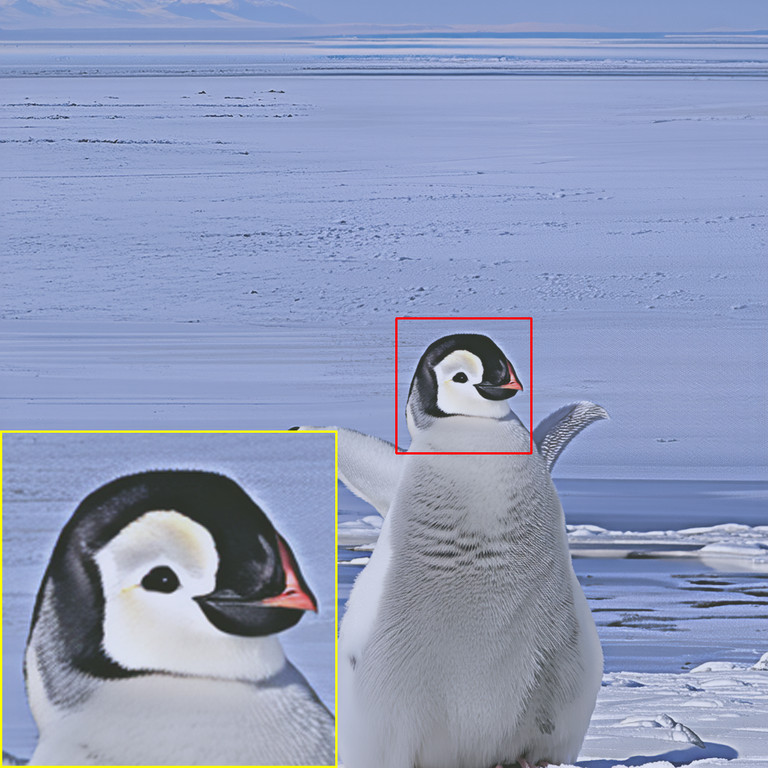} \hspace{-1.2mm} &
    \includegraphics[width=0.188\textwidth]{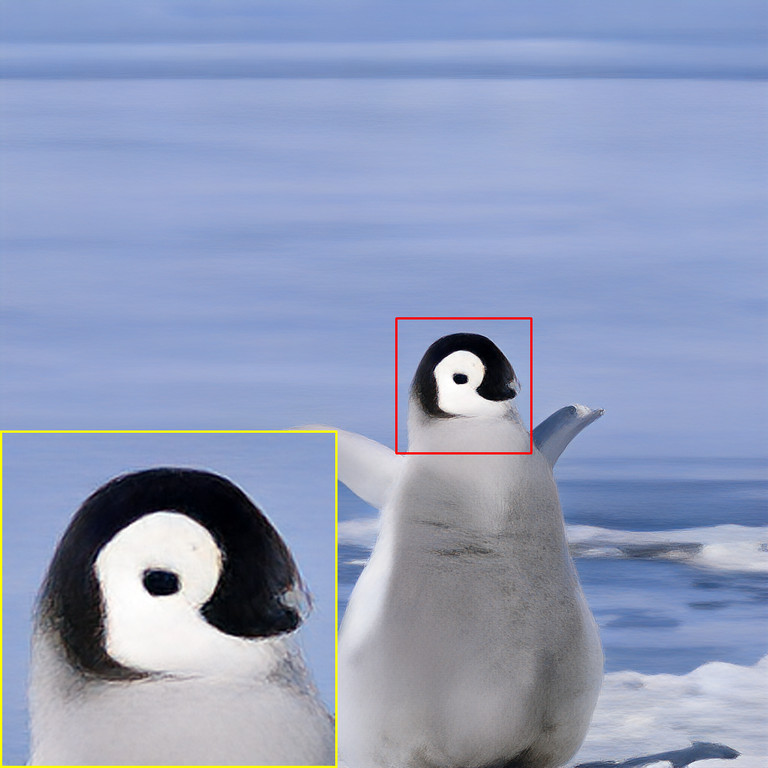} \hspace{-1.2mm} &
    \includegraphics[width=0.188\textwidth]{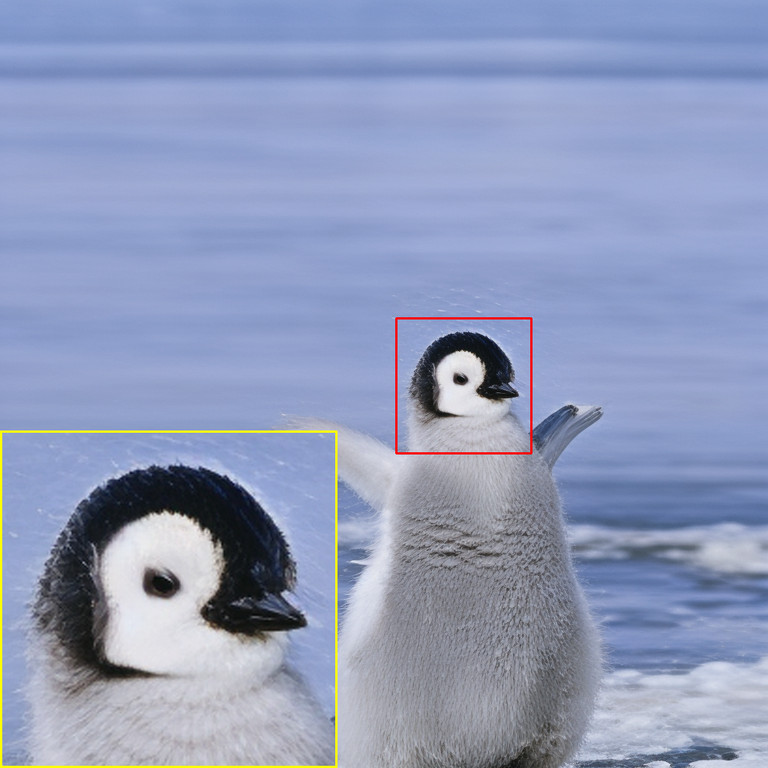} \hspace{-1.2mm} &
    \includegraphics[width=0.188\textwidth]{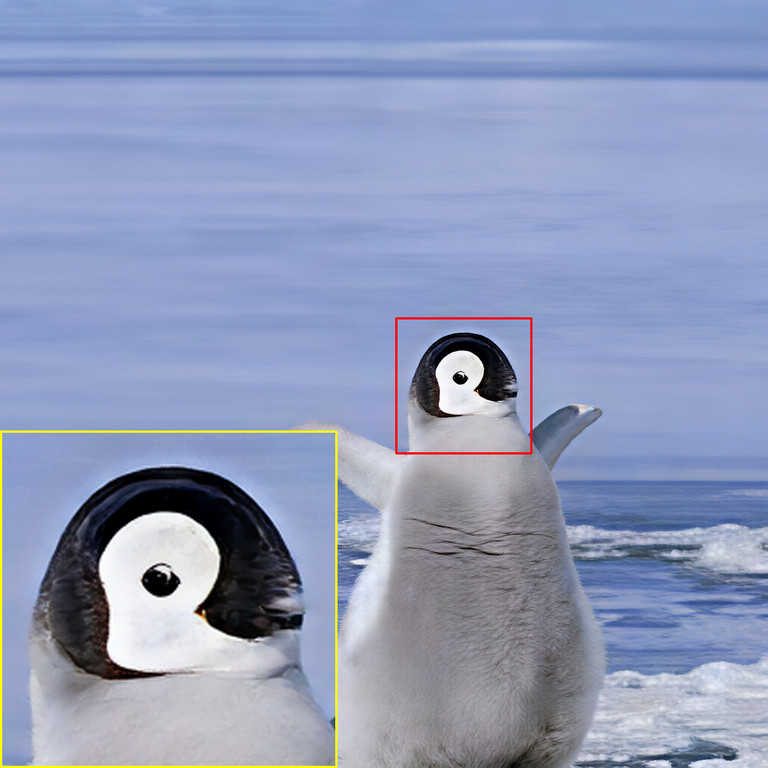} \hspace{-1.2mm} &
    \includegraphics[width=0.188\textwidth]{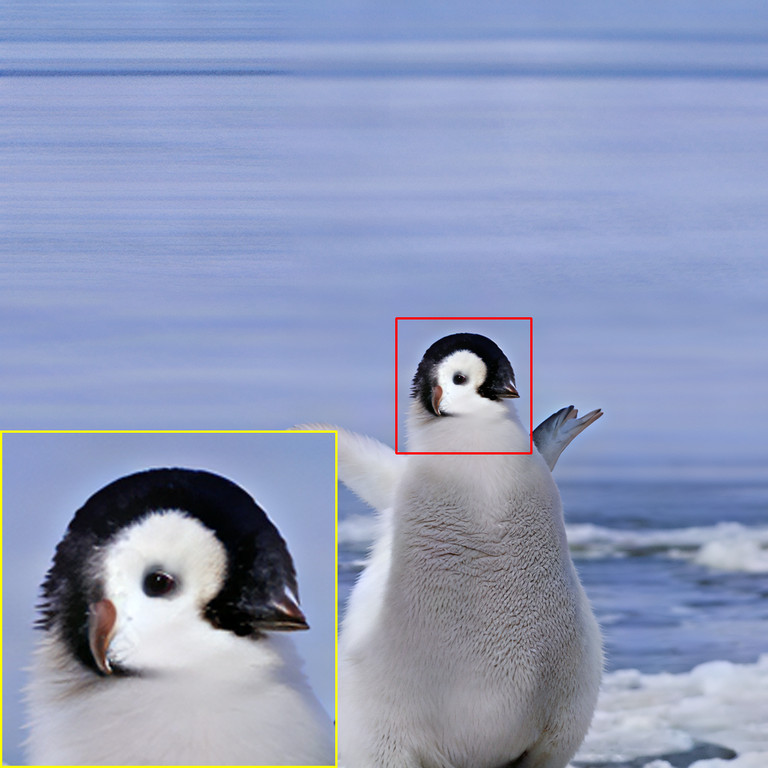} \hspace{-1.2mm} \\
    DiT4SR\cite{duan2025dit4sr} \hspace{-1.2mm} &
    CTMSR~\cite{you2025ctmsr} \hspace{-1.2mm} &
    PiSA-SR~\cite{sun2024pisasr} \hspace{-1.2mm} &
    TSD-SR~\cite{dong2025tsdsr} \hspace{-1.2mm} &
    HYPIR~\cite{lin2025hypir}  \hspace{-1.2mm} \\

    \includegraphics[width=0.188\textwidth]{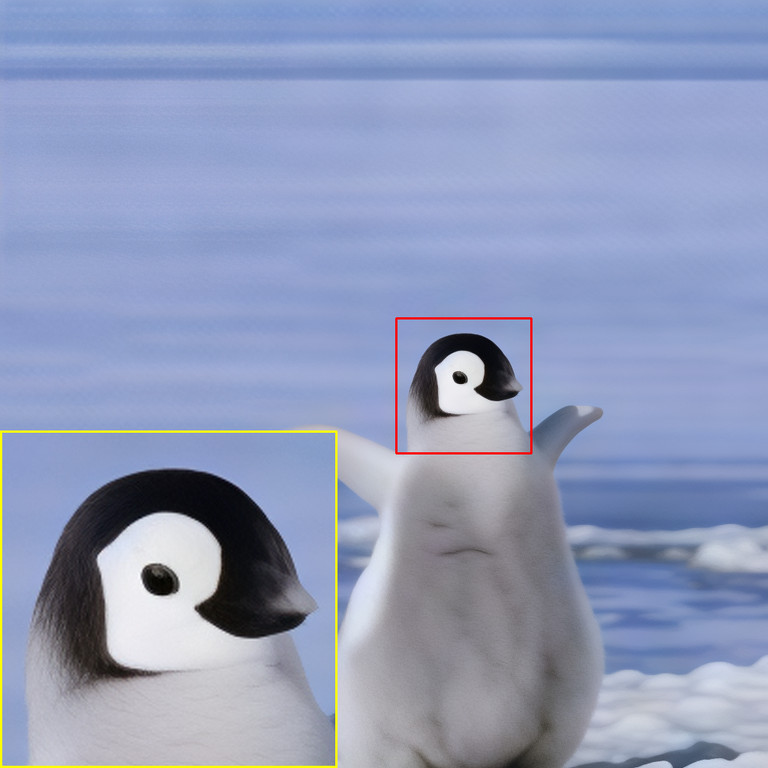} \hspace{-1.2mm} &
    \includegraphics[width=0.188\textwidth]{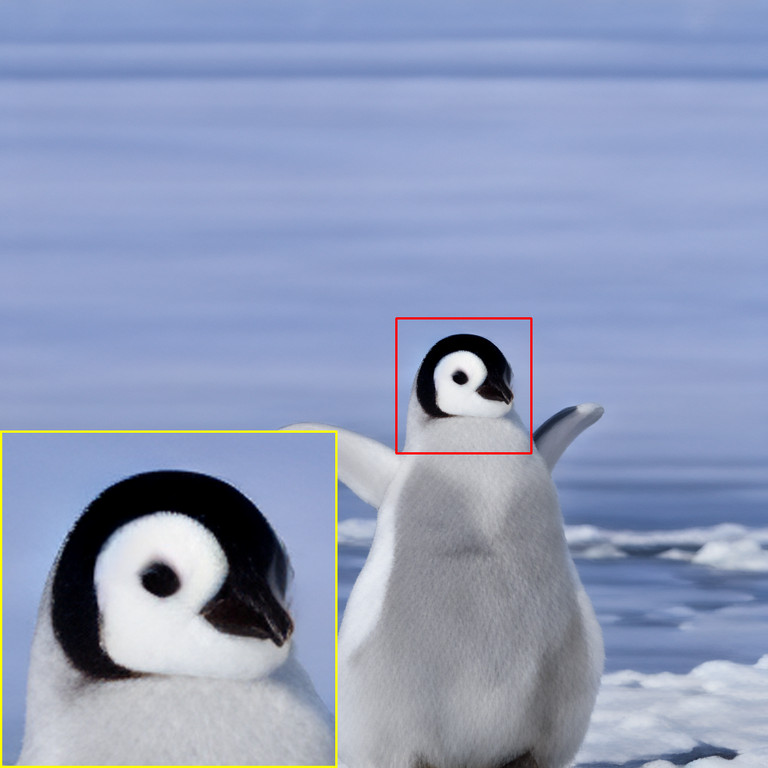} \hspace{-1.2mm} &
    \includegraphics[width=0.188\textwidth]{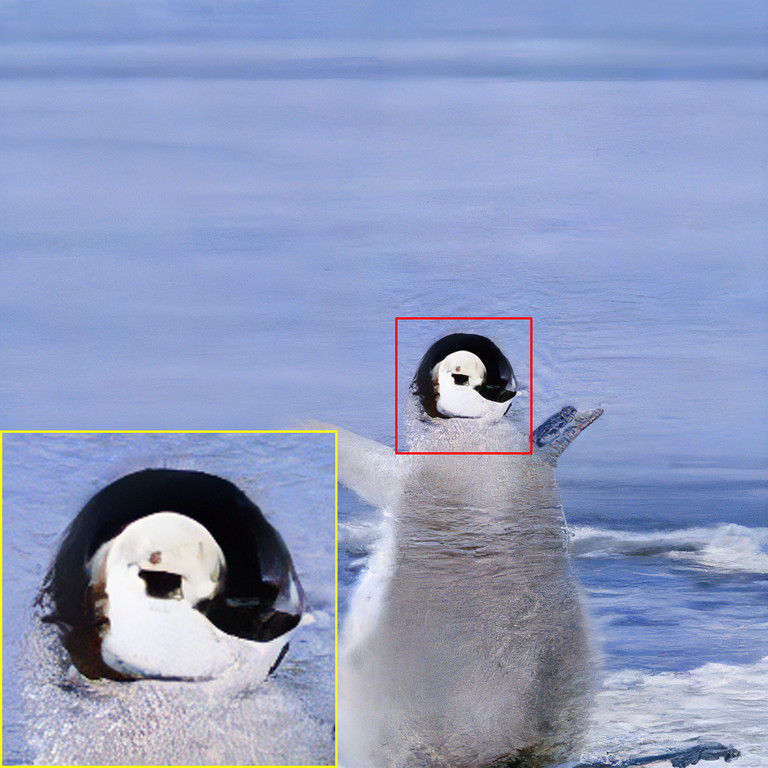} \hspace{-1.2mm} &
    \includegraphics[width=0.188\textwidth]{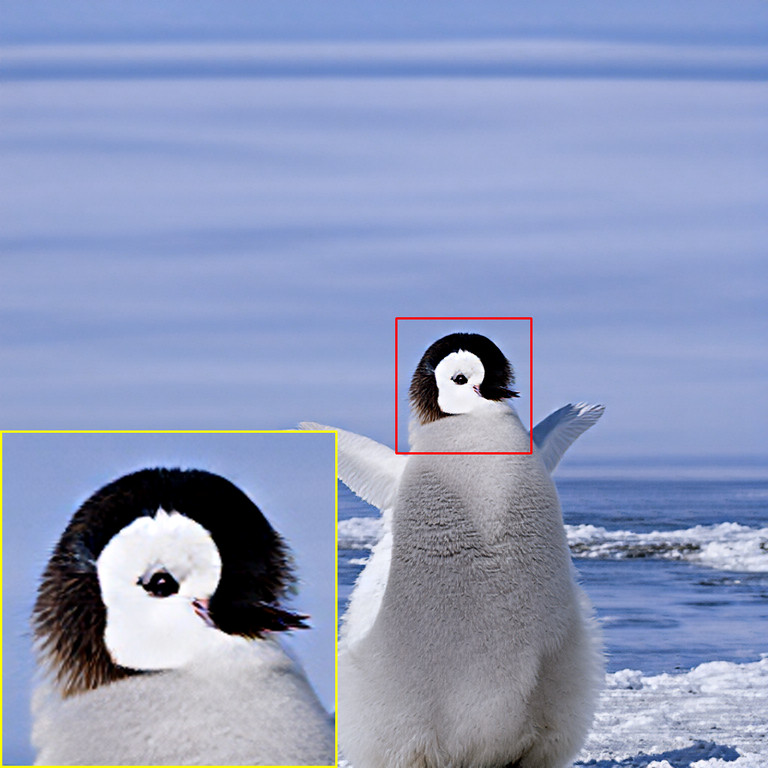} \hspace{-1.2mm} &
    \includegraphics[width=0.188\textwidth]{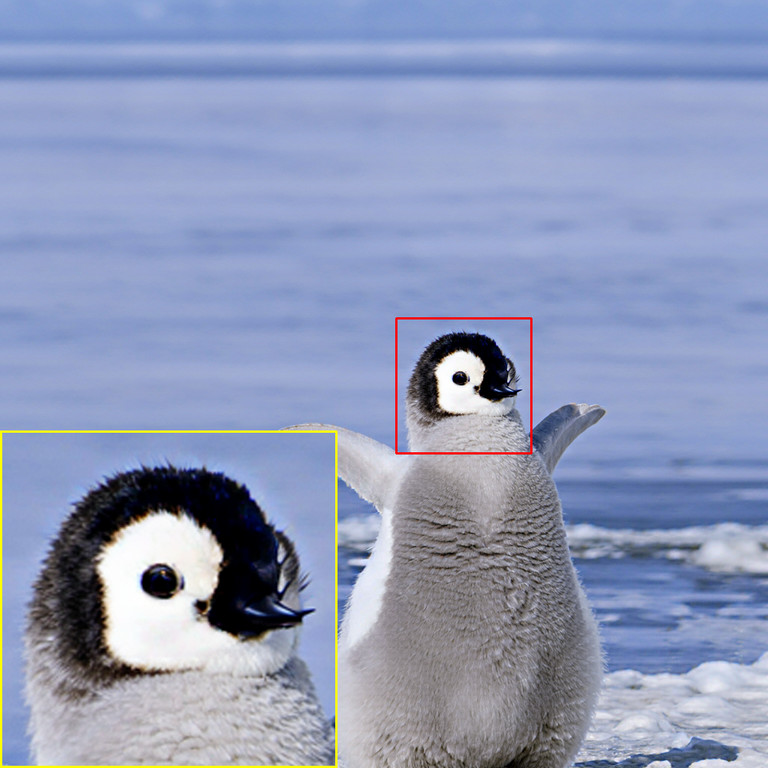} \hspace{-1.2mm} \\
    InvSR~\cite{yue2025invsr} \hspace{-1.2mm} &
    OSEDiff~\cite{wu2024osediff} \hspace{-1.2mm} &
    SinSR~\cite{wang2024sinsr} \hspace{-1.2mm} &
    StrSR - Z-Image  \hspace{-1.2mm} &
    StrSR - FLUX \hspace{-1.2mm}  \\

    \end{tabular}
    \end{adjustbox}
}

\vspace{6mm}

\renewcommand{\imgid}{244}
\scalebox{1.0}{
    \hspace{-0.4cm}
    \begin{adjustbox}{valign=t}
    \begin{tabular}{cccccc}
    \includegraphics[width=0.188\textwidth]{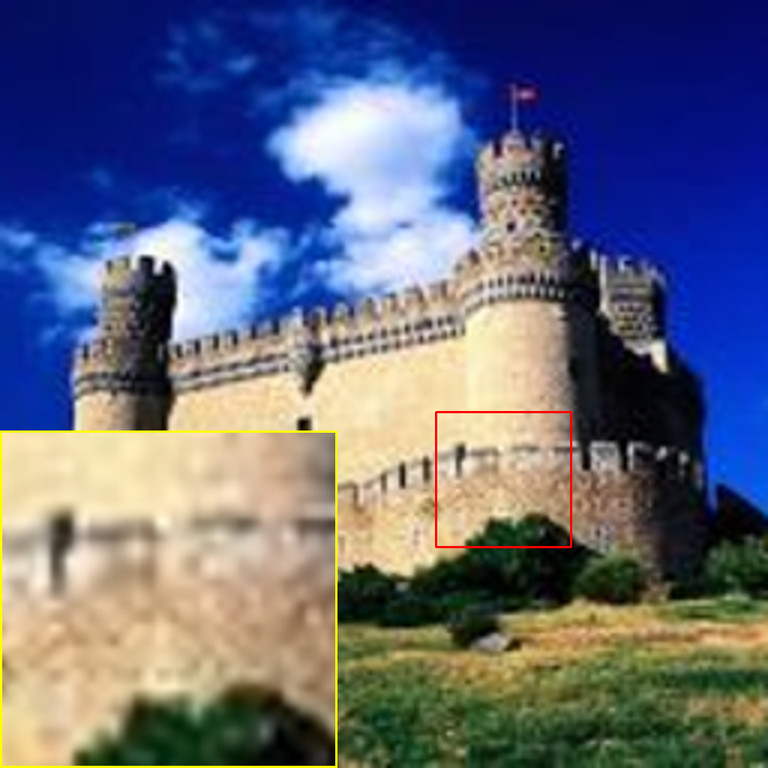} \hspace{-1.2mm} &
    \includegraphics[width=0.188\textwidth]{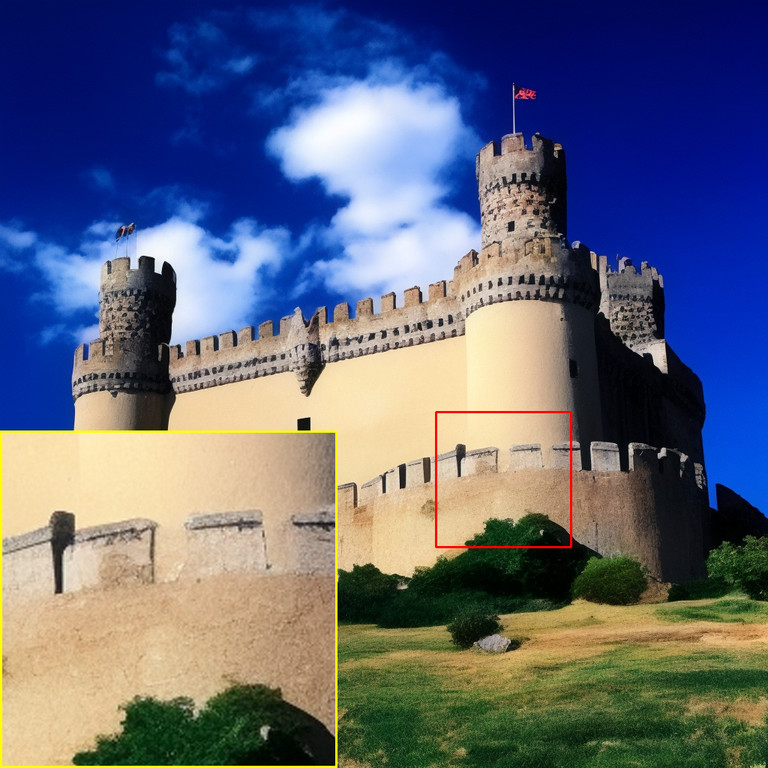} \hspace{-1.2mm} &
    \includegraphics[width=0.188\textwidth]{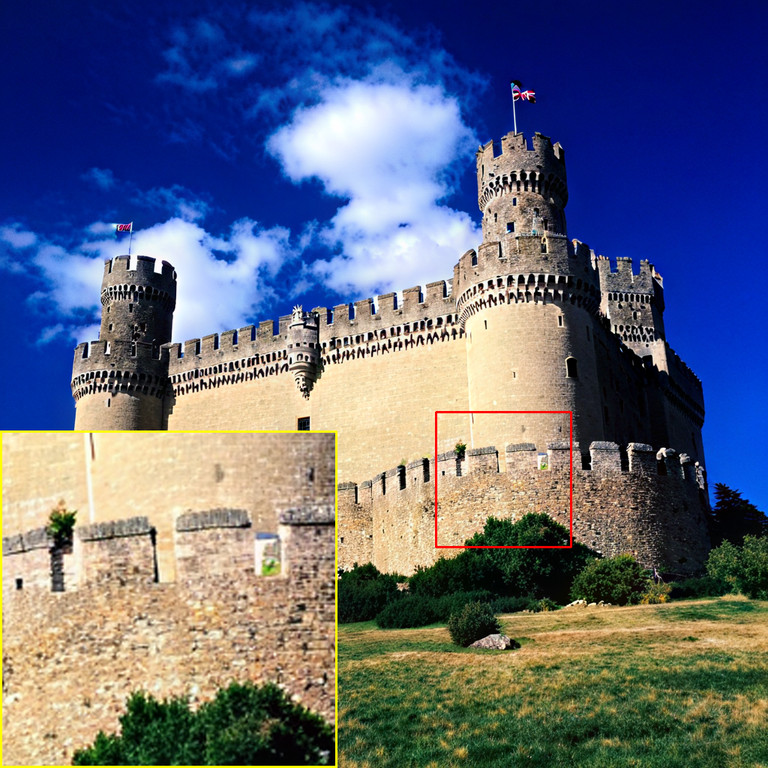} \hspace{-1.2mm} &
    \includegraphics[width=0.188\textwidth]{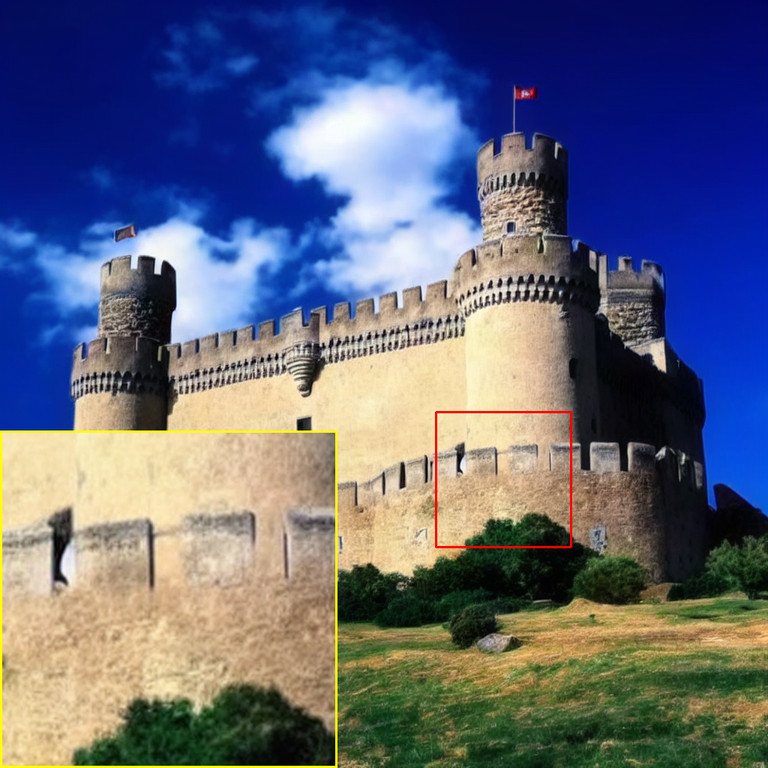} \hspace{-1.2mm} &
    \includegraphics[width=0.188\textwidth]{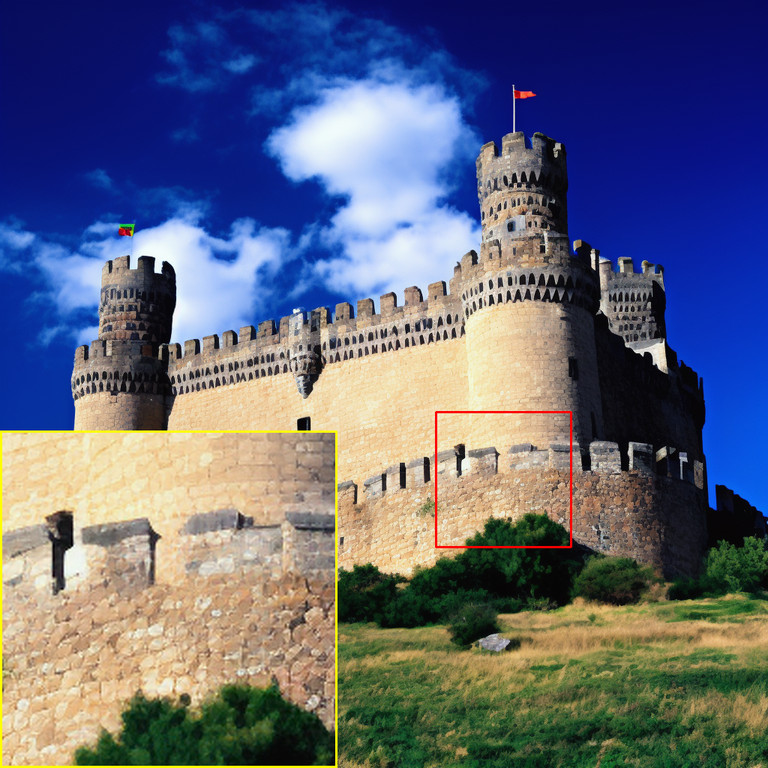} \hspace{-1.2mm} \\
    RealLQ250: \texttt{\imgid} \hspace{-1.2mm} &
    DiffBIR~\cite{lin2024diffbir} \hspace{-1.2mm} &
    SUPIR~\cite{yu2024supir} \hspace{-1.2mm} &
    PASD~\cite{yang2024pasd} \hspace{-1.2mm} &
    SeeSR~\cite{wu2024seesr}  \hspace{-1.2mm} \\
    
    \includegraphics[width=0.188\textwidth]{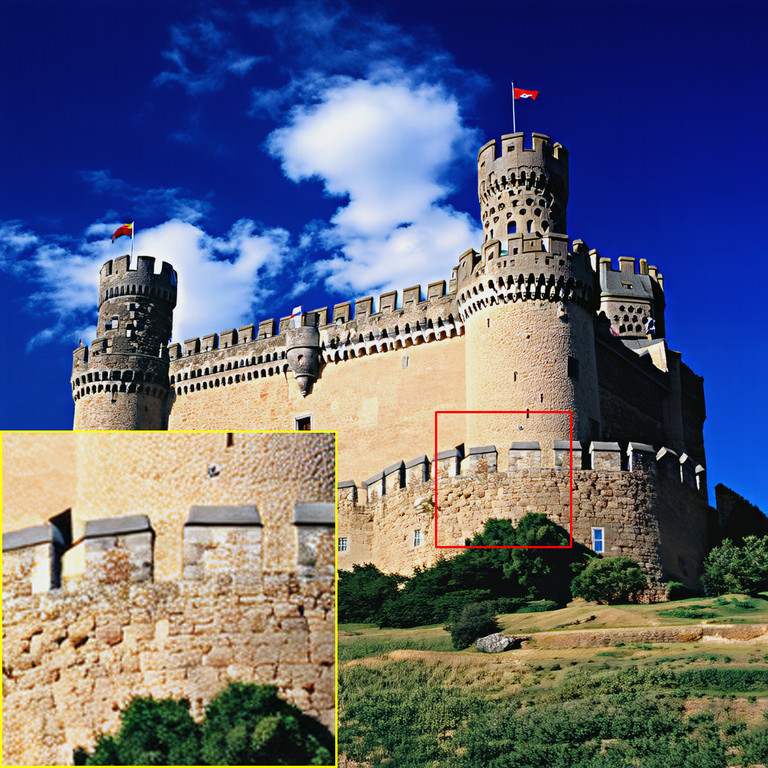} \hspace{-1.2mm} &
    \includegraphics[width=0.188\textwidth]{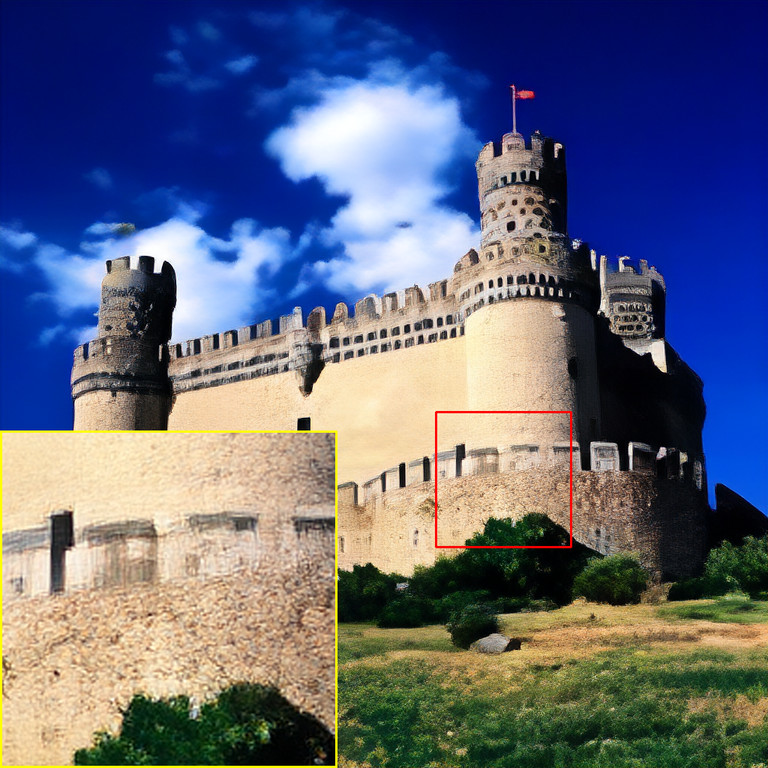} \hspace{-1.2mm} &
    \includegraphics[width=0.188\textwidth]{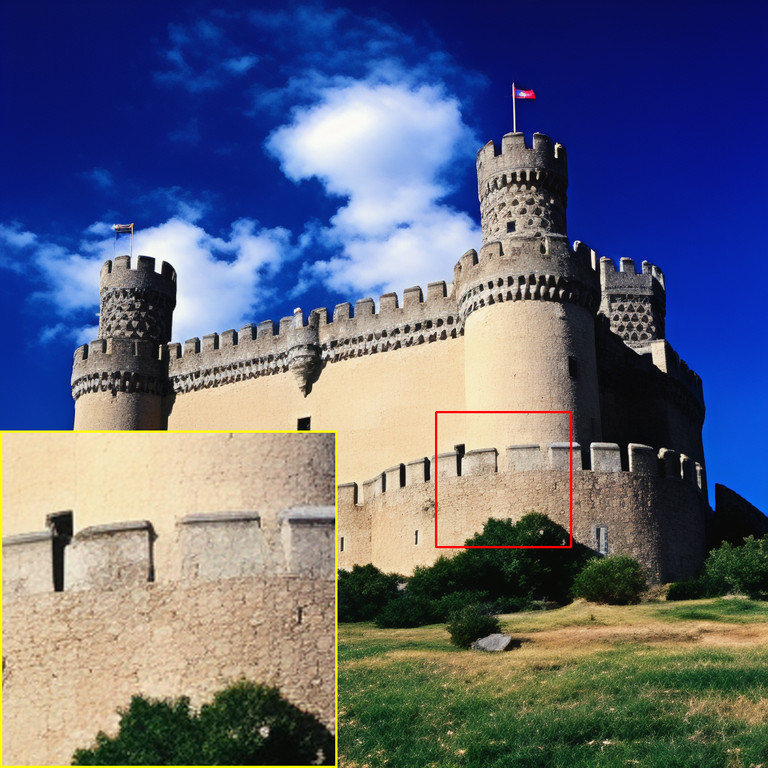} \hspace{-1.2mm} &
    \includegraphics[width=0.188\textwidth]{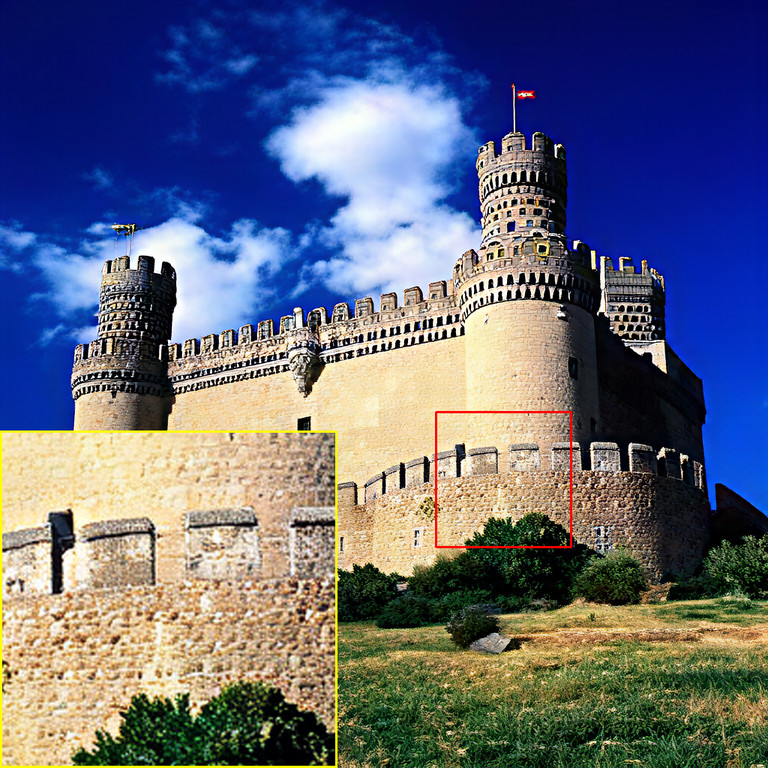} \hspace{-1.2mm} &
    \includegraphics[width=0.188\textwidth]{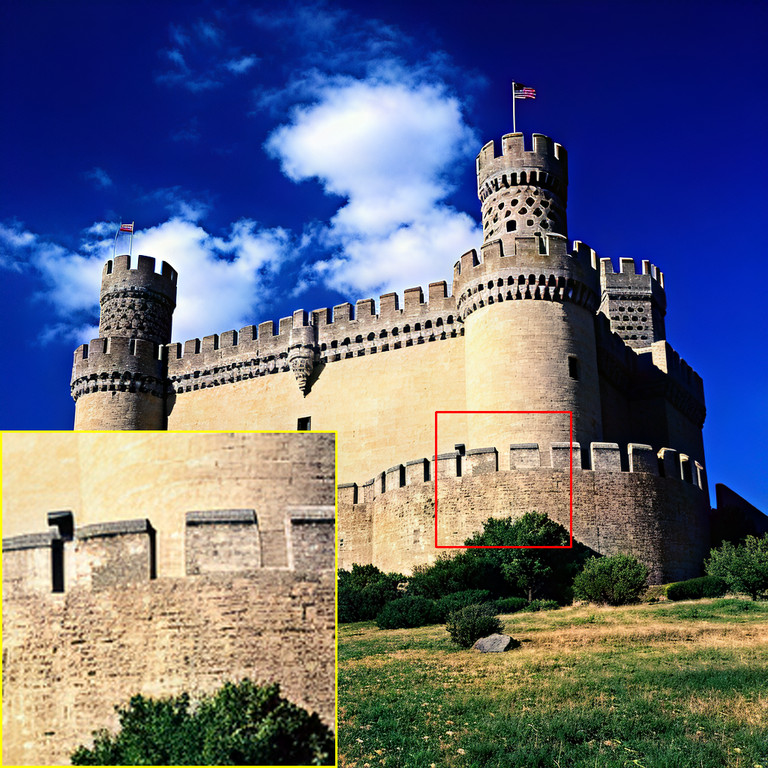} \hspace{-1.2mm} \\
    DiT4SR\cite{duan2025dit4sr} \hspace{-1.2mm} &
    CTMSR~\cite{you2025ctmsr} \hspace{-1.2mm} &
    PiSA-SR~\cite{sun2024pisasr} \hspace{-1.2mm} &
    TSD-SR~\cite{dong2025tsdsr} \hspace{-1.2mm} &
    HYPIR~\cite{lin2025hypir}  \hspace{-1.2mm} \\

    \includegraphics[width=0.188\textwidth]{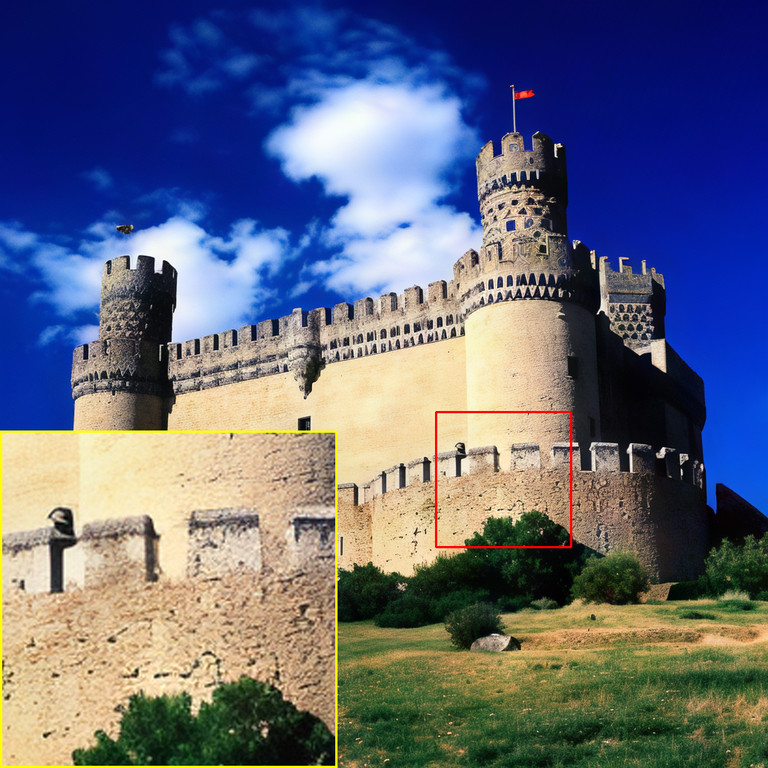} \hspace{-1.2mm} &
    \includegraphics[width=0.188\textwidth]{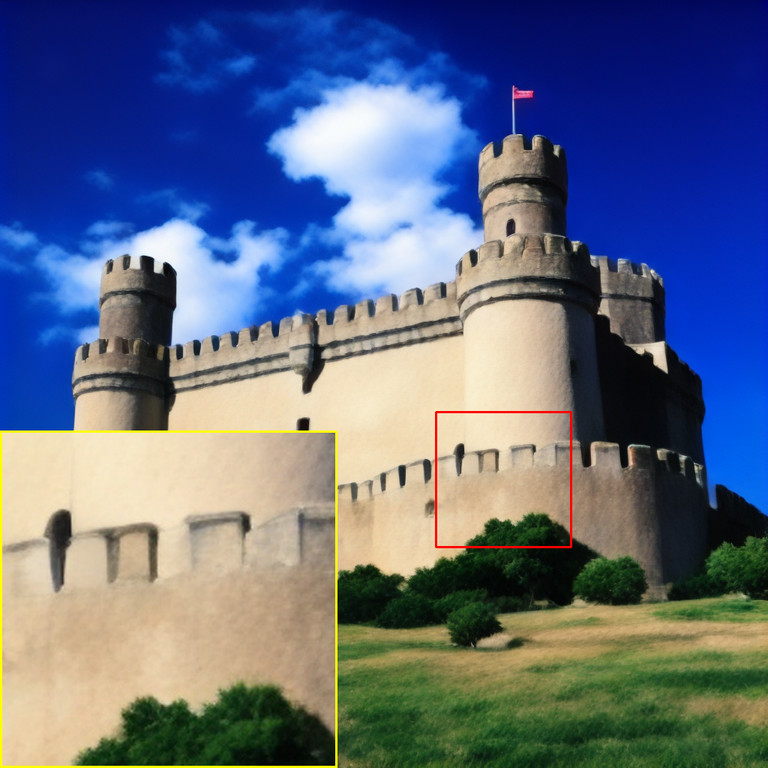} \hspace{-1.2mm} &
    \includegraphics[width=0.188\textwidth]{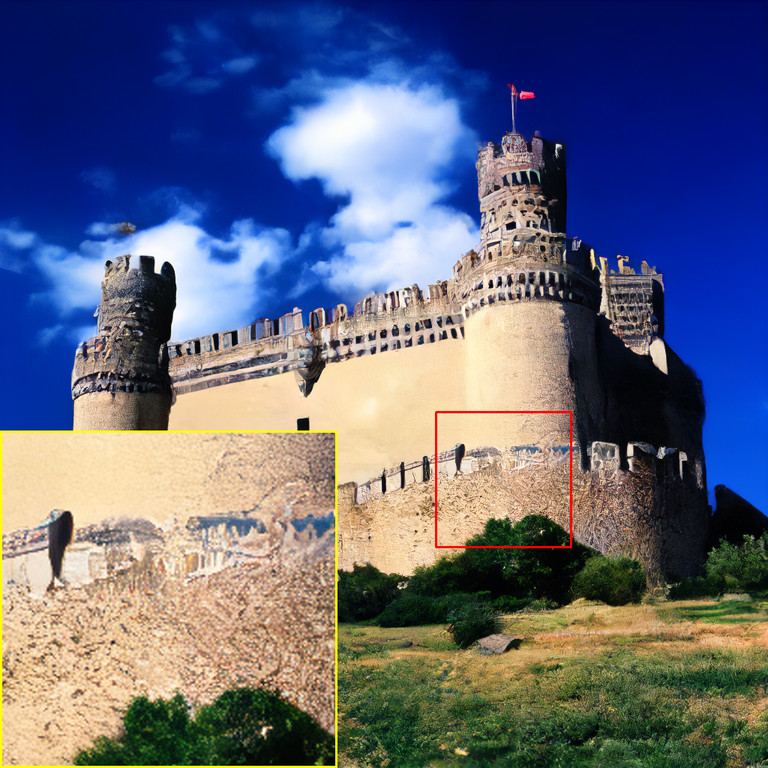} \hspace{-1.2mm} &
    \includegraphics[width=0.188\textwidth]{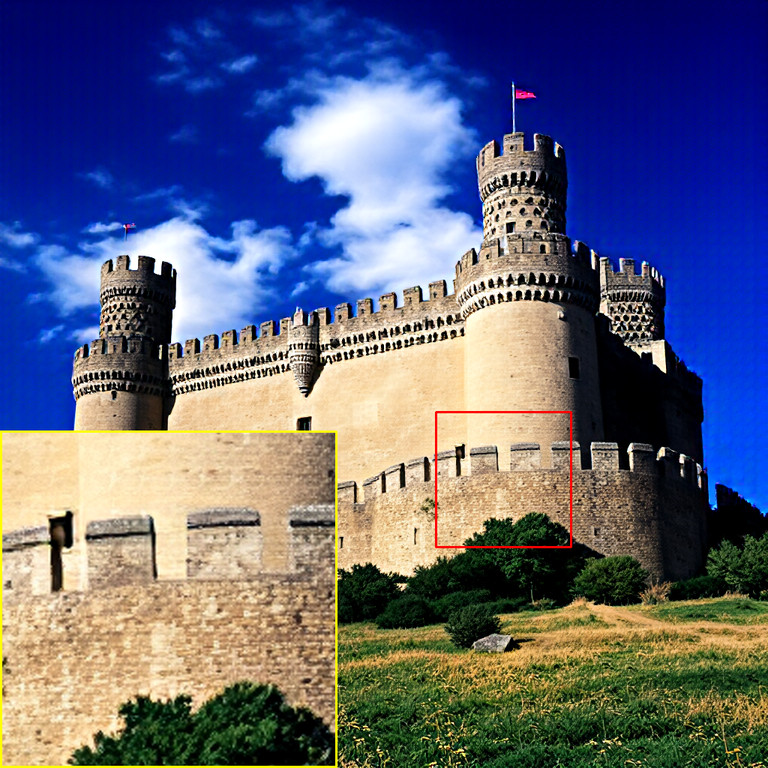} \hspace{-1.2mm} &
    \includegraphics[width=0.188\textwidth]{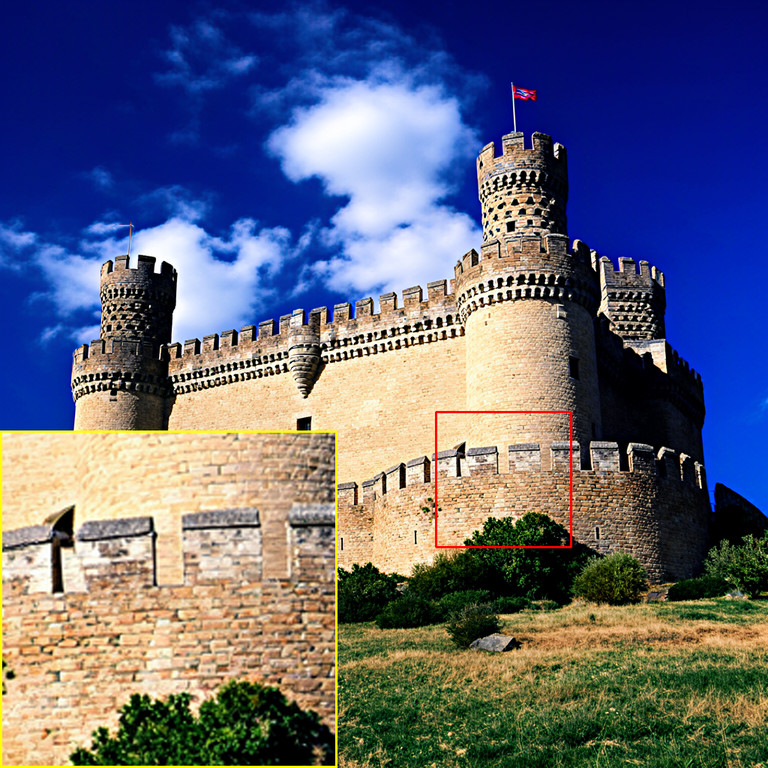} \hspace{-1.2mm} \\
    InvSR~\cite{yue2025invsr} \hspace{-1.2mm} &
    OSEDiff~\cite{wu2024osediff} \hspace{-1.2mm} &
    SinSR~\cite{wang2024sinsr} \hspace{-1.2mm} &
    StrSR - Z-Image  \hspace{-1.2mm} &
    StrSR - FLUX \hspace{-1.2mm}  \\

    \end{tabular}
    \end{adjustbox}
}

\vspace{6mm}
\caption{More visual comparison for image SR ($\times$4) in RealLQ250 dataset. }
\label{fig:vis-reallq250-3}
\end{figure*}

In addition, we conduct visual comparisons with more multi-step diffusion models on RealLQ250 dataset, such as DiffBIR~\cite{lin2024diffbir}, SUPIR~\cite{yu2024supir}, PASD~\cite{yang2024pasd}, SeeSR~\cite{wu2024seesr}, and DiT4SR~\cite{duan2025dit4sr}. We evaluate these models' performance on the real-world RealLQ250 dataset. As shown in Figs.~\ref{fig:vis-reallq250-1},~\ref{fig:vis-reallq250-2}, and ~\ref{fig:vis-reallq250-3}, our models show superior capability to handle real-world degradation compared with other one-step and multi-step models, especially when dealing with high-frequency regions. For instance, StrSR can generate detailed brick textures in Fig.~\ref{fig:vis-reallq250-1}, while most other methods only reconstruct shallow tile-like stripes or irregular tiles. Additionally, StrSR effectively utilizes semantic information for detail reconstruction, such as stone carvings in Fig.~\ref{fig:vis-reallq250-2} and penguin feathers in Fig.~\ref{fig:vis-reallq250-3}.

\bibliographystyle{splncs04}
\bibliography{main}